\documentclass{article}

 \usepackage[main, final]{neurips_2026}

\usepackage[utf8]{inputenc}
\usepackage[T1]{fontenc}

\usepackage{amsmath,amssymb,amsfonts,amsthm,bbm}
\usepackage{mathtools}
\usepackage{microtype}
\usepackage{xcolor}
\usepackage{pifont}

\usepackage{url}
\usepackage{booktabs}
\usepackage{nicefrac}
\usepackage{enumerate,enumitem}
\usepackage{epsfig}
\usepackage{algorithm,algorithmic}
\usepackage{framed}
\usepackage{setspace}
\usepackage[title]{appendix}
\usepackage{tikz}
\usetikzlibrary{positioning,calc,arrows.meta}
\usepackage[most]{tcolorbox}
\usepackage{pifont}
\usepackage{multirow}
\usepackage{tabulary}
\usepackage{colortbl}
\usepackage{tablefootnote}
\usepackage{threeparttable}
\allowdisplaybreaks[4]

\usepackage{hyperref}

\usepackage{aliascnt}
\usepackage{makecell}
\theoremstyle{plain}
\newtheorem{theorem}{Theorem}[section]

\newaliascnt{lemma}{theorem}
\newtheorem{lemma}[lemma]{Lemma}
\aliascntresetthe{lemma}

\newaliascnt{corollary}{theorem}

\aliascntresetthe{corollary}

\newaliascnt{proposition}{theorem}

\aliascntresetthe{proposition}

\theoremstyle{definition}
\newaliascnt{definition}{theorem}

\aliascntresetthe{definition}

\newaliascnt{assumption}{theorem}
\newtheorem{assumption}[assumption]{Assumption}
\aliascntresetthe{assumption}

\theoremstyle{remark}
\newaliascnt{remark}{theorem}
\newtheorem{remark}[remark]{Remark}
\aliascntresetthe{remark}

\usepackage[capitalize,noabbrev]{cleveref}
\crefname{equation}{Eq.}{Eqs.}
\Crefname{equation}{Eq.}{Eqs.}

\crefname{theorem}{Theorem}{Theorems}
\Crefname{theorem}{Theorem}{Theorems}

\crefname{lemma}{Lemma}{Lemmas}
\Crefname{lemma}{Lemma}{Lemmas}

\crefname{corollary}{Corollary}{Corollaries}
\Crefname{corollary}{Corollary}{Corollaries}

\crefname{proposition}{Proposition}{Propositions}
\Crefname{proposition}{Proposition}{Propositions}

\crefname{definition}{Definition}{Definitions}
\Crefname{definition}{Definition}{Definitions}

\crefname{assumption}{Assumption}{Assumptions}
\Crefname{assumption}{Assumption}{Assumptions}

\crefname{remark}{Remark}{Remarks}
\Crefname{remark}{Remark}{Remarks}

\renewcommand{\appendixname}

\usepackage[textsize=tiny]{todonotes}

\usepackage{subcaption}

\title{Orthogonal Updates for the Win: Towards Accelerated \\ Adaptive Minimax Optimization}

\author{%
  Zhiwei Zhai$^{1}$,
  Xinyu Wang$^{2}$,
  Wenjing Yan$^{1}$,
  Lei Ding$^{3}$,
  Ying-Jun Angela Zhang$^{1}$ \\[0.5em]
  $^{1}$The Chinese University of Hong Kong
  \quad
  $^{2}$McGill University
  \quad
  $^{3}$University of Manitoba \\[0.6em]
  {\small\texttt{zz024@ie.cuhk.edu.hk}},
  {\small\texttt{xinyu.wang5@mail.mcgill.ca}},
  {\small\texttt{wjyan@ie.cuhk.edu.hk}}, \\[0.2em]
  {\small\texttt{lei.ding@umanitoba.ca}},
  {\small\texttt{yjzhang@ie.cuhk.edu.hk}}
}

\begin{document}

\maketitle

\vspace{-0.25cm}
\begin{abstract}
Single-loop minimax methods are appealing for modern machine learning, but standard stochastic gradient updates can be unstable and oscillatory under tightly coupled primal--dual dynamics, making performance notoriously sensitive to stepsizes. To fundamentally improve stability, we develop an orthogonal update framework tailored to minimax optimization, which applies updates along orthogonalized directions to reduce update anisotropy. However, orthogonalization reshapes the optimization geometry and makes coordinating primal--dual stepsizes even more delicate.
To overcome this challenge, we develop \textbf{AdaSGDA}, which equips orthogonal updates with an adaptive stepsize rule driven by accumulated gradient norms. This mechanism automatically balances coupled primal--dual progress under orthogonalization, yielding an effectively parameter-free method for nonconvex-strongly-concave minimax optimization.
We prove that AdaSGDA avoids problem-dependent tuning while attaining the state-of-the-art $\mathcal{O}\left(T^{-1/4}\right)$ convergence rate. We further propose \textbf{AdaSGDA-VR}, which incorporates a carefully designed variance reduction scheme and achieves the faster $\mathcal{O}\left(T^{-1/3}\right)$ rate. Extensive experiments demonstrate that our methods deliver competitive performance compared to existing minimax optimization methods.
\end{abstract}

\vspace{-0.3cm}
\section{Introduction}
\vspace{-0.2cm}
With the rapid growth of machine learning research, many fundamental problems go beyond simple minimization. While classical training objectives can often be written as minimizing an empirical risk, modern learning pipelines frequently involve two competing objectives that naturally lead to minimax formulations. In these problems, one typically {minimizes} a model-side loss over the primal variable (e.g., network parameters), while {maximizing} over an adversary or a reweighting distribution that captures worst-case behaviors. Representative examples include generative adversarial networks (GAN) \citep{goodfellow2014generative,stephanovitch2023wasserstein}, robust optimization \citep{gulrajani2017improved, mohri2019agnostic}, adversarial training of neural networks \citep{wang2021adversarial}, and fair machine learning \citep{madras2018learning, shekhar2021adaptive}, etc. These applications highlight the necessity of reliable minimax optimization methods as core building blocks in modern machine learning.

A standard baseline for minimax optimization is Gradient Descent Ascent (GDA) \citep{dem1972numerical,nemirovski2009robust}, which performs a descent step on the primal variable and an ascent step on the dual variable using stochastic gradients. However, reliable primal progress often requires the dual variable to closely track the inner maximizer, which typically necessitates multiple ascent steps or an explicit inner procedure before each primal update \citep{yang2022nest}. Such inner-loop computations substantially increase cost and complicate implementation, making them a major obstacle to scalable minimax optimization.
This has spurred interest in {single-loop} methods \citep{yang2022faster}, which use only one stochastic gradient per iteration and are therefore easier to deploy. Yet, without an inner procedure, the dual variable may drift away from its instantaneous optimum, and the coupled dynamics become highly sensitive to time-scale separation. Consequently, performance can depend critically on the choice of stepsizes and their relative scaling to prevent instability, oscillation, or stagnation.

Existing theory indicates that optimal stepsize choices in minimax optimization depend on problem-specific quantities such as smoothness, strong concavity, and gradient noise levels \citep{zhang2021complexity,yang2020global}. Even with adaptive methods such as AdaGrad \citep{ward2020adagrad}, current minimax guarantees are only near-optimal, or they still require tuning guided by these problem parameters to keep the primal and dual updates on compatible time scales \citep{yang2022nest,li2022tiada,huang2024achieving}. Without such guidance, standard adaptive minimax updates often fail to properly balance the primal--dual progress, resulting in slow convergence and leading to near-optimal rates. This exposes a double bind: existing methods rarely achieve both optimal convergence rates and problem-parameter-free deployment at the same time. 

To fundamentally address this limitation, we consider developing an orthogonal update framework for minimax optimization, which applies updates along orthogonalized directions to {regularize the update geometry} and improve stability. A recent example highlighting the promise of such updates is the Muon optimizer \citep{Muon2024}, which has shown strong empirical success in large language model training \citep{liu2025muon}. Its key idea is orthogonalized momentum for matrix parameters, which distributes updates more evenly across directions and often leads to more stable progress.
However, orthogonalization can also substantially reshape the optimization dynamics \citep{jordan2024muon}, making stepsize selection nontrivial even for standard minimization---and even more delicate in minimax settings where the primal and dual variables must be coordinated. Yet, the very mechanism that makes orthogonalized updates effective also obscures how their stepsizes should be chosen, especially under coupled minimax dynamics. This raises a fundamental question:

\vspace{-0.14cm}
\begin{tcolorbox}[colback=gray!5,
    colframe=black!70, 
    frame style={line width=10pt}, arc=2mm  ]
\textbf{Can we design a single-loop, parameter-free minimax algorithm with orthogonal updates that achieves the optimal convergence rate, matching well-tuned counterparts?
}
\end{tcolorbox}
\vspace{-0.14cm}

\vspace{-0.2cm}
\subsection{Main Contribution}
\vspace{-0.25cm}
We answer this question affirmatively. We develop \textbf{AdaSGDA}, a single-loop algorithm for nonconvex-strongly-concave (NC--SC) minimax optimization that leverages the complementary strengths of orthogonal updates and adaptive stepsizes. AdaSGDA stabilizes the coupled primal--dual dynamics by updating along orthogonalized directions, while an accumulated-gradient-norm-based scaling rule automatically balances the relative progress of the two variables. As a result, AdaSGDA is {problem-parameter-free} and provably attains the {optimal} convergence rate. Our main contributions are summarized as follows:
\begin{itemize}[topsep=0pt, itemsep=2pt, parsep=2pt, leftmargin=0.25cm]

\item We propose \textbf{AdaSGDA}, the first \emph{parameter-free} algorithm that attains the \emph{optimal convergence rate} for minimax optimization. 
AdaSGDA is built upon orthogonalized momentum with adaptive scaling, but making these ingredients work for stochastic minimax is highly nontrivial: orthogonalization can substantially amplify stochastic noise when gradient matrices are ill conditioned \citep{zhang2025sharp}, making naive norm accumulation ineffective.
Moreover, the coupled minimax dynamics require the primal and dual updates to be coordinated within a single-loop scheme. To address these issues, AdaSGDA introduces two key mechanisms: \textbf{(i) orthogonalization-calibrated adaptive stepsizes}, which adapt to historical gradient information while applying updates in a geometry that preserves orthogonality; and \textbf{(ii) hierarchical stepsize design} that respects the interdependence between the two variables while preserving the autonomy of adaptive stepsizes.

\item We conduct a comprehensive theoretical analysis for our adaptive minimax scheme. This setting is technically challenging because adaptive scaling and orthogonalized directions make the effective geometry and stochastic error evolve over time, while the primal and dual updates remain tightly coupled. Establishing parameter-free convergence therefore calls for a new proof framework beyond existing minimax analyses. In our analysis, we show that AdaSGDA achieves a convergence rate of $\mathcal{O}(T^{-1/4})$, matching well-tuned methods, and that incorporating variance reduction further improves this rate to $\mathcal{O}(T^{-1/3})$. 
Notably, in contrast to existing parameter-free minimax approaches \citep{yang2022nest,li2022tiada,huang2024achieving}, which typically obtain only near-optimal rates, our guarantees attain the best-known optimal rate.

\item 
We evaluate our method on several benchmark tasks, including distributionally robust image classification and large language model pretraining, as well as GAN training, and show that our method performs competitive with existing well-tuned baselines. Moreover, our method is markedly less sensitive to the initial stepsize, remaining stable across a broad range of choices, which supports the effectiveness of our adaptive minimax approach.
\end{itemize}  

\begin{table*}\small
\vspace{-0.1cm}
\centering
\renewcommand{\arraystretch}{1.27} 
\setlength{\tabcolsep}{4.1pt} 
\setlength{\heavyrulewidth}{1.5pt} 
\setlength{\lightrulewidth}{1pt} 
\setlength{\cmidrulewidth}{0.6pt} 
\caption{Comparison between different minimax optimization algorithms.\\ \(T\) denotes the number of iterations; $\widetilde{\mathcal{O}}(\cdot)$ suppresses logarithmic factors in the convergence bound; $\alpha$ and $\beta$ are stepsize-decay exponents used in TiAda and D-AdaST, satisfying $0<\beta<\alpha<1$.}
\vspace{-0.1cm}
\begin{threeparttable}
\begin{tabular}{@{}lccccc@{}}
\toprule
\textbf{Algorithm} & \textbf{Single-Loop} & \textbf{Parameter-Free} & \textbf{Optimal-Rate$^\dagger$} &  \textbf{Convergence Rate$^\diamond$} \\ \midrule
GDA \cite{jin2019minmax} & \ding{56} & \ding{56}& \ding{52} & ${\mathcal{O}}\!\left(\frac{1}{T^{1/4}}\right)$\\
TTGDA \cite{lin2025two} & \ding{52} & \ding{56}& \ding{52} & ${\mathcal{O}}\!\left(\frac{1}{T^{1/4}}\right)$\\
NeAda \cite{yang2022nest}   & \ding{56} & \ding{52} & \ding{56} &  $\widetilde{\mathcal{O}}\!\left(\frac{1}{T^{1/4}}\right)$\\
TiAda \cite{li2022tiada}  & \ding{52} & \ding{52} & \ding{56}  & $\mathcal{O}\left(\frac{1}{T^{(1-\alpha)/2}}+\frac{1}{T^{\beta/2}}\right)$ \\
D-AdaST \cite{huang2024achieving}   & \ding{52} & \ding{52} & \ding{56} & $\widetilde{\mathcal{O}}\left(\frac{1}{T^{(1-\alpha)/2}}+\frac{1}{T^{\beta/2}}\right)$  \\ \rowcolor[gray]{0.9}
\textbf{AdaSGDA (This paper)} & \ding{52} & \ding{52} & \ding{52} & $\mathcal{O}\left(\frac{1}{T^{1/4}}\right)$ \\ \rowcolor[gray]{0.9}
\textbf{AdaSGDA-VR (This paper)} & \ding{52} & \ding{52} & \ding{52} & $\mathcal{O}\left(\frac{1}{T^{1/3}}\right)$ \\ \bottomrule
\end{tabular}
\vspace{0cm}
\label{tab:comparison}
 \begin{tablenotes}  
   \footnotesize
   \item[$\dagger$] Optimal-Rate indicates whether the convergence rate matches the best-known scaling in $T$ for the considered setting, without extra residual terms or logarithmic overheads (up to constant factors).
   \item[$\diamond$] Convergence rate refers to an upper bound on the expected first-order stationarity measure averaged over $T$ iterations.
  \end{tablenotes}
\end{threeparttable}
\label{table1}
\vspace{-0.5cm}
\end{table*}

\vspace{-0.3cm}
\subsection{Comparisons with Prior Approaches}
\vspace{-0.2cm}
We compare AdaSGDA with representative minimax methods in \cref{table1}. Compared with non–parameter-free baselines, including non-loopless GDA \citep{jin2019minmax} and loopless TTGDA \citep{lin2025two}, AdaSGDA matches their optimal rate $\mathcal{O}(T^{-1/4})$ while removing problem-parameter-based tuning. This matters in practice, since such parameters are rarely known and typically lead to substantial manual tuning. Moreover, AdaSGDA-VR further reduces stochastic errors via variance reduction and improves the rate to $\mathcal{O}(T^{-1/3})$, outperforming these well-tuned counterparts.
Turning to adaptive methods, NeAda \citep{yang2022nest} is parameter-free but nested and near-optimal with logarithmic overheads, while loopless TiAda \citep{li2022tiada} and D-AdaST \citep{huang2024achieving} are also near-optimal due to residual terms. In contrast, AdaSGDA attains the optimal rate in a single loop by carefully controlling the primal–dual time scales via adaptive scaling and a hierarchical coordination rule. 
Moreover, we pair orthogonalized updates with orthogonalization-calibrated adaptivity, which stabilizes the induced matrix dynamics and prevents adversarial growth of update directions. This structure enables a sharper control of stochastic errors, leading to parameter-free guarantees that achieve the best-known optimal rate. As a result, our method is the first single-loop minimax method that is simultaneously parameter-free and rate-optimal. This represents a significant advancement over existing minimax methods, facilitating robust deployment across diverse environments.

\vspace{-0.3cm}
\section{Preliminaries}
\vspace{-0.2cm}
\subsection{Problem Model}\label{problemmodel}
\vspace{-0.2cm}
In this paper, we consider the following minimax optimization problem:
{
\setlength{\abovedisplayskip}{3pt}
\setlength{\belowdisplayskip}{-2pt}
\setlength{\abovedisplayshortskip}{3pt}
\setlength{\belowdisplayshortskip}{-2pt}
\begin{equation}
\min_{X \in \mathbb{R}^{m_x\times n_x}} \max_{Y \in \mathbb{R}^{m_y\times n_y}} f(X, Y) \triangleq \mathbb{E}_{\xi \sim P}\!\left[F(X, Y; \xi)\right],
\label{P1}
\end{equation}}

where $P$ is an unknown distribution from which i.i.d.\ samples can be drawn, and $\xi$ denotes a random variable. We assume that $f$ is jointly $L$-smooth in $(X,Y)$, possibly nonconvex in $X$, and $\mu$-strongly concave in $Y$.

\textbf{Notations.} We denote $L$ as the smoothness parameter and $\mu$ as the strong-concavity parameter, whose formal definitions are given in Assumptions~\ref{smoothness} and~\ref{concavity}.
We assume access to a stochastic first-order oracle that returns
$\big[\nabla_X F(X,Y;\xi),\, \nabla_Y F(X,Y;\xi)\big]$.
For problem~\eqref{P1}, we denote
$Y^\ast(X) := \arg\max_{Y \in \mathbb{R}^{m_y\times n_y}} f(X,Y)$ as the solution of the inner maximization and
$\Phi(X) := f\big(X, Y^\ast(X)\big)$ as the induced primal function. For a matrix $A \in \mathbb{R}^{m \times n}$, $\|A\|_F$ denotes the Frobenius norm,
$\|A\|_\ast$ denotes the nuclear norm, and $\|A\|_{\mathrm{op}}$ denotes the spectral (operator) norm. We will repeatedly use the standard bound
$\|A\|_\ast \le \sqrt{\mathrm{rank}(A)}\,\|A\|_F \le \sqrt{r}\,\|A\|_F$,
where $r$ is a global rank upper bound for all matrices considered (e.g., $r:=\min\{m,n\}$ for $A\in\mathbb{R}^{m\times n}$).
In addition, for $A, B \in \mathbb{R}^{m \times n}$, we define
$\langle A, B \rangle := \mathrm{Tr}(A^\top B)$, where $\mathrm{Tr}(\cdot)$ denotes the trace.

\vspace{-0.25cm}
\subsection{Orthogonal Updates}
\vspace{-0.15cm}

Recent progress on orthogonalized optimization, exemplified by Muon \citep{jordan2024muon},
advocates exploiting the intrinsic matrix structure of network parameters rather than treating
them as long vectors. The key operation is to transform a momentum (or gradient) matrix
$M_t \in \mathbb{R}^{m \times n}$ into an orthogonalized search direction $O_t$ that preserves
directional information while discarding magnitude effects. To streamline the exposition, we focus on the case $m \ge n$; the remaining case can be handled analogously. Concretely, Muon computes the closest
matrix with orthonormal columns:
{
\setlength{\abovedisplayskip}{2pt}
\setlength{\belowdisplayskip}{-5pt}
\setlength{\abovedisplayshortskip}{2pt}
\setlength{\belowdisplayshortskip}{-5pt}
\begin{equation}
  O_t \;:=\; \arg\min_{O \in \mathbb{R}^{m \times n}:\; O^\top O = I_n}\ \|O - M_t\|_F .
  \label{eq:muon-problem}
\end{equation}
}

By classical matrix theory, if $M_t = U_t \Sigma_t V_t^\top$ is the singular value decomposition (SVD), then the solution is the polar factor $O_t = U_t V_t^\top$. An equivalent characterization
is $O_t \;=\; M_t (M_t^\top M_t)^{-1/2}$,
which highlights that orthogonalization reduces to applying an inverse square root rather than
recovering all singular components.
Computing \cref{eq:muon-problem} exactly via SVD is often impractical for the large matrices
encountered in modern deep networks. 
In practice, Muon instead approximates the polar factor by applying a few Newton--Schulz (NS) iterations to the inverse-square-root term in the polar decomposition \citep{bernstein2024oldoptimizer,higham2008functionsofmatrices,bjorck1971iterative,kovarik1970iterative,bernstein2025deriving}. This yields an efficient and numerically stable approximation that is typically close to the SVD-based solution while being far less computationally demanding, thereby making orthogonalized updates feasible at scale. We adopt this orthogonalization scheme for implementation in our adaptive minimax algorithm.
\vspace{-0.3cm}
\section{Algorithm Development}
\vspace{-0.2cm}
\begin{algorithm}[tb]
   \caption{Adaptive Minimax Optimization with Orthogonal Updates (AdaSGDA)}
   \label{alg1}
\begin{algorithmic}[1]\small
    \STATE {\bfseries Initialization:} $X_{0}$, $Y_{0}$,
    $M_0^{\scriptscriptstyle X}=0$, $M_0^{\scriptscriptstyle Y}=0$,
    buffers $m^{\scriptscriptstyle X}_{0} = m^{\scriptscriptstyle Y}_{0} = c > 0$,
    hyperparameters $\eta_{\scriptscriptstyle X}, \eta_{\scriptscriptstyle Y}, \epsilon_{\scriptscriptstyle X}, \epsilon_{\scriptscriptstyle Y}, \gamma > 0$, and
    momentum parameter $\beta$.
    \FOR{iteration $t = 0, 1,  \cdots, T-1$}
        \STATE Sample mini-batch of size $b_t$ and compute stochastic gradients:
        \\ $G^{\scriptscriptstyle X}_{t} \!=\! \nabla_X F(X_{t}, Y_{t}; \xi^{\scriptscriptstyle X}_{t})$, $G^{\scriptscriptstyle Y}_{t} \!=\! \nabla_Y F(X_{t}, Y_{t}; \xi^{\scriptscriptstyle Y}_{t})$.
        \STATE Momentum updates:\\
        $M^{\scriptscriptstyle X}_{t+1}=\beta M^{\scriptscriptstyle X}_{t}+(1-\beta)G^{\scriptscriptstyle X}_{t}$,\\
        $M^{\scriptscriptstyle Y}_{t+1}=\beta M^{\scriptscriptstyle Y}_{t}+(1-\beta)G^{\scriptscriptstyle Y}_{t}$.
        \STATE Accumulate the gradient norms:\\
        $[m^{\scriptscriptstyle X}_{t+1}]^2 = [m^{\scriptscriptstyle X}_{t}]^2 + \min\{\|G^{\scriptscriptstyle X}_{t}\|_F^2,\gamma^2\}$,\\
        $[m^{\scriptscriptstyle Y}_{t+1}]^2 = [m^{\scriptscriptstyle Y}_{t}]^2 + \min\{\|G^{\scriptscriptstyle Y}_{t}\|_F^2,\gamma^2\}$.
        \STATE Orthogonalize the momentum matrices:\\
        $O_{t+1}^{\scriptscriptstyle X} =\operatorname*{argmin}_{O\in\mathbb{R}^{m_x\times n_x}}\!\|O-M_{t+1}^{\scriptscriptstyle X}\|_F$,\\
        s.t. $O^\top O=I_{n_x}$ if $m_x\ge n_x$, else $OO^\top\!=I_{m_{x}}$,\\
        $O_{t+1}^{\scriptscriptstyle Y} =\operatorname*{argmin}_{O\in\mathbb{R}^{m_y\times n_y}}\!\|O-M_{t+1}^{\scriptscriptstyle Y}\|_F$,\\
        s.t. $O^\top O=I_{n_y}$ if $m_y\ge n_y$, else $OO^\top\!=I_{m_y}$.
        \STATE Define adaptive stepsizes:\\
        $\alpha^{\scriptscriptstyle Y}_{t+1}=\max\!\left\{\epsilon_{\scriptscriptstyle Y},\;\eta_{\scriptscriptstyle Y} \frac{\min\{\|G_t^{\scriptscriptstyle Y}\|_F,\gamma\}}{m_{t+1}^{\scriptscriptstyle Y}}\right\}$,\\
        $\alpha^{\scriptscriptstyle X}_{t+1}=\max\!\left\{\epsilon_{\scriptscriptstyle X},\;\eta_{\scriptscriptstyle X} \frac{\min\{\|G_t^{\scriptscriptstyle X}\|_F,\gamma\}}{\max\{m_{t+1}^{\scriptscriptstyle X},m_{t+1}^{\scriptscriptstyle Y}\}}\right\}$.
        \STATE Update primal and dual variables:\\
        $Y_{t+1} = {Y_{t} + \alpha^{\scriptscriptstyle Y}_{t+1} O^{\scriptscriptstyle Y}_{t+1}}$,\\
        $X_{t+1} = X_{t} - \alpha^{\scriptscriptstyle X}_{t+1} O^{\scriptscriptstyle X}_{t+1}$.
    \ENDFOR
\end{algorithmic}
\end{algorithm}

In this section, we present the design of our adaptive minimax algorithm, which is driven by the need to address two key challenges:
\begin{enumerate}[topsep=0pt, itemsep=2pt, parsep=0pt, leftmargin=0.25cm]
\item \textbf{Time-scale adaptivity in single-loop minimax.}
Unlike minimization, minimax dynamics require properly separated primal--dual time scales. As shown by \citet{yang2022nest}, directly applying standard adaptive stepsizes to GDA---where each variable adapts only to the (moving) statistics of its own past gradients---can fail to converge unless the initial stepsize ratio between $x$ and $y$ is carefully calibrated. This implies that we must precisely control the relative progress of $x$ and $y$ to maintain the required time-scale separation and ensure stable convergence.

\item \textbf{Adaptivity under orthogonalized directions.}
Orthogonal updates can substantially reshape the optimization dynamics \citep{jordan2024muon}. The main difficulty is that orthogonalization is a nonlinear transformation of the gradient: two gradients that are close in Frobenius norm can lead to orthogonalized directions that differ much more, especially when the underlying gradient matrices are ill-conditioned \citep{zhang2025sharp,liu2025muon}. Consequently, stochastic noise can have a disproportionately large impact on the update direction, and simply accumulating raw gradient norms (e.g., AdaGrad-Norm \citep{xie2020linear}) may not provide a reliable scale for stepsize adaptation in this setting.
\end{enumerate}

We address these challenges by proposing \textbf{AdaSGDA}, a single-loop method with explicit time-scale adaptivity and orthogonalization-calibrated adaptive stepsizes. To further improve stochastic efficiency, we develop \textbf{AdaSGDA-VR} by incorporating variance reduction. We prove that both algorithms converge to an $\epsilon$-stationary point of problem~\eqref{P1} with state-of-the-art rates.

\vspace{-0.25cm}
\subsection{Algorithm Development of AdaSGDA}
\vspace{-0.15cm}
We first present the development of our adaptive minimax algorithm with orthogonal updates, \textbf{AdaSGDA}; the detailed procedure is outlined in \cref{alg1}. AdaSGDA is built on three key mechanisms: orthogonalized momentum, a hierarchical stepsize design, and orthogonalization-calibrated adaptive stepsizes. We elaborate on each mechanism below.

\textbf{Orthogonalized Momentum.}
The momentum is incorporated to smooth stochastic fluctuations and to stabilize the optimization trajectory. At iteration $t$, we maintain momentum matrices for the primal and dual updates:
{
\setlength{\abovedisplayskip}{3pt}
\setlength{\belowdisplayskip}{-3pt}
\setlength{\abovedisplayshortskip}{3pt}
\setlength{\belowdisplayshortskip}{-3pt}
\begin{align}
M^{\scriptscriptstyle X}_{t+1} = \beta M^{\scriptscriptstyle X}_t + (1-\beta)\nabla_X F(X_t, Y_t; \xi_t^{\scriptscriptstyle X}),\quad
M^{\scriptscriptstyle Y}_{t+1} = \beta M^{\scriptscriptstyle Y}_t + (1-\beta)\nabla_Y F(X_t, Y_t; \xi_t^{\scriptscriptstyle Y}), \notag
\end{align}}

where $\beta \!\in\! (0,1)$ is the momentum coefficient, $\nabla_X F(\!X_t, Y_t; \xi_t^{\scriptscriptstyle X}\!)$ and
$\nabla_Y F(\!X_t, Y_t; \xi_t^{\scriptscriptstyle Y}\!)$ are stochastic gradient estimators evaluated at the current iterate
$(\!X_t,Y_t\!)$ (based on randomness $\xi_t^{\scriptscriptstyle X}$ and $\xi_t^{\scriptscriptstyle Y}$), and $M^{\scriptscriptstyle X}_t, M^{\scriptscriptstyle Y}_t$
store the corresponding momentum matrices from the previous iteration.

Different from standard momentum methods that update parameters directly along $M^{\scriptscriptstyle X}_{t+1}$ and
$M^{\scriptscriptstyle Y}_{t+1}$, we subsequently orthogonalize these momentum matrices and use the orthogonalized directions for updating $X$ and $Y$. In this way, the update
direction leverages the accumulated matrix-structured signal in the momentum while being less
sensitive to its magnitude, which leads to more stable orthogonal updates in our minimax framework.

\textbf{Hierarchical Stepsize Design.}
A key objective of AdaSGDA is to avoid stepsize choices that depend on unknown problem constants. To this end, we use AdaGrad-type accumulators that track the magnitude of past stochastic gradients. In addition, we introduce a clamping constant $\gamma>0$ and accumulate $\min\{\|\nabla_X F(X_t,Y_t;\xi_t^{\scriptscriptstyle X})\|_F^2,\gamma^2\}$ and $\min\{\|\nabla_Y F(X_t,Y_t;\xi_t^{\scriptscriptstyle Y})\|_F^2,\gamma^2\}$ instead of the raw squared gradient norms. This cap limits the contribution of each iteration to at most $\gamma^2$, so the accumulators $m_t^{\scriptscriptstyle X}$ and $m_t^{\scriptscriptstyle Y}$ grow in a more controlled way. Practically, clamping acts as a safeguard against occasional large stochastic gradients, preventing the adaptive denominators from being dominated by rare spikes and thereby stabilizing the resulting stepsizes and the primal–dual time-scale coordination.

Concretely, we define the accumulators for the dual and primal variables as
{
\setlength{\abovedisplayskip}{3pt}
\setlength{\belowdisplayskip}{-3pt}
\setlength{\abovedisplayshortskip}{3pt}
\setlength{\belowdisplayshortskip}{-3pt}
\begin{align}
[m^{\scriptscriptstyle Y}_{t+1}]^2 \!\!=\!\! [m^{\scriptscriptstyle Y}_t]^2 \!\!+\!\! \min\!\left\{\!\big\|\!\nabla_Y F(X_t, Y_t; \xi_t^{\scriptscriptstyle Y})\!\big\|_F^2,\gamma^2\!\right\}\!, 
[m^{\scriptscriptstyle X}_{t+1}]^2 \!\!=\!\! [m^{\scriptscriptstyle X}_t]^2 \!\!+\!\! \min\!\left\{\!\big\|\!\nabla_X F(X_t, Y_t; \xi_t^{\scriptscriptstyle X})\!\big\|_F^2,\gamma^2\!\right\}\!. \notag
\end{align}}

Based on these accumulators, we first set the adaptive scale for the dual variable as
$\frac{\eta^{}_{\scriptscriptstyle Y}}{m^{\scriptscriptstyle Y}_{t+1}}$,
where $\eta_{\scriptscriptstyle Y}>0$ is a control coefficient independent of problem parameters.

For the primal variable $X$, using the same form $\frac{\eta^{}_{\scriptscriptstyle X}}{m^{\scriptscriptstyle X}_{t+1}}$ is generally insufficient in minimax
optimization, because the primal update is sensitive to how accurately $Y_t$ tracks the current
maximizer $Y^\ast(X_t)$. If $Y_t$ is still far from $Y^\ast(X_t)$, an overly aggressive primal step can destabilize the coupled dynamics. To account for this
interdependence, we adopt a hierarchical scale for $X$: $\frac{\eta^{}_{\scriptscriptstyle X}}{\max\{m^{\scriptscriptstyle X}_{t+1},\,m^{\scriptscriptstyle Y}_{t+1}\}}$,
where $\eta_{\scriptscriptstyle X}>0$ is also problem-parameter-free. This design automatically makes the primal step more conservative when the dual variable is still changing rapidly.
It enforces the desired time-scale separation within a single loop and promotes stable convergence, without requiring any problem-parameter-based tuning.

\textbf{Orthogonalization-Calibrated Adaptive Stepsizes.}
The hierarchical scales specify how the stepsizes should
evolve over time to coordinate the primal--dual progress. We now turn to the remaining question:
how should these scales be applied when the update {directions} are orthogonalized.

In AdaSGDA, we update $X$ and $Y$ along orthogonalized directions computed from the momentum
matrices. 
Concretely, we compute the orthogonalized momentum directions $O_{t+1}^{\scriptscriptstyle X}$ and $O_{t+1}^{\scriptscriptstyle Y}$ to update $X$ and $Y$, respectively, as
{
\setlength{\abovedisplayskip}{6pt}
\setlength{\belowdisplayskip}{-3pt}
\setlength{\abovedisplayshortskip}{6pt}
\setlength{\belowdisplayshortskip}{-3pt}
\begin{align*}
&O_{t}^{\scriptscriptstyle X} \!=\!\!\operatorname*{argmin}_{O\in\mathbb{R}^{m_x\times n_x}}\!\! \{\|O\!-\!M_{t}^{\scriptscriptstyle X}\|_F\!: O O^\top\!\!=\!I_{m_x}\!\!\!\!\!\quad \text{or}\quad\!\!\!\!  O^\top \!O\!=\!I_{n_x}\}, \notag\\
&O_{t}^{\scriptscriptstyle Y} \!=\!\!\operatorname*{argmin}_{O\in\mathbb{R}^{m_y\times n_y}}\!\! \{\|O\!-\!M_{t}^{\scriptscriptstyle Y}\|_F\!: O O^\top\!\!=\!I_{m_y}\!\!\!\!\quad \text{or}\quad\!\!\!\!  O^\top \!O\!=\!I_{n_y}\},\notag
\end{align*}}

where the orthogonality constraint is chosen according to the matrix shape: for $O_t^{\scriptscriptstyle X}\in\mathbb{R}^{m_x\times n_x}$ we impose $(O_t^{\scriptscriptstyle X})^\top O_t^{\scriptscriptstyle X}=I_{n_x}$ when $m_x\ge n_x$, and $O_t^{\scriptscriptstyle X}(O_t^{\scriptscriptstyle X})^\top=I_{m_x}$ otherwise; the same convention applies to $O_t^{\scriptscriptstyle Y}\in\mathbb{R}^{m_y\times n_y}$.
Moreover, since the orthogonalized directions satisfy $\|O_{t+1}^{\scriptscriptstyle X}\|_{\mathrm{op}}=\|O_{t+1}^{\scriptscriptstyle Y}\|_{\mathrm{op}}=1$, we calibrate the effective stepsizes using the current stochastic gradient norms by scaling them with the clamped factors $\min\{\|\nabla_X F(X_t,Y_t;\xi_t^{\scriptscriptstyle X})\|_F,\gamma\}$ and
$\min\{\|\nabla_Y F(X_t,Y_t;\xi_t^{\scriptscriptstyle Y})\|_F,\gamma\}$.
This calibration makes the update magnitude shrink
as the gradients shrink.
In particular, near a stationary point where the gradients are small, the per-iteration updates
naturally decay to zero, which is a desirable null-gradient consistency property in first-order
optimization \citep{lau2025polargrad}.

To avoid overly small steps caused by growing accumulators, we also impose lower bounds
$\epsilon_{\scriptscriptstyle X},\epsilon_{\scriptscriptstyle Y}>0$ on the effective stepsizes. This ensures nontrivial progress and guarantees
that AdaSGDA is at least as aggressive as using orthogonal updates with a small constant stepsize.
Importantly, $\epsilon_{\scriptscriptstyle X}$ and $\epsilon_{\scriptscriptstyle Y}$ can be chosen without any problem-dependent constants.

Putting these components together, we define the calibrated stepsizes as
{
\setlength{\abovedisplayskip}{3pt}
\setlength{\belowdisplayskip}{-3pt}
\setlength{\abovedisplayshortskip}{3pt}
\setlength{\belowdisplayshortskip}{-3pt}
\begin{align}
\alpha^{\scriptscriptstyle X}_{t+1}\!\!
=\!\!\max\!\!\left\{\!\epsilon_{\scriptscriptstyle X}\!,\eta_{\scriptscriptstyle X}\! \frac{\min\{\!\|\nabla_{\!X} \!F(X_t,Y_t;\xi_t^{\scriptscriptstyle X})\|_F\!,\!\gamma\}}{\max\{m_{t+1}^{\scriptscriptstyle X},m_{t+1}^{\scriptscriptstyle Y}\}}\!\!\right\}\!,
\alpha^{\scriptscriptstyle Y}_{t+1}\!\!
=\!\!\max\!\!\left\{\!\epsilon_{\scriptscriptstyle Y}\!,\eta_{\scriptscriptstyle Y}\! \frac{\min\{\!\|\nabla_{\!Y}\! F(X_t,Y_t;\xi_t^{\scriptscriptstyle Y})\|_F\!,\gamma\}}{m_{t+1}^{\scriptscriptstyle Y}}\!\!\right\}\!.
\notag
\end{align}}

The resulting updates are
$Y_{t+1} = Y_t + \alpha^{\scriptscriptstyle Y}_{t+1} O^{\scriptscriptstyle Y}_{t+1}$ and $X_{t+1} = X_t - \alpha^{\scriptscriptstyle X}_{t+1} O^{\scriptscriptstyle X}_{t+1}$.

Overall, this orthogonalization-calibrated design couples the hierarchical scaling with the orthogonal geometry,
leading to stable single-loop minimax updates without problem-specific tuning.

\vspace{-0.25cm}
\subsection{Algorithm Development of AdaSGDA-VR}
\vspace{-0.15cm}
To further accelerate convergence under stochastic gradients, we integrate variance reduction into
AdaSGDA and propose \textbf{AdaSGDA-VR} (\cref{alg2} in \cref{VRproof}). The key idea is to construct a
momentum-based variance-reduced gradient surrogate that reuses information from the previous
iterate, thereby reducing the variance of stochastic updates.
Our design is inspired by the variance-reduction principles behind SPIDER \citep{fang2018spider},
STORM \citep{cutkosky2019momentum}, and SVRG-type updates \citep{zhou2020stochastic}.

Specifically, AdaSGDA-VR augments the momentum updates with a variance-reduction correction computed from the
same mini-batch evaluated at the current and previous iterates. For $t\ge 1$, the momentum matrices for the primal and dual variables are updated as
{
\setlength{\abovedisplayskip}{5pt}
\setlength{\belowdisplayskip}{-2pt}
\setlength{\abovedisplayshortskip}{5pt}
\setlength{\belowdisplayshortskip}{-2pt}
\begin{align}
M^{\scriptscriptstyle X}_{t+1}
&=\beta M^{\scriptscriptstyle X}_t + (1-\beta)\nabla\!_X\! F(X_t, Y_t; \xi_t^{\scriptscriptstyle X})+ \beta\left(\nabla\!_X\! F(X_t, Y_t; \xi_t^{\scriptscriptstyle X})-\nabla\!_X\! F(X_{t\!-\!1}, Y_{t\!-\!1}; \xi_t^{\scriptscriptstyle X})\right), \notag\\
M^{\scriptscriptstyle Y}_{t+1}
&=\beta M^{\scriptscriptstyle Y}_t + (1-\beta)\nabla\!_Y\! F(X_t, Y_t; \xi_t^{\scriptscriptstyle Y})+ \beta\left(\nabla\!_Y\! F(X_t, Y_t; \xi_t^{\scriptscriptstyle Y})-\nabla\!_Y\! F(X_{t\!-\!1}, Y_{t\!-\!1}; \xi_t^{\scriptscriptstyle Y})\right). \notag
\end{align}}

This structured correction is designed to cancel a large portion of the noise introduced by mini-batch
sampling, thereby yielding a lower-variance update direction for subsequent orthogonalization and
updates \citep{cutkosky2019momentum}.
After forming the variance-reduced momentum matrices, AdaSGDA-VR applies the same
orthogonalization procedure and calibrated stepsize rules as AdaSGDA. This results in a fully
single-loop algorithm that remains parameter-free and time-scale adaptive, while achieving faster
convergence by reducing stochastic errors.

\vspace{-0.3cm}
\section{Theoretical Analysis}
\vspace{-0.2cm}
\subsection{Technical challenges}
\vspace{-0.15cm}
Establishing optimal convergence guarantees for {single-loop, parameter-free} minimax optimization
with orthogonal updates poses several technical difficulties:
\begin{itemize}[topsep=0pt, itemsep=2pt, parsep=0pt, leftmargin=0.25cm]
    \item \textbf{Coupled primal--dual dynamics.}
    The minimax scheme tightly couples the evolution of $(X_t,Y_t)$: the primal progress depends on how well the dual variable tracks $Y\!^\star\!(X_t)$, while the dual update is driven by a moving target that changes with $X_t$. This dependence makes the analysis more delicate than in single-variable minimization.

    \item \textbf{Adaptive time-scale control under orthogonal geometry.}
In a single-loop scheme, the stepsizes for $X$ and $Y$ cannot be analyzed separately: their
history-dependent denominators must jointly enforce the required time-scale separation. Moreover,
updates are applied along unit-magnitude orthogonalized directions rather than raw gradients, which
changes the effective geometry and requires careful stepsize calibration. Establishing convergence in
this coupled, stochastic two-time-scale setting calls for a refined analysis beyond prior minimax work.

\item \textbf{Parameter-free stepsizes with optimal rates.}
  Achieving the best-known minimax rate typically relies on stepsizes tuned to problem-specific constants. In the parameter-free regime, the analysis must show that our adaptive stepsizes still realize the correct balance between descent in the primal objective and sufficient ascent in the dual variable, so that the overall error terms telescope to the optimal $\mathcal{O}(T^{-1/4})$ rate (and $\mathcal{O}(T^{-1/3})$ with variance reduction).
\end{itemize}

\vspace{-0.28cm}
\subsection{Assumptions}
\vspace{-0.15cm}
We consider the NC-SC setting of Problem \eqref{P1} with the following assumptions that are commonly used in the existing works \citep{yang2022faster,yang2022nest,li2022tiada,huang2024achieving, lin2025two, he2025low}.
\begin{assumption}[Joint smoothness]
The stochastic objective function $F(X,Y;\xi)$ is jointly
$L$-smooth ($L>0$) in $(X,Y)$. That is, for any
$X_1,X_2 \in \mathbb{R}^{m_x\times n_x}$ and
$Y_1,Y_2 \in \mathbb{R}^{m_y\times n_y}$, we have
{
\setlength{\abovedisplayskip}{1pt}
\setlength{\belowdisplayskip}{1pt}
\setlength{\abovedisplayshortskip}{1pt}
\setlength{\belowdisplayshortskip}{1pt}
\begin{align}
&\mathbb{E}_{\xi}\![
  \left\|
    \nabla_X F(X_1,Y_1;\xi)
    - \nabla_X F(X_2,Y_2;\xi)
  \right\|_*^2
]\le
L^2\left(
  \|X_1-X_2\|_{\mathrm{op}}^2
  + \|Y_1-Y_2\|_{\mathrm{op}}^2
\right),
\notag\\
&\mathbb{E}_{\xi}\![
  \left\|
    \nabla_Y F(X_1,Y_1;\xi)
    - \nabla_Y F(X_2,Y_2;\xi)
  \right\|_*^2
]\le
L^2\left(
  \|X_1-X_2\|_{\mathrm{op}}^2
  + \|Y_1-Y_2\|_{\mathrm{op}}^2
\right)\notag.
\end{align}}
\label{smoothness}
\vspace{-0.433cm}
\end{assumption}

\begin{assumption}[$\mu$-Strong concavity in $Y$]
Function $f(X,Y)$ is $\mu$-strongly concave ($\mu>0$) in $Y$, that is, for any
$X \in  \mathbb{R}^{m_x\times n_x}$ and $Y_1,Y_2 \in \mathbb{R}^{m_y\times n_y}$, we have
{
\setlength{\abovedisplayskip}{0pt}
\setlength{\belowdisplayskip}{3pt}
\setlength{\abovedisplayshortskip}{0pt}
\setlength{\belowdisplayshortskip}{3pt}
\begin{align}
f(X,Y_1)\ge
f(X,Y_2)
+
\langle \nabla_Y f(X,Y_1), Y_1 - Y_2 \rangle+
\frac{\mu}{2}\|Y_1 - Y_2\|_F^2\notag.
\end{align}}
\vspace{-0.41cm}
\label{concavity}
\end{assumption}

\begin{assumption}[Stochastic gradients]
$\nabla_X F(X_t,Y_t;\xi^{\scriptscriptstyle X}_t)$ and $\nabla_Y F(X_t,Y_t;\xi^{\scriptscriptstyle Y}_t)$ are unbiased stochastic gradient estimators such that
$\mathbb{E}[\nabla_X F(X_t,Y_t;\xi^{\scriptscriptstyle X}_t)] = \nabla_X f(X_t,Y_t)$ and $\mathbb{E}[\nabla_Y F(X_t,Y_t;\xi^{\scriptscriptstyle Y}_t)] = \nabla_Y f(X_t,Y_t)$,
with a uniformly bounded variance:
{
\setlength{\abovedisplayskip}{-1pt}
\setlength{\belowdisplayskip}{-5pt}
\setlength{\abovedisplayshortskip}{-1pt}
\setlength{\belowdisplayshortskip}{-5pt}
\begin{align}
 \mathbb{E} \left[\|\nabla_X F(X_t,Y_t;\xi^{\scriptscriptstyle X}_t)-\nabla_X f(X_t,Y_t)\|_F^2\right]\leq\frac{\kappa^2}{b_t}\!,
 \mathbb{E} \left[\|\nabla_Y F(X_t,Y_t;\xi^{\scriptscriptstyle Y}_t)-\nabla_Y f(X_t,Y_t)\|_F^2\right]\leq\frac{\kappa^2}{b_t}\!,\notag
\end{align}}

where $b_t\geq 1$ is the batch size. Moreover, there exists $G>0$ such that
$\|\nabla_Z F(X_s,Y_s;\xi_t^{\scriptscriptstyle Z})\|_F\leq G$
almost surely for $Z\in\{X,Y\}$ and every stochastic oracle
evaluation made by \cref{alg1,alg2} along their generated
trajectories.
\label{stochasticgradient}
\end{assumption}
\begin{remark}
By the first-order optimality condition for the inner maximization defining $Y^\star(X)$, we have $\nabla_Y f(X, Y^\star(X))=0$. This property is important for AdaGrad-type stepsizes whose denominators accumulate past gradient norms. If the dual gradient does not vanish at $Y^\star(X)$, the accumulator may keep growing even near optimality, forcing the stepsize to keep shrinking and slowing convergence \citep{bach2019universal,li2022tiada}.
In addition, bounded gradients (\cref{stochasticgradient}) are a standard assumption for controlling adaptive denominators in AdaGrad- and Adam-type analyses \citep{levy2018online,zou2019sufficient,ivgi2023dog}, and have also been commonly adopted in minimax optimization \citep{li2022tiada,huang2024achieving}.
\end{remark}

\vspace{-0.28cm}
\subsection{Theoretical Results}
\vspace{-0.15cm}
In this subsection, we present the main theoretical results for the proposed \cref{alg1} and \cref{alg2}. As outlined in \cref{descentlemma}, the convergence behavior is governed by the following key terms: (i)~~$\frac{1}{T}\sum_{t=1}^T \mathbb{E}\big[\|M_t^{\scriptscriptstyle X}-\nabla\Phi(X_{t-1})\|_*\big]$, (ii) $\frac{\eta^{}_{\scriptscriptstyle X}\sqrt{r}}{\epsilon^{}_{\scriptscriptstyle X} T}\sqrt{\ln\!\Big(\frac{\gamma^2T}{z_0^2}\Big)+1}
\sqrt{\sum_{t=1}^T \mathbb{E}\big[\|M_t^{\scriptscriptstyle X}-\nabla\Phi(X_{t-1})\|_F^2\big]}$,
where $z_0=\max\{m_0^{\scriptscriptstyle X},m_0^{\scriptscriptstyle Y}\}$. Term (i) is the average gradient estimation error measured in the nuclear norm $\|\cdot\|_*$, while term (ii) depends on the cumulative squared estimation error measured in the Frobenius norm $\|\cdot\|_F$.

\smallskip
\noindent\textbf{\cref{alg1}.} Under the hyperparameter choices in \cref{theorem1}, our analysis shows that term (i) is upper bounded by $\mathcal{O}\!\left(T^{-\frac{1}{4}}\right)$ and term (ii) is upper bounded by $\mathcal{O}\!\left(T^{-\left(\frac{3}{8}+q\right)}\right)$. Plugging these bounds into \cref{descentlemma} yields \cref{theorem1}.

\begin{theorem}
Under \cref{smoothness,concavity,stochasticgradient}, when we set $1\!-\!\beta\!=\!\frac{1}{T^{\frac{1}{4}}}$, $\epsilon_{\scriptscriptstyle X}\!=\!\frac{1}{T^{\frac{3}{4}}}$, $\epsilon_{\scriptscriptstyle Y}\!=\!\frac{1}{T^{\frac{1}{2}}}$, $\eta_{\scriptscriptstyle X}\!=\!\frac{1}{T^{\frac{7}{8}+q}}$, $\eta_{\scriptscriptstyle Y}\!=\!\frac{1}{T^{\frac{5}{8}+q}}$, where $q\!\ge\!0$ is a fixed exponent offset used in the theoretical schedules, together with $\gamma\!>\!0$, $z_0\!>\!0$, and $b_t\!\ge\! T^{1/4}$ for all $t\!\in\!\{1,\ldots,T\}$, the iterates generated by \cref{alg1} satisfy:
{
\setlength{\abovedisplayskip}{0pt}
\setlength{\belowdisplayskip}{-6pt}
\setlength{\abovedisplayshortskip}{0pt}
\setlength{\belowdisplayshortskip}{-6pt}
\begin{align*}
\frac{1}{T}\!\!\sum_{t=1}^T \!\mathbb{E}[\|\!\nabla\Phi(X_{t})\!\|_*\!]
\!\leq\!\mathcal{O}\!\!\left(\!\!\frac{\Delta\!\!+\!\!\kappa\!\sqrt{r}}{T^\frac{1}{4}}\!\!+\!\!\frac{GL\!\sqrt r(1\!\!+\!\!\lambda)\!\!+\!\!\kappa L\!\sqrt{r}\!\!+\!\!L^2\!{r}}{\mu T^\frac{1}{4}}\!\right)\!\!+\!\!\mathcal{O}\!\!\left(\!\!\frac{Lr^\frac{3}{4}\!\sqrt{\kappa G}\!\!+\!\!L^\frac{3}{2}\!r\!\sqrt{G}}{\mu T^{\frac{3}{8}+q}}\!\!\sqrt{\ln\!\!\left(\!\!\frac{\gamma^2T}{z_0^2}\!\right)\!\!+\!\!1}\right)\!,
\end{align*}}\label{theorem1}

where $\Delta\!:=\!\Phi(X_0)- \Phi_*$, $\Phi_* \!:=\! \min_X \Phi(X)$, and $\lambda$ denotes the Lipschitz constant of $Y^\ast(\cdot)$ as defined in \cref{additionallemma}.

where $\Delta:=\Phi(X_0)-\Phi_*$, $\Phi_*:=\min_X \Phi(X)$, $\lambda$ denotes the Lipschitz constant of $Y^\ast(\cdot)$ as defined in \cref{additionallemma}, and $q>0$ is any fixed arbitrarily small constant used in the stepsize exponents; it only affects the higher-order term and does not change the leading $\mathcal{O}(T^{-1/4})$ rate.
\end{theorem}

\smallskip
\noindent\textbf{\cref{alg2}.} Under the hyperparameter choices in \cref{theoremVR}, we obtain that term (i) is upper bounded by $\mathcal{O}\!\left(T^{-\frac{1}{3}}\right)$ and term (ii) is upper bounded by $\mathcal{O}\!\left(T^{-\left(\frac{1}{2}+q\right)}\right)$. Substituting them into \cref{descentlemma} establishes \cref{theoremVR}. 

\begin{theorem}
Under \cref{smoothness,concavity,stochasticgradient}, when we set $1\!-\!\beta\!=\!\frac{1}{T^{\frac{1}{3}}}$, $\epsilon_{\scriptscriptstyle X}\!=\!\frac{1}{T^{\frac{2}{3}}}$, $\epsilon_{\scriptscriptstyle Y}\!=\!\frac{1}{T^{\frac{1}{3}}}$, $\eta_{\scriptscriptstyle X}\!=\!\frac{1}{T^{\frac{2}{3}+q}}$, $\eta_{\scriptscriptstyle Y}\!=\!\frac{1}{T^{\frac{1}{2}+q}}$, $\gamma\!>\!0$, $z_0\!>\!0$, $b_t\!\geq\! T^\frac{1}{3}$ for all $t\!\in\!\{1,\cdots,T\}$, the iterates generated by \cref{alg2} satisfy:
{
\setlength{\abovedisplayskip}{3pt}
\setlength{\belowdisplayskip}{3pt}
\setlength{\abovedisplayshortskip}{3pt}
\setlength{\belowdisplayshortskip}{3pt}
\begin{align*}
\frac{1}{T}\!\!\sum_{t=1}^T \!\mathbb{E}[\|\!\nabla\Phi(X_{t})\!\|_*\!]
\!\leq\!\mathcal{O}\!\!\left(\!\!\frac{\Delta\!\!+\!\!rL\!\!+\!\!\kappa\sqrt r}{T^\frac{1}{3}} \!\!+\!\!\frac{GL\sqrt r(1\!\!+\!\!\lambda)\!\!+\!\!\kappa L\sqrt{r}\!\!+\!\!L^2\!r}{{\mu}T^\frac{1}{3}}\!\right)\!\!+\!\!\mathcal{O}\!\!\left(\!\!\frac{{Lr}}{T^{\frac{1}{2}+q}}\!\!\sqrt{\!\ln\!\!\left(\!\!\frac{\gamma^2T}{z_0^2}\!\!\right)\!\!+\!\!1}\right)\!.
\end{align*}}\label{theoremVR}
\end{theorem}
\vspace{-0.3cm}
\begin{remark}
Our theoretical analysis establishes optimal convergence guarantees for adaptive minimax optimization with orthogonal updates. Existing adaptive minimax methods only provide near-optimal rates \citep{yang2022nest,li2022tiada,huang2024achieving}, and attaining optimal rates often requires stepsize tuning based on problem-specific constants such as smoothness or strong concavity parameters (e.g., \citealp{jin2019minmax,lin2025two}). In sharp contrast, our orthogonal adaptive design is inherently parameter-free yet attains the optimal rate in \cref{theorem1}. In addition, our carefully designed variance reduction scheme, together with a refined control of stochastic errors in the theoretical analysis, leads to the faster guarantee in \cref{theoremVR}, matching the best-known rate for NC-SC minimax optimization.
\end{remark}

\vspace{-0.35cm}
\section{Numerical Experiments}
\vspace{-0.2cm}

In this section, we evaluate \textbf{AdaSGDA} and \textbf{AdaSGDA-VR} against representative minimax baselines, including SGDA \citep{lin2025two}, NeAda \citep{yang2022nest}, D-AdaST \citep{huang2024achieving}, AdamW \citep{loshchilov2017decoupled}, and Muon-based minimax method \citep{Muon2024}. Our experiments cover two representative minimax tasks: (i) NC--SC distributional robustness optimization (DRO) \citep{sinha2017certifiable}, and (ii) training the nonconvex--nonconcave (NC--NC) Wasserstein GAN with gradient penalty \citep{gulrajani2017improved}. Experiments are conducted on MNIST \citep{lecun2002gradient}, Fashion-MNIST (FMNIST) \citep{xiao2017fashion}, CIFAR-10 and CIFAR-100 \citep{krizhevsky2009learning}, and on FineWeb \citep{penedo2024fineweb} using OPT-125M, OPT-350M, and OPT-1.3B. Additional results and configurations are provided in \cref{experimentsec}.

\vspace{-0.25cm}
\subsection{Distributional Robustness Optimization}
\vspace{-0.2cm}
In this subsection, we consider DRO following the adversarial-perturbation formulation in \citet{sinha2017certifiable}. The goal is to learn a model that is robust to worst-case perturbations of the training inputs. We treat the model parameters as the primal variable and the perturbations as the dual variable.
Let $\{\xi_i\}_{i=1}^n$ be the training samples, and let $V_i$ denote the corresponding input representation (e.g., a raw image for vision tasks, or an embedding sequence for language modeling). For each sample, we introduce a perturbed input $Y_i$ and solve
$\min_{X}\ \max_{Y=[Y_1,\ldots,Y_n]}\ \frac{1}{n}\sum_{i=1}^n \ell(X; Y_i)\;-\;\rho \|Y_i - V_i\|_F^2$,
where $\ell(X; Y_i)$ is the task loss evaluated at the perturbed input, and $\rho>0$ controls the perturbation strength by penalizing deviations from $V_i$. The inner maximization searches for worst-case perturbations within a local neighborhood, while the outer minimization learns parameters that perform well under these worst-case inputs. When $\rho$ is sufficiently large relative to the smoothness of $\ell(X;\cdot)$ in $Y_i$, the inner objective becomes strongly concave in $Y$, yielding an NC--SC minimax instance.
We evaluate this DRO setup on MNIST, FMNIST, CIFAR-10/100, and FineWeb datasets.

\begin{figure*}[t]
\centering
\begin{minipage}[t]{0.496\textwidth}
    \centering
    \captionsetup{type=figure}
    \setcounter{subfigure}{0}
    \begin{subfigure}[t]{0.493\linewidth}
        \centering
        \includegraphics[width=\linewidth]{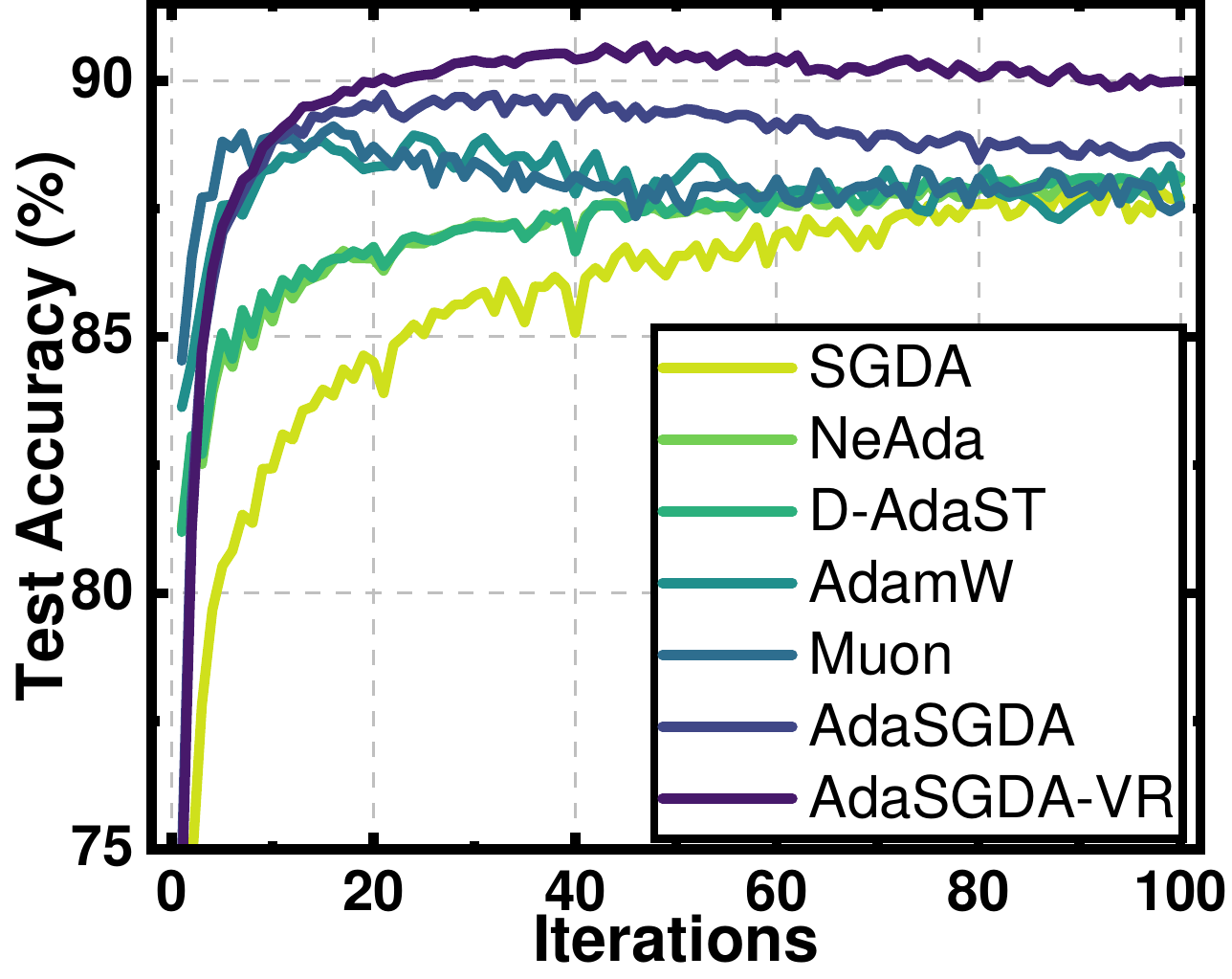}
        \vspace{-0.65cm}
        \caption{Test accuracy}
        \label{fmnisttest}
    \end{subfigure}
    \hfill
    \begin{subfigure}[t]{0.493\linewidth}
        \centering
        \includegraphics[width=\linewidth]{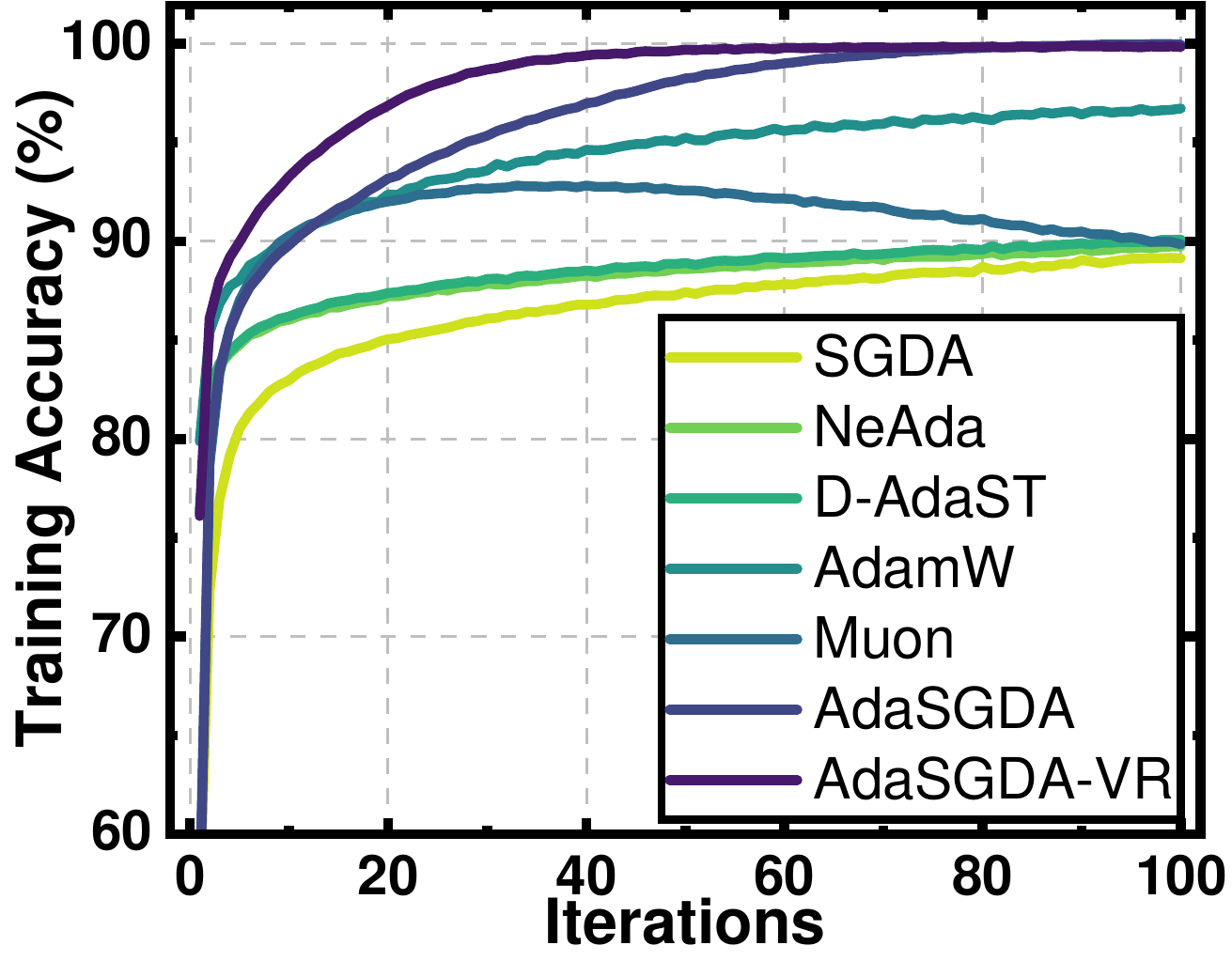}
        \vspace{-0.65cm}
        \caption{Training accuracy}
        \label{fmnisttrain}
    \end{subfigure}
    \vspace{-0.32cm}
    \caption{Performance comparison on FMNIST.}
    \label{fmnisttt}
\end{minipage}
\hfill
\begin{minipage}[t]{0.496\textwidth}
    \centering
    \captionsetup{type=figure}
    \setcounter{subfigure}{0}
    \begin{subfigure}[t]{0.493\linewidth}
        \centering
        \includegraphics[width=\linewidth]{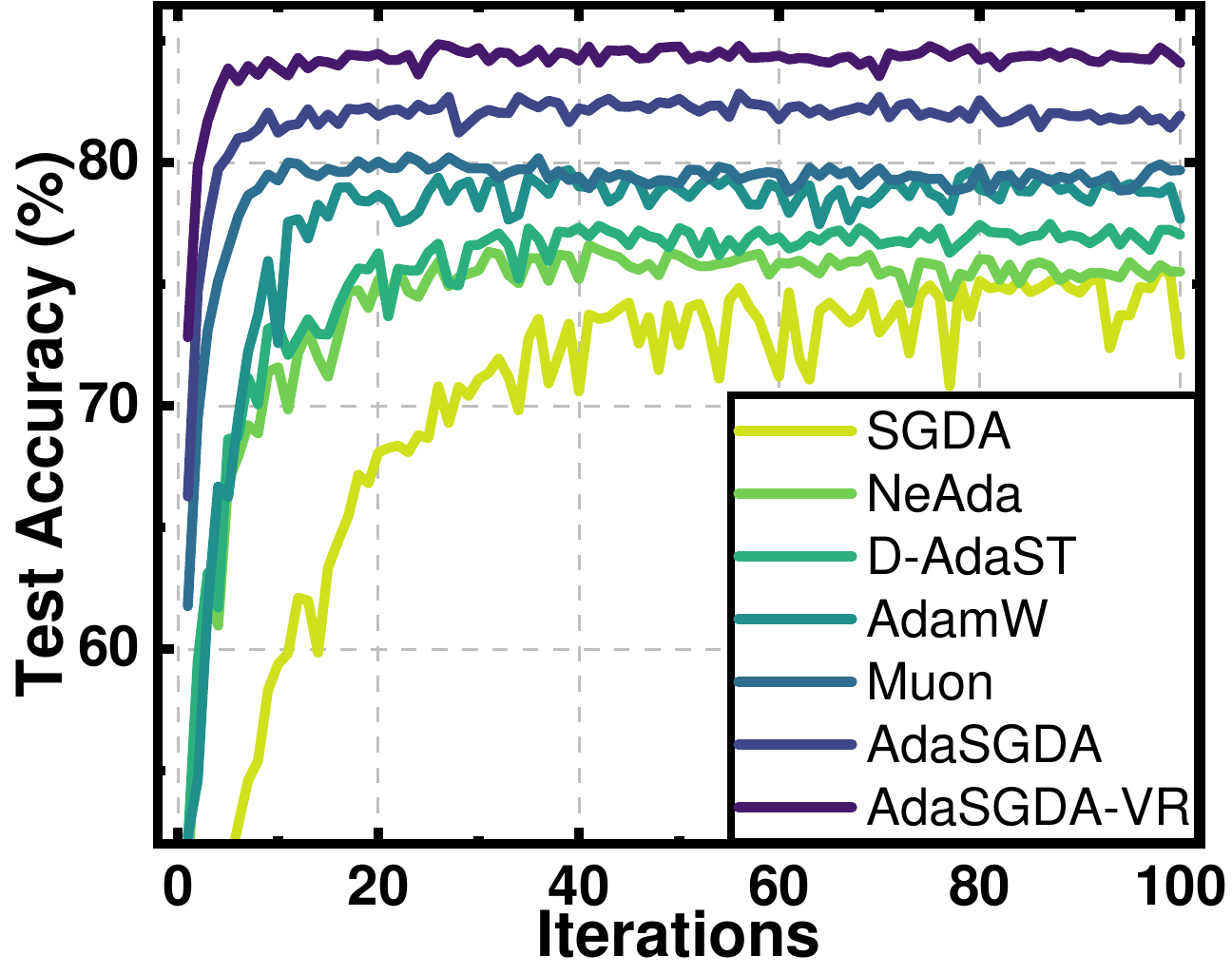}
        \vspace{-0.65cm}
        \caption{Test accuracy}
        \label{cifar10test}
    \end{subfigure}
    \hfill
    \begin{subfigure}[t]{0.493\linewidth}
        \centering
        \includegraphics[width=\linewidth]{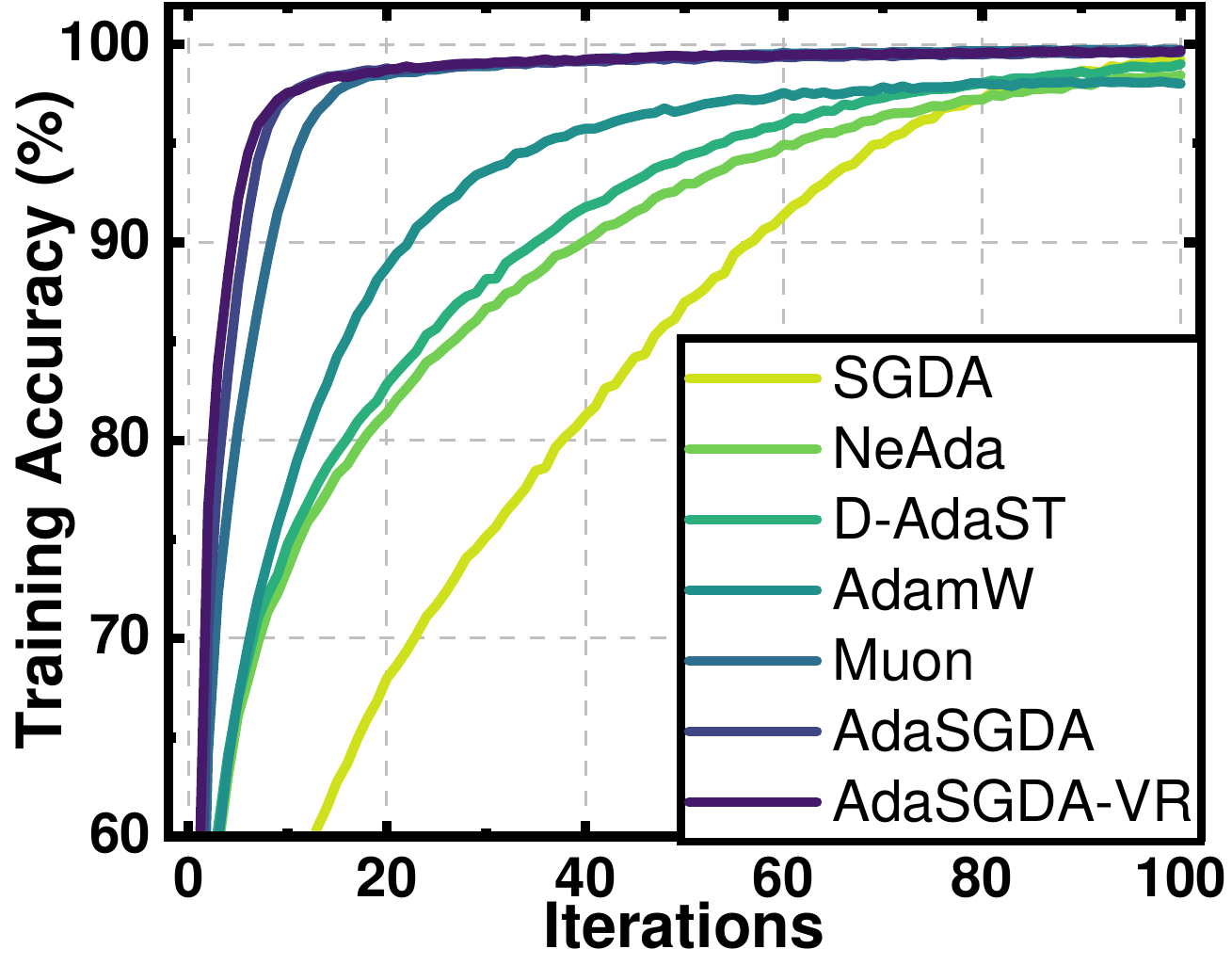}
        \vspace{-0.65cm}
        \caption{Training accuracy}
        \label{cifar10train}
    \end{subfigure}
    \vspace{-0.32cm}
    \caption{Performance comparison on CIFAR-10.}
    \label{cifar10tt}
\end{minipage}
\vspace{-0.2cm}
\end{figure*}

\begin{figure*}[t]
\centering
\begin{minipage}[t]{0.496\textwidth}
    \centering
    \captionsetup{type=figure}
    \setcounter{subfigure}{0}
    \begin{subfigure}[t]{0.493\linewidth}
        \centering
        \includegraphics[width=\linewidth]{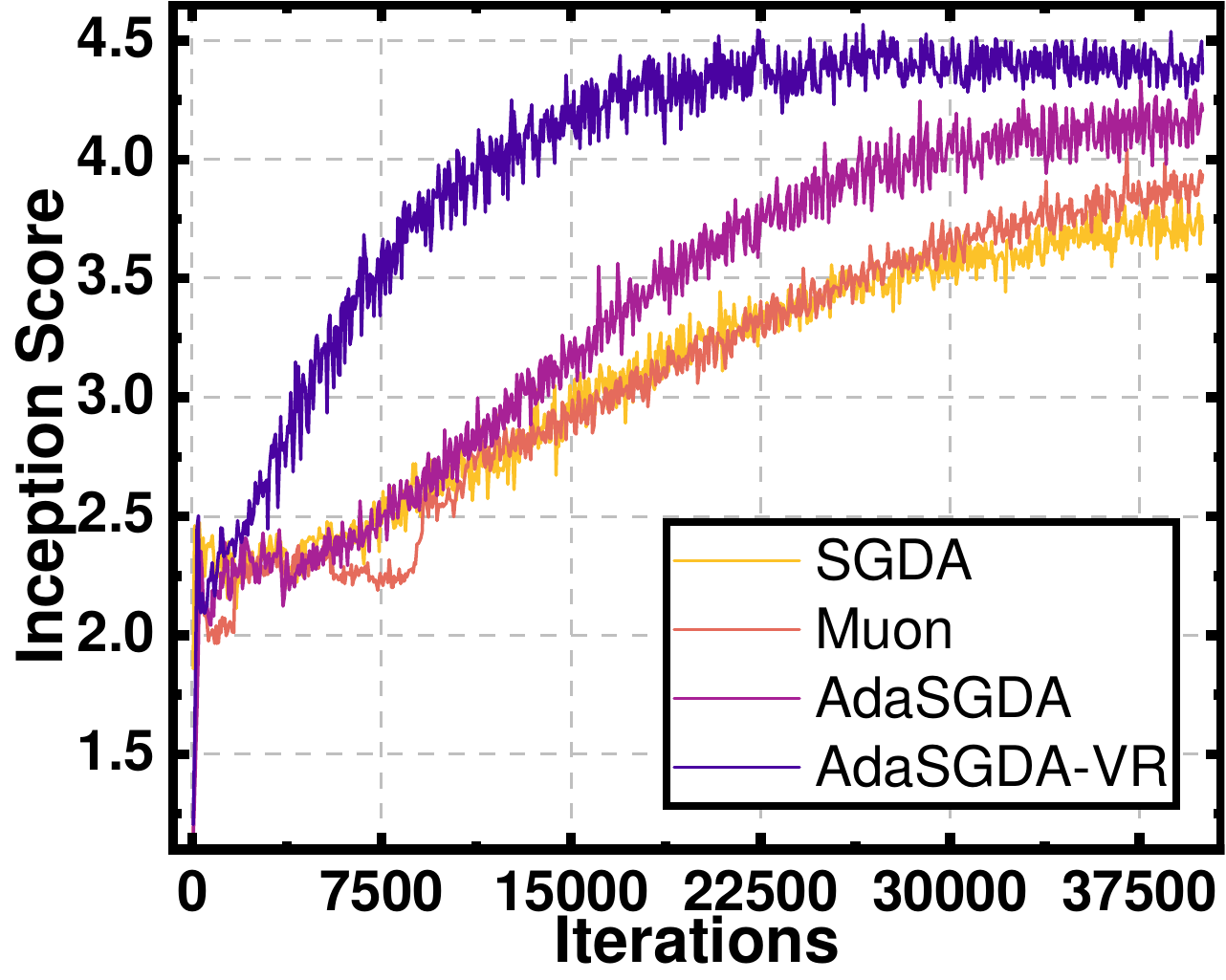}
        \vspace{-0.65cm}
        \caption{Inception score comparison among baselines}
        \label{GANINCEPTION}
    \end{subfigure}
    \hfill
    \begin{subfigure}[t]{0.493\linewidth}
        \centering
        \includegraphics[width=\linewidth]{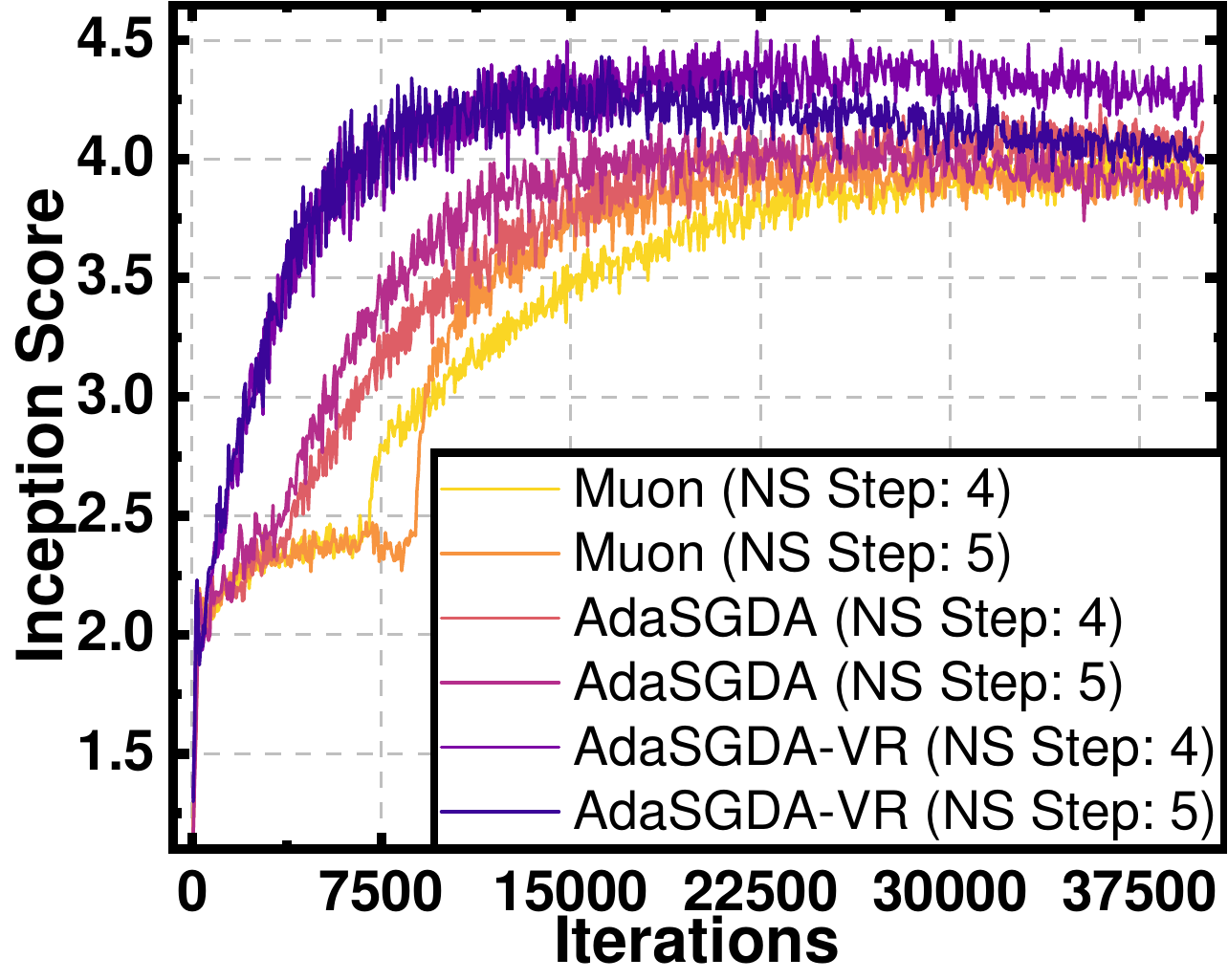}
        \vspace{-0.65cm}
        \caption{Inception score with different NS steps}
        \label{GANVARY}
    \end{subfigure}
    \vspace{-0.32cm}
    \caption{Inception score on WGAN-GP.}
    \label{GAN}
\end{minipage}
\hfill
\begin{minipage}[t]{0.496\textwidth}
    \centering
    \captionsetup{type=figure}
    \setcounter{subfigure}{0}
    \begin{subfigure}[t]{0.493\linewidth}
        \centering
        \includegraphics[width=\linewidth]{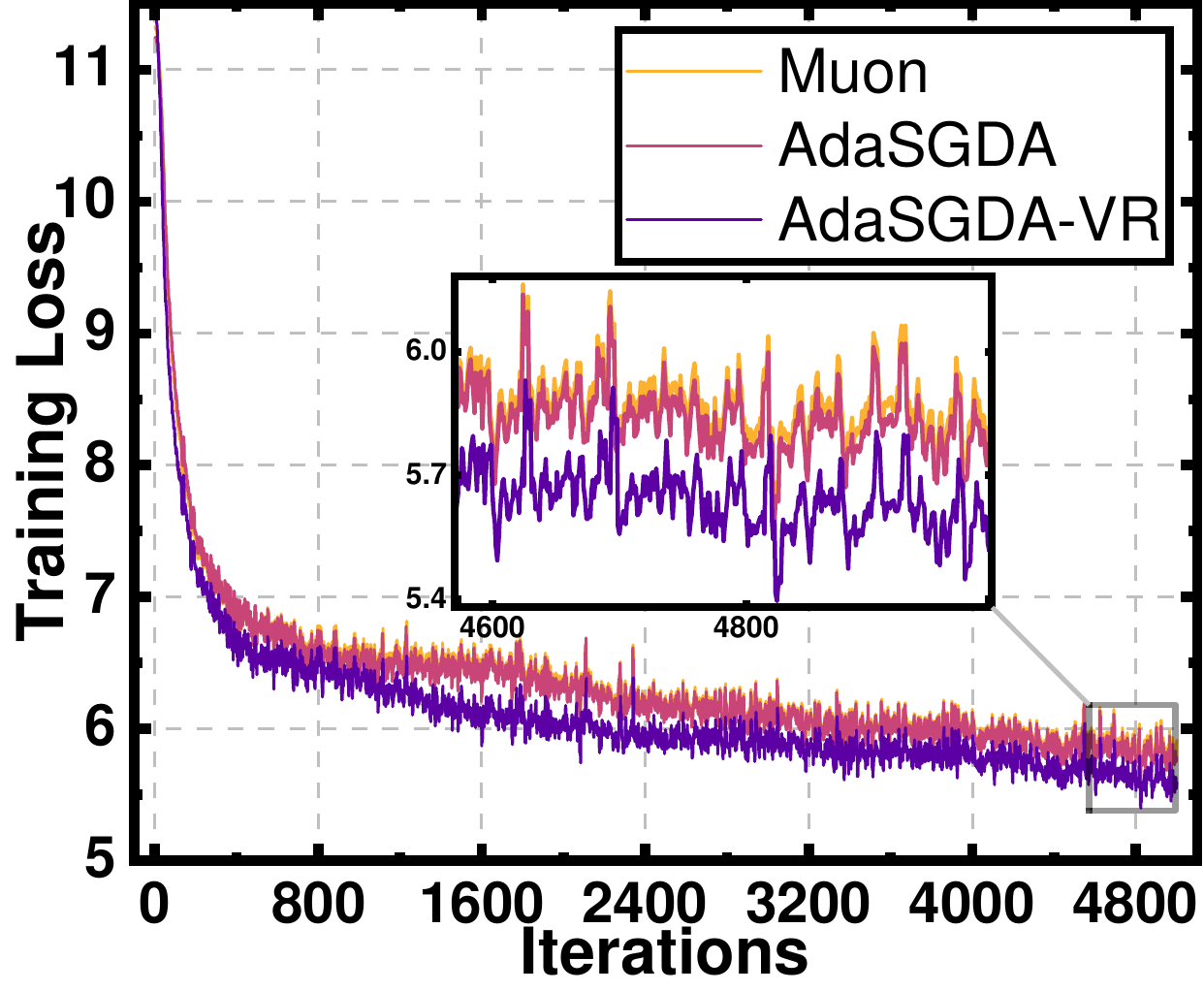}
        \vspace{-0.65cm}
        \caption{Training loss with OPT-125M model}
        \label{llm125}
    \end{subfigure}
    \hfill
    \begin{subfigure}[t]{0.493\linewidth}
        \centering
        \includegraphics[width=\linewidth]{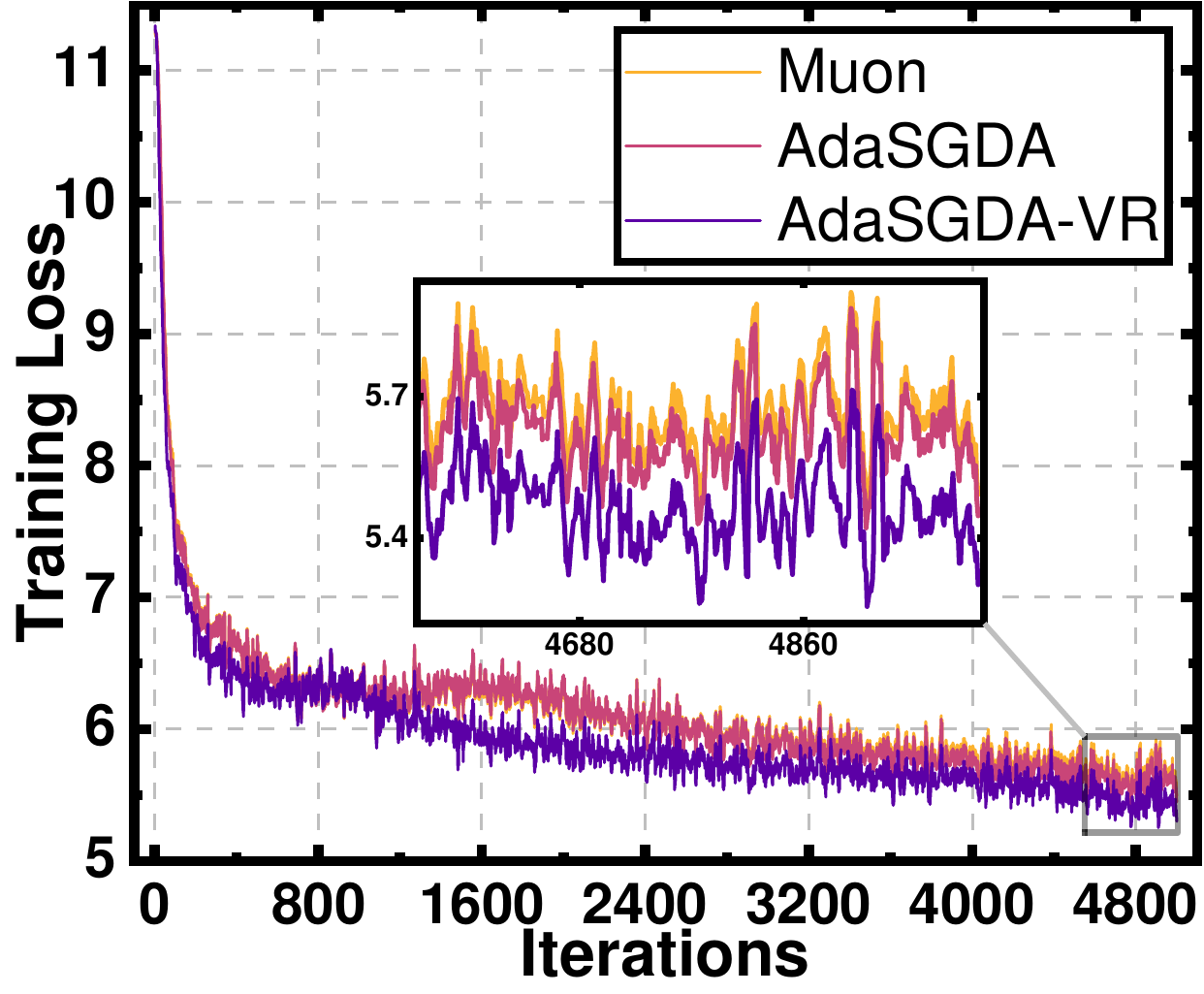}
        \vspace{-0.65cm}
        \caption{Training loss with OPT-1.3B model}
        \label{llm1.2}
    \end{subfigure}
    \vspace{-0.32cm}
    \caption{Training loss on the FineWeb dataset.}
    \label{llmtt}
\end{minipage}
\vspace{-0.6cm}
\end{figure*}

In \cref{fmnisttt,cifar10tt}, we report the training and test accuracy on FMNIST and CIFAR-10; additional results on MNIST and CIFAR-100 are provided in \cref{experimentsec}. Overall, \textbf{AdaSGDA} achieves the best performance across all datasets and baselines. Moreover, \textbf{AdaSGDA-VR} further improves over AdaSGDA, as variance reduction reduces the stochastic gradient noise and leads to more stable and faster progress in practice.
Notably, our methods also consistently outperform Muon. This advantage is attributed to our parameter-free design under orthogonal updates: the accumulated-gradient-norm-based stepsizes automatically adapt to the training dynamics, eliminating manual stepsize tuning while enabling convergence at optimal rates under orthogonalization.
In addition, we evaluate on FineWeb using OPT-125M and OPT-1.3B in \cref{llmtt}. Our methods again outperform the Muon-based minimax optimization method, demonstrating strong performance in more challenging large-scale settings.

\vspace{-0.25cm}
\subsection{Generative Adversarial Networks}
\vspace{-0.2cm}
Another widely studied application of minimax optimization is generative adversarial networks. In GAN training, the discriminator (critic) learns to distinguish real from generated samples, while the generator is trained simultaneously to produce samples that can fool the critic, resulting in a coupled minimax game. In our experiments, we adopt the WGAN-GP objective \citep{gulrajani2017improved}, which enforces the critic to be approximately $1$-Lipschitz via a gradient penalty, and evaluate performance on the CIFAR-10 and CIFAR-100 datasets.

In \cref{GANINCEPTION}, we plot the Inception Scores \citep{salimans2016improved} of different baselines. Our methods achieve the highest scores throughout training, demonstrating clear practical advantages on this GAN task. In \cref{GANVARY}, we further compare Muon and our methods under different numbers of NS steps, which affect the accuracy of the orthogonalization (i.e., the quality of the SVD/polar-factor approximation). Across all NS settings, our methods consistently outperform Muon, indicating that the performance gains persist even when the orthogonalization accuracy varies. We attribute this robustness to our adaptive design, where norm-accumulation-based stepsizes automatically adapt under orthogonal updates, maintaining strong performance across different NS configurations in line with our theoretical insights.
Additional results on sensitivity of NS steps, stepsize choices, and the momentum coefficient $\beta$, as well as wall-clock time performance, are provided in \cref{experimentsec}.
\vspace{-0.3cm}
\section{Conclusions and Limitations}
\vspace{-0.25cm}
We proposed \textbf{AdaSGDA}, a single-loop method with orthogonal updates for NC--SC minimax optimization. Specifically, our method introduces an adaptive update rule that removes problem-parameter-based stepsize tuning, while stabilizing the coupled primal--dual dynamics inherent to minimax optimization. Through rigorous theoretical analysis, we proved that AdaSGDA achieves the $\mathcal{O}(T^{-{1}/{4}})$ convergence guarantee. We further developed \textbf{AdaSGDA-VR} by integrating variance reduction, and showed that it attains the faster $\mathcal{O}(T^{-{1}/{3}})$ rate, achieving the best-known rate for NC--SC minimax optimization. Extensive experiments validate the effectiveness and scalability of our proposed methods. Nevertheless, despite the problem-parameter-free design of our methods, the theoretical guarantees still rely on horizon-dependent schedules and batch-size requirements. Reducing these dependencies remains an open direction. We aim to address these limitations in future work to further improve the practical flexibility and efficiency of our methods.

\bibliography{ref}
\bibliographystyle{plainnat}

\newpage
\appendix
\section{Related Works}\label{secrelated work}
\vspace{-0.1cm}

\textbf{Minimax Optimization.} Minimax problems are an important and tractable optimization class with broad interest in optimization and machine learning \citep{ostrovskii2021efficient,kong2021accelerated,zhao2024primal}. A classical line of work studies stochastic algorithms with stationarity guarantees; for instance, \citet{rafique2022weakly} propose methods that rely on nested loops to approximately maximize the dual variable, which can be computationally costly. Motivated by large-scale training, many recent works develop single-loop schemes that enforce time-scale separation implicitly while keeping updates simple \citep{zheng2023universal,lin2025two}. While more deployable, these methods typically require stepsize choices based on problem-dependent constants (e.g., smoothness, strong concavity, or gradient-noise level), which are often unknown and thus require substantial manual tuning.

To mitigate dependence on problem constants, \citet{yang2022nest} propose a nested adaptive method that achieves parameter-free via an inner loop, and \citet{li2022tiada} propose a single-loop adaptive method using separated adaptive factors for the primal and dual stepsizes; adaptive minimax optimization in federated and decentralized settings has also been explored in \citep{huang2024adaptive,huang2024achieving}. Despite this progress, existing parameter-free guarantees are typically near-optimal, as the rates incur residual terms or logarithmic penalties that prevent matching the best-known scaling with the iteration budget. Recently, orthogonal update techniques have emerged as an effective mechanism for stabilizing optimization \citep{jordan2024muon}. However, their role in stochastic minimax optimization has not been systematically studied, and a rigorous convergence theory in this setting has yet to be established. This gap is nontrivial because orthogonalization decouples update magnitudes from gradient norms, and the resulting stepsize tuning becomes even more delicate in minimax optimization due to the coupled primal and dual dynamics. To the best of our knowledge, it remains open to design a single-loop, parameter-free minimax method with orthogonal updates that achieves the optimal convergence rate. 

\textbf{Orthogonal Update Techniques.} \citet{Muon2024} introduced Muon, which orthogonalizes a momentum matrix using a few Newton--Schulz iterations, and reported strong empirical performance at large language model scale \citep{liu2025muon}. On the theoretical side, Muon has been analyzed under an exact polar step computed via singular value decomposition \citep{shen2025convergence}, under Frobenius-smoothness with dimension-dependent constants \citep{li2025note}, and when the polar step is approximated by Newton--Schulz iterations \citep{he2025low}. However, these guarantees typically rely on stepsize choices that depend on problem-specific parameters to attain the best-known rates, which can hinder practical deployment when such parameters are unknown.
Several adaptive variants of Muon have also been explored, including combining Muon with element-wise adaptivity \citep{si2025adamuon}, SOAP \citep{liu2025cosmos}, and related adaptive mechanisms \citep{zhang2025adagrad}. 

\vspace{-0.05cm}

While these hybrids often improve performance in practice, existing orthogonal-update methods and theory primarily focus on minimization settings. In contrast, orthogonalized updates for minimax optimization are still largely unexplored, and theoretical guidance is scarce regarding how to choose stepsizes or design parameter-free rules. These limitations motivate a more principled treatment of orthogonalized updates in minimax optimization that is both theoretically grounded and practically robust.

\vspace{-0.2cm}
\paragraph{Adaptive Methods.} Adaptive gradient methods have played a central role in modern optimization. Starting from AdaGrad \citep{mcmahan2010adaptive,duchi2011adaptive}, which introduced coordinate-wise scaling in online convex optimization, a large family of adaptive optimizers has become standard in deep learning, including Adadelta \citep{zeiler2012adadelta}, RMSprop \citep{tieleman2017divide}, and Adam \citep{kingma2014adam, luo2019adaptive,xie2024adan}. A particularly relevant line develops {parameter-free} adaptivity via normalized or self-tuned stepsizes, e.g., AdaNGD \citep{levy2017online}, AcceleGrad \citep{levy2018online}, and AdaGrad-Norm \citep{xie2020linear, zhai2026problem}, which reduce or eliminate reliance on problem constants in stepsize selection.
Recent works further clarify the power and limits of adaptive stepsizes through lower bounds and refined designs, including norm-based objectives \citep{jiang2024convergence}, on-the-fly local curvature estimation \citep{malitsky2019adaptive}, and restart-based parameter-free acceleration \citep{marumo2024parameter}.

\vspace{-0.05cm}

Beyond element-wise adaptivity, another line of work exploits the matrix structure of
gradients via matrix-aware preconditioning. Shampoo \citep{gupta2018shampoo} predates Muon and becomes equivalent to Muon when momentum and accumulation are removed. ASGO \citep{an2025asgo} proposes an adaptive one-sided preconditioner and reduces to Muon under the same simplifications. PolarGrad \citep{lau2025polargrad} provides a unifying view of matrix-aware preconditioned optimizers and studies polar-decomposition updates that subsume Muon. 
More broadly, parameter-free adaptivity and geometry-modifying updates have distinct advantages: parameter-free adaptivity automatically calibrates stepsizes to stochasticity and evolving training dynamics, improving robustness and reducing manual tuning, whereas orthogonalization reshapes the update geometry, often enhancing stability and scalability in large-scale, high-dimensional training. However, developing a single-loop method for stochastic minimax optimization that addresses both aspects remains nontrivial, since the primal--dual coupling amplifies instability and makes stepsize choices particularly delicate. Establishing provable guarantees without relying on problem-dependent tuning under orthogonalized updates therefore remains an important open problem.

\vspace{-0.2cm}
\section{Supporting Lemmas}
\begin{lemma}[Smoothness of $\Phi(\cdot)$ and Lipschitzness of $Y^*(\cdot)$. Lemma 4.3 in \citet{lin2020gradient}]
Under Assumptions \ref{smoothness} and \ref{concavity}, we have that $Y^*(\cdot)$ is $\lambda$-Lipschitz with $\lambda=\frac{L}{\mu}$, and $\Phi(\cdot)$ is $L_\Phi=(L + \lambda L)$-smooth with $\nabla \Phi(X) = \nabla_X f(X, Y^*(X))$.
\label{additionallemma}
\end{lemma}
\vspace{-0.127cm}

\begin{lemma}[A logarithmic bound for cumulative ratios]
\label{lem:log_cum_ratio}
Let $\{a_t\}_{t=1}^T$ be a sequence of nonnegative numbers with $a_1>0$.
Define the cumulative sums
$A_t := \sum_{i=1}^t a_i$, where $t=1,\dots,T.$
Then it holds that
\[
\sum_{t=1}^T \frac{a_t}{A_t}
\;\le\;
1+\ln\!\left(\frac{A_T}{a_1}\right)
\;=\;
1+\ln\!\left(\frac{\sum_{t=1}^T a_t}{a_1}\right).
\]
\end{lemma}

\begin{proof}
Note that $A_1=a_1>0$ and $A_t$ is nondecreasing since $a_t\ge 0$.
For any $t\ge 2$, we write
\[
\ln(A_t)-\ln(A_{t-1})
=
\int_{A_{t-1}}^{A_t}\frac{1}{u}\,du.
\]
Since $u\mapsto 1/u$ is decreasing on $(0,\infty)$, $u\in[A_{t-1},A_t]$ implies $\frac{1}{u}\ge \frac{1}{A_t}$. Therefore,
\[
\ln(A_t)-\ln(A_{t-1})
=
\int_{A_{t-1}}^{A_t}\frac{1}{u}\,du
\;\ge\;
\int_{A_{t-1}}^{A_t}\frac{1}{A_t}\,du
=
\frac{A_t-A_{t-1}}{A_t}
=
\frac{a_t}{A_t}.
\]
Therefore,
\[
\sum_{t=1}^T \frac{a_t}{A_t}
=
\frac{a_1}{A_1}+\sum_{t=2}^T \frac{a_t}{A_t}
\le
1+\sum_{t=2}^T \big(\ln(A_t)-\ln(A_{t-1})\big)
=
1+\ln(A_T)-\ln(A_1),
\]
which gives the desired bound since $A_1=a_1$.
\end{proof}

\section{Proof of \cref{theorem1}}
\begin{lemma}[Stationarity Bound]
    Suppose \cref{smoothness,concavity,stochasticgradient} hold, we have:
\begin{align*}
&\frac{1}{T}\sum_{t=1}^T \mathbb{E}[\|\nabla\Phi(X_{t-1})\|_*]\notag\\
&\leq\frac{\Delta}{T\epsilon_{\scriptscriptstyle X}} + {L_\Phi\epsilon_{\scriptscriptstyle X}}  + \frac{L_\Phi\eta^2_{\scriptscriptstyle X}}{T\epsilon_{\scriptscriptstyle X}}\left(\ln\left(\frac{\gamma^2 T}{z_0^2}\right) + 1\right)\notag\\
&\quad+ \frac{2}{T}\sum_{t=1}^T \mathbb{E}[\|M_t^{\scriptscriptstyle X} - \nabla\Phi(X_{t-1})\|_*] + \frac{2\sqrt{r}\eta_{\scriptscriptstyle X}}{T\epsilon_{\scriptscriptstyle X}}\sqrt{\ln\left(\frac{\gamma^2 T}{z_0^2}\right) + 1}\sqrt{\sum_{t=1}^T \mathbb{E}[\|M_t^{\scriptscriptstyle X} - \nabla\Phi(X_{t-1})\|_F^2]} ,
\end{align*}
\label{descentlemma}
\end{lemma}
where $\Delta:=\Phi(X_0) - \Phi_*$, $\Phi_* := \min_X \Phi(X)$, and $z_0:=\max\{m_0^{\scriptscriptstyle X},m_0^{\scriptscriptstyle Y}\}$.
\begin{proof}
Define the scalar function $g:[0,1]\to\mathbb{R}$ by
\(
g(s):=\Phi\bigl(X_t+s(X_{t+1}-X_t)\bigr).
\)
By the fundamental theorem of calculus,
\begin{equation}
\label{eq:ftc}
\Phi(X_{t+1})-\Phi(X_t)=g(1)-g(0)=\int_0^1 g'(s)\,ds.
\end{equation}
By the chain rule, we have
\begin{equation}
\label{eq6ftc}
g'(s)=\left\langle \nabla \Phi\bigl(X_t+s(X_{t+1}-X_t)\bigr),\,X_{t+1}-X_t\right\rangle.
\end{equation}
Substituting \cref{eq6ftc} into \cref{eq:ftc}, we get
\begin{align*}
&\Phi(X_{t+1})-\Phi(X_t)\notag\\
&= \langle \nabla\Phi(X_t),X_{t+1}-X_t\rangle
 + \int_0^1 \left\langle \nabla \Phi\bigl(X_t+s(X_{t+1}-X_t)\bigr)-\nabla\Phi(X_t),\,X_{t+1}-X_t\right\rangle ds.
\end{align*}
Using Hölder's inequality \(
\left\langle A,B\right\rangle\le \|A\|_*\,\|B\|_{\mathrm{op}},
\)
we bound the integrand as
\begin{align*}
&\left\langle \nabla \Phi\bigl(X_t+s(X_{t+1}-X_t)\bigr)-\nabla\Phi(X_t),\,X_{t+1}-X_t\right\rangle\notag\\
&\le
\left\|\nabla \Phi\bigl(X_t+s(X_{t+1}-X_t)\bigr)-\nabla\Phi(X_t)\right\|_*\;\|X_{t+1}-X_t\|_{\mathrm{op}}.
\end{align*}
By the smoothness of $\Phi$ in \cref{additionallemma}, we have
\[
\left\|\nabla \Phi\bigl(X_t+s(X_{t+1}-X_t)\bigr)-\nabla\Phi(X_t)\right\|_*
\le L_\Phi \|s(X_{t+1}-X_t)\|_{\mathrm{op}}
= sL_\Phi \|X_{t+1}-X_t\|_{\mathrm{op}}.
\]
Therefore,
\[
\left\langle \nabla \Phi\bigl(X_t+s(X_{t+1}-X_t)\bigr)-\nabla\Phi(X_t),\,X_{t+1}-X_t\right\rangle
\le sL_\Phi \|X_{t+1}-X_t\|_{\mathrm{op}}^2.
\]
Plugging back and integrating yields
\begin{align}
\Phi(X_{t+1})-\Phi(X_t)
&\le \langle \nabla\Phi(X_t),X_{t+1}-X_t\rangle + \int_0^1 sL_\Phi \|X_{t+1}-X_t\|_{\mathrm{op}}^2\,ds \notag\\
&\leq \langle \nabla\Phi(X_t),X_{t+1}-X_t\rangle + {L_\Phi}\|X_{t+1}-X_t\|_{\mathrm{op}}^2.
\label{descentequation}
\end{align}

Then define
$\theta_{t+1} := \frac{\min\{\|\nabla_X F(X_{t},Y_{t};\xi_{t}^{\scriptscriptstyle X})\|_F, \gamma\}}{\max\{m_{t+1}^{\scriptscriptstyle X},m_{t+1}^{\scriptscriptstyle Y}\}}$ and let $\mathbb{E}_t[\cdot] := \mathbb{E}[\cdot|X_{t-1}]$ denote the conditional expectation given the previous iterates $X_0, \ldots, X_{t-1}$.
By \cref{descentequation}, we have
\begin{align*}
&\mathbb{E}_t[\Phi(X_t) - \Phi(X_{t-1})] &\\
&\leq \mathbb{E}_t\left[-\langle\nabla\Phi(X_{t-1}), \max\{\epsilon_{\scriptscriptstyle X}, \eta_{\scriptscriptstyle X}\theta_t\} O_t^{\scriptscriptstyle X}\rangle\right] + L_\Phi\mathbb{E}_t\left[\max\{\epsilon_{\scriptscriptstyle X}, \eta_{\scriptscriptstyle X}\theta_t\}^2\right]\\
&\leq \mathbb{E}_t\left[-\langle\nabla\Phi(X_{t-1}) - M_t^{\scriptscriptstyle X}, \max\{\epsilon_{\scriptscriptstyle X}, \eta_{\scriptscriptstyle X}\theta_t\} O_t^{\scriptscriptstyle X}\rangle\right]\notag\\
&\quad- \mathbb{E}_t[\max\{\epsilon_{\scriptscriptstyle X}, \eta_{\scriptscriptstyle X}\theta_t\} \|M_t^{\scriptscriptstyle X}\|_*] + {L_\Phi\epsilon_{\scriptscriptstyle X}^2} + {L_\Phi\eta^2_{\scriptscriptstyle X} }\mathbb{E}_t[[\theta_t]^2]\\
&\overset{(a)}{\leq} \mathbb{E}_t[\max\{\epsilon_{\scriptscriptstyle X}, \eta_{\scriptscriptstyle X}\theta_t\} \|\nabla\Phi(X_{t-1}) - M_t^{\scriptscriptstyle X}\|_*] - \mathbb{E}_t[\max\{\epsilon_{\scriptscriptstyle X}, \eta_{\scriptscriptstyle X}\theta_t\} \|M_t^{\scriptscriptstyle X}\|_*]\notag\\
&\quad+ {L_\Phi\epsilon_{\scriptscriptstyle X}^2} + {L_\Phi\eta^2_{\scriptscriptstyle X} }\mathbb{E}_t[[\theta_t]^2]\\
&\leq \mathbb{E}_t[2\max\{\epsilon_{\scriptscriptstyle X}, \eta_{\scriptscriptstyle X}\theta_t\} \|\nabla\Phi(X_{t-1}) - M_t^{\scriptscriptstyle X}\|_*] - \mathbb{E}_t[\max\{\epsilon_{\scriptscriptstyle X}, \eta_{\scriptscriptstyle X}\theta_t\} \|\nabla\Phi(X_{t-1})\|_*] \notag\\
&\quad+ {L_\Phi\epsilon_{\scriptscriptstyle X}^2} + {L_\Phi\eta^2_{\scriptscriptstyle X} }\mathbb{E}_t[[\theta_t]^2],
\end{align*}
where (a) is due to $-\langle A,B\rangle \le |\langle A,B\rangle| \le \|A\|_{*}\,\|B\|_\mathrm{op}$.
Then by the law of total expectation,
\begin{align}
&\sum_{t=1}^T \mathbb{E}[\max\{\epsilon_{\scriptscriptstyle X}, \eta_{\scriptscriptstyle X}\theta_t\} \|\nabla\Phi(X_{t-1})\|_*]\notag\\
&\leq \Phi(X_0) - \Phi_* + 2\sum_{t=1}^T \mathbb{E}[\max\{\epsilon_{\scriptscriptstyle X}, \eta_{\scriptscriptstyle X}\theta_t\} \|\nabla\Phi(X_{t-1}) - M_t^{\scriptscriptstyle X}\|_*] + {L_\Phi\epsilon_{\scriptscriptstyle X}^2 T} + {L_\Phi\eta^2_{\scriptscriptstyle X}}\sum_{t=1}^T \mathbb{E}[\theta_t^2]\notag\\
&\leq \Phi(X_0) - \Phi_* + 2\epsilon_{\scriptscriptstyle X}\sum_{t=1}^T \mathbb{E}[\|E_t\|_*] + 2\eta_{\scriptscriptstyle X}\sum_{t=1}^T \mathbb{E}[\theta_t \|E_t\|_*] + {L_\Phi\epsilon_{\scriptscriptstyle X}^2 T}  + {L_\Phi\eta^2_{\scriptscriptstyle X}}\sum_{t=1}^T \mathbb{E}[\theta_t^2]\notag\\
&\leq \Phi(X_0) - \Phi_* + 2\epsilon_{\scriptscriptstyle X}\sum_{t=1}^T \mathbb{E}[\|E_t\|_*] + 2\eta_{\scriptscriptstyle X}\sum_{t=1}^T \sqrt{\mathbb{E}[\|E_t\|_*^2]} \sqrt{\mathbb{E}[\theta_t^2]} + {L_\Phi\epsilon_{\scriptscriptstyle X}^2 T}  + {L_\Phi\eta^2_{\scriptscriptstyle X}}\sum_{t=1}^T \mathbb{E}[\theta_t^2]\notag\\
&\leq \Phi(X_0) - \Phi_* + 2\epsilon_{\scriptscriptstyle X}\sum_{t=1}^T \mathbb{E}[\|E_t\|_*] + 2\eta_{\scriptscriptstyle X}\sqrt{\sum_{t=1}^T \mathbb{E}[\|E_t\|_*^2]} \sqrt{\sum_{t=1}^T \mathbb{E}[\theta_t^2]}\notag\\
&\quad+ {L_\Phi\epsilon_{\scriptscriptstyle X}^2 T}  + {L_\Phi\eta^2_{\scriptscriptstyle X}}\sum_{t=1}^T \mathbb{E}[\theta_t^2]\notag\\
&\overset{(a)}{\leq} \Phi(X_0) - \Phi_* + 2\epsilon_{\scriptscriptstyle X}\sum_{t=1}^T \mathbb{E}[\|E_t\|_*] + 2\eta_{\scriptscriptstyle X}\sqrt{\sum_{t=1}^T \mathbb{E}[\|E_t\|_*^2]} \sqrt{\ln\left(\frac{\gamma^2 T}{z_0^2}\right) + 1} \notag\\
&\quad+ {L_\Phi\epsilon_{\scriptscriptstyle X}^2 T}  + {L_\Phi\eta^2_{\scriptscriptstyle X}}\left(\ln\left(\frac{\gamma^2 T}{z_0^2}\right) + 1\right)\notag\\
&\leq \Phi(X_0) - \Phi_* + 2\epsilon_{\scriptscriptstyle X}\sum_{t=1}^T \mathbb{E}[\|E_t\|_*] + 2\eta_{\scriptscriptstyle X}\sqrt{r\sum_{t=1}^T \mathbb{E}[\|E_t\|_F^2]} \sqrt{\ln\left(\frac{\gamma^2 T}{z_0^2}\right) + 1} \notag\\
&\quad+ {L_\Phi\epsilon_{\scriptscriptstyle X}^2 T}  + {L_\Phi\eta^2_{\scriptscriptstyle X}}\left(\ln\left(\frac{\gamma^2 T}{z_0^2}\right) + 1\right)\notag, \label{eq7}
\end{align}
where $E_t := M_t^{\scriptscriptstyle X} - \nabla\Phi(X_{t-1})$ and $\Phi_* := \min_X \Phi(X)$. The last inequality follows from the inequality
$\|A\|_\ast \le \sqrt{r}\,\|A\|_F$. Moreover, (a) follows from \cref{lem:log_cum_ratio}.
Specifically, let $u_t=\min\{\|\nabla_X F(X_{t-1}, Y_{t-1}; \xi^{\scriptscriptstyle X}_{t-1})\|_F,\gamma\}$ and $S_t=z_0^2+\sum_{s=1}^t u_s^2$. 
Since $\max\{m_{t}^{\scriptscriptstyle X},m_{t}^{\scriptscriptstyle Y}\}^2 \ge S_t$, we have
$\theta_t^2=\frac{u_t^2}{\max\{m_{t}^{\scriptscriptstyle X},m_{t}^{\scriptscriptstyle Y}\}^2}\le \frac{u_t^2}{S_t}$.
Applying \cref{lem:log_cum_ratio} yields
$\sum_{t=1}^T \theta_t^2 \le \sum_{t=1}^T \frac{u_t^2}{S_t} \le \ln\!\left(\frac{S_T}{S_0}\right)
= \ln\!\left(1+\frac{\sum_{t=1}^T u_t^2}{z_0^2}\right)$.
Finally, using $u_t\le \gamma$ so that $\sum_{t=1}^T u_t^2 \le \gamma^2 T$ and $\ln(1+v)\le \ln v + 1$ for $v>0$, we obtain
$\sum_{t=1}^T \theta_t^2 \le \ln\!\left(1+\frac{\gamma^2 T}{z_0^2}\right)
\le \ln\!\left(\frac{\gamma^2 T}{z_0^2}\right)+1$. Then we can get
\begin{align}
    &\frac{1}{T} \sum_{t=1}^T \mathbb{E}[\|\nabla\Phi(X_{t-1})\|_*]\notag\\
&\leq \frac{1}{T\epsilon_{\scriptscriptstyle X}} \sum_{t=1}^T \mathbb{E}[\max\{\epsilon_{\scriptscriptstyle X}, \eta_{\scriptscriptstyle X}\theta_t\} \|\nabla\Phi(X_{t-1})\|_*]\notag\\
&\leq\frac{\Delta}{T\epsilon_{\scriptscriptstyle X}} + \frac{2}{T}\sum_{t=1}^T \mathbb{E}[\|E_t\|_*] + \frac{2\sqrt{r}\eta_{\scriptscriptstyle X}}{T\epsilon_{\scriptscriptstyle X}}\sqrt{\ln\left(\frac{\gamma^2 T}{z_0^2}\right) + 1}\sqrt{\sum_{t=1}^T \mathbb{E}[\|E_t\|_F^2]} \notag\\
&\quad+ {L_\Phi\epsilon_{\scriptscriptstyle X}}  + \frac{L_\Phi\eta^2_{\scriptscriptstyle X}}{T\epsilon_{\scriptscriptstyle X}}\left(\ln\left(\frac{\gamma^2 T}{z_0^2}\right) + 1\right),\notag
\end{align}
where $\Delta:=\Phi(X_0) - \Phi_*$. The proof is thus completed.
\end{proof}

We have the following results on the terms $\frac{1}{T}\sum_{t=1}^T \mathbb{E}[\|M_t^{\scriptscriptstyle X} - \nabla\Phi(X_{t-1})\|_*]$ and $\sum_{t=1}^T \mathbb{E}[\|M_t^{\scriptscriptstyle X} - \nabla\Phi(X_{t-1})\|_F^2]$ in \cref{descentlemma}.
\begin{lemma}
Under \cref{smoothness,concavity,stochasticgradient}, for \cref{alg1}, suppose the total iteration rounds is $T$, the term $\frac{1}{T}\sum_{t=1}^T \mathbb{E}[\|M_t^{\scriptscriptstyle X} - \nabla\Phi(X_{t-1})\|_*]$ is upper bounded by:
\begin{align}
&\frac{1}{T}\sum_{t=1}^T \mathbb{E}[\|M_t^{\scriptscriptstyle X} - \nabla\Phi(X_{t-1})\|_*]\notag\\
&\leq \frac{\kappa\sqrt{r}}{T(1-\beta)\sqrt{b_1}} + \frac{\kappa}{T}\sum_{t=0}^{T-1} \sqrt{\frac{r(1-\beta)}{b_t}} + \frac{\beta L_\Phi\epsilon_{\scriptscriptstyle X}}{1-\beta} + \frac{\beta\eta_{\scriptscriptstyle X} L_\Phi}{(1-\beta) \sqrt{T}} \sqrt{\ln\left(\frac{\gamma^2 T}{\left[\max\{m_0^{\scriptscriptstyle X},m_0^{\scriptscriptstyle Y}\}\right]^2}\right) + 1}\notag\\
&\quad + L\Bigg(\frac{\Delta_0^{\scriptscriptstyle Y}}{T\mu\epsilon_{\scriptscriptstyle Y}}
+\frac{G\epsilon_{\scriptscriptstyle X}\sqrt{r}(2+\lambda)}{\mu\epsilon_{\scriptscriptstyle Y}}+\frac{G\eta_{\scriptscriptstyle X}\sqrt{r}(2+\lambda)}{\mu\epsilon_{\scriptscriptstyle Y}\sqrt{T}}\sqrt{\ln\left(\frac{\gamma^2 T}{\left[\max\{m_0^{\scriptscriptstyle X},m_0^{\scriptscriptstyle Y}\}\right]^2}\right) + 1}\notag\\
&\quad+\frac{L\epsilon_X^2(2+\lambda^2)}{2\mu\epsilon_Y}+\frac{L\eta_X^2(2+\lambda^2)}{2T\mu\epsilon_Y}
\ln\left(\frac{\gamma^2T}{\left[\max\{m_0^{\scriptscriptstyle X},m_0^{\scriptscriptstyle Y}\}\right]^2}+1\right)\notag\\
&\quad+\frac{L}{\mu\epsilon_Y}
\sqrt{\left(\epsilon_X^2+\frac{\eta_X^2}{T}\ln\left(\frac{\gamma^2T}{\left[\max\{m_0^{\scriptscriptstyle X},m_0^{\scriptscriptstyle Y}\}\right]^2}+1\right)\right)
\left(\epsilon_Y^2+\frac{\eta_Y^2}{T}\ln\left(\frac{\gamma^2T}{\left[m_0^{\scriptscriptstyle Y}\right]^2}+1\right)\right)}\notag\\
&\quad+\frac{L r\epsilon_{\scriptscriptstyle Y}}{2\mu}+\frac{L r\eta_{\scriptscriptstyle Y}^2}{2T\mu\epsilon_{\scriptscriptstyle Y}}\ln\!\left(\frac{\gamma^2 T}{[m_0^{\scriptscriptstyle Y}]^2}+1\right)+\frac{2\sqrt{r}}{\mu\epsilon_{\scriptscriptstyle Y}}\sqrt{\epsilon_{\scriptscriptstyle Y}^2+\frac{\eta_{\scriptscriptstyle Y}^2}{T}\ln\!\left(\frac{\gamma^2 T}{[m_0^{\scriptscriptstyle Y}]^2}+1\right)}\notag\\
&\quad\cdot\left(\frac{3\kappa^2}{T(1-\beta)b_1}
+\frac{6 L^2}{(1-\beta)^2}\left(
{r\epsilon_{\scriptscriptstyle X}^2}+\frac{r\eta_{\scriptscriptstyle X}^2}{T}\ln\!\left(\frac{\gamma^2 T}{\left[\max\{m_0^{\scriptscriptstyle X},m_0^{\scriptscriptstyle Y}\}\right]^2}+1\right)\right.\right.\notag\\
&\quad\left.\left.+
{r\epsilon_{\scriptscriptstyle Y}^2}+\frac{r\eta_{\scriptscriptstyle Y}^2}{T}\ln\!\left(\frac{\gamma^2 T}{[m_0^{\scriptscriptstyle Y}]^2}+1\right)
\right)
+\frac{3(1-\beta)}{T(1+\beta)}\sum_{t=0}^{T-1}\frac{\kappa^2}{b_t}\right)^{\frac{1}{2}}\Bigg),
\end{align}
where $\Delta_0^{\scriptscriptstyle Y}:=\Phi(X_0)-f(X_0,Y_0).$
\label{e1lemma}
\end{lemma}

\begin{lemma}
Under \cref{smoothness,concavity,stochasticgradient}, for \cref{alg1}, suppose the total iteration rounds is $T$, the term $\sum_{t=1}^T \mathbb{E}[\|M_t^{\scriptscriptstyle X} - \nabla\Phi(X_{t-1})\|_F^2]$ can be upper bounded by:
\begin{align}
&\sum_{t=1}^T \mathbb{E}[\|M_t^{\scriptscriptstyle X} - \nabla\Phi(X_{t-1})\|_F^2]\notag\\
&\leq\frac{4}{1-\beta^2} \frac{\kappa^2}{b_1}+ \frac{4(1-\beta)}{1+\beta}\sum_{t=1}^{T}\frac{\kappa^2}{b_t} +  \frac{4L_\Phi^2r\beta^2 \epsilon_{\scriptscriptstyle X}^2T}{1-\beta^2} + \frac{4L_\Phi^2r\eta_{\scriptscriptstyle X}^2 \beta^2}{1-\beta^2} T\ln\!\left(\frac{\gamma^2T}{\left[\max\{m_0^{\scriptscriptstyle X},m_0^{\scriptscriptstyle Y}\}\right]^2}+1\right)\notag\\
&\quad+ 4L^2T(1-\beta) \Bigg[\frac{8 G\Delta_0^{\scriptscriptstyle Y}}{7\mu^2\epsilon_{\scriptscriptstyle Y}}
+\frac{4Gr\epsilon_{\scriptscriptstyle X}^2T(L_\Phi+L)}{7\mu^2\epsilon_{\scriptscriptstyle Y}}+\frac{16r^2L^2G\epsilon_{\scriptscriptstyle X}^2T}{7\mu^3\epsilon_{\scriptscriptstyle Y}}+\frac{16rG^2L^2\epsilon_{\scriptscriptstyle X}^2T}{7\mu^4\epsilon_{\scriptscriptstyle Y}^2}\notag\\
&\quad+\left(\frac{4Gr\eta_{\scriptscriptstyle X}^2(L_\Phi+L)}{7\mu^2\epsilon_{\scriptscriptstyle Y}}+\frac{16r^2L^2G\eta_{\scriptscriptstyle X}^2}{7\mu^3\epsilon_{\scriptscriptstyle Y}}+\frac{16rG^2L^2\eta_{\scriptscriptstyle X}^2}{7\mu^4\epsilon_{\scriptscriptstyle Y}^2}\right)\ln\!\left(\frac{\gamma^2 T}{\left[\max\{m_0^{\scriptscriptstyle X},m_0^{\scriptscriptstyle Y}\}\right]^2}+1\right)\notag\\
&\quad
+\frac{G\epsilon_{\scriptscriptstyle Y}T}{7\mu}+\frac{4G L r\epsilon_{\scriptscriptstyle Y}T}{7\mu^2}+\left(\frac{G\eta_{\scriptscriptstyle Y}^2}{7\mu\epsilon_{\scriptscriptstyle Y}}+\frac{4\eta_{\scriptscriptstyle Y}^2G L r}{7\mu^2\epsilon_{\scriptscriptstyle Y}}\right)\ln\!\left(\frac{\gamma^2 T}{[m_0^{\scriptscriptstyle Y}]^2}+1\right)
\notag\\
&\quad+
\frac{16GT\sqrt r}{7\mu^2\epsilon_{\scriptscriptstyle Y}}
\sqrt{\epsilon_{\scriptscriptstyle Y}^2+\frac{\eta_{\scriptscriptstyle Y}^2}{T}\ln\!\left(\frac{\gamma^2 T}{[m_0^{\scriptscriptstyle Y}]^2}+1\right)}\notag\\
&\quad\cdot\left(\frac{3\kappa^2}{T(1-\beta)b_1}
+\frac{6 L^2}{(1-\beta)^2}\left(
{r\epsilon_{\scriptscriptstyle X}^2}+\frac{r\eta_{\scriptscriptstyle X}^2}{T}\ln\!\left(\frac{\gamma^2 T}{\left[\max\{m_0^{\scriptscriptstyle X},m_0^{\scriptscriptstyle Y}\}\right]^2}+1\right)\right.\right.\notag\\
&\quad\left.\left.+
{r\epsilon_{\scriptscriptstyle Y}^2}+\frac{r\eta_{\scriptscriptstyle Y}^2}{T}\ln\!\left(\frac{\gamma^2 T}{[m_0^{\scriptscriptstyle Y}]^2}+1\right)
\right)
+\frac{3(1-\beta)}{T(1+\beta)}\sum_{t=0}^{T-1}\frac{\kappa^2}{b_t}\right)^{\frac{1}{2}}\Bigg],\label{eq111Fnorm}
\end{align}
where $\Delta_0^{\scriptscriptstyle Y}:=\Phi(X_0)-f(X_0,Y_0).$
\label{e2lemma}
\end{lemma}

The proof of \cref{e1lemma} is presented in \cref{sece1} and the proof of \cref{e2lemma} is presented in \cref{sece2}. In deriving these results, we will rely on the following lemma.
\begin{lemma}
Under Assumptions \ref{smoothness}-\ref{stochasticgradient}, for \cref{alg1}, suppose the total iteration rounds is $T$, we have:
\begin{align}
&\frac{1}{T}\sum_{t=1}^{T}\mathbb E[\|Y_t-Y^\star(X_t)\|_F]\notag\\
&\le
\frac{\Delta_0^{\scriptscriptstyle Y}}{T\mu\epsilon_{\scriptscriptstyle Y}}
+\frac{G\epsilon_{\scriptscriptstyle X}\sqrt{r}(2+\lambda)}{\mu\epsilon_{\scriptscriptstyle Y}}+\frac{G\eta_{\scriptscriptstyle X}\sqrt{r}(2+\lambda)}{\mu\epsilon_{\scriptscriptstyle Y}\sqrt{T}}\sqrt{\ln\left(\frac{\gamma^2 T}{\left[\max\{m_0^{\scriptscriptstyle X},m_0^{\scriptscriptstyle Y}\}\right]^2}\right) + 1}\notag\\
&\quad+\frac{L\epsilon_X^2(2+\lambda^2)}{2\mu\epsilon_Y}+\frac{L\eta_X^2(2+\lambda^2)}{2T\mu\epsilon_Y}
\ln\left(\frac{\gamma^2T}{\left[\max\{m_0^{\scriptscriptstyle X},m_0^{\scriptscriptstyle Y}\}\right]^2}+1\right)\notag\\
&\quad+\frac{L}{\mu\epsilon_Y}
\sqrt{\left(\epsilon_X^2+\frac{\eta_X^2}{T}\ln\left(\frac{\gamma^2T}{\left[\max\{m_0^{\scriptscriptstyle X},m_0^{\scriptscriptstyle Y}\}\right]^2}+1\right)\right)
\left(\epsilon_Y^2+\frac{\eta_Y^2}{T}\ln\left(\frac{\gamma^2T}{\left[m_0^{\scriptscriptstyle Y}\right]^2}+1\right)\right)}\notag\\
&\quad+\frac{L r\epsilon_{\scriptscriptstyle Y}}{2\mu}+\frac{L r\eta_{\scriptscriptstyle Y}^2}{2T\mu\epsilon_{\scriptscriptstyle Y}}\ln\!\left(\frac{\gamma^2 T}{[m_0^{\scriptscriptstyle Y}]^2}+1\right)+\frac{2\sqrt{r}}{\mu\epsilon_{\scriptscriptstyle Y}}\sqrt{\epsilon_{\scriptscriptstyle Y}^2+\frac{\eta_{\scriptscriptstyle Y}^2}{T}\ln\!\left(\frac{\gamma^2 T}{[m_0^{\scriptscriptstyle Y}]^2}+1\right)}\notag\\
&\quad\cdot\left(\frac{3\kappa^2}{T(1-\beta)b_1}
+\frac{6 L^2}{(1-\beta)^2}\left(
{r\epsilon_{\scriptscriptstyle X}^2}+\frac{r\eta_{\scriptscriptstyle X}^2}{T}\ln\!\left(\frac{\gamma^2 T}{\left[\max\{m_0^{\scriptscriptstyle X},m_0^{\scriptscriptstyle Y}\}\right]^2}+1\right)\right.\right.\notag\\
&\quad\left.\left.+
{r\epsilon_{\scriptscriptstyle Y}^2}+\frac{r\eta_{\scriptscriptstyle Y}^2}{T}\ln\!\left(\frac{\gamma^2 T}{[m_0^{\scriptscriptstyle Y}]^2}+1\right)
\right)
+\frac{3(1-\beta)}{T(1+\beta)}\sum_{t=0}^{T-1}\frac{\kappa^2}{b_t}\right)^{\frac{1}{2}},
\end{align}
and
\begin{align}
&\frac1T\sum_{t=1}^{T}\mathbb E[\|Y_t-Y^\star(X_t)\|_F^2]\notag\\
&\le
\frac{8 G\Delta_0^{\scriptscriptstyle Y}}{7\mu^2\epsilon_{\scriptscriptstyle Y}T}
+\frac{4Gr\epsilon_{\scriptscriptstyle X}^2(L_\Phi+L)}{7\mu^2\epsilon_{\scriptscriptstyle Y}}+\frac{16r^2L^2G\epsilon_{\scriptscriptstyle X}^2}{7\mu^3\epsilon_{\scriptscriptstyle Y}}+\frac{16rG^2L^2\epsilon_{\scriptscriptstyle X}^2}{7\mu^4\epsilon_{\scriptscriptstyle Y}^2}\notag\\
&\quad+\left(\frac{4Gr\eta_{\scriptscriptstyle X}^2(L_\Phi+L)}{7T\mu^2\epsilon_{\scriptscriptstyle Y}}+\frac{16r^2L^2G\eta_{\scriptscriptstyle X}^2}{7T\mu^3\epsilon_{\scriptscriptstyle Y}}+\frac{16rG^2L^2\eta_{\scriptscriptstyle X}^2}{7T\mu^4\epsilon_{\scriptscriptstyle Y}^2}\right)\ln\!\left(\frac{\gamma^2 T}{\left[\max\{m_0^{\scriptscriptstyle X},m_0^{\scriptscriptstyle Y}\}\right]^2}+1\right)\notag\\
&\quad
+\frac{G\epsilon_{\scriptscriptstyle Y}}{7\mu}+\frac{4G L r\epsilon_{\scriptscriptstyle Y}}{7\mu^2}+\left(\frac{G\eta_{\scriptscriptstyle Y}^2}{7T\mu\epsilon_{\scriptscriptstyle Y}}+\frac{4\eta_{\scriptscriptstyle Y}^2G L r}{7T\mu^2\epsilon_{\scriptscriptstyle Y}}\right)\ln\!\left(\frac{\gamma^2 T}{[m_0^{\scriptscriptstyle Y}]^2}+1\right)
\notag\\
&\quad+
\frac{16G\sqrt r}{7\mu^2\epsilon_{\scriptscriptstyle Y}}
\sqrt{\epsilon_{\scriptscriptstyle Y}^2+\frac{\eta_{\scriptscriptstyle Y}^2}{T}\ln\!\left(\frac{\gamma^2 T}{[m_0^{\scriptscriptstyle Y}]^2}+1\right)}\notag\\
&\quad\cdot\left(\frac{3\kappa^2}{T(1-\beta)b_1}
+\frac{6 L^2}{(1-\beta)^2}\left(
{r\epsilon_{\scriptscriptstyle X}^2}+\frac{r\eta_{\scriptscriptstyle X}^2}{T}\ln\!\left(\frac{\gamma^2 T}{\left[\max\{m_0^{\scriptscriptstyle X},m_0^{\scriptscriptstyle Y}\}\right]^2}+1\right)\right.\right.\notag\\
&\quad\left.\left.+
{r\epsilon_{\scriptscriptstyle Y}^2}+\frac{r\eta_{\scriptscriptstyle Y}^2}{T}\ln\!\left(\frac{\gamma^2 T}{[m_0^{\scriptscriptstyle Y}]^2}+1\right)
\right)
+\frac{3(1-\beta)}{T(1+\beta)}\sum_{t=0}^{T-1}\frac{\kappa^2}{b_t}\right)^{\frac{1}{2}},
\label{eq:dt2-avg-final-tight0}
\end{align}
where $\Delta_0^{\scriptscriptstyle Y}:=\Phi(X_0)-f(X_0,Y_0).$
\label{lemmay}
\end{lemma}

\begin{proof}
Let $\sigma_{t+1}:=\frac{\min\{\|\nabla_Y F(X_{t},Y_{t};\xi_{t}^{\scriptscriptstyle Y})\|_F,\gamma\}}{m_{t+1}^{\scriptscriptstyle Y}}$, $\Delta_t^{\scriptscriptstyle Y}:=\Phi(X_t)-f(X_t,Y_t)\ge 0$, and
$\bar F_{t+1}^{\scriptscriptstyle Y}:=M_{t+1}^{\scriptscriptstyle Y}-\nabla_Y f(X_t,Y_t)$.
Then we have
\begin{align}
\langle \nabla_Y f(X_t,Y_t),\,O_{t+1}^{\scriptscriptstyle Y}\rangle
&=\langle M_{t+1}^{\scriptscriptstyle Y}, O_{t+1}^{\scriptscriptstyle Y}\rangle - \langle \bar F_{t+1}^{\scriptscriptstyle Y}, O_{t+1}^{\scriptscriptstyle Y}\rangle \notag\\
&\overset{(a)}{=} \|M_{t+1}^{\scriptscriptstyle Y}\|_* - \langle \bar F_{t+1}^{\scriptscriptstyle Y}, O_{t+1}^{\scriptscriptstyle Y}\rangle \notag\\
&\overset{(b)}{\geq} \|\nabla_Y f(X_t,Y_t)+\bar F_{t+1}^{\scriptscriptstyle Y}\|_* - \|\bar F_{t+1}^{\scriptscriptstyle Y}\|_* \notag\\
&\overset{(c)}{\geq} \|\nabla_Y f(X_t,Y_t)\|_* -2\|\bar F_{t+1}^{\scriptscriptstyle Y}\|_* \notag\\
&\overset{(d)}{\geq} \mu\|Y_t-Y^\star(X_t)\|_F -2\|\bar F_{t+1}^{\scriptscriptstyle Y}\|_*,
\label{eq:align-noVR}
\end{align}
where (a) uses the orthogonalization rule $O_{t+1}^{\scriptscriptstyle Y}=\mathrm{Orth}(M_{t+1}^{\scriptscriptstyle Y})$, defined as the Frobenius projection onto the Stiefel manifold
$\mathrm{St}(m_y,n_y):=\{Q\in\mathbb{R}^{m_y\times n_y}:Q^\top Q=I_{n_y}\}$, namely
\[
O_{t+1}^{\scriptscriptstyle Y}\in\arg\min_{Q\in\mathrm{St}(m_y,n_y)}\|M_{t+1}^{\scriptscriptstyle Y}-Q\|_F^2.
\]
Since $\|Q\|_F^2=\mathrm{tr}(Q^\top Q)=n_y$ for all $Q\in\mathrm{St}(m_y,n_y)$, we have
\[
\|M_{t+1}^{\scriptscriptstyle Y}-Q\|_F^2=\|M_{t+1}^{\scriptscriptstyle Y}\|_F^2+n_y-2\langle M_{t+1}^{\scriptscriptstyle Y},Q\rangle,
\]
and hence the above minimization is equivalent to the following inner-product maximization over $\mathrm{St}(m_y,n_y)$:
\[
O_{t+1}^{\scriptscriptstyle Y}\in\arg\max_{Q\in\mathrm{St}(m_y,n_y)}\langle M_{t+1}^{\scriptscriptstyle Y},Q\rangle.
\]
Consequently,
\[
\langle M_{t+1}^{\scriptscriptstyle Y},O_{t+1}^{\scriptscriptstyle Y}\rangle=\|M_{t+1}^{\scriptscriptstyle Y}\|_*=\|\nabla_Y f(X_t,Y_t)+\bar F_{t+1}^{\scriptscriptstyle Y}\|_*.
\]
Moreover, (b) follows from the Hölder inequality
$|\langle \bar F_{t+1}^{\scriptscriptstyle Y}, O_{t+1}^{\scriptscriptstyle Y}\rangle|\le \|\bar F_{t+1}^{\scriptscriptstyle Y}\|_*\,\|O_{t+1}^{\scriptscriptstyle Y}\|_{\mathrm{op}}$,
together with $\|O_{t+1}^{\scriptscriptstyle Y}\|_{\mathrm{op}}=1$ for $O_{t+1}^{\scriptscriptstyle Y}\in\mathrm{St}(m_y,n_y)$, which yields
\[
\langle \bar F_{t+1}^{\scriptscriptstyle Y}, O_{t+1}^{\scriptscriptstyle Y}\rangle
\le |\langle \bar F_{t+1}^{\scriptscriptstyle Y}, O_{t+1}^{\scriptscriptstyle Y}\rangle|
\le \|\bar F_{t+1}^{\scriptscriptstyle Y}\|_*.
\]
Inequality (c) follows from the reverse triangle inequality for the nuclear norm, and (d) follows from \cref{concavity}, which implies
$\|\nabla_Y f(X_t,Y_t)\|_F\ge \mu\|Y_t-Y^\star(X_t)\|_F$, together with the norm domination
$\|\nabla_Y f(X_t,Y_t)\|_*\ge \|\nabla_Y f(X_t,Y_t)\|_F$.
For notational simplicity, the above argument is presented for the case $m_y\ge n_y$; the complementary case follows by an analogous argument. In either case, 
$\|O_{t+1}^{\scriptscriptstyle Y}\|_F^2=\min\{m_y,n_y\}\le r$ under the global rank bound $r$.
Consequently, by $L$-smoothness, we obtain
\begin{align}
f(X_t,Y_{t+1})
&\ge f(X_t,Y_t)+\max\{\epsilon_{\scriptscriptstyle Y},\eta_{\scriptscriptstyle Y}\sigma_t\}\langle \nabla_Y f(X_t,Y_t),O_{t+1}^{\scriptscriptstyle Y}\rangle-\frac{L}{2}[\max\{\epsilon_{\scriptscriptstyle Y},\eta_{\scriptscriptstyle Y}\sigma_t\}]^2\|O_{t+1}^{\scriptscriptstyle Y}\|_F^2 \notag\\
&= f(X_t,Y_t)+\max\{\epsilon_{\scriptscriptstyle Y},\eta_{\scriptscriptstyle Y}\sigma_t\}\langle \nabla_Y f(X_t,Y_t),O_{t+1}^{\scriptscriptstyle Y}\rangle-\frac{L r}{2}[\max\{\epsilon_{\scriptscriptstyle Y},\eta_{\scriptscriptstyle Y}\sigma_t\}]^2.
\label{eqyy10}
\end{align}
Equivalently, we have
\begin{align}
&\Phi(X_t)-f(X_t,Y_{t+1})\notag\\
&\le
\Phi(X_t)-f(X_t,Y_{t}) - \max\{\epsilon_{\scriptscriptstyle Y},\eta_{\scriptscriptstyle Y}\sigma_t\}\langle \nabla_Y f(X_t,Y_t),O_{t+1}^{\scriptscriptstyle Y}\rangle+\frac{L r}{2}(\max\{\epsilon_{\scriptscriptstyle Y},\eta_{\scriptscriptstyle Y}\sigma_t\})^2.
\label{eq:gap-rec-noVR}
\end{align}
Plugging \cref{eq:align-noVR} into \cref{eq:gap-rec-noVR} yields
\begin{align}
&\max\{\epsilon_{\scriptscriptstyle Y},\eta_{\scriptscriptstyle Y}\sigma_t\}\,\mu\|Y_t-Y^\star(X_t)\|_F\notag\\
&\le
\left(\Phi(X_t)-f(X_t,Y_t)\right)-\left(\Phi(X_t)-f(X_t,Y_{t+1})\right)\notag\\
&\quad+\frac{L r}{2}(\max\{\epsilon_{\scriptscriptstyle Y},\eta_{\scriptscriptstyle Y}\sigma_t\})^2
+2\max\{\epsilon_{\scriptscriptstyle Y},\eta_{\scriptscriptstyle Y}\sigma_t\}\|\bar F_{t+1}^{\scriptscriptstyle Y}\|_* .
\label{eq:dual-linear-fixedx-noVR}
\end{align}

\paragraph{Upper Bound for $\frac{1}{T}\sum_{t=0}^{T-1}\mathbb E[\|Y_t-Y^\star(X_t)\|_F]$.} 
Using $\Delta_t^{\scriptscriptstyle Y}=\Phi(X_t)-f(X_t,Y_t)$ and inserting $\Phi(X_{t+1})-f(X_{t+1},Y_{t+1})$, we obtain
\begin{align}
&\big(\Phi(X_t)-f(X_t,Y_t)\big)-\big(\Phi(X_t)-f(X_t,Y_{t+1})\big)\notag\\
&=
(\Delta_t^{\scriptscriptstyle Y}-\Delta_{t+1}^{\scriptscriptstyle Y})
+\big(\Phi(X_{t+1})-\Phi(X_t)\big)
-\big(f(X_{t+1},Y_{t+1})-f(X_t,Y_{t+1})\big).
\label{eq:gap_decomp}
\end{align}

Summing \cref{eq:dual-linear-fixedx-noVR} for $t=0,\dots,T-1$ and substituting \cref{eq:gap_decomp} yields
\begin{align}
&\mu\sum_{t=0}^{T-1} \max\{\epsilon_{\scriptscriptstyle Y},\eta_{\scriptscriptstyle Y}\sigma_t\}\|Y_t-Y^\star(X_t)\|_F\notag\\
&\le
\sum_{t=0}^{T-1}(\Delta_t^{\scriptscriptstyle Y}-\Delta_{t+1}^{\scriptscriptstyle Y})
+\sum_{t=0}^{T-1}\Big[\Phi(X_{t+1})-\Phi(X_t)-\big(f(X_{t+1},Y_{t+1})-f(X_t,Y_{t+1})\big)\Big]\notag\\
&\quad+\frac{L r}{2}\sum_{t=0}^{T-1} [\max\{\epsilon_{\scriptscriptstyle Y},\eta_{\scriptscriptstyle Y}\sigma_t\}]^2
+2\sum_{t=0}^{T-1} \max\{\epsilon_{\scriptscriptstyle Y},\eta_{\scriptscriptstyle Y}\sigma_t\}\|\bar F_{t+1}^{\scriptscriptstyle Y}\|_* \notag\\
&\le
\Delta_0^{\scriptscriptstyle Y}
+\sum_{t=0}^{T-1}\Big[\Phi(X_{t+1})-\Phi(X_t)-\big(f(X_{t+1},Y_{t+1})-f(X_t,Y_{t+1})\big)\Big]\notag\\
&\quad+\frac{L r}{2}\sum_{t=0}^{T-1} [\max\{\epsilon_{\scriptscriptstyle Y},\eta_{\scriptscriptstyle Y}\sigma_t\}]^2
+2\sum_{t=0}^{T-1} \max\{\epsilon_{\scriptscriptstyle Y},\eta_{\scriptscriptstyle Y}\sigma_t\}\|\bar F_{t+1}^{\scriptscriptstyle Y}\|_* ,
\label{eq:sum_bt_dt1}
\end{align}
where the last inequality uses $-\Delta_T^{\scriptscriptstyle Y}\le 0$.
We then decompose
\begin{align}
&\Phi(X_{t+1})-\Phi(X_t)\notag\\
&=
f(X_{t+1},Y^*(X_{t+1}))-f(X_t,Y^*(X_t)) \notag\\
&=
\underbrace{f(X_{t+1},Y^*(X_t))-f(X_t,Y^*(X_t))}_{(a)}
+
\underbrace{f(X_{t+1},Y^*(X_{t+1}))-f(X_{t+1},Y^*(X_t))}_{(b)}.
\label{eq:phi-split}
\end{align}

By \cref{stochasticgradient}, unbiasedness, and Jensen's
inequality, we have
$\|\nabla_Z f(X_t,Y_t)\|_F\leq G$
for $Z\in\{X,Y\}$ almost surely.
Moreover, strong concavity and
$\nabla_Y f(X_t,Y^\ast(X_t))=0$ imply
\[
\mu\|Y_t-Y^\ast(X_t)\|_F
\leq \|\nabla_Y f(X_t,Y_t)\|_F
\leq G.
\]

For term $(a)$, joint smoothness and the duality between
the nuclear and operator norms give
\begin{align}
&f(X_{t+1},Y^\ast(X_t))
-f(X_t,Y^\ast(X_t))\notag\\
&\leq
\big\langle
\nabla_X f(X_t,Y^\ast(X_t)),X_{t+1}-X_t
\big\rangle
+\frac{L}{2}\|X_{t+1}-X_t\|_{\mathrm{op}}^2\notag\\
&\leq
\|\nabla_X f(X_t,Y^\ast(X_t))\|_\ast \|X_{t+1}-X_t\|_{\mathrm{op}}
+\frac{L}{2}\|X_{t+1}-X_t\|_{\mathrm{op}}^2\notag\\
&\leq
\left(\sqrt{r}G+\frac{LG}{\mu}\right)\|X_{t+1}-X_t\|_{\mathrm{op}}
+\frac{L}{2}\|X_{t+1}-X_t\|_{\mathrm{op}}^2,
\label{eq:term-II}
\end{align}
where the last inequality follows from
\begin{align*}
\|\nabla_X f(X_t,Y^\ast(X_t))\|_\ast
&\leq
\|\nabla_X f(X_t,Y_t)\|_\ast
+\|\nabla_X f(X_t,Y^\ast(X_t))
-\nabla_X f(X_t,Y_t)\|_\ast\\
&\leq
\sqrt{r}G+L\|Y^\ast(X_t)-Y_t\|_{\mathrm{op}}\\
&\leq
\sqrt{r}G+L\|Y^\ast(X_t)-Y_t\|_F\notag\\
&\leq \sqrt{r}G+\frac{LG}{\mu}.
\end{align*}
Here, we use the global rank bound $r$ and
$\|A\|_\ast\leq\sqrt{r}\|A\|_F$.

For term $(b)$, applying smoothness at
$(X_{t+1},Y^\ast(X_{t+1}))$ and using
$\nabla_Y f(X_{t+1},Y^\ast(X_{t+1}))=0$, we obtain
\begin{align}
f(X_{t+1},Y^\ast(X_{t+1}))
-f(X_{t+1},Y^\ast(X_t))
&\leq
\frac{L}{2}
\|Y^\ast(X_{t+1})-Y^\ast(X_t)\|_{\mathrm{op}}^2
\notag\\
&\leq
\frac{L\lambda^2}{2}\|X_{t+1}-X_t\|_{\mathrm{op}}^2,
\label{eq:term-I}
\end{align}
where the last inequality follows from the
$\lambda$-Lipschitz continuity of $Y^\ast(\cdot)$
in \cref{additionallemma}.
Combining \cref{eq:phi-split,eq:term-II,eq:term-I} yields
\begin{equation}
\Phi(X_{t+1})-\Phi(X_t)
\leq
\left(\sqrt{r}G+\frac{LG}{\mu}\right)\|X_{t+1}-X_t\|_{\mathrm{op}}
+\frac{L}{2}(1+\lambda^2)\|X_{t+1}-X_t\|_{\mathrm{op}}^2.
\label{eq:phi-linear}
\end{equation}

Next, joint smoothness implies
\begin{align*}
\|\nabla_X f(X_t,Y_{t+1})\|_\ast
&\leq
\|\nabla_X f(X_t,Y_t)\|_\ast+
\|\nabla_X f(X_t,Y_{t+1})
-\nabla_X f(X_t,Y_t)\|_\ast\\
&\leq \sqrt{r}G+L\|Y_{t+1}-Y_t\|_{\mathrm{op}}.
\end{align*}
Consequently, another application of smoothness gives
\begin{align}
&-\left(
f(X_{t+1},Y_{t+1})-f(X_t,Y_{t+1})
\right)\notag\\
&\leq
-\big\langle
\nabla_X f(X_t,Y_{t+1}),X_{t+1}-X_t
\big\rangle
+\frac{L}{2}\|X_{t+1}-X_t\|_{\mathrm{op}}^2\notag\\
&\leq
\|\nabla_X f(X_t,Y_{t+1})\|_\ast \|X_{t+1}-X_t\|_{\mathrm{op}}
+\frac{L}{2}\|X_{t+1}-X_t\|_{\mathrm{op}}^2\notag\\
&\leq
(\sqrt{r}G+L\|Y_{t+1}-Y_t\|_{\mathrm{op}})\|X_{t+1}-X_t\|_{\mathrm{op}}
+\frac{L}{2}\|X_{t+1}-X_t\|_{\mathrm{op}}^2.
\label{eq:f-linear}
\end{align}

Adding \cref{eq:phi-linear,eq:f-linear}, we conclude that
\begin{align}
&\Phi(X_{t+1})-\Phi(X_t)
-\left(
f(X_{t+1},Y_{t+1})-f(X_t,Y_{t+1})
\right)\notag\\
&\leq
G\sqrt{r}\left(2+\lambda\right)\|X_{t+1}-X_t\|_{\mathrm{op}}
+L\|X_{t+1}-X_t\|_{\mathrm{op}}\|Y_{t+1}-Y_t\|_{\mathrm{op}}\notag\\
&\quad+\frac{L}{2}(2+\lambda^2)\|X_{t+1}-X_t\|_{\mathrm{op}}^2.
\label{eq:drift-linear}
\end{align}

Since $\max\{\epsilon_{\scriptscriptstyle Y},\eta_{\scriptscriptstyle Y}\sigma_t\}\ge \epsilon_{\scriptscriptstyle Y}>0$, we have $\sum_{t=0}^{T-1} \max\{\epsilon_{\scriptscriptstyle Y},\eta_{\scriptscriptstyle Y}\sigma_t\}\|Y_t-Y^\star(X_t)\|_F \ge \epsilon_{\scriptscriptstyle Y}\sum_{t=0}^{T-1}\|Y_t-Y^\star(X_t)\|_F$.
Taking expectation in \cref{eq:sum_bt_dt1} and dividing both sides by $T\mu\epsilon_{\scriptscriptstyle Y}$ yields
\begin{align}
&\frac{1}{T}\sum_{t=0}^{T-1}\mathbb E[\|Y_t-Y^\star(X_t)\|_F]\notag\\
&\le
\frac{\Delta_0^{\scriptscriptstyle Y}}{T\mu\epsilon_{\scriptscriptstyle Y}}
+\frac{G\sqrt{r}(2+\lambda)}{T\mu\epsilon_{\scriptscriptstyle Y}}\sum_{t=0}^{T-1}\mathbb{E}\|X_{t+1}-X_t\|_{\mathrm{op}}+
\frac{L}{T\mu\epsilon_Y}
\sum_{t=0}^{T-1}\mathbb E[\|X_{t+1}-X_t\|_{\mathrm{op}}\|Y_{t+1}-Y_t\|_{\mathrm{op}}]
\notag\\
&\quad+
\frac{L(2+\lambda^2)}{2T\mu\epsilon_Y}
\sum_{t=0}^{T-1}\mathbb E[\|X_{t+1}-X_t\|_{\mathrm{op}}^2]+\frac{L r}{2T\mu\epsilon_{\scriptscriptstyle Y}}\sum_{t=0}^{T-1}\mathbb{E}[\left(\max\{\epsilon_{\scriptscriptstyle Y},\eta_{\scriptscriptstyle Y}\sigma_t\}\right)^2] \notag\\
&\quad+\frac{2}{T\mu\epsilon_{\scriptscriptstyle Y}}\sum_{t=0}^{T-1}\mathbb{E}\big[\max\{\epsilon_{\scriptscriptstyle Y},\eta_{\scriptscriptstyle Y}\sigma_t\}\|\bar F_{t+1}^{\scriptscriptstyle Y}\|_*\big]\notag\\
&\le
\frac{\Delta_0^{\scriptscriptstyle Y}}{T\mu\epsilon_{\scriptscriptstyle Y}}
+\frac{G\sqrt{r}(2+\lambda)}{T\mu\epsilon_{\scriptscriptstyle Y}}\sum_{t=0}^{T-1}\mathbb{E}[\max\{\epsilon_{\scriptscriptstyle X},\eta_{\scriptscriptstyle X}\theta_t\}]+
\frac{L}{T\mu\epsilon_Y}
\sum_{t=0}^{T-1}\mathbb E[\max\{\epsilon_{\scriptscriptstyle X},\eta_{\scriptscriptstyle X}\theta_t\}\max\{\epsilon_{\scriptscriptstyle Y},\eta_{\scriptscriptstyle Y}\sigma_t\}]\notag\\
&\quad+
\frac{L(2+\lambda^2)}{2T\mu\epsilon_Y}
\sum_{t=0}^{T-1}\mathbb E[\left(\max\{\epsilon_{\scriptscriptstyle X},\eta_{\scriptscriptstyle X}\theta_t\}\right)^2]+\frac{L r}{2T\mu\epsilon_{\scriptscriptstyle Y}}\sum_{t=0}^{T-1}\mathbb{E}[\left(\max\{\epsilon_{\scriptscriptstyle Y},\eta_{\scriptscriptstyle Y}\sigma_t\}\right)^2]\notag\\
&\quad+\frac{2\sqrt{r}}{\mu\epsilon_{\scriptscriptstyle Y}}\sqrt{\frac{1}{T}\sum_{t=0}^{T-1}\mathbb E\left[(\max\{\epsilon_{\scriptscriptstyle Y},\eta_{\scriptscriptstyle Y}\sigma_t\})^2\right]}\,
\sqrt{\frac{1}{T}\sum_{t=0}^{T-1}\mathbb E[\|\bar F_{t+1}^{\scriptscriptstyle Y}\|_F^2]},
\label{firstordery}
\end{align}
where the last inequality follows by applying the Cauchy--Schwarz inequality, together with the bound $\|A\|_* \le \sqrt r\,\|A\|_F$.

\paragraph{Upper Bound for $\frac{1}{T}\sum_{t=0}^{T-1}\mathbb E[\|Y_t-Y^\star(X_t)\|_F^2]$.} According to \cref{concavity,stochasticgradient}, we have
\begin{equation}
\mu\|Y_t-Y^\star(X_t)\|_F \le \|\nabla_Y f(X_t,Y_t)\|_F \le G.
\label{eq:d-bdd}
\end{equation}
Then we have $\mu\|Y_t-Y^\star(X_t)\|_F^2 \le G\|Y_t-Y^\star(X_t)\|_F.$ Since $\max\{\epsilon_{\scriptscriptstyle Y},\eta_{\scriptscriptstyle Y}\sigma_t\}\geq\epsilon_{\scriptscriptstyle Y}>0$, we can get
\begin{equation}
\mu\|Y_t-Y^\star(X_t)\|_F^2\leq G\|Y_t-Y^\star(X_t)\|_F
\le
\frac{G}{\mu\epsilon_{\scriptscriptstyle Y}}\big( \max\{\epsilon_{\scriptscriptstyle Y},\eta_{\scriptscriptstyle Y}\sigma_t\}\mu\|Y_t-Y^\star(X_t)\|_F\big).
\label{eq:d-to-bd}
\end{equation}

Combining \cref{eq:dual-linear-fixedx-noVR} and \cref{eq:d-to-bd}, we have
\begin{align}
&\mu\|Y_t-Y^\star(X_t)\|_F^2\notag\\
&\le
\frac{G}{\mu\epsilon_{\scriptscriptstyle Y}}\left[\left(\Phi(X_t)-f(X_t,Y_t)\right)-\left(\Phi(X_t)-f(X_t,Y_{t+1})\right)\right]
+\frac{G L r}{2\mu\epsilon_{\scriptscriptstyle Y}}(\max\{\epsilon_{\scriptscriptstyle Y},\eta_{\scriptscriptstyle Y}\sigma_t\})^2\notag\\
&\quad+\frac{2G}{\mu\epsilon_{\scriptscriptstyle Y}}\max\{\epsilon_{\scriptscriptstyle Y},\eta_{\scriptscriptstyle Y}\sigma_t\}\|\bar F_{t+1}^{\scriptscriptstyle Y}\|_* .
\label{eq:dt2-perstep-noVR}
\end{align}

Taking expectation and summing $t=0,\dots,T-1$ gives
\begin{align}
&\mu\sum_{t=0}^{T-1}\mathbb E[\|Y_t-Y^\star(X_t)\|_F^2]\notag\\
&\le
\frac{G}{\mu\epsilon_{\scriptscriptstyle Y}}\sum_{t=0}^{T-1}\mathbb E\!\left[\left(\Phi(X_t)-f(X_t,Y_t)\right)-\left(\Phi(X_t)-f(X_t,Y_{t+1})\right)\right]
+\frac{G L r}{2\mu\epsilon_{\scriptscriptstyle Y}}\sum_{t=0}^{T-1}\mathbb E\left[(\max\{\epsilon_{\scriptscriptstyle Y},\eta_{\scriptscriptstyle Y}\sigma_t\})^2\right]\notag\\
&\quad+\frac{2G}{\mu\epsilon_{\scriptscriptstyle Y}}\sum_{t=0}^{T-1}\mathbb E\ \left[\max\{\epsilon_{\scriptscriptstyle Y},\eta_{\scriptscriptstyle Y}\sigma_t\}\|\bar F_{t+1}^{\scriptscriptstyle Y}\|_*\right].
\label{eq:dt2-sum-skeleton-noVR}
\end{align}

Since
\begin{align}
&\left(\Phi(X_t)-f(X_t,Y_t)\right)-\left(\Phi(X_t)-f(X_t,Y_{t+1})\right)\notag\\
&=(\Delta_t^{\scriptscriptstyle Y}-\Delta_{t+1}^{\scriptscriptstyle Y})+\big(\Phi(X_{t+1})-\Phi(X_t)\big)
-\big(f(X_{t+1},Y_{t+1})-f(X_t,Y_{t+1})\big),
\label{eqyy15}
\end{align}
then
\begin{align}
&\sum_{t=0}^{T-1}\left[\left(\Phi(X_t)-f(X_t,Y_t)\right)-\left(\Phi(X_t)-f(X_t,Y_{t+1})\right)\right]\notag\\
&=
\Delta_0^{\scriptscriptstyle Y}-\Delta_T^{\scriptscriptstyle Y}+\sum_{t=0}^{T-1}\left[\big(\Phi(X_{t+1})-\Phi(X_t)\big)
-\big(f(X_{t+1},Y_{t+1})-f(X_t,Y_{t+1})\big)\right]\notag\\
&\overset{(a)}{\le}
\Delta_0^{\scriptscriptstyle Y}+\sum_{t=0}^{T-1}\left[\big(\Phi(X_{t+1})-\Phi(X_t)\big)
-\big(f(X_{t+1},Y_{t+1})-f(X_t,Y_{t+1})\big)\right],
\label{eq:telescope-drift-noVR}
\end{align}
where (a) is due to $\Delta_T^{\scriptscriptstyle Y}\geq0$. By smoothness of $\Phi$ and $f(\cdot,Y)$, we have
\begin{align}
&\big(\Phi(X_{t+1})-\Phi(X_t)\big)
-
\big(f(X_{t+1},Y_{t+1})-f(X_t,Y_{t+1})\big)\notag\\
&\le
\langle \nabla\Phi(X_t)-\nabla_X f(X_t,Y_{t+1}),X_{t+1}-X_t\rangle
+\frac{L_\Phi+L}{2}\|X_{t+1}-X_t\|_F^2 \notag\\
&\le
\|\nabla\Phi(X_t)-\nabla_X f(X_t,Y_{t+1})\|_F\,\|X_{t+1}-X_t\|_F
+\frac{L_\Phi+L}{2}\|X_{t+1}-X_t\|_F^2 \notag\\
&\le
L\|Y_{t+1}-Y^\star(X_t)\|_F\,\|X_{t+1}-X_t\|_F
+\frac{L_\Phi+L}{2}\|X_{t+1}-X_t\|_F^2 \notag\\
&\le
L\big(\|Y_t-Y^\star(X_t)\|_F+\max\{\epsilon_{\scriptscriptstyle Y},\eta_{\scriptscriptstyle Y}\sigma_t\}\|O_{t+1}^{\scriptscriptstyle Y}\|_F\big)\|X_{t+1}-X_t\|_F
+\frac{L_\Phi+L}{2}\|X_{t+1}-X_t\|_F^2 \notag\\
&\overset{(a)}{\leq}
L\|Y_t-Y^\star(X_t)\|_F\|X_{t+1}-X_t\|_F
+L\sqrt r\,\max\{\epsilon_{\scriptscriptstyle Y},\eta_{\scriptscriptstyle Y}\sigma_t\}\|X_{t+1}-X_t\|_F\notag\\
&\quad+\frac{L_\Phi+L}{2}\|X_{t+1}-X_t\|_F^2\notag\\
&\overset{(b)}{\leq}\frac{\epsilon_{\scriptscriptstyle Y}\mu^2}{8G}\|Y_t-Y^\star(X_t)\|_F^2
+\frac{\mu}{8}(\max\{\epsilon_{\scriptscriptstyle Y},\eta_{\scriptscriptstyle Y}\sigma_t\})^2\notag\\
&\quad+\left(\frac{L_\Phi+L}{2}+\frac{2rL^2}{\mu}+\frac{2L^2G}{\epsilon_{\scriptscriptstyle Y}\mu^2}\right)\|X_{t+1}-X_t\|_F^2.
\label{eq:drift-bound-noVR}
\end{align}
where (a) used $\|O_{t+1}^{\scriptscriptstyle Y}\|_F\le \sqrt r$ under the global rank bound $r$ and (b) comes from the Young's inequality:
$L\|Y_t-Y^\star(X_t)\|_F\|X_{t+1}-X_t\|_F
\le
\frac{\epsilon^{}_{\scriptscriptstyle Y}\mu^2}{8G}\|Y_t-Y^\star(X_t)\|_F^2+\frac{2L^2G}{\epsilon^{}_{\scriptscriptstyle Y}\mu^2}\|X_{t+1}-X_t\|_F^2$ and
$L\sqrt r\,\max\{\epsilon_{\scriptscriptstyle Y},\eta_{\scriptscriptstyle Y}\sigma_t\}\|X_{t+1}-X_t\|
\le
\frac{\mu}{8}(\max\{\epsilon_{\scriptscriptstyle Y},\eta_{\scriptscriptstyle Y}\sigma_t\})^2+\frac{2rL^2}{\mu}\|X_{t+1}-X_t\|^2_F.$

Combining \cref{eq:telescope-drift-noVR} and \cref{eq:drift-bound-noVR} gives
\begin{align}
&\sum_{t=0}^{T-1}\left[\left(\Phi(X_t)-f(X_t,Y_t)\right)-\left(\Phi(X_t)-f(X_t,Y_{t+1})\right)\right]\notag\\
&\le
\Delta_0^{\scriptscriptstyle Y}+\sum_{t=0}^{T-1}\mathbb E\left[\big(\Phi(X_{t+1})-\Phi(X_t)\big)
-
\big(f(X_{t+1},Y_{t+1})-f(X_t,Y_{t+1})\big)\right] \notag\\
&\le
\Delta_0^{\scriptscriptstyle Y}
+\frac{\epsilon_{\scriptscriptstyle Y}\mu^2}{8G}\sum_{t=0}^{T-1}\mathbb E[\|Y_{t}-Y^\star(X_t)\|_F^2]
+\frac{\mu}{8}\sum_{t=0}^{T-1}\mathbb E\left[(\max\{\epsilon_{\scriptscriptstyle Y},\eta_{\scriptscriptstyle Y}\sigma_t\})^2\right]
\notag\\
&\quad+\left(\frac{L_\Phi+L}{2}+\frac{2rL^2}{\mu}+\frac{2L^2G}{\epsilon_{\scriptscriptstyle Y}\mu^2}\right)\sum_{t=0}^{T-1}\mathbb E[\|X_{t+1}-X_t\|_F^2].
\label{eq:gapdiff-sum-noVR}
\end{align}

Insert \cref{eq:gapdiff-sum-noVR} into \cref{eq:dt2-sum-skeleton-noVR} to obtain
\begin{align}
&\mu\sum_{t=0}^{T-1}\mathbb E[\|Y_{t}-Y^\star(X_t)\|_F^2]\notag\\
&\le
\frac{G\Delta_0^{\scriptscriptstyle Y}}{\mu\epsilon_{\scriptscriptstyle Y}}
+\frac{\mu}{8}\sum_{t=0}^{T-1}\mathbb E[\|Y_{t}-Y^\star(X_t)\|_F^2]
+\left(\frac{G}{8\epsilon_{\scriptscriptstyle Y}}+\frac{G L r}{2\mu\epsilon_{\scriptscriptstyle Y}}\right)\sum_{t=0}^{T-1}\mathbb E\left[(\max\{\epsilon_{\scriptscriptstyle Y},\eta_{\scriptscriptstyle Y}\sigma_t\})^2\right]\notag\\
&\quad+\left(\frac{G(L_\Phi+L)}{2\mu\epsilon_{\scriptscriptstyle Y}}+\frac{2rL^2G}{\mu^2\epsilon_{\scriptscriptstyle Y}}+\frac{2G^2L^2}{\mu^3\epsilon_{\scriptscriptstyle Y}^2}\right)\sum_{t=0}^{T-1}\mathbb E[\|X_{t+1}-X_t\|_F^2]\notag\\
&\quad+\frac{2G}{\mu\epsilon_{\scriptscriptstyle Y}}\sum_{t=0}^{T-1}\mathbb E\ \left[\max\{\epsilon_{\scriptscriptstyle Y},\eta_{\scriptscriptstyle Y}\sigma_t\}\|\bar F_{t+1}^{\scriptscriptstyle Y}\|_*\right].
\label{eq:dt2-sum-final-noVR0}
\end{align}
Then we have
\begin{align}
&\sum_{t=0}^{T-1}\mathbb E[\|Y_{t}-Y^\star(X_t)\|_F^2]\notag\\
&\le
\frac{8 G\Delta_0^{\scriptscriptstyle Y}}{7\mu^2\epsilon_{\scriptscriptstyle Y}}
+\left(\frac{G}{7\mu\epsilon_{\scriptscriptstyle Y}}+\frac{4G L r}{7\mu^2\epsilon_{\scriptscriptstyle Y}}\right)\sum_{t=0}^{T-1}\mathbb E\left[(\max\{\epsilon_{\scriptscriptstyle Y},\eta_{\scriptscriptstyle Y}\sigma_t\})^2\right]\notag\\
&\quad+\left(\frac{4G(L_\Phi+L)}{7\mu^2\epsilon_{\scriptscriptstyle Y}}+\frac{16rL^2G}{7\mu^3\epsilon_{\scriptscriptstyle Y}}+\frac{16G^2L^2}{7\mu^4\epsilon_{\scriptscriptstyle Y}^2}\right)\sum_{t=0}^{T-1}\mathbb E[\|X_{t+1}-X_t\|_F^2]\notag\\
&\quad+\frac{16G}{7\mu^2\epsilon_{\scriptscriptstyle Y}}\sum_{t=0}^{T-1}\mathbb E\ \left[\max\{\epsilon_{\scriptscriptstyle Y},\eta_{\scriptscriptstyle Y}\sigma_t\}\|\bar F_{t+1}^{\scriptscriptstyle Y}\|_*\right].
\label{eq:dt2-sum-final-noVR}
\end{align}

Dividing by $T$ and using Cauchy--Schwarz with $\|A\|_*\le \sqrt r\,\|A\|_F$ yields
\begin{align}
&\frac{1}{T}\sum_{t=0}^{T-1}\mathbb E[\|Y_t-Y^\star(X_t)\|_F^2]\notag\\
&\le
\frac{8 G\Delta_0^{\scriptscriptstyle Y}}{7\mu^2\epsilon_{\scriptscriptstyle Y}T}
+\left(\frac{G}{7\mu\epsilon_{\scriptscriptstyle Y}}+\frac{4G L r}{7\mu^2\epsilon_{\scriptscriptstyle Y}}\right)\frac{1}{T}\sum_{t=0}^{T-1}\mathbb E\left[(\max\{\epsilon_{\scriptscriptstyle Y},\eta_{\scriptscriptstyle Y}\sigma_t\})^2\right]
\notag\\
&\quad+\left(\frac{4G(L_\Phi+L)}{7\mu^2\epsilon_{\scriptscriptstyle Y}}+\frac{16rL^2G}{7\mu^3\epsilon_{\scriptscriptstyle Y}}+\frac{16G^2L^2}{7\mu^4\epsilon_{\scriptscriptstyle Y}^2}\right)\frac{1}{T}\sum_{t=0}^{T-1}\mathbb E[\|X_{t+1}-X_t\|_F^2]\notag\\
&\quad+\frac{16G}{7T\mu^2\epsilon_{\scriptscriptstyle Y}}\sum_{t=0}^{T-1}\mathbb E\ \left[\max\{\epsilon_{\scriptscriptstyle Y},\eta_{\scriptscriptstyle Y}\sigma_t\}\|\bar F_{t+1}^{\scriptscriptstyle Y}\|_*\right]\notag\\
&\le
\frac{8 G\Delta_0^{\scriptscriptstyle Y}}{7\mu^2\epsilon_{\scriptscriptstyle Y}T}
+\left(\frac{G}{7\mu\epsilon_{\scriptscriptstyle Y}}+\frac{4G L r}{7\mu^2\epsilon_{\scriptscriptstyle Y}}\right)\frac{1}{T}\sum_{t=0}^{T-1}\mathbb E\left[(\max\{\epsilon_{\scriptscriptstyle Y},\eta_{\scriptscriptstyle Y}\sigma_t\})^2\right]
\notag\\
&\quad+\left(\frac{4Gr(L_\Phi+L)}{7\mu^2\epsilon_{\scriptscriptstyle Y}}+\frac{16r^2L^2G}{7\mu^3\epsilon_{\scriptscriptstyle Y}}+\frac{16rG^2L^2}{7\mu^4\epsilon_{\scriptscriptstyle Y}^2}\right)\frac{1}{T}\sum_{t=0}^{T-1}\mathbb E\left[(\max\{\epsilon_{\scriptscriptstyle X},\eta_{\scriptscriptstyle X}\theta_t\})^2\right]\notag\\
&\quad+\frac{16G\sqrt r}{7\mu^2\epsilon_{\scriptscriptstyle Y}}
\sqrt{\frac{1}{T}\sum_{t=0}^{T-1}\mathbb E\left[(\max\{\epsilon_{\scriptscriptstyle Y},\eta_{\scriptscriptstyle Y}\sigma_t\})^2\right]}\,
\sqrt{\frac{1}{T}\sum_{t=0}^{T-1}\mathbb E[\|\bar F_{t+1}^{\scriptscriptstyle Y}\|_F^2]}.
\label{eq:dual-avg-dt2-final-noVR}
\end{align}

To proceed with the derivations following \cref{firstordery} and \cref{eq:dual-avg-dt2-final-noVR}, we will need an upper bound on $\frac{1}{T}\sum_{t=0}^{T-1}\mathbb E[\|\bar F_{t+1}^{\scriptscriptstyle Y}\|_F^2]$. We therefore establish such a bound in the following.

\paragraph{Upper Bound for $\frac{1}{T}\sum_{t=0}^{T-1}\mathbb E[\|\bar F_{t+1}^{\scriptscriptstyle Y}\|_F^2]$.}
We have $\bar F_{t+1}^{\scriptscriptstyle Y}=M_{t+1}^{\scriptscriptstyle Y}-\nabla_Y f(X_t,Y_t)$ satisfies
\begin{align}\label{eq:Fbar_rec}
\bar F_{t+1}^{\scriptscriptstyle Y}= &\beta (M_{t}^{\scriptscriptstyle Y}-\nabla_Y f(X_{t-1},Y_{t-1})) + \beta(\nabla_Y f(X_{t-1},Y_{t-1})-\nabla_Y f(X_t,Y_t)) \notag\\
&+ (1-\beta)(\nabla_Y F(X_{t}, Y_{t}; \xi^{\scriptscriptstyle Y}_{t})-\nabla_Y f(X_t,Y_t)).
\end{align}
Unrolling \cref{eq:Fbar_rec} yields
\begin{align}\label{eq:Fbar_unroll}
\bar F_{t+1}^{\scriptscriptstyle Y}
=&\beta^{t}\bar F_1^{\scriptscriptstyle Y}+\sum_{\tau=0}^{t-1}\beta^{\tau}\Big(\beta \big(\nabla_Y f(X_{t-\tau-1},Y_{t-\tau-1})-\nabla_Y f(X_{t-\tau},Y_{t-\tau})\big)\notag\\
&+(1-\beta)\big(\nabla_Y F(X_{t-\tau}, Y_{t-\tau}; \xi^{\scriptscriptstyle Y}_{t-\tau})-\nabla_Y f(X_{t-\tau},Y_{t-\tau})\big)\Big).
\end{align}
Using $\|A+B+C\|_F^2\le 3(\|A\|_F^2+\|B\|_F^2+\|C\|_F^2)$, the weighted Cauchy--Schwarz inequality
\(
\Big\|\sum_{\tau=0}^{t-1}\beta^\tau V_\tau\Big\|_F^2
\le \Big(\sum_{\tau=0}^{t-1}\beta^\tau\Big)\Big(\sum_{\tau=0}^{t-1}\beta^\tau\|V_\tau\|_F^2\Big)
\le \frac{1}{1-\beta}\sum_{\tau=0}^{t-1}\beta^\tau\|V_\tau\|_F^2,
\)
and the mini-batch sampling property $\mathbb E[\nabla_Y F(X_t,Y_t;\xi_t^{\scriptscriptstyle Y})]=\nabla_Y f(X_t,Y_t)$ with independent draws across $t$, we obtain for any $t\ge 0$,
\begin{align}
\!\!\mathbb E\|\bar F_{t+1}^{\scriptscriptstyle Y}\|_F^2
&\le 3\beta^{2t}\mathbb E\|\bar F_1^{\scriptscriptstyle Y}\|_F^2
\!+\!\frac{3\beta^2}{1-\beta}\sum_{\tau=0}^{t-1}\beta^{\tau}\,
\mathbb E\!\left[\|\nabla_Y f(X_{t-\tau-1},Y_{t-\tau-1})-\nabla_Y f(X_{t-\tau},Y_{t-\tau})\|_F^2\right]\notag\\
&\quad+3(1-\beta)^2\sum_{\tau=0}^{t-1}\beta^{2\tau}\,
\mathbb E\!\left[\|\nabla_Y F(X_{t-\tau}, Y_{t-\tau}; \xi^{\scriptscriptstyle Y}_{t-\tau})-\nabla_Y f(X_{t-\tau},Y_{t-\tau})\|_F^2\right].
\label{eq:Fbar_split_weighted_single_t_minibatch}
\end{align}
Summing \eqref{eq:Fbar_split_weighted_single_t_minibatch} over $t=0,\ldots,T-1$ and dividing by $T$, we use the bound $\sum_{t=0}^{T-1}\beta^{2t}\le \frac{1}{1-\beta^2}$ and the geometric-series rearrangement
\(
\sum_{t=0}^{T-1}\sum_{\tau=0}^{t-1}\beta^\tau a_{t-\tau}
=\sum_{s=1}^{T-1}\Big(\sum_{j=0}^{T-1-s}\beta^j\Big)a_s
\le \frac{1}{1-\beta}\sum_{s=1}^{T-1}a_s,
\)
and similarly,
\(
\sum_{t=0}^{T-1}\sum_{\tau=0}^{t-1}\beta^{2\tau} a_{t-\tau}
=\sum_{s=1}^{T-1}\Big(\sum_{j=0}^{T-1-s}\beta^{2j}\Big)a_s
\le \frac{1}{1-\beta^2}\sum_{s=1}^{T-1}a_s.
\)
This yields
\begin{align}
&\frac{1}{T}\sum_{t=0}^{T-1}\mathbb E\|\bar F_{t+1}^{\scriptscriptstyle Y}\|_F^2\notag\\
&\le \frac{3}{T(1-\beta^2)}\mathbb E\|\bar F_1^{\scriptscriptstyle Y}\|_F^2
+\frac{3\beta^2}{T(1-\beta)^2}\sum_{t=1}^{T-1}\mathbb E\!\left[\|\nabla_Y f(X_{t-1},Y_{t-1})-\nabla_Y f(X_t,Y_t)\|_F^2\right]\notag\\
&\quad+\frac{3(1-\beta)^2}{T(1-\beta^2)}\sum_{t=0}^{T-1}\mathbb E\!\left[\|\nabla_Y F(X_t,Y_t;\xi_t^{\scriptscriptstyle Y})-\nabla_Y f(X_t,Y_t)\|_F^2\right]\notag\\
&= \frac{3}{T(1-\beta^2)}\mathbb E\|\bar F_1^{\scriptscriptstyle Y}\|_F^2
+\frac{3\beta^2}{T(1-\beta)^2}\sum_{t=1}^{T-1}\mathbb E\!\left[\|\nabla_Y f(X_{t-1},Y_{t-1})-\nabla_Y f(X_t,Y_t)\|_F^2\right]\notag\\
&\quad+\frac{3(1-\beta)}{T(1+\beta)}\sum_{t=0}^{T-1}\mathbb E\!\left[\|\nabla_Y F(X_t,Y_t;\xi_t^{\scriptscriptstyle Y})-\nabla_Y f(X_t,Y_t)\|_F^2\right]\notag\\
&\overset{(a)}{\le} \frac{3}{T(1-\beta^2)}\mathbb E\|\bar F_1^{\scriptscriptstyle Y}\|_F^2
+\frac{6\beta^2L^2}{T(1-\beta)^2}\sum_{t=1}^{T-1}\mathbb E\!\left[\|X_t-X_{t-1}\|_F^2+\|Y_t-Y_{t-1}\|_F^2\right]
+\frac{3(1-\beta)}{T(1+\beta)}\sum_{t=0}^{T-1}\frac{\kappa^2}{b_t}\notag\\
&\overset{(b)}{\le}
\frac{3\kappa^2}{Tb_1(1-\beta^2)}
+\frac{3(1-\beta)}{T(1+\beta)}\sum_{t=0}^{T-1}\frac{\kappa^2}{b_t}\notag\\
&\quad+\frac{6\beta^2L^2r}{(1-\beta)^2}\left(\frac{1}{T}\sum_{t=1}^{T-1}\mathbb E\!\left[(\max\{\epsilon_{\scriptscriptstyle X},\eta_{\scriptscriptstyle X}\theta_t\})^2\right]
+\frac{1}{T}\sum_{t=1}^{T-1}\mathbb E\!\left[(\max\{\epsilon_{\scriptscriptstyle Y},\eta_{\scriptscriptstyle Y}\sigma_t\})^2\right]\right)\notag\\
&\overset{(c)}{\le}
\frac{3\kappa^2}{Tb_1(1-\beta)}
+\frac{3(1-\beta)}{T(1+\beta)}\sum_{t=0}^{T-1}\frac{\kappa^2}{b_t}\notag\\
&\quad +\frac{6L^2r}{(1-\beta)^2}\left(
\epsilon_{\scriptscriptstyle X}^2+\epsilon_{\scriptscriptstyle Y}^2
+\frac{\eta_{\scriptscriptstyle X}^2}{T}\ln\!\left(\frac{\gamma^2T}{[\max\{m_0^{\scriptscriptstyle X},m_0^{\scriptscriptstyle Y}\}]^2}+1\right)
+\frac{\eta_{\scriptscriptstyle Y}^2}{T}\ln\!\left(\frac{\gamma^2T}{[m_0^{\scriptscriptstyle Y}]^2}+1\right)\right),
\label{eq:Fbar-avg-noVR-compact-weighted}
\end{align}
where (a) follows from \cref{smoothness}; (b) follows from \cref{stochasticgradient} and the update rules
$X_t-X_{t-1}=\max\{\epsilon_{\scriptscriptstyle X},\eta_{\scriptscriptstyle X}\theta_t\}O_t^{\scriptscriptstyle X}$ and
$Y_t-Y_{t-1}=\max\{\epsilon_{\scriptscriptstyle Y},\eta_{\scriptscriptstyle Y}\sigma_t\}O_t^{\scriptscriptstyle Y}$, together with the bounds
$\|O_t^{\scriptscriptstyle X}\|_F^2\le r$ and $\|O_t^{\scriptscriptstyle Y}\|_F^2\le r$, where $r$ denotes the global rank upper bound as defined in \cref{problemmodel}; and (c) uses \cref{lem:log_cum_ratio}.

By substituting \cref{eq:Fbar-avg-noVR-compact-weighted} into \cref{firstordery}, we obtain
\begin{align}
&\frac{1}{T}\sum_{t=0}^{T-1}\mathbb E[\|Y_t-Y^\star(X_t)\|_F]\notag\\
&\le
\frac{\Delta_0^{\scriptscriptstyle Y}}{T\mu\epsilon_{\scriptscriptstyle Y}}
+\frac{G\sqrt{r}(2+\lambda)}{T\mu\epsilon_{\scriptscriptstyle Y}}\sum_{t=0}^{T-1}\mathbb{E}[\max\{\epsilon_{\scriptscriptstyle X},\eta_{\scriptscriptstyle X}\theta_t\}]+
\frac{L}{T\mu\epsilon_Y}
\sum_{t=0}^{T-1}\mathbb E[\max\{\epsilon_{\scriptscriptstyle X},\eta_{\scriptscriptstyle X}\theta_t\}\max\{\epsilon_{\scriptscriptstyle Y},\eta_{\scriptscriptstyle Y}\sigma_t\}]\notag\\
&\quad+
\frac{L(2+\lambda^2)}{2T\mu\epsilon_Y}
\sum_{t=0}^{T-1}\mathbb E[\left(\max\{\epsilon_{\scriptscriptstyle X},\eta_{\scriptscriptstyle X}\theta_t\}\right)^2]+\frac{L r}{2T\mu\epsilon_{\scriptscriptstyle Y}}\sum_{t=0}^{T-1}\mathbb{E}[\left(\max\{\epsilon_{\scriptscriptstyle Y},\eta_{\scriptscriptstyle Y}\sigma_t\}\right)^2]\notag\\
&\quad+\frac{2\sqrt{r}}{\mu\epsilon_{\scriptscriptstyle Y}}\sqrt{\frac{1}{T}\sum_{t=0}^{T-1}\mathbb E\left[(\max\{\epsilon_{\scriptscriptstyle Y},\eta_{\scriptscriptstyle Y}\sigma_t\})^2\right]}\,
\sqrt{\frac{1}{T}\sum_{t=0}^{T-1}\mathbb E[\|\bar F_{t+1}^{\scriptscriptstyle Y}\|_F^2]}\notag\\
&\le
\frac{\Delta_0^{\scriptscriptstyle Y}}{T\mu\epsilon_{\scriptscriptstyle Y}}
+\frac{G\epsilon_{\scriptscriptstyle X}\sqrt{r}(2+\lambda)}{\mu\epsilon_{\scriptscriptstyle Y}}+\frac{G\eta_{\scriptscriptstyle X}\sqrt{r}(2+\lambda)}{\mu\epsilon_{\scriptscriptstyle Y}\sqrt{T}}\sqrt{\ln\left(\frac{\gamma^2 T}{\left[\max\{m_0^{\scriptscriptstyle X},m_0^{\scriptscriptstyle Y}\}\right]^2}\right) + 1}\notag\\ 
&\quad+\frac{L r\epsilon_{\scriptscriptstyle Y}}{2\mu}+\frac{L r\eta_{\scriptscriptstyle Y}^2}{2T\mu\epsilon_{\scriptscriptstyle Y}}\ln\!\left(\frac{\gamma^2 T}{[m_0^{\scriptscriptstyle Y}]^2}+1\right)+\frac{L\epsilon_X^2(2+\lambda^2)}{2\mu\epsilon_Y}+\frac{L\eta_X^2(2+\lambda^2)}{2T\mu\epsilon_Y}
\ln\left(\frac{\gamma^2T}{\left[\max\{m_0^{\scriptscriptstyle X},m_0^{\scriptscriptstyle Y}\}\right]^2}+1\right)\notag\\
&\quad+\frac{L}{\mu\epsilon_Y}
\sqrt{\left(\epsilon_X^2+\frac{\eta_X^2}{T}\ln\left(\frac{\gamma^2T}{\left[\max\{m_0^{\scriptscriptstyle X},m_0^{\scriptscriptstyle Y}\}\right]^2}+1\right)\right)
\left(\epsilon_Y^2+\frac{\eta_Y^2}{T}\ln\left(\frac{\gamma^2T}{\left[m_0^{\scriptscriptstyle Y}\right]^2}+1\right)\right)}\notag\\
&\quad+\frac{2\sqrt{r}}{\mu\epsilon_{\scriptscriptstyle Y}}\sqrt{\frac{1}{T}\sum_{t=0}^{T-1}\mathbb E\left[(\max\{\epsilon_{\scriptscriptstyle Y},\eta_{\scriptscriptstyle Y}\sigma_t\})^2\right]}\,
\sqrt{\frac{1}{T}\sum_{t=0}^{T-1}\mathbb E[\|\bar F_{t+1}^{\scriptscriptstyle Y}\|_F^2]}\notag\\
&\le
\frac{\Delta_0^{\scriptscriptstyle Y}}{T\mu\epsilon_{\scriptscriptstyle Y}}
+\frac{G\epsilon_{\scriptscriptstyle X}\sqrt{r}(2+\lambda)}{\mu\epsilon_{\scriptscriptstyle Y}}+\frac{G\eta_{\scriptscriptstyle X}\sqrt{r}(2+\lambda)}{\mu\epsilon_{\scriptscriptstyle Y}\sqrt{T}}\sqrt{\ln\left(\frac{\gamma^2 T}{\left[\max\{m_0^{\scriptscriptstyle X},m_0^{\scriptscriptstyle Y}\}\right]^2}\right) + 1}\notag\\
&\quad+\frac{L\epsilon_X^2(2+\lambda^2)}{2\mu\epsilon_Y}+\frac{L\eta_X^2(2+\lambda^2)}{2T\mu\epsilon_Y}
\ln\left(\frac{\gamma^2T}{\left[\max\{m_0^{\scriptscriptstyle X},m_0^{\scriptscriptstyle Y}\}\right]^2}+1\right)\notag\\
&\quad+\frac{L}{\mu\epsilon_Y}
\sqrt{\left(\epsilon_X^2+\frac{\eta_X^2}{T}\ln\left(\frac{\gamma^2T}{\left[\max\{m_0^{\scriptscriptstyle X},m_0^{\scriptscriptstyle Y}\}\right]^2}+1\right)\right)
\left(\epsilon_Y^2+\frac{\eta_Y^2}{T}\ln\left(\frac{\gamma^2T}{\left[m_0^{\scriptscriptstyle Y}\right]^2}+1\right)\right)}\notag\\
&\quad+\frac{L r\epsilon_{\scriptscriptstyle Y}}{2\mu}+\frac{L r\eta_{\scriptscriptstyle Y}^2}{2T\mu\epsilon_{\scriptscriptstyle Y}}\ln\!\left(\frac{\gamma^2 T}{[m_0^{\scriptscriptstyle Y}]^2}+1\right)+\frac{2\sqrt{r}}{\mu\epsilon_{\scriptscriptstyle Y}}\sqrt{\epsilon_{\scriptscriptstyle Y}^2+\frac{\eta_{\scriptscriptstyle Y}^2}{T}\ln\!\left(\frac{\gamma^2 T}{[m_0^{\scriptscriptstyle Y}]^2}+1\right)}\notag\\
&\quad\cdot\left(\frac{3\kappa^2}{T(1-\beta)b_1}
+\frac{6 L^2}{(1-\beta)^2}\left(
{r\epsilon_{\scriptscriptstyle X}^2}+\frac{r\eta_{\scriptscriptstyle X}^2}{T}\ln\!\left(\frac{\gamma^2 T}{\left[\max\{m_0^{\scriptscriptstyle X},m_0^{\scriptscriptstyle Y}\}\right]^2}+1\right)\right.\right.\notag\\
&\quad\left.\left.+
{r\epsilon_{\scriptscriptstyle Y}^2}+\frac{r\eta_{\scriptscriptstyle Y}^2}{T}\ln\!\left(\frac{\gamma^2 T}{[m_0^{\scriptscriptstyle Y}]^2}+1\right)
\right)
+\frac{3(1-\beta)}{T(1+\beta)}\sum_{t=0}^{T-1}\frac{\kappa^2}{b_t}\right)^{\frac{1}{2}}.
\end{align}
Similarly, substituting \cref{eq:Fbar-avg-noVR-compact-weighted} into \cref{eq:dual-avg-dt2-final-noVR}, we have
\begin{align}
&\frac1T\sum_{t=0}^{T-1}\mathbb E[\|Y_t-Y^\star(X_t)\|_F^2]\notag\\
&\le
\frac{8 G\Delta_0^{\scriptscriptstyle Y}}{7\mu^2\epsilon_{\scriptscriptstyle Y}T}
+\left(\frac{G}{7\mu\epsilon_{\scriptscriptstyle Y}}+\frac{4G L r}{7\mu^2\epsilon_{\scriptscriptstyle Y}}\right)\frac{1}{T}\sum_{t=0}^{T-1}\mathbb E\left[(\max\{\epsilon_{\scriptscriptstyle Y},\eta_{\scriptscriptstyle Y}\sigma_t\})^2\right]
\notag\\
&\quad+\left(\frac{4Gr(L_\Phi+L)}{7\mu^2\epsilon_{\scriptscriptstyle Y}}+\frac{16r^2L^2G}{7\mu^3\epsilon_{\scriptscriptstyle Y}}+\frac{16rG^2L^2}{7\mu^4\epsilon_{\scriptscriptstyle Y}^2}\right)\frac{1}{T}\sum_{t=0}^{T-1}\mathbb E\left[(\max\{\epsilon_{\scriptscriptstyle X},\eta_{\scriptscriptstyle X}\theta_t\})^2\right]\notag\\
&\quad+\frac{16G}{7\mu^2\epsilon_{\scriptscriptstyle Y}}
\sqrt{\frac{1}{T}\sum_{t=0}^{T-1}\mathbb E\left[(\max\{\epsilon_{\scriptscriptstyle Y},\eta_{\scriptscriptstyle Y}\sigma_t\})^2\right]}\,
\sqrt{\frac{1}{T}\sum_{t=0}^{T-1}\mathbb E[\|\bar F_{t+1}^{\scriptscriptstyle Y}\|_F^2]}\notag\\
&\leq
\frac{8 G\Delta_0^{\scriptscriptstyle Y}}{7\mu^2\epsilon_{\scriptscriptstyle Y}T}
+\left(\frac{G}{7\mu\epsilon_{\scriptscriptstyle Y}}+\frac{4G L r}{7\mu^2\epsilon_{\scriptscriptstyle Y}}\right)\frac{1}{T}\sum_{t=0}^{T-1}\mathbb E\left[(\max\{\epsilon_{\scriptscriptstyle Y},\eta_{\scriptscriptstyle Y}\sigma_t\})^2\right]
\notag\\
&\quad+\left(\frac{4Gr(L_\Phi+L)}{7\mu^2\epsilon_{\scriptscriptstyle Y}}+\frac{16r^2L^2G}{7\mu^3\epsilon_{\scriptscriptstyle Y}}+\frac{16rG^2L^2}{7\mu^4\epsilon_{\scriptscriptstyle Y}^2}\right)\frac{1}{T}\sum_{t=0}^{T-1}\mathbb E\left[(\max\{\epsilon_{\scriptscriptstyle X},\eta_{\scriptscriptstyle X}\theta_t\})^2\right]\notag\\
&\quad+\frac{16G\sqrt r}{7\mu^2\epsilon_{\scriptscriptstyle Y}}
\sqrt{\frac{1}{T}\sum_{t=0}^{T-1}\mathbb E\left[(\max\{\epsilon_{\scriptscriptstyle Y},\eta_{\scriptscriptstyle Y}\sigma_t\})^2\right]}\notag\\
&\quad\cdot\left(\frac{3\kappa^2}{T(1-\beta)b_1}
+\frac{6 L^2}{(1-\beta)^2}\left(
{r\epsilon_{\scriptscriptstyle X}^2}+\frac{r\eta_{\scriptscriptstyle X}^2}{T}\ln\!\left(\frac{\gamma^2 T}{\left[\max\{m_0^{\scriptscriptstyle X},m_0^{\scriptscriptstyle Y}\}\right]^2}+1\right)\right.\right.\notag\\
&\quad\left.\left.+
{r\epsilon_{\scriptscriptstyle Y}^2}+\frac{r\eta_{\scriptscriptstyle Y}^2}{T}\ln\!\left(\frac{\gamma^2 T}{[m_0^{\scriptscriptstyle Y}]^2}+1\right)
\right)
+\frac{3(1-\beta)}{T(1+\beta)}\sum_{t=0}^{T-1}\frac{\kappa^2}{b_t}\right)^{\frac{1}{2}}\notag\\
&\le
\frac{8 G\Delta_0^{\scriptscriptstyle Y}}{7\mu^2\epsilon_{\scriptscriptstyle Y}T}
+\frac{4Gr\epsilon_{\scriptscriptstyle X}^2(L_\Phi+L)}{7\mu^2\epsilon_{\scriptscriptstyle Y}}+\frac{16r^2L^2G\epsilon_{\scriptscriptstyle X}^2}{7\mu^3\epsilon_{\scriptscriptstyle Y}}+\frac{16rG^2L^2\epsilon_{\scriptscriptstyle X}^2}{7\mu^4\epsilon_{\scriptscriptstyle Y}^2}\notag\\
&\quad+\left(\frac{4Gr\eta_{\scriptscriptstyle X}^2(L_\Phi+L)}{7T\mu^2\epsilon_{\scriptscriptstyle Y}}+\frac{16r^2L^2G\eta_{\scriptscriptstyle X}^2}{7T\mu^3\epsilon_{\scriptscriptstyle Y}}+\frac{16rG^2L^2\eta_{\scriptscriptstyle X}^2}{7T\mu^4\epsilon_{\scriptscriptstyle Y}^2}\right)\ln\!\left(\frac{\gamma^2 T}{\left[\max\{m_0^{\scriptscriptstyle X},m_0^{\scriptscriptstyle Y}\}\right]^2}+1\right)\notag\\
&\quad
+\frac{G\epsilon_{\scriptscriptstyle Y}}{7\mu}+\frac{4G L r\epsilon_{\scriptscriptstyle Y}}{7\mu^2}+\left(\frac{G\eta_{\scriptscriptstyle Y}^2}{7T\mu\epsilon_{\scriptscriptstyle Y}}+\frac{4\eta_{\scriptscriptstyle Y}^2G L r}{7T\mu^2\epsilon_{\scriptscriptstyle Y}}\right)\ln\!\left(\frac{\gamma^2 T}{[m_0^{\scriptscriptstyle Y}]^2}+1\right)
\notag\\
&\quad+
\frac{16G\sqrt r}{7\mu^2\epsilon_{\scriptscriptstyle Y}}
\sqrt{\epsilon_{\scriptscriptstyle Y}^2+\frac{\eta_{\scriptscriptstyle Y}^2}{T}\ln\!\left(\frac{\gamma^2 T}{[m_0^{\scriptscriptstyle Y}]^2}+1\right)}\notag\\
&\quad\cdot\left(\frac{3\kappa^2}{T(1-\beta)b_1}
+\frac{6 L^2}{(1-\beta)^2}\left(
{r\epsilon_{\scriptscriptstyle X}^2}+\frac{r\eta_{\scriptscriptstyle X}^2}{T}\ln\!\left(\frac{\gamma^2 T}{\left[\max\{m_0^{\scriptscriptstyle X},m_0^{\scriptscriptstyle Y}\}\right]^2}+1\right)\right.\right.\notag\\
&\quad\left.\left.+
{r\epsilon_{\scriptscriptstyle Y}^2}+\frac{r\eta_{\scriptscriptstyle Y}^2}{T}\ln\!\left(\frac{\gamma^2 T}{[m_0^{\scriptscriptstyle Y}]^2}+1\right)
\right)
+\frac{3(1-\beta)}{T(1+\beta)}\sum_{t=0}^{T-1}\frac{\kappa^2}{b_t}\right)^{\frac{1}{2}}.
\label{eq:dt2-avg-final-tight}
\end{align}
The proof is thus completed.
\end{proof}

\subsection{Proof of \cref{e1lemma}}\label{sece1}
We first define $\tilde{E}_t := \nabla_X F(X_{t-1}, Y_{t-1}; \xi^{\scriptscriptstyle X}_{t-1}) - \nabla_X f(X_{t-1},Y_{t-1})$ and $E_{t}:=M_{t}^{\scriptscriptstyle X}-\nabla\Phi(X_{t-1})$ for $t \geq 1$. Then
\begin{align*}
M_{t+1}^{\scriptscriptstyle X} &= \beta M_t^{\scriptscriptstyle X} + (1-\beta)\nabla_X F(X_{t}, Y_{t}; \xi^{\scriptscriptstyle X}_{t})\\
&= \beta(E_t + \nabla\Phi(X_{t-1})) + (1-\beta)\left(\tilde{E}_{t+1} + \nabla_X f(X_t,Y_t)\right)\\
&= \nabla_X f(X_t,Y_t)) + \beta(\nabla\Phi(X_{t-1}) - \nabla_X f(X_t,Y_t)) + \beta E_t + (1-\beta)\tilde{E}_{t+1}.
\end{align*}

Hence, for $t \geq 0$,
\begin{align*}
E_{t+1}
&=M_{t+1}^{\scriptscriptstyle X}-\nabla\Phi(X_t)\\
&=\beta E_t
+\beta\big(\nabla\Phi(X_{t-1})-\nabla\Phi(X_t)\big)
+(1-\beta)\tilde E_{t+1}
+(1-\beta)\big(\nabla_X f(X_t,Y_t)-\nabla\Phi(X_t)\big).
\end{align*}

A recursive formula can be derived as:
\begin{align}
E_{t+1}&
\leq \beta^t E_1 + (1-\beta)\sum_{\tau=0}^{t-1} \beta^\tau \tilde{E}_{t+1-\tau} + \beta\sum_{\tau=0}^{t-1} \beta^\tau (\nabla\Phi(X_{t-\tau-1}) - \nabla\Phi(X_{t-\tau}))\notag\\
&\quad+(1-\beta)\sum_{\tau=0}^{t-1}\beta^\tau\big(\nabla_X f(X_{t-\tau},Y_{t-\tau})-\nabla\Phi(X_{t-\tau})\big).\label{iterationequatiln}
\end{align}

Assume that, for each $t=1,\ldots,T$, an independent mini-batch of size $b_t\geq1$ is sampled. Then, by \cref{additionallemma}, it follows that
\begin{align*}
&\mathbb{E}[\|E_{t+1}\|_*]\\
&\leq \beta^t \mathbb{E}[\|E_1\|_*] + (1-\beta)\mathbb{E}\left[\left\|\sum_{\tau=0}^{t-1} \beta^\tau \tilde{E}_{t+1-\tau}\right\|_*\right] + \beta L_\Phi \sum_{\tau=0}^{t-1} \beta^\tau \mathbb{E}[\|X_{t-\tau}-X_{t-\tau-1}\|_{\mathrm{op}}]\\
&\quad+ (1-\beta) L\sum_{\tau=0}^{t-1} \beta^\tau \mathbb{E}[\|Y_{t-\tau}-Y^*(X_{t-\tau})\|_{\mathrm{op}}]\\
&\overset{(a)}{\leq} \beta^t \mathbb{E}[\|E_1\|_*] + (1-\beta)\mathbb{E}\left[\left\|\sum_{\tau=0}^{t-1} \beta^\tau \tilde{E}_{t+1-\tau}\right\|_*\right] + \beta L_\Phi \sum_{\tau=0}^{t-1} \beta^\tau \mathbb{E}[\max\{\epsilon_{\scriptscriptstyle X}, \eta_{\scriptscriptstyle X}\theta_{t-\tau}\}] \\
&\quad+ (1-\beta) L\sum_{\tau=0}^{t-1} \beta^\tau \mathbb{E}[\|Y_{t-\tau}-Y^*(X_{t-\tau})\|_F]\\
&\overset{(b)}{\leq} \frac{\beta^t \kappa \sqrt{r}}{\sqrt{b_1}} + (1-\beta)\sqrt{r}\mathbb{E}\left[\left\|\sum_{\tau=0}^{t-1} \beta^\tau \tilde{E}_{t+1-\tau}\right\|_F\right] + \beta L_\Phi\epsilon_{\scriptscriptstyle X} \frac{1-\beta^t}{1-\beta} + \beta\eta_{\scriptscriptstyle X} L_\Phi \sum_{\tau=0}^{t-1} \beta^\tau \mathbb{E}[\theta_{t-\tau}]\\
&\quad+ (1-\beta) L\sum_{\tau=0}^{t-1} \beta^\tau \mathbb{E}[\|Y_{t-\tau}-Y^*(X_{t-\tau})\|_F]\\
&\leq \frac{\beta^t \kappa \sqrt{r}}{\sqrt{b_1}} + (1-\beta)\sqrt{r}\mathbb{E}\left[\left\|\sum_{\tau=0}^{t-1} \beta^\tau \tilde{E}_{t+1-\tau}\right\|_F^2\right]^{\frac{1}{2}} + \beta L_\Phi\epsilon_{\scriptscriptstyle X} \frac{1-\beta^t}{1-\beta} + \beta\eta_{\scriptscriptstyle X} L_\Phi \sum_{\tau=0}^{t-1} \beta^\tau \mathbb{E}[\theta_{t-\tau}]\\
&\quad+ (1-\beta) L\sum_{\tau=0}^{t-1} \beta^\tau \mathbb{E}[\|Y_{t-\tau}-Y^*(X_{t-\tau})\|_F]\\
&\overset{(c)}{\leq} \frac{\beta^t \kappa \sqrt{r}}{\sqrt{b_1}} + (1-\beta)\frac{\kappa\sqrt{r}}{\sqrt{b_t}} \sqrt{\sum_{\tau=0}^{t-1} \beta^{2\tau}} + \beta L_\Phi \epsilon_{\scriptscriptstyle X}\frac{1-\beta^t}{1-\beta} + \beta\eta_{\scriptscriptstyle X} L_\Phi \sum_{\tau=0}^{t-1} \beta^\tau \mathbb{E}[\theta_{t-\tau}]\\
&\quad+ (1-\beta) L\sum_{\tau=0}^{t-1} \beta^\tau \mathbb{E}[\|Y_{t-\tau}-Y^*(X_{t-\tau})\|_F]\\
&= \frac{\beta^t \kappa \sqrt{r}}{\sqrt{b_1}} + (1-\beta)\frac{\kappa\sqrt{r}}{\sqrt{b_t}} \sqrt{\frac{1-\beta^{2t}}{1-\beta^2}} + \beta L_\Phi \epsilon_{\scriptscriptstyle X}\frac{1-\beta^t}{1-\beta} + \beta\eta_{\scriptscriptstyle X} L_\Phi \sum_{\tau=0}^{t-1} \beta^\tau \mathbb{E}[\theta_{t-\tau}]\\
&\quad+ (1-\beta) L\sum_{\tau=0}^{t-1} \beta^\tau \mathbb{E}[\|Y_{t-\tau}-Y^*(X_{t-\tau})\|_F],
\end{align*}
where (a) uses $\|A\|_{\mathrm{op}}\le \|A\|_{F}$, (b) uses \cref{stochasticgradient} together with the inequality $\|A\|_\ast \le \sqrt{r}\,\|A\|_F$, and (c) follows from \cref{stochasticgradient}.
Therefore,
\begin{align*}
&\sum_{t=0}^{T-1} \mathbb{E}[\|E_{t+1}\|_*] \\
&\leq \frac{\kappa\sqrt{r}}{(1-\beta)\sqrt{b_1}} + \sum_{t=0}^{T-1} \kappa\sqrt{\frac{r(1-\beta)}{b_t}} + \frac{T\beta L_\Phi\epsilon_{\scriptscriptstyle X}}{1-\beta} + \beta\eta_{\scriptscriptstyle X} L_\Phi\sum_{t=1}^T \sum_{\tau=0}^{t-1} \beta^\tau \mathbb{E}[\theta_{t-\tau}]\\
&\quad + (1-\beta) L\sum_{t=1}^T \sum_{\tau=0}^{t-1} \beta^\tau \mathbb{E}[\|Y_{t-\tau}-Y^*(X_{t-\tau})\|_F]\\
&\leq \frac{\kappa\sqrt{r}}{(1-\beta)\sqrt{b_1}} +\sum_{t=0}^{T-1} \kappa\sqrt{\frac{r(1-\beta)}{b_t}} + \frac{T\beta L_\Phi\epsilon_{\scriptscriptstyle X}}{1-\beta} + \frac{\beta\eta_{\scriptscriptstyle X} L_\Phi}{1-\beta} \sum_{t=1}^T \mathbb{E}[\theta_t]\\
&\quad + (1-\beta) L\sum_{t=1}^T \sum_{\tau=0}^{t-1} \beta^\tau \mathbb{E}[\|Y_{t-\tau}-Y^*(X_{t-\tau})\|_F]\\
&\leq \frac{\kappa\sqrt{r}}{(1-\beta)\sqrt{b_1}} + \sum_{t=0}^{T-1} \kappa\sqrt{\frac{r(1-\beta)}{b_t}} + \frac{T\beta L_\Phi\epsilon_{\scriptscriptstyle X}}{1-\beta} + \frac{\beta\eta_{\scriptscriptstyle X} L_\Phi}{1-\beta} \sqrt{T} \left(\sum_{t=1}^T \mathbb{E}[\theta_t^2]\right)^{\frac{1}{2}}\\
&\quad + (1-\beta) L\sum_{t=1}^T \sum_{\tau=0}^{t-1} \beta^\tau \mathbb{E}[\|Y_{t-\tau}-Y^*(X_{t-\tau})\|_F]\\
&\leq L\sum_{t=0}^{T-1}  \mathbb{E}[\|Y_{t}-Y^*(X_{t})\|_F] + \frac{T\beta L_\Phi\epsilon_{\scriptscriptstyle X}}{1-\beta} + \frac{\beta\eta_{\scriptscriptstyle X} L_\Phi}{1-\beta} \sqrt{T} \sqrt{\ln\left(\frac{\gamma^2 T}{\left[\max\{m_0^{\scriptscriptstyle X},m_0^{\scriptscriptstyle Y}\}\right]^2}\right) + 1}\notag\\
&\quad+\frac{\kappa\sqrt{r}}{(1-\beta)\sqrt{b_1}} + \sum_{t=0}^{T-1} \kappa\sqrt{\frac{r(1-\beta)}{b_t}},
\end{align*}
where the last inequality uses \cref{lem:log_cum_ratio}.
Then, according to \cref{lemmay}, we have
\begin{align}
&\frac{1}{T}\sum_{t=0}^{T-1} \mathbb{E}[\|E_{t+1}\|_*] \notag\\
&\leq \frac{\kappa\sqrt{r}}{T(1-\beta)\sqrt{b_1}} + \frac{\kappa}{T}\sum_{t=0}^{T-1} \sqrt{\frac{r(1-\beta)}{b_t}} + \frac{\beta L_\Phi\epsilon_{\scriptscriptstyle X}}{1-\beta} + \frac{\beta\eta_{\scriptscriptstyle X} L_\Phi}{(1-\beta) \sqrt{T}} \sqrt{\ln\left(\frac{\gamma^2 T}{\left[\max\{m_0^{\scriptscriptstyle X},m_0^{\scriptscriptstyle Y}\}\right]^2}\right) + 1}\notag\\
&\quad + L\Bigg(\frac{\Delta_0^{\scriptscriptstyle Y}}{T\mu\epsilon_{\scriptscriptstyle Y}}
+\frac{G\epsilon_{\scriptscriptstyle X}\sqrt{r}(2+\lambda)}{\mu\epsilon_{\scriptscriptstyle Y}}+\frac{G\eta_{\scriptscriptstyle X}\sqrt{r}(2+\lambda)}{\mu\epsilon_{\scriptscriptstyle Y}\sqrt{T}}\sqrt{\ln\left(\frac{\gamma^2 T}{\left[\max\{m_0^{\scriptscriptstyle X},m_0^{\scriptscriptstyle Y}\}\right]^2}\right) + 1}\notag\\
&\quad+\frac{L\epsilon_X^2(2+\lambda^2)}{2\mu\epsilon_Y}+\frac{L\eta_X^2(2+\lambda^2)}{2T\mu\epsilon_Y}
\ln\left(\frac{\gamma^2T}{\left[\max\{m_0^{\scriptscriptstyle X},m_0^{\scriptscriptstyle Y}\}\right]^2}+1\right)\notag\\
&\quad+\frac{L}{\mu\epsilon_Y}
\sqrt{\left(\epsilon_X^2+\frac{\eta_X^2}{T}\ln\left(\frac{\gamma^2T}{\left[\max\{m_0^{\scriptscriptstyle X},m_0^{\scriptscriptstyle Y}\}\right]^2}+1\right)\right)
\left(\epsilon_Y^2+\frac{\eta_Y^2}{T}\ln\left(\frac{\gamma^2T}{\left[m_0^{\scriptscriptstyle Y}\right]^2}+1\right)\right)}\notag\\
&\quad+\frac{L r\epsilon_{\scriptscriptstyle Y}}{2\mu}+\frac{L r\eta_{\scriptscriptstyle Y}^2}{2T\mu\epsilon_{\scriptscriptstyle Y}}\ln\!\left(\frac{\gamma^2 T}{[m_0^{\scriptscriptstyle Y}]^2}+1\right)+\frac{2\sqrt{r}}{\mu\epsilon_{\scriptscriptstyle Y}}\sqrt{\epsilon_{\scriptscriptstyle Y}^2+\frac{\eta_{\scriptscriptstyle Y}^2}{T}\ln\!\left(\frac{\gamma^2 T}{[m_0^{\scriptscriptstyle Y}]^2}+1\right)}\notag\\
&\quad\cdot\left(\frac{3\kappa^2}{T(1-\beta)b_1}
+\frac{6 L^2}{(1-\beta)^2}\left(
{r\epsilon_{\scriptscriptstyle X}^2}+\frac{r\eta_{\scriptscriptstyle X}^2}{T}\ln\!\left(\frac{\gamma^2 T}{\left[\max\{m_0^{\scriptscriptstyle X},m_0^{\scriptscriptstyle Y}\}\right]^2}+1\right)\right.\right.\notag\\
&\quad\left.\left.+
{r\epsilon_{\scriptscriptstyle Y}^2}+\frac{r\eta_{\scriptscriptstyle Y}^2}{T}\ln\!\left(\frac{\gamma^2 T}{[m_0^{\scriptscriptstyle Y}]^2}+1\right)
\right)
+\frac{3(1-\beta)}{T(1+\beta)}\sum_{t=0}^{T-1}\frac{\kappa^2}{b_t}\right)^{\frac{1}{2}}\Bigg).\label{eq102}
\end{align}

The proof for \cref{e1lemma} is thus completed.

\subsection{Proof of \cref{e2lemma}}\label{sece2}
According to \cref{iterationequatiln}, we have
\begin{align*}
&\mathbb{E}\left[\|E_{t+1}\|_F^2\right]\\
&\leq 4\beta^{2t} \mathbb{E}\left[\|E_1\|_F^2\right]
+ 4(1-\beta)^2 \sum_{\tau=0}^{t-1}  \mathbb{E}\left[\left\|\beta^{\tau}\tilde{E}_{t+1-\tau}\right\|_F^2\right]\notag\\
&\quad+ 4\beta^2 \mathbb{E}\left[\left\|\sum_{\tau=0}^{t-1} \beta^\tau (\nabla\Phi(X_{t-\tau-1}) - \nabla\Phi(X_{t-\tau}))\right\|_F^2\right]\\
&\quad + 4(1-\beta)^2 \mathbb{E}\left[\left\|\sum_{\tau=0}^{t-1} \beta^\tau (\nabla\Phi(X_{t-\tau}) - \nabla_X f(X_{t-\tau},Y_{t-\tau}))\right\|_F^2\right]\\
&\leq \frac{4\beta^{2t} \kappa^2}{b_1}+ 4(1-\beta)^2 \frac{\kappa^2}{b_t} \sum_{\tau=0}^{t-1} \beta^{2\tau} + 4\beta^2 \mathbb{E}\left[\left(\sum_{\tau=0}^{t-1} \beta^\tau \|\nabla\Phi(X_{t-\tau-1}) - \nabla\Phi(X_{t-\tau})\|_F\right)^2\right]\\
&\quad + 4(1-\beta)^2 \mathbb{E}\left[\left(\sum_{\tau=0}^{t-1} \beta^\tau \left\|\nabla\Phi(X_{t-\tau}) - \nabla_X f(X_{t-\tau},Y_{t-\tau})\right\|_F\right)^2\right].
\end{align*}
The last inequality follows from \cref{stochasticgradient}. Then, by the Cauchy--Schwarz inequality, \cref{smoothness}, \cref{additionallemma}, and the definition of the global rank bound $r$, we have
\begin{align}
&\mathbb{E}\left[\|E_{t+1}\|_F^2\right] \notag\\
&\leq \frac{4\beta^{2t}\kappa^2}{b_1} + \frac{4\kappa^2}{b_t}\frac{(1-\beta)(1-\beta^{2t})}{1+\beta} + 4L_\Phi^2r\beta^2 \epsilon_{\scriptscriptstyle X}^2 \left(\frac{1-\beta^{2t}}{1-\beta^2}\right)\notag\\
&\quad+ 4L_\Phi^2r\eta_{\scriptscriptstyle X}^2 \beta^2 \left(\sum_{\tau=0}^{t-1} \beta^{2\tau}\right) \mathbb{E}\left[\sum_{\tau=0}^{t-1} \theta_{t-\tau}^2\right]\notag\\
&\quad+ 4L^2(1-\beta)^2 \left(\sum_{\tau=0}^{t-1} \beta^{2\tau}\right) \mathbb{E}\left[\sum_{\tau=0}^{t-1} \|Y^*(X_{t-\tau})-Y_{t-\tau}\|_F^2\right] \notag \\
&\leq \frac{4\beta^{2t}\kappa^2}{b_1} + \frac{4\kappa^2}{b_t}\frac{(1-\beta)(1-\beta^{2t})}{1+\beta}+ 4L_\Phi^2r\beta^2 \epsilon_{\scriptscriptstyle X}^2 \left(\frac{1-\beta^{2t}}{1-\beta^2}\right)\notag\\
&\quad+ 4L_\Phi^2r\eta_{\scriptscriptstyle X}^2 \beta^2\left(\frac{1-\beta^{2t}}{1-\beta^2}\right) \mathbb{E}\left[\sum_{\tau=0}^{t-1} \theta_{t-\tau}^2\right]\notag\\
&\quad+ 4L^2 \left(\frac{1-\beta}{1+\beta}\right) \mathbb{E}\left[\sum_{\tau=0}^{t-1} \|Y^*(X_{t-\tau})-Y_{t-\tau}\|_F^2\right].\label{eq11}
\end{align}

Then, by applying Lemma 3.2 in \cite{ward2020adagrad}, we have
\begin{align}
\sum_{t=0}^{T-1}\sum_{\tau=0}^{t-1} \theta_{t-\tau}^2
&=
\sum_{t=0}^{T-1}\sum_{\tau=1}^{t} \theta_{\tau}^2
=
\sum_{t=1}^{T} (T-t+1)\theta_t^2
\;\le\;
T \sum_{t=1}^{T}\theta_t^2
\;\le\;
T\ln\!\left(\frac{\gamma^2T}{\left[\max\{m_0^{\scriptscriptstyle X},m_0^{\scriptscriptstyle Y}\}\right]^2}+1\right).
\label{sumloguseful}
\end{align} 
Summing \cref{eq11} over $t=0,\dots,T-1$ and combining the result with \cref{sumloguseful} and \cref{lemmay}, we obtain
\begin{align}
&\sum_{t=0}^{T-1}
\mathbb{E}\big[\|E_{t+1}\|_F^2\big]\notag\\
&\leq\frac{4}{1-\beta^2} \frac{\kappa^2}{b_1}+ \frac{4(1-\beta)}{1+\beta}\sum_{t=1}^{T}\frac{\kappa^2}{b_t}+  \frac{4L_\Phi^2r\beta^2 \epsilon_{\scriptscriptstyle X}^2T}{1-\beta^2} + \frac{4L_\Phi^2r\eta_{\scriptscriptstyle X}^2 \beta^2}{1-\beta^2} T\ln\!\left(\frac{\gamma^2T}{\left[\max\{m_0^{\scriptscriptstyle X},m_0^{\scriptscriptstyle Y}\}\right]^2}+1\right)\notag\\
&\quad+ {4L^2(1-\beta) } \sum_{t=1}^{T}\sum_{\tau=0}^{t-1}\mathbb{E}[\|Y^*(X_{t-\tau})-Y_{t-\tau}\|_F^2]\notag\\
&\leq\frac{4}{1-\beta^2} \frac{\kappa^2}{b_1} + \frac{4(1-\beta)}{1+\beta}\sum_{t=1}^{T}\frac{\kappa^2}{b_t}+4L^2T(1-\beta)\sum_{t=1}^T\mathbb{E}\left[\|Y_t-Y^*(X_t)\|_F^2\right] \notag \\
&\quad + \frac{4L_\Phi^2r\beta^2 \epsilon_{\scriptscriptstyle X}^2T}{1-\beta^2} + \frac{4L_\Phi^2r\eta_{\scriptscriptstyle X}^2 \beta^2}{1-\beta^2} T\ln\!\left(\frac{\gamma^2T}{\left[\max\{m_0^{\scriptscriptstyle X},m_0^{\scriptscriptstyle Y}\}\right]^2}+1\right)\notag\\
&\leq\frac{4}{1-\beta^2} \frac{\kappa^2}{b_1}+ \frac{4(1-\beta)}{1+\beta}\sum_{t=1}^{T}\frac{\kappa^2}{b_t} +  \frac{4L_\Phi^2r\beta^2 \epsilon_{\scriptscriptstyle X}^2T}{1-\beta^2} + \frac{4L_\Phi^2r\eta_{\scriptscriptstyle X}^2 \beta^2}{1-\beta^2} T\ln\!\left(\frac{\gamma^2T}{\left[\max\{m_0^{\scriptscriptstyle X},m_0^{\scriptscriptstyle Y}\}\right]^2}+1\right)\notag\\
&\quad+ 4L^2T(1-\beta) \Bigg[\frac{8 G\Delta_0^{\scriptscriptstyle Y}}{7\mu^2\epsilon_{\scriptscriptstyle Y}}
+\frac{4Gr\epsilon_{\scriptscriptstyle X}^2T(L_\Phi+L)}{7\mu^2\epsilon_{\scriptscriptstyle Y}}+\frac{16r^2L^2G\epsilon_{\scriptscriptstyle X}^2T}{7\mu^3\epsilon_{\scriptscriptstyle Y}}+\frac{16rG^2L^2\epsilon_{\scriptscriptstyle X}^2T}{7\mu^4\epsilon_{\scriptscriptstyle Y}^2}\notag\\
&\quad+\left(\frac{4Gr\eta_{\scriptscriptstyle X}^2(L_\Phi+L)}{7\mu^2\epsilon_{\scriptscriptstyle Y}}+\frac{16r^2L^2G\eta_{\scriptscriptstyle X}^2}{7\mu^3\epsilon_{\scriptscriptstyle Y}}+\frac{16rG^2L^2\eta_{\scriptscriptstyle X}^2}{7\mu^4\epsilon_{\scriptscriptstyle Y}^2}\right)\ln\!\left(\frac{\gamma^2 T}{\left[\max\{m_0^{\scriptscriptstyle X},m_0^{\scriptscriptstyle Y}\}\right]^2}+1\right)\notag\\
&\quad
+\frac{G\epsilon_{\scriptscriptstyle Y}T}{7\mu}+\frac{4G L r\epsilon_{\scriptscriptstyle Y}T}{7\mu^2}+\left(\frac{G\eta_{\scriptscriptstyle Y}^2}{7\mu\epsilon_{\scriptscriptstyle Y}}+\frac{4\eta_{\scriptscriptstyle Y}^2G L r}{7\mu^2\epsilon_{\scriptscriptstyle Y}}\right)\ln\!\left(\frac{\gamma^2 T}{[m_0^{\scriptscriptstyle Y}]^2}+1\right)
\notag\\
&\quad+
\frac{16GT\sqrt r}{7\mu^2\epsilon_{\scriptscriptstyle Y}}
\sqrt{\epsilon_{\scriptscriptstyle Y}^2+\frac{\eta_{\scriptscriptstyle Y}^2}{T}\ln\!\left(\frac{\gamma^2 T}{[m_0^{\scriptscriptstyle Y}]^2}+1\right)}\notag\\
&\quad\cdot\left(\frac{3\kappa^2}{T(1-\beta)b_1}
+\frac{6 L^2}{(1-\beta)^2}\left(
{r\epsilon_{\scriptscriptstyle X}^2}+\frac{r\eta_{\scriptscriptstyle X}^2}{T}\ln\!\left(\frac{\gamma^2 T}{\left[\max\{m_0^{\scriptscriptstyle X},m_0^{\scriptscriptstyle Y}\}\right]^2}+1\right)\right.\right.\notag\\
&\quad\left.\left.+
{r\epsilon_{\scriptscriptstyle Y}^2}+\frac{r\eta_{\scriptscriptstyle Y}^2}{T}\ln\!\left(\frac{\gamma^2 T}{[m_0^{\scriptscriptstyle Y}]^2}+1\right)
\right)
+\frac{3(1-\beta)}{T(1+\beta)}\sum_{t=0}^{T-1}\frac{\kappa^2}{b_t}\right)^{\frac{1}{2}}\Bigg].\label{eq11Fnorm}
\end{align}

The proof for \cref{e2lemma} is thus completed.

Under the parameter choices $1-\beta=\frac{1}{T^{\frac{1}{4}}}$, $\epsilon_{\scriptscriptstyle X}=\frac{1}{T^{\frac{3}{4}}}$, $\epsilon_{\scriptscriptstyle Y}=\frac{1}{T^{\frac{1}{2}}}$, $\eta_{\scriptscriptstyle X}=\frac{1}{T^{\frac{7}{8}+q}}$, $\eta_{\scriptscriptstyle Y}=\frac{1}{T^{\frac{5}{8}+q}}$, $\gamma>0$, $m^{\scriptscriptstyle X}_0>0$, $m^{\scriptscriptstyle Y}_0>0$, and $b_t\geq T^\frac{1}{4}$ for $t\in\{1,\cdots,T\}$, it follows from \cref{e1lemma} that
\begin{align}
&\frac{1}{T}\sum_{t=0}^T \mathbb{E}[\|E_{t+1}\|_*] \notag\\
&\leq \frac{\kappa\sqrt{r}}{{T}^\frac{7}{8}} + \frac{\kappa\sqrt r}{T^\frac{1}{4}} + \frac{L_\Phi}{\sqrt T} + \frac{L_\Phi}{T^{\frac{9}{8}+q}} \sqrt{\ln\left(\frac{\gamma^2 T}{[\max\{m_0^{\scriptscriptstyle X},m_0^{\scriptscriptstyle Y}\}]^2}\right) + 1}\notag\\
&\quad +L\Bigg(\frac{\Delta_0^{\scriptscriptstyle Y}}{\sqrt{T}\mu}
+\frac{G\sqrt r(2+\lambda)}{\mu T^\frac{1}{4}}+\frac{G\sqrt r(2+\lambda)}{\mu{T}^{\frac{7}{8}+q}}\sqrt{\ln\left(\frac{\gamma^2 T}{\left[\max\{m_0^{\scriptscriptstyle X},m_0^{\scriptscriptstyle Y}\}\right]^2}\right) + 1} \notag\\
&\quad+
\frac{L(2+\lambda^2)}{2\mu T}
\left(1+\frac{1}{T^{5/4+2q}}\ln\left(\frac{\gamma^2 T}{[\max\{m_0^{\scriptscriptstyle X},m_0^{\scriptscriptstyle Y}\}]^2}+1\right)\right)\notag\\
&\quad+
\frac{L}{\mu T^{3/4}}
\sqrt{
\left(1+\frac{1}{T^{5/4+2q}}\ln\left(\frac{\gamma^2 T}{[\max\{m_0^{\scriptscriptstyle X},m_0^{\scriptscriptstyle Y}\}]^2}+1\right)\right)
\left(1+\frac{1}{T^{5/4+2q}}\ln\!\left(\frac{\gamma^2 T}{[m_0^{\scriptscriptstyle Y}]^2}+1\right)\right)
}\notag\\
&\quad+\frac{L r}{2\mu\sqrt{T}}+\frac{L r}{2T^{\frac{7}{4}+2q}\mu}\ln\!\left(\frac{\gamma^2 T}{[m_0^{\scriptscriptstyle Y}]^2}+1\right)+\frac{2\sqrt{rT}}{\mu}\sqrt{\frac{1}{T}+\frac{1}{T^{\frac{9}{4}+2q}}\ln\!\left(\frac{\gamma^2 T}{[m_0^{\scriptscriptstyle Y}]^2}+1\right)}\notag\\
&\quad\cdot\left(\frac{3\kappa^2}{{T}}
+{6 L^2}\sqrt{T}\left(
\frac{r}{T^\frac{3}{2}}+\frac{r}{T^{\frac{11}{4}+2q}}\ln\!\left(\frac{\gamma^2 T}{\left[\max\{m_0^{\scriptscriptstyle X},m_0^{\scriptscriptstyle Y}\}\right]^2}+1\right)\right.\right.\notag\\
&\quad\left.\left.+
\frac{r}{T}+\frac{r}{T^{\frac{9}{4}+2q}}\ln\!\left(\frac{\gamma^2 T}{[m_0^{\scriptscriptstyle Y}]^2}+1\right)
\right)
+\frac{3\kappa^2}{\sqrt{T}}\right)^{\frac{1}{2}}\Bigg)\notag\\
&\lesssim\frac{\kappa\sqrt r}{T^\frac{1}{4}}+\frac{LG\sqrt r(2+\lambda)+\kappa L\sqrt{r}+L^2r}{\mu T^\frac{1}{4}}+\frac{L_\Phi}{\sqrt T}+\frac{\kappa L\sqrt{r}+L^2 r+L\Delta_0^{\scriptscriptstyle Y}}{\mu\sqrt T}
.\label{eq1027}
\end{align}

Moreover, \cref{e2lemma} gives
\begin{align}
&\sum_{t=0}^{T-1}
\mathbb{E}\big[\|E_{t+1}\|_F^2\big]\notag\\
&\leq{4\kappa^2}+{4\sqrt T\kappa^2} + \frac{4L_\Phi^2r}{T^\frac{1}{4}} + \frac{4L_\Phi^2r}{T^{\frac{1}{2}+2q}}\ln\!\left(\frac{\gamma^2T}{\left[\max\{m_0^{\scriptscriptstyle X},m_0^{\scriptscriptstyle Y}\}\right]^2}+1\right)\notag\\
&\quad+ 4L^2 T^\frac{3}{4} \Bigg[\frac{8 G\Delta_0^{\scriptscriptstyle Y}\sqrt{T}}{7\mu^2}+\frac{4Gr(L_\Phi+L)}{7\mu^2}+\frac{16r^2L^2G}{7\mu^3}+\frac{16rG^2L^2 \sqrt T}{7\mu^4}\notag\\
&\quad+\left(\frac{4Gr(L_\Phi+L)}{7\mu^2T^{\frac{5}{4}+2q}}+\frac{16r^2L^2G}{7\mu^3T^{\frac{5}{4}+2q}}+\frac{16rG^2L^2}{7\mu^4T^{\frac{3}{4}+2q}}\right)\ln\!\left(\frac{\gamma^2 T}{\left[\max\{m_0^{\scriptscriptstyle X},m_0^{\scriptscriptstyle Y}\}\right]^2}+1\right)\notag\\
&\quad
+\frac{G\sqrt{T}}{7\mu}+\frac{4G L r\sqrt{T}}{7\mu^2}+\left(\frac{G}{7\mu T^{\frac{3}{4}+2q}}+\frac{4G L r}{7\mu^2 T^{\frac{3}{4}+2q}}\right)\ln\!\left(\frac{\gamma^2 T}{[m_0^{\scriptscriptstyle Y}]^2}+1\right)
\notag\\
&\quad+
\frac{16T^\frac{3}{2} G\sqrt{r}}{7\mu^2}
\sqrt{\frac{1}{T}+\frac{1}{T^{\frac{9}{4}+2q}}\ln\!\left(\frac{\gamma^2 T}{[m_0^{\scriptscriptstyle Y}]^2}+1\right)}\notag\\
&\quad\cdot\left(\frac{3\kappa^2}{{T}}
+{6 L^2}\sqrt{T}\left(
\frac{r}{T^\frac{3}{2}}+\frac{r}{T^{\frac{11}{4}+2q}}\ln\!\left(\frac{\gamma^2 T}{\left[\max\{m_0^{\scriptscriptstyle X},m_0^{\scriptscriptstyle Y}\}\right]^2}+1\right)\right.\right.\notag\\
&\quad\left.\left.+
\frac{r}{T}+\frac{r}{T^{\frac{9}{4}+2q}}\ln\!\left(\frac{\gamma^2 T}{[m_0^{\scriptscriptstyle Y}]^2}+1\right)
\right)
+\frac{3\kappa^2}{\sqrt{T}}\right)^{\frac{1}{2}}\Bigg].\label{eq150}
\end{align}

Hence, we obtain
\begin{align}
&\frac{2\sqrt{r}}{T\epsilon_{\scriptscriptstyle X}}\left(\eta_{\scriptscriptstyle X}\sqrt{\ln\!\left(\frac{\gamma^2T}{[\max\{m_0^{\scriptscriptstyle X},m_0^{\scriptscriptstyle Y}\}]^2}\right)+1} \right)\sqrt{\sum_{t=0}^{T-1} \mathbb{E}[ \|E_t\|_F^2]}\notag\\
&\leq\frac{2\sqrt{r}}{T^{\frac{9}{8}+q}}\left(\sqrt{\ln\!\left(\frac{\gamma^2T}{[\max\{m_0^{\scriptscriptstyle X},m_0^{\scriptscriptstyle Y}\}]^2}\right)+1} \right)\notag\\
&\quad\cdot\Bigg[{4\kappa^2}+{4\sqrt T\kappa^2} + \frac{4L_\Phi^2r}{T^\frac{1}{4}} + \frac{4L_\Phi^2r}{T^{\frac{1}{2}+2q}}\ln\!\left(\frac{\gamma^2T}{\left[\max\{m_0^{\scriptscriptstyle X},m_0^{\scriptscriptstyle Y}\}\right]^2}+1\right)\notag\\
&\quad+ 4L^2 T^\frac{3}{4} \Bigg[\frac{8 G\Delta_0^{\scriptscriptstyle Y}\sqrt{T}}{7\mu^2}+\frac{4Gr(L_\Phi+L)}{7\mu^2}+\frac{16r^2L^2G}{7\mu^3}+\frac{16rG^2L^2 \sqrt T}{7\mu^4}\notag\\
&\quad+\left(\frac{4Gr(L_\Phi+L)}{7\mu^2T^{\frac{5}{4}+2q}}+\frac{16r^2L^2G}{7\mu^3T^{\frac{5}{4}+2q}}+\frac{16rG^2L^2}{7\mu^4T^{\frac{3}{4}+2q}}\right)\ln\!\left(\frac{\gamma^2 T}{\left[\max\{m_0^{\scriptscriptstyle X},m_0^{\scriptscriptstyle Y}\}\right]^2}+1\right)\notag\\
&\quad
+\frac{G\sqrt{T}}{7\mu}+\frac{4G L r\sqrt{T}}{7\mu^2}+\left(\frac{G}{7\mu T^{\frac{3}{4}+2q}}+\frac{4G L r}{7\mu^2 T^{\frac{3}{4}+2q}}\right)\ln\!\left(\frac{\gamma^2 T}{[m_0^{\scriptscriptstyle Y}]^2}+1\right)
\notag\\
&\quad+
\frac{16T^\frac{3}{2} G\sqrt{r}}{7\mu^2}
\sqrt{\frac{1}{T}+\frac{1}{T^{\frac{9}{4}+2q}}\ln\!\left(\frac{\gamma^2 T}{[m_0^{\scriptscriptstyle Y}]^2}+1\right)}\notag\\
&\quad\cdot\left(\frac{3\kappa^2}{{T}}
+{6 L^2}\sqrt{T}\left(
\frac{r}{T^\frac{3}{2}}+\frac{r}{T^{\frac{11}{4}+2q}}\ln\!\left(\frac{\gamma^2 T}{\left[\max\{m_0^{\scriptscriptstyle X},m_0^{\scriptscriptstyle Y}\}\right]^2}+1\right)\right.\right.\notag\\
&\quad\left.\left.+
\frac{r}{T}+\frac{r}{T^{\frac{9}{4}+2q}}\ln\!\left(\frac{\gamma^2 T}{[m_0^{\scriptscriptstyle Y}]^2}+1\right)
\right)
+\frac{3\kappa^2}{\sqrt{T}}\right)^{\frac{1}{2}}\Bigg]\Bigg]^\frac{1}{2}\notag\\
&\lesssim \left[\frac{L\sqrt{r\Delta_0^{\scriptscriptstyle Y} G}+L^\frac{3}{2}r\sqrt{G}+L\sqrt{\kappa G}r^\frac{3}{4}+L^\frac{3}{2}r\sqrt{G}}{\mu T^{\frac{1}{2}+q}}+\frac{L^2r{G}}{\mu^2 T^{\frac{1}{2}+q}}+\frac{L\sqrt{r G}}{\sqrt{\mu} T^{\frac{1}{2}+q}}\right.\notag\\
&\left.\quad+\frac{L\sqrt{\kappa G}r^\frac{3}{4}+L^\frac{3}{2}r\sqrt{G}}{\mu T^{\frac{3}{8}+q}}\right]\sqrt{\ln\!\left(\frac{\gamma^2T}{[\max\{m_0^{\scriptscriptstyle X},m_0^{\scriptscriptstyle Y}\}]^2}\right)+1}.
\label{eq1029}
\end{align}

Plugging \cref{eq1027} and \cref{eq1029} into \cref{descentlemma}, we have
\begin{align}
&\frac{1}{T} \sum_{t=1}^T \mathbb{E}[\|\nabla\Phi(X_{t})\|_*]\notag\\
&\leq \frac{\Delta}{T^\frac{1}{4}} + 2\Bigg(\frac{\kappa\sqrt{r}}{{T}^\frac{7}{8}} + \frac{\kappa\sqrt r}{T^\frac{1}{4}} + \frac{L_\Phi}{\sqrt T} + \frac{L_\Phi}{T^{\frac{9}{8}+q}} \sqrt{\ln\left(\frac{\gamma^2 T}{[\max\{m_0^{\scriptscriptstyle X},m_0^{\scriptscriptstyle Y}\}]^2}\right) + 1}\notag\\
&\quad +L\Bigg(\frac{\Delta_0^{\scriptscriptstyle Y}}{\sqrt{T}\mu}
+\frac{G\sqrt r(2+\lambda)}{\mu T^\frac{1}{4}}+\frac{G\sqrt r(2+\lambda)}{\mu{T}^{\frac{7}{8}+q}}\sqrt{\ln\left(\frac{\gamma^2 T}{\left[\max\{m_0^{\scriptscriptstyle X},m_0^{\scriptscriptstyle Y}\}\right]^2}\right) + 1} \notag\\
&\quad+
\frac{L(2+\lambda^2)}{2\mu T}
\left(1+\frac{1}{T^{5/4+2q}}\ln\left(\frac{\gamma^2 T}{[\max\{m_0^{\scriptscriptstyle X},m_0^{\scriptscriptstyle Y}\}]^2}+1\right)\right)\notag\\
&\quad+
\frac{L}{\mu T^{3/4}}
\sqrt{
\left(1+\frac{1}{T^{5/4+2q}}\ln\left(\frac{\gamma^2 T}{[\max\{m_0^{\scriptscriptstyle X},m_0^{\scriptscriptstyle Y}\}]^2}+1\right)\right)
\left(1+\frac{1}{T^{5/4+2q}}\ln\!\left(\frac{\gamma^2 T}{[m_0^{\scriptscriptstyle Y}]^2}+1\right)\right)
}\notag\\
&\quad+\frac{L r}{2\mu\sqrt{T}}+\frac{L r}{2T^{\frac{7}{4}+2q}\mu}\ln\!\left(\frac{\gamma^2 T}{[m_0^{\scriptscriptstyle Y}]^2}+1\right)+\frac{2\sqrt{rT}}{\mu}\sqrt{\frac{1}{T}+\frac{1}{T^{\frac{9}{4}+2q}}\ln\!\left(\frac{\gamma^2 T}{[m_0^{\scriptscriptstyle Y}]^2}+1\right)}\notag\\
&\quad\cdot\left(\frac{3\kappa^2}{{T}}
+{6 L^2}\sqrt{T}\left(
\frac{r}{T^\frac{3}{2}}+\frac{r}{T^{\frac{11}{4}+2q}}\ln\!\left(\frac{\gamma^2 T}{\left[\max\{m_0^{\scriptscriptstyle X},m_0^{\scriptscriptstyle Y}\}\right]^2}+1\right)\right.\right.\notag\\
&\quad\left.\left.+
\frac{r}{T}+\frac{r}{T^{\frac{9}{4}+2q}}\ln\!\left(\frac{\gamma^2 T}{[m_0^{\scriptscriptstyle Y}]^2}+1\right)
\right)
+\frac{3\kappa^2}{\sqrt{T}}\right)^{\frac{1}{2}}\Bigg)\Bigg)\notag\\
&\quad+ \frac{2\sqrt{r}}{T^{\frac{9}{8}+q}}\left(\sqrt{\ln\!\left(\frac{\gamma^2T}{[\max\{m_0^{\scriptscriptstyle X},m_0^{\scriptscriptstyle Y}\}]^2}\right)+1} \right)\notag\\
&\quad\cdot\Bigg[{4\kappa^2}+{4\sqrt T\kappa^2} + \frac{4L_\Phi^2r}{T^\frac{1}{4}} + \frac{4L_\Phi^2r}{T^{\frac{1}{2}+2q}}\ln\!\left(\frac{\gamma^2T}{\left[\max\{m_0^{\scriptscriptstyle X},m_0^{\scriptscriptstyle Y}\}\right]^2}+1\right)\notag\\
&\quad+ 4L^2 T^\frac{3}{4} \Bigg[\frac{8 G\Delta_0^{\scriptscriptstyle Y}\sqrt{T}}{7\mu^2}+\frac{4Gr(L_\Phi+L)}{7\mu^2}+\frac{16r^2L^2G}{7\mu^3}+\frac{16rG^2L^2 \sqrt T}{7\mu^4}\notag\\
&\quad+\left(\frac{4Gr(L_\Phi+L)}{7\mu^2T^{\frac{5}{4}+2q}}+\frac{16r^2L^2G}{7\mu^3T^{\frac{5}{4}+2q}}+\frac{16rG^2L^2}{7\mu^4T^{\frac{3}{4}+2q}}\right)\ln\!\left(\frac{\gamma^2 T}{\left[\max\{m_0^{\scriptscriptstyle X},m_0^{\scriptscriptstyle Y}\}\right]^2}+1\right)\notag\\
&\quad
+\frac{G\sqrt{T}}{7\mu}+\frac{4G L r\sqrt{T}}{7\mu^2}+\left(\frac{G}{7\mu T^{\frac{3}{4}+2q}}+\frac{4G L r}{7\mu^2 T^{\frac{3}{4}+2q}}\right)\ln\!\left(\frac{\gamma^2 T}{[m_0^{\scriptscriptstyle Y}]^2}+1\right)
\notag\\
&\quad+
\frac{16T^\frac{3}{2} G\sqrt{r}}{7\mu^2}
\sqrt{\frac{1}{T}+\frac{1}{T^{\frac{9}{4}+2q}}\ln\!\left(\frac{\gamma^2 T}{[m_0^{\scriptscriptstyle Y}]^2}+1\right)}\notag\\
&\quad\cdot\left(\frac{3\kappa^2}{{T}}
+{6 L^2}\sqrt{T}\left(
\frac{r}{T^\frac{3}{2}}+\frac{r}{T^{\frac{11}{4}+2q}}\ln\!\left(\frac{\gamma^2 T}{\left[\max\{m_0^{\scriptscriptstyle X},m_0^{\scriptscriptstyle Y}\}\right]^2}+1\right)\right.\right.\notag\\
&\quad\left.\left.+
\frac{r}{T}+\frac{r}{T^{\frac{9}{4}+2q}}\ln\!\left(\frac{\gamma^2 T}{[m_0^{\scriptscriptstyle Y}]^2}+1\right)
\right)
+\frac{3\kappa^2}{\sqrt{T}}\right)^{\frac{1}{2}}\Bigg]\Bigg]^\frac{1}{2}\notag\\
&\quad+ \frac{L_\Phi}{ T^\frac{3}{4}} + \frac{ L_\Phi}{T^{\frac{5}{4}+2q}}\left(\ln\left(\frac{\gamma^2 T}{[\max\{m_0^{\scriptscriptstyle X},m_0^{\scriptscriptstyle Y}\}]^2}\right) + 1\right) \notag\\
&\lesssim\frac{\Delta+\kappa\sqrt r}{T^\frac{1}{4}}+\frac{LG\sqrt r(2+\lambda)+\kappa L\sqrt{r}+L^2r}{\mu T^\frac{1}{4}}+\frac{L_\Phi}{\sqrt T}+\frac{\kappa L\sqrt{r}+L^2 r+L\Delta_0^{\scriptscriptstyle Y}}{\mu\sqrt T}\notag\\
&\quad+\left[\frac{L\sqrt{r\Delta_0^{\scriptscriptstyle Y} G}+L^\frac{3}{2}r\sqrt{G}+L\sqrt{\kappa G}r^\frac{3}{4}+L^\frac{3}{2}r\sqrt{G}}{\mu T^{\frac{1}{2}+q}}+\frac{L^2r{G}}{\mu^2 T^{\frac{1}{2}+q}}+\frac{L\sqrt{r G}}{\sqrt{\mu} T^{\frac{1}{2}+q}}\right.\notag\\
&\left.\quad+\frac{L\sqrt{\kappa G}r^\frac{3}{4}+L^\frac{3}{2}r\sqrt{G}}{\mu T^{\frac{3}{8}+q}}\right]\sqrt{\ln\!\left(\frac{\gamma^2T}{[\max\{m_0^{\scriptscriptstyle X},m_0^{\scriptscriptstyle Y}\}]^2}\right)+1}\notag\\
&\leq  \mathcal{O}\left(\frac{\Delta+\kappa\sqrt{r}}{T^\frac{1}{4}}+\frac{LG\sqrt r(1+\lambda)+\kappa L\sqrt{r}+L^2r}{\mu T^\frac{1}{4}}+\frac{Lr^\frac{3}{4}\sqrt{\kappa G}+L^\frac{3}{2}r\sqrt{G}}{\mu T^{\frac{3}{8}+q}}\sqrt{\ln\!\left(\frac{\gamma^2T}{z_0^2}\right)+1}\right),
\end{align}
where $\Delta:=\Phi(X_0) - \Phi_*$ and $z_0:=\max\{m_0^{\scriptscriptstyle X},m_0^{\scriptscriptstyle Y}\}$.

The proof for \cref{theorem1} is thus completed.

\newpage
\section{Proof of \cref{theoremVR}}\label{VRproof}
\begin{algorithm}[tb]
   \caption{Adaptive Minimax Optimization with Variance-Reduced Orthogonal Updates (AdaSGDA-VR)}
   \label{alg2}
\begin{algorithmic}[1]
    \STATE {\bfseries Initialization:} $X_{0}$, $Y_{0}$, $X_{-1}=X_0$, $Y_{-1}=Y_0$, 
    $M_0^{\scriptscriptstyle X}=0$, $M_0^{\scriptscriptstyle Y}=0$,
    buffers $m^{\scriptscriptstyle X}_{0} = m^{\scriptscriptstyle Y}_{0} = c > 0$,
    hyperparameters $\eta_{\scriptscriptstyle X}, \eta_{\scriptscriptstyle Y}, \epsilon_{\scriptscriptstyle X}, \epsilon_{\scriptscriptstyle Y}, \gamma > 0$, and
    momentum parameter $\beta$.
    \FOR{iteration $t = 0, 1,  \cdots, T-1$}
        \STATE Sample a mini-batch of size $b_t$ and compute the stochastic gradients:
$\nabla_X F(X_t, Y_t; \xi_t^{\scriptscriptstyle X})$ and $\nabla_Y F(X_t, Y_t; \xi_t^{\scriptscriptstyle Y})$.
        \STATE Momentum updates:\\
        $M^{\scriptscriptstyle X}_{t+1}=\beta M^{\scriptscriptstyle X}_{t}+(1-\beta)\nabla_X F(X_{t}, Y_{t}; \xi^{\scriptscriptstyle X}_{t})+\beta(\nabla_X F(X_{t}, Y_{t}; \xi^{\scriptscriptstyle X}_{t})-\nabla_X F(X_{t-1}, Y_{t-1}; \xi^{\scriptscriptstyle X}_{t}))$,\\
        $M^{\scriptscriptstyle Y}_{t+1}=\beta M^{\scriptscriptstyle Y}_{t}+(1-\beta)\nabla_Y F(X_{t}, Y_{t}; \xi^{\scriptscriptstyle Y}_{t})+\beta(\nabla_Y F(X_{t}, Y_{t}; \xi^{\scriptscriptstyle Y}_{t})-\nabla_Y F(X_{t-1},Y_{t-1}; \xi^{\scriptscriptstyle Y}_{t}))$.\STATE Accumulate the gradient norms:\\
        $[m^{\scriptscriptstyle X}_{t+1}]^2 = [m^{\scriptscriptstyle X}_{t}]^2 + \min\{\|\nabla_X F(X_{t}, Y_{t}; \xi^{\scriptscriptstyle X}_{t})\|_F^2,\gamma^2\}$,\\
        $[m^{\scriptscriptstyle Y}_{t+1}]^2 = [m^{\scriptscriptstyle Y}_{t}]^2 + \min\{\|\nabla_Y F(X_{t}, Y_{t}; \xi^{\scriptscriptstyle Y}_{t})\|_F^2,\gamma^2\}$.
        \STATE Orthogonalize the momentum matrices:\\
         $O_{t+1}^{\scriptscriptstyle X} =\operatorname*{argmin}_{O\in\mathbb{R}^{m_x\times n_x}}\|O-M_{t+1}^{\scriptscriptstyle X}\|_F$,
        s.t. $O^\top O=I_{n_x}$ if $m_x\ge n_x$, else $OO^\top=I_{m_x}$,\\
        $O_{t+1}^{\scriptscriptstyle Y} =\operatorname*{argmin}_{O\in\mathbb{R}^{m_y\times n_y}}\|O-M_{t+1}^{\scriptscriptstyle Y}\|_F$,
        s.t. $O^\top O=I_{n_y}$ if $m_y\ge n_y$, else $OO^\top=I_{m_y}$.
        \STATE Define adaptive stepsizes:\\
        $\alpha^{\scriptscriptstyle Y}_{t+1}=\max\!\left\{\epsilon_{\scriptscriptstyle Y},\;\eta_{\scriptscriptstyle Y} \frac{\min\{\|\nabla_Y F(X_{t}, Y_{t}; \xi^{\scriptscriptstyle Y}_{t})\|_F,\gamma\}}{m_{t+1}^{\scriptscriptstyle Y}}\right\}$,\\
        $\alpha^{\scriptscriptstyle X}_{t+1}=\max\!\left\{\epsilon_{\scriptscriptstyle X},\;\eta_{\scriptscriptstyle X} \frac{\min\{\|\nabla_X F(X_{t}, Y_{t}; \xi^{\scriptscriptstyle X}_{t})\|_F,\gamma\}}{\max\{m_{t+1}^{\scriptscriptstyle X},m_{t+1}^{\scriptscriptstyle Y}\}}\right\}$.
        \STATE Update primal and dual variables:\\
        $Y_{t+1} = Y_{t} + \alpha^{\scriptscriptstyle Y}_{t+1} O^{\scriptscriptstyle Y}_{t+1}$,\\
        $X_{t+1} = X_{t} - \alpha^{\scriptscriptstyle X}_{t+1} O^{\scriptscriptstyle X}_{t+1}$.
    \ENDFOR
\end{algorithmic}
\end{algorithm}

According to \cref{descentlemma}, we have
\begin{align*}
&\frac{1}{T}\sum_{t=1}^T \mathbb{E}[\|\nabla\Phi(X_{t-1})\|_*]\notag\\
&\leq\frac{\Delta}{T\epsilon_{\scriptscriptstyle X}}+ {L_\Phi\epsilon_{\scriptscriptstyle X}}  + \frac{L_\Phi\eta^2_{\scriptscriptstyle X}}{T\epsilon_{\scriptscriptstyle X}}\left(\ln\left(\frac{\gamma^2 T}{z_0^2}\right) + 1\right) \notag\\
&\quad+ \frac{2}{T}\sum_{t=1}^T \mathbb{E}[\|M_t^{\scriptscriptstyle X} - \nabla\Phi(X_{t-1})\|_*] + \frac{2\sqrt{r}\eta_{\scriptscriptstyle X}}{T\epsilon_{\scriptscriptstyle X}}\sqrt{\ln\left(\frac{\gamma^2 T}{z_0^2}\right) + 1}\sqrt{\sum_{t=1}^T \mathbb{E}[\|M_t^{\scriptscriptstyle X} - \nabla\Phi(X_{t-1})\|_F^2]} ,
\end{align*}
where $\Delta:=\Phi(X_0) - \Phi_*$, $\Phi_* := \min_X \Phi(X)$, and $z_0:=\max\{m_0^{\scriptscriptstyle X},m_0^{\scriptscriptstyle Y}\}$. Then we derive
the following results on the terms $\frac{1}{T}\sum_{t=1}^T \mathbb{E}[\|M_t^{\scriptscriptstyle X} - \nabla\Phi(X_{t-1})\|_*]$ and $\sum_{t=1}^T \mathbb{E}[\|M_t^{\scriptscriptstyle X} - \nabla\Phi(X_{t-1})\|_F^2]$.

\begin{lemma}
Under \cref{smoothness,concavity,stochasticgradient}, for \cref{alg2}, suppose the total iteration rounds is $T$, the term $\frac{1}{T}\sum_{t=1}^T \mathbb{E}[\|M_t^{\scriptscriptstyle X} - \nabla\Phi(X_{t-1})\|_*]$ is upper bounded by:
\begin{align}
&\frac{1}{T}\sum_{t=1}^T \mathbb{E}[\|M_t^{\scriptscriptstyle X} - \nabla\Phi(X_{t-1})\|_*]\notag\\
&\le
\frac{\beta\sqrt r\,G}{T(1-\beta)}
+
\frac{\kappa\sqrt r}{T\sqrt{b_0}}+
\sqrt{\frac{r}{1-\beta^2}}\,
\Bigg(
\frac{2\beta^2L^2r
\epsilon_{\scriptscriptstyle X}^2}
{\min_{0\le s<T}b_s}+
\frac{2\beta^2L^2r
\eta_{\scriptscriptstyle X}^2}
{T\min_{0\le s<T}b_s}
\ln\left(
1+\frac{\gamma^2T}
{[\max\{m_0^{\scriptscriptstyle X},
         m_0^{\scriptscriptstyle Y}\}]^2}
\right)
\notag\\
&\quad+
\frac{2\beta^2L^2r
\epsilon_{\scriptscriptstyle Y}^2}
{\min_{0\le s<T}b_s}+
\frac{2\beta^2L^2r
\eta_{\scriptscriptstyle Y}^2}
{T\min_{0\le s<T}b_s}
\ln\left(
1+\frac{\gamma^2T}
{[m_0^{\scriptscriptstyle Y}]^2}
\right)+
\frac{2(1-\beta)^2}{T}
\sum_{t=1}^{T-1}\frac{\kappa^2}{b_t}
\Bigg)^{\frac12}
\notag\\
&\quad+
L\Bigg[
\frac{\Delta_0^{\scriptscriptstyle Y}}
{T\mu\epsilon_{\scriptscriptstyle Y}}
+
\frac{G\epsilon_{\scriptscriptstyle X}
\sqrt r(2+\lambda)}
{\mu\epsilon_{\scriptscriptstyle Y}}+
\frac{G\eta_{\scriptscriptstyle X}
\sqrt r(2+\lambda)}
{\mu\epsilon_{\scriptscriptstyle Y}\sqrt T}
\sqrt{
\ln\left(
1+\frac{\gamma^2T}
{[\max\{m_0^{\scriptscriptstyle X},
         m_0^{\scriptscriptstyle Y}\}]^2}
\right)
}
\notag\\
&\quad+
\frac{L\epsilon_{\scriptscriptstyle X}^2
(2+\lambda^2)}
{2\mu\epsilon_{\scriptscriptstyle Y}}
+
\frac{L\eta_{\scriptscriptstyle X}^2
(2+\lambda^2)}
{2T\mu\epsilon_{\scriptscriptstyle Y}}
\ln\left(
1+\frac{\gamma^2T}
{[\max\{m_0^{\scriptscriptstyle X},
         m_0^{\scriptscriptstyle Y}\}]^2}
\right)
\notag\\
&\quad+
\frac{L}{\mu\epsilon_{\scriptscriptstyle Y}}
\sqrt{
\left(
\epsilon_{\scriptscriptstyle X}^2
+\frac{\eta_{\scriptscriptstyle X}^2}{T}
\ln\left(
1+\frac{\gamma^2T}
{[\max\{m_0^{\scriptscriptstyle X},
         m_0^{\scriptscriptstyle Y}\}]^2}
\right)
\right)
\left(
\epsilon_{\scriptscriptstyle Y}^2
+\frac{\eta_{\scriptscriptstyle Y}^2}{T}
\ln\left(
1+\frac{\gamma^2T}
{[m_0^{\scriptscriptstyle Y}]^2}
\right)
\right)
}
\notag\\
&\quad+
\frac{Lr\epsilon_{\scriptscriptstyle Y}}{2\mu}
+
\frac{Lr\eta_{\scriptscriptstyle Y}^2}
{2T\mu\epsilon_{\scriptscriptstyle Y}}
\ln\left(
1+\frac{\gamma^2T}
{[m_0^{\scriptscriptstyle Y}]^2}
\right)+
\frac{2\sqrt r}{\mu\epsilon_{\scriptscriptstyle Y}}
\sqrt{
\epsilon_{\scriptscriptstyle Y}^2
+\frac{\eta_{\scriptscriptstyle Y}^2}{T}
\ln\left(
1+\frac{\gamma^2T}
{[m_0^{\scriptscriptstyle Y}]^2}
\right)
}
\notag\\
&\quad\cdot
\Bigg(
\frac{3\beta^2G^2}{T(1-\beta)}
+
\frac{3(1-\beta)\kappa^2}{Tb_0}
+
3\epsilon_{\scriptscriptstyle X}^2L^2r+
\frac{3\eta_{\scriptscriptstyle X}^2L^2r}{T}
\ln\left(
1+\frac{\gamma^2T}
{[\max\{m_0^{\scriptscriptstyle X},
         m_0^{\scriptscriptstyle Y}\}]^2}
\right)
+
3\epsilon_{\scriptscriptstyle Y}^2L^2r
\notag\\
&\qquad+
\frac{3\eta_{\scriptscriptstyle Y}^2L^2r}{T}
\ln\left(
1+\frac{\gamma^2T}
{[m_0^{\scriptscriptstyle Y}]^2}
\right)+
\frac{6\beta^2
\epsilon_{\scriptscriptstyle X}^2L^2r}
{(1-\beta^2)\min_{0\le s<T}b_s}
\notag\\
&\qquad+
\frac{6\beta^2
\eta_{\scriptscriptstyle X}^2L^2r}
{T(1-\beta^2)\min_{0\le s<T}b_s}
\ln\left(
1+\frac{\gamma^2T}
{[\max\{m_0^{\scriptscriptstyle X},
         m_0^{\scriptscriptstyle Y}\}]^2}
\right)+
\frac{6\beta^2
\epsilon_{\scriptscriptstyle Y}^2L^2r}
{(1-\beta^2)\min_{0\le s<T}b_s}
\notag\\
&\qquad+
\frac{6\beta^2
\eta_{\scriptscriptstyle Y}^2L^2r}
{T(1-\beta^2)\min_{0\le s<T}b_s}
\ln\left(
1+\frac{\gamma^2T}
{[m_0^{\scriptscriptstyle Y}]^2}
\right)+
\frac{6(1-\beta)}{T(1+\beta)}
\sum_{t=1}^{T-1}\frac{\kappa^2}{b_t}
\Bigg)^{\frac12}
\Bigg],
\end{align}
where $\Delta_0^{\scriptscriptstyle Y}:=\Phi(X_0)-f(X_0,Y_0).$
\label{e3lemma}
\end{lemma}

\begin{lemma}
Under \cref{smoothness,concavity,stochasticgradient}, for \cref{alg2}, suppose the total iteration rounds is $T$, the term $\sum_{t=1}^T \mathbb{E}[\|M_t^{\scriptscriptstyle X} - \nabla\Phi(X_{t-1})\|_F^2]$ can be upper bounded by:
\begin{align}
&\sum_{t=1}^T \mathbb{E}[\|M_t^{\scriptscriptstyle X} - \nabla\Phi(X_{t-1})\|_F^2]\notag\\
&\le
\frac{2\beta^2G^2}{1-\beta}
+
\frac{4L^2\Delta_0^{\scriptscriptstyle Y}}
{\mu(1-\beta)}
+
\frac{(1-\beta)\kappa^2}{b_0}
\notag\\
&\quad+
\left(
\frac{3\beta^2L_\Phi^2r}{(1-\beta)^2}
+
\frac{3\beta^2L^2r}{(1-\beta)^2}
+
\frac{2\beta^2L^2r}
{(1-\beta)\min_{0\le s<T}b_s}
\right)
\left(
T\epsilon_{\scriptscriptstyle X}^2
+
\eta_{\scriptscriptstyle X}^2
\ln\left(
1+\frac{\gamma^2T}
{[\max\{m_0^{\scriptscriptstyle X},
         m_0^{\scriptscriptstyle Y}\}]^2}
\right)
\right)
\notag\\
&\quad+
\left(
\frac{3\beta^2L^2r}{(1-\beta)^2}
+
\frac{2\beta^2L^2r}
{(1-\beta)\min_{0\le s<T}b_s}
\right)
\left(
T\epsilon_{\scriptscriptstyle Y}^2
+
\eta_{\scriptscriptstyle Y}^2
\ln\left(
1+\frac{\gamma^2T}
{[m_0^{\scriptscriptstyle Y}]^2}
\right)
\right)
+
2(1-\beta)
\sum_{t=1}^{T-1}\frac{\kappa^2}{b_t}
\notag\\
&\quad+
\frac{24GL^2\Delta_0^{\scriptscriptstyle Y}}
{7\mu^2\epsilon_{\scriptscriptstyle Y}}+
\frac{12GrL^2(L_\Phi+L)
\epsilon_{\scriptscriptstyle X}^2T}
{7\mu^2\epsilon_{\scriptscriptstyle Y}}
+
\frac{48r^2L^4G
\epsilon_{\scriptscriptstyle X}^2T}
{7\mu^3\epsilon_{\scriptscriptstyle Y}}
+
\frac{48rG^2L^4
\epsilon_{\scriptscriptstyle X}^2T}
{7\mu^4\epsilon_{\scriptscriptstyle Y}^2}
\notag\\
&\quad+
\left(
\frac{12GrL^2(L_\Phi+L)
\eta_{\scriptscriptstyle X}^2}
{7\mu^2\epsilon_{\scriptscriptstyle Y}}
+
\frac{48r^2L^4G
\eta_{\scriptscriptstyle X}^2}
{7\mu^3\epsilon_{\scriptscriptstyle Y}}
+
\frac{48rG^2L^4
\eta_{\scriptscriptstyle X}^2}
{7\mu^4\epsilon_{\scriptscriptstyle Y}^2}
\right)
\ln\left(
1+\frac{\gamma^2T}
{[\max\{m_0^{\scriptscriptstyle X},
         m_0^{\scriptscriptstyle Y}\}]^2}
\right)
\notag\\
&\quad+
\frac{3GL^2\epsilon_{\scriptscriptstyle Y}T}
{7\mu}
+
\frac{12GL^3r\epsilon_{\scriptscriptstyle Y}T}
{7\mu^2}+
\left(
\frac{3GL^2\eta_{\scriptscriptstyle Y}^2}
{7\mu\epsilon_{\scriptscriptstyle Y}}
+
\frac{12GL^3r\eta_{\scriptscriptstyle Y}^2}
{7\mu^2\epsilon_{\scriptscriptstyle Y}}
\right)
\ln\left(
1+\frac{\gamma^2T}
{[m_0^{\scriptscriptstyle Y}]^2}
\right)
\notag\\
&\quad+
\frac{48G\sqrt r\,L^2T}
{7\mu^2\epsilon_{\scriptscriptstyle Y}}
\sqrt{
\epsilon_{\scriptscriptstyle Y}^2
+\frac{\eta_{\scriptscriptstyle Y}^2}{T}
\ln\left(
1+\frac{\gamma^2T}
{[m_0^{\scriptscriptstyle Y}]^2}
\right)
}\cdot
\Bigg(
\frac{3\beta^2G^2}{T(1-\beta)}
+
\frac{3(1-\beta)\kappa^2}{Tb_0}
+
3\epsilon_{\scriptscriptstyle X}^2L^2r
\notag\\
&\quad+
\frac{3\eta_{\scriptscriptstyle X}^2L^2r}{T}
\ln\left(
1+\frac{\gamma^2T}
{[\max\{m_0^{\scriptscriptstyle X},
         m_0^{\scriptscriptstyle Y}\}]^2}
\right)
+
3\epsilon_{\scriptscriptstyle Y}^2L^2r+
\frac{3\eta_{\scriptscriptstyle Y}^2L^2r}{T}
\ln\left(
1+\frac{\gamma^2T}
{[m_0^{\scriptscriptstyle Y}]^2}
\right)
\notag\\
&\quad+
\frac{6\beta^2
\epsilon_{\scriptscriptstyle X}^2L^2r}
{(1-\beta^2)\min_{0\le s<T}b_s}+
\frac{6\beta^2
\eta_{\scriptscriptstyle X}^2L^2r}
{T(1-\beta^2)\min_{0\le s<T}b_s}
\ln\left(
1+\frac{\gamma^2T}
{[\max\{m_0^{\scriptscriptstyle X},
         m_0^{\scriptscriptstyle Y}\}]^2}
\right)
\notag\\
&\quad+
\frac{6\beta^2
\epsilon_{\scriptscriptstyle Y}^2L^2r}
{(1-\beta^2)\min_{0\le s<T}b_s}+
\frac{6\beta^2
\eta_{\scriptscriptstyle Y}^2L^2r}
{T(1-\beta^2)\min_{0\le s<T}b_s}
\ln\left(
1+\frac{\gamma^2T}
{[m_0^{\scriptscriptstyle Y}]^2}
\right)+
\frac{6(1-\beta)}{T(1+\beta)}
\sum_{t=1}^{T-1}\frac{\kappa^2}{b_t}
\Bigg)^{\frac12},\label{eq111Fnorm2}
\end{align}
where $\Delta_0^{\scriptscriptstyle Y}:=\Phi(X_0)-f(X_0,Y_0).$
\label{e4lemma}
\end{lemma}

The proof of \cref{e3lemma} is presented in \cref{sece3} and the proof of \cref{e4lemma} is presented in \cref{sece4}. In deriving these results, we will rely on the following lemma.
\begin{lemma}
Under \cref{smoothness,concavity,stochasticgradient}, for \cref{alg2}, suppose the total iteration rounds is $T$, we have:
\begin{align}
&\frac{1}{T}\sum_{t=1}^{T}\mathbb E[\|Y_t-Y^\star(X_t)\|_F]\notag\\
&\le
\frac{\Delta_0^{\scriptscriptstyle Y}}
{T\mu\epsilon_{\scriptscriptstyle Y}}
+
\frac{G\epsilon_{\scriptscriptstyle X}
\sqrt r(2+\lambda)}
{\mu\epsilon_{\scriptscriptstyle Y}}
+
\frac{G\eta_{\scriptscriptstyle X}
\sqrt r(2+\lambda)}
{\mu\epsilon_{\scriptscriptstyle Y}\sqrt T}
\sqrt{
\ln\left(
1+\frac{\gamma^2T}
{[\max\{m_0^{\scriptscriptstyle X},
         m_0^{\scriptscriptstyle Y}\}]^2}
\right)
}
\notag\\
&\quad+
\frac{L\epsilon_{\scriptscriptstyle X}^2
(2+\lambda^2)}
{2\mu\epsilon_{\scriptscriptstyle Y}}
+
\frac{L\eta_{\scriptscriptstyle X}^2
(2+\lambda^2)}
{2T\mu\epsilon_{\scriptscriptstyle Y}}
\ln\left(
1+\frac{\gamma^2T}
{[\max\{m_0^{\scriptscriptstyle X},
         m_0^{\scriptscriptstyle Y}\}]^2}
\right)
\notag\\
&\quad+
\frac{L}{\mu\epsilon_{\scriptscriptstyle Y}}
\sqrt{
\left(
\epsilon_{\scriptscriptstyle X}^2
+\frac{\eta_{\scriptscriptstyle X}^2}{T}
\ln\left(
1+\frac{\gamma^2T}
{[\max\{m_0^{\scriptscriptstyle X},
         m_0^{\scriptscriptstyle Y}\}]^2}
\right)
\right)
\left(
\epsilon_{\scriptscriptstyle Y}^2
+\frac{\eta_{\scriptscriptstyle Y}^2}{T}
\ln\left(
1+\frac{\gamma^2T}
{[m_0^{\scriptscriptstyle Y}]^2}
\right)
\right)
}
\notag\\
&\quad+
\frac{Lr\epsilon_{\scriptscriptstyle Y}}{2\mu}
+
\frac{Lr\eta_{\scriptscriptstyle Y}^2}
{2T\mu\epsilon_{\scriptscriptstyle Y}}
\ln\left(
1+\frac{\gamma^2T}
{[m_0^{\scriptscriptstyle Y}]^2}
\right)+
\frac{2\sqrt r}{\mu\epsilon_{\scriptscriptstyle Y}}
\sqrt{
\epsilon_{\scriptscriptstyle Y}^2
+\frac{\eta_{\scriptscriptstyle Y}^2}{T}
\ln\left(
1+\frac{\gamma^2T}
{[m_0^{\scriptscriptstyle Y}]^2}
\right)
}
\notag\\
&\quad\cdot
\Bigg(
\frac{3\beta^2G^2}{T(1-\beta)}
+
\frac{3(1-\beta)\kappa^2}{Tb_0}
+
3\epsilon_{\scriptscriptstyle X}^2L^2r
+
\frac{3\eta_{\scriptscriptstyle X}^2L^2r}{T}
\ln\left(
1+\frac{\gamma^2T}
{[\max\{m_0^{\scriptscriptstyle X},
         m_0^{\scriptscriptstyle Y}\}]^2}
\right)
+
3\epsilon_{\scriptscriptstyle Y}^2L^2r
\notag\\
&\qquad+
\frac{3\eta_{\scriptscriptstyle Y}^2L^2r}{T}
\ln\left(
1+\frac{\gamma^2T}
{[m_0^{\scriptscriptstyle Y}]^2}
\right)+
\frac{6\beta^2
\epsilon_{\scriptscriptstyle X}^2L^2r}
{(1-\beta^2)\min_{0\le s<T}b_s}
\notag\\
&\qquad+
\frac{6\beta^2
\eta_{\scriptscriptstyle X}^2L^2r}
{T(1-\beta^2)\min_{0\le s<T}b_s}
\ln\left(
1+\frac{\gamma^2T}
{[\max\{m_0^{\scriptscriptstyle X},
         m_0^{\scriptscriptstyle Y}\}]^2}
\right)
+
\frac{6\beta^2
\epsilon_{\scriptscriptstyle Y}^2L^2r}
{(1-\beta^2)\min_{0\le s<T}b_s}
\notag\\
&\qquad+
\frac{6\beta^2
\eta_{\scriptscriptstyle Y}^2L^2r}
{T(1-\beta^2)\min_{0\le s<T}b_s}
\ln\left(
1+\frac{\gamma^2T}
{[m_0^{\scriptscriptstyle Y}]^2}
\right)
+
\frac{6(1-\beta)}{T(1+\beta)}
\sum_{t=1}^{T-1}\frac{\kappa^2}{b_t}
\Bigg)^{\frac12},
\end{align}
and
\begin{align}
&\frac1T\sum_{t=1}^{T}\mathbb E[\|Y_t-Y^\star(X_t)\|_F^2]\notag\\
&\le
\frac{8G\Delta_0^{\scriptscriptstyle Y}}
{7\mu^2\epsilon_{\scriptscriptstyle Y}T}
+
\frac{4Gr\epsilon_{\scriptscriptstyle X}^2
(L_\Phi+L)}
{7\mu^2\epsilon_{\scriptscriptstyle Y}}
+
\frac{16r^2L^2G
\epsilon_{\scriptscriptstyle X}^2}
{7\mu^3\epsilon_{\scriptscriptstyle Y}}
+
\frac{16rG^2L^2
\epsilon_{\scriptscriptstyle X}^2}
{7\mu^4\epsilon_{\scriptscriptstyle Y}^2}
\notag\\
&\quad+
\left(
\frac{4Gr\eta_{\scriptscriptstyle X}^2
(L_\Phi+L)}
{7T\mu^2\epsilon_{\scriptscriptstyle Y}}
+
\frac{16r^2L^2G
\eta_{\scriptscriptstyle X}^2}
{7T\mu^3\epsilon_{\scriptscriptstyle Y}}
+
\frac{16rG^2L^2
\eta_{\scriptscriptstyle X}^2}
{7T\mu^4\epsilon_{\scriptscriptstyle Y}^2}
\right)
\ln\left(
1+\frac{\gamma^2T}
{[\max\{m_0^{\scriptscriptstyle X},
         m_0^{\scriptscriptstyle Y}\}]^2}
\right)
\notag\\
&\quad+
\frac{G\epsilon_{\scriptscriptstyle Y}}{7\mu}
+
\frac{4GLr\epsilon_{\scriptscriptstyle Y}}{7\mu^2}+
\left(
\frac{G\eta_{\scriptscriptstyle Y}^2}
{7T\mu\epsilon_{\scriptscriptstyle Y}}
+
\frac{4GLr\eta_{\scriptscriptstyle Y}^2}
{7T\mu^2\epsilon_{\scriptscriptstyle Y}}
\right)
\ln\left(
1+\frac{\gamma^2T}
{[m_0^{\scriptscriptstyle Y}]^2}
\right)
\notag\\
&\quad+
\frac{16G\sqrt r}
{7\mu^2\epsilon_{\scriptscriptstyle Y}}
\sqrt{
\epsilon_{\scriptscriptstyle Y}^2
+\frac{\eta_{\scriptscriptstyle Y}^2}{T}
\ln\left(
1+\frac{\gamma^2T}
{[m_0^{\scriptscriptstyle Y}]^2}
\right)
}
\notag\\
&\quad\cdot
\Bigg(
\frac{3\beta^2G^2}{T(1-\beta)}
+\frac{3(1-\beta)\kappa^2}{Tb_0}
+
3\epsilon_{\scriptscriptstyle X}^2L^2r
+
\frac{3\eta_{\scriptscriptstyle X}^2L^2r}{T}
\ln\left(
1+\frac{\gamma^2T}
{[\max\{m_0^{\scriptscriptstyle X},
         m_0^{\scriptscriptstyle Y}\}]^2}
\right)
+
3\epsilon_{\scriptscriptstyle Y}^2L^2r
\notag\\
&\qquad+
\frac{3\eta_{\scriptscriptstyle Y}^2L^2r}{T}
\ln\left(
1+\frac{\gamma^2T}
{[m_0^{\scriptscriptstyle Y}]^2}
\right)
+
\frac{6\beta^2
\epsilon_{\scriptscriptstyle X}^2L^2r}
{(1-\beta^2)\min_{0\le s<T}b_s}
\notag\\
&\qquad+
\frac{6\beta^2
\eta_{\scriptscriptstyle X}^2L^2r}
{T(1-\beta^2)\min_{0\le s<T}b_s}
\ln\left(
1+\frac{\gamma^2T}
{[\max\{m_0^{\scriptscriptstyle X},
         m_0^{\scriptscriptstyle Y}\}]^2}
\right)
+
\frac{6\beta^2
\epsilon_{\scriptscriptstyle Y}^2L^2r}
{(1-\beta^2)\min_{0\le s<T}b_s}
\notag\\
&\qquad+
\frac{6\beta^2
\eta_{\scriptscriptstyle Y}^2L^2r}
{T(1-\beta^2)\min_{0\le s<T}b_s}
\ln\left(
1+\frac{\gamma^2T}
{[m_0^{\scriptscriptstyle Y}]^2}
\right)
+
\frac{6(1-\beta)}{T(1+\beta)}
\sum_{t=1}^{T-1}\frac{\kappa^2}{b_t}
\Bigg)^{\frac12},
\end{align}
where $\Delta_0^{\scriptscriptstyle Y}:=\Phi(X_0)-f(X_0,Y_0).$
\label{lemmay2}
\end{lemma}
\begin{proof}
We first define
$\sigma_{t+1}:=
\frac{\min\{\|\nabla_Y F(X_t,Y_t;\xi_t^{\scriptscriptstyle Y})\|_F,\gamma\}}
{m_{t+1}^{\scriptscriptstyle Y}}$
and
$\Delta_t^{\scriptscriptstyle Y}:=
\Phi(X_t)-f(X_t,Y_t)\ge 0$.
Then we use the tracker
\begin{equation}
M_{t+1}^{\scriptscriptstyle Y}
=
\beta M_t^{\scriptscriptstyle Y}
+(1-\beta)\nabla_Y F(X_t,Y_t;\xi_t^{\scriptscriptstyle Y})
+\beta\big(
\nabla_Y F(X_t,Y_t;\xi_t^{\scriptscriptstyle Y})
-\nabla_Y F(X_{t-1},Y_{t-1};\xi_t^{\scriptscriptstyle Y})
\big),
\label{eq:storm-tracker-y0}
\end{equation}
and define
\begin{equation}
\tilde\Delta_t^{\scriptscriptstyle Y}
:=
\big(
\nabla_Y F(X_t,Y_t;\xi_t^{\scriptscriptstyle Y})
-\nabla_Y f(X_t,Y_t)
\big)
-\beta\big(
\nabla_Y F(X_{t-1},Y_{t-1};\xi_t^{\scriptscriptstyle Y})
-\nabla_Y f(X_{t-1},Y_{t-1})
\big).
\label{eq:Delta-storm-y}
\end{equation}
Then we have
\begin{align}
&M_{t+1}^{\scriptscriptstyle Y}\notag\\
&=
\beta M_t^{\scriptscriptstyle Y}
+(1-\beta)\nabla_Y F(X_t,Y_t;\xi_t^{\scriptscriptstyle Y})
+\beta\big(
\nabla_Y F(X_t,Y_t;\xi_t^{\scriptscriptstyle Y})
-\nabla_Y F(X_{t-1},Y_{t-1};\xi_t^{\scriptscriptstyle Y})
\big)\notag\\
&=
\beta M_t^{\scriptscriptstyle Y}
+\nabla_Y f(X_t,Y_t)
-\beta\nabla_Y f(X_{t-1},Y_{t-1})\notag\\
&\quad+
\left(
\big(
\nabla_Y F(X_t,Y_t;\xi_t^{\scriptscriptstyle Y})
-\nabla_Y f(X_t,Y_t)
\big)
-\beta\big(
\nabla_Y F(X_{t-1},Y_{t-1};\xi_t^{\scriptscriptstyle Y})
-\nabla_Y f(X_{t-1},Y_{t-1})
\big)
\right)\notag\\
&=
\beta M_t^{\scriptscriptstyle Y}
+\nabla_Y f(X_t,Y_t)
-\beta\nabla_Y f(X_{t-1},Y_{t-1})
+\tilde\Delta_t^{\scriptscriptstyle Y}.
\label{eq:storm-tracker-y}
\end{align}

Define
\(
\bar F_{t+1}^{\scriptscriptstyle Y}
:=
M_{t+1}^{\scriptscriptstyle Y}
-\nabla_Y f(X_t,Y_t),
\)
so that
\(
M_{t+1}^{\scriptscriptstyle Y}
=
\nabla_Y f(X_t,Y_t)
+\bar F_{t+1}^{\scriptscriptstyle Y}.
\)
Under the same orthogonalization property of \(O_{t+1}^{\scriptscriptstyle Y}\) and the same smoothness/concavity assumptions as in \cref{lemmay}, the subsequent analysis follows the same line of reasoning. We first provide an upper bound for
\(
\frac{1}{T}\sum_{t=0}^{T-1}\mathbb{E}[\|\bar F_{t+1}^{\scriptscriptstyle Y}\|_F^2].
\)

\paragraph{Upper Bound for $\frac{1}{T}\sum_{t=0}^{T-1}\mathbb E[\|\bar F_{t+1}^{\scriptscriptstyle Y}\|_F^2]$.}
Let us denote
$\bar F_{t+1}^{\scriptscriptstyle Y}
:=M_{t+1}^{\scriptscriptstyle Y}-\nabla_Y f(X_t,Y_t)$.
Then we can get, for $t\ge 1$,
\begin{align}
\bar F_{t+1}^{\scriptscriptstyle Y}
=
\beta\big(
M_t^{\scriptscriptstyle Y}
-\nabla_Y f(X_{t-1},Y_{t-1})
\big)
+\tilde\Delta_t^{\scriptscriptstyle Y}=
\beta\bar F_t^{\scriptscriptstyle Y}
+\tilde\Delta_t^{\scriptscriptstyle Y}.
\label{eq:Fbar-rec}
\end{align}
Iterating \cref{eq:Fbar-rec} gives, for any $t\ge 0$,
\begin{equation}
\bar F_{t+1}^{\scriptscriptstyle Y}
=
\beta^t\bar F_1^{\scriptscriptstyle Y}
+\sum_{\tau=0}^{t-1}
\beta^\tau\tilde\Delta_{t-\tau}^{\scriptscriptstyle Y}.
\label{eq:Fbar-unroll}
\end{equation}
Then we have
\begin{align}
&\mathbb E\big[
\|\tilde\Delta_t^{\scriptscriptstyle Y}\|_F^2
\big]
\notag\\
&=
\mathbb E\Big[
\Big\|
\big(
\nabla_Y F(X_t,Y_t;\xi_t^{\scriptscriptstyle Y})
-\nabla_Y f(X_t,Y_t)
\big)
-\beta\big(
\nabla_Y F(X_{t-1},Y_{t-1};\xi_t^{\scriptscriptstyle Y})
-\nabla_Y f(X_{t-1},Y_{t-1})
\big)
\Big\|_F^2
\Big]
\notag\\
&=
\mathbb E\Big[
\Big\|
(1-\beta)\big(
\nabla_Y F(X_t,Y_t;\xi_t^{\scriptscriptstyle Y})
-\nabla_Y f(X_t,Y_t)
\big)
+\beta\Big(
\nabla_Y F(X_t,Y_t;\xi_t^{\scriptscriptstyle Y})
-\nabla_Y F(X_{t-1},Y_{t-1};\xi_t^{\scriptscriptstyle Y})
\notag\\
&\qquad\quad
-\nabla_Y f(X_t,Y_t)
+\nabla_Y f(X_{t-1},Y_{t-1})
\Big)
\Big\|_F^2
\Big]
\notag\\
&\le
2(1-\beta)^2
\mathbb E\Big[
\big\|
\nabla_Y F(X_t,Y_t;\xi_t^{\scriptscriptstyle Y})
-\nabla_Y f(X_t,Y_t)
\big\|_F^2
\Big]
\notag\\
&\quad+
2\beta^2
\mathbb E\Big[
\Big\|
\nabla_Y F(X_t,Y_t;\xi_t^{\scriptscriptstyle Y})
-\nabla_Y F(X_{t-1},Y_{t-1};\xi_t^{\scriptscriptstyle Y})
-\nabla_Y f(X_t,Y_t)
+\nabla_Y f(X_{t-1},Y_{t-1})
\Big\|_F^2
\Big]
\notag\\
&=
2(1-\beta)^2
\mathbb E\Big[
\big\|
\nabla_Y F(X_t,Y_t;\xi_t^{\scriptscriptstyle Y})
-\nabla_Y f(X_t,Y_t)
\big\|_F^2
\Big]
\notag\\
&\quad+
\frac{2\beta^2}{b_t}
\mathbb E\Big[
\Big\|
\nabla_Y F(X_t,Y_t;\xi_{t,1}^{\scriptscriptstyle Y})
-\nabla_Y F(X_{t-1},Y_{t-1};
             \xi_{t,1}^{\scriptscriptstyle Y})
-\nabla_Y f(X_t,Y_t)
+\nabla_Y f(X_{t-1},Y_{t-1})
\Big\|_F^2
\Big]
\notag\\
&\overset{(a)}{\le}
2(1-\beta)^2
\mathbb E\Big[
\big\|
\nabla_Y F(X_t,Y_t;\xi_t^{\scriptscriptstyle Y})
-\nabla_Y f(X_t,Y_t)
\big\|_F^2
\Big]
\notag\\
&\quad+
\frac{2\beta^2}{b_t}
\mathbb E\Big[
\big\|
\nabla_Y F(X_t,Y_t;\xi_{t,1}^{\scriptscriptstyle Y})
-\nabla_Y F(X_{t-1},Y_{t-1};
             \xi_{t,1}^{\scriptscriptstyle Y})
\big\|_F^2
\Big]
\notag\\
&\overset{(b)}{\le}
\frac{2\beta^2L^2}{b_t}
\mathbb E[\|X_t-X_{t-1}\|_F^2]
+
\frac{2\beta^2L^2}{b_t}
\mathbb E[\|Y_t-Y_{t-1}\|_F^2]
+
\frac{2(1-\beta)^2\kappa^2}{b_t}
\notag\\
&\overset{(c)}{\le}
\frac{2\beta^2L^2r}{b_t}
\mathbb E\left[
\big(\max\{
\epsilon_{\scriptscriptstyle X},
\eta_{\scriptscriptstyle X}\theta_t
\}\big)^2
\right]+
\frac{2\beta^2L^2r}{b_t}
\mathbb E\left[
\big(\max\{
\epsilon_{\scriptscriptstyle Y},
\eta_{\scriptscriptstyle Y}\sigma_t
\}\big)^2
\right]
+
\frac{2(1-\beta)^2\kappa^2}{b_t}.
\label{eq:Delta-y-step1}
\end{align}
Inequality (a) follows from the variance decomposition
\(
\mathbb E[\|Z\|^2]
-\|\mathbb E[Z]\|^2
=\mathrm{Var}(Z)\ge 0,
\)
and hence
\(
\mathbb E[\|Z\|^2]
-\|\mathbb E[Z]\|^2
\le \mathbb E[\|Z\|^2].
\)
This decomposition is applied conditionally on the history
before sampling the current mini-batch, and then averaged.
Inequality (b) follows from the mean-square smoothness in
\cref{smoothness} and the variance bound in
\cref{stochasticgradient}.
Inequality (c) follows from the update forms
$X_t-X_{t-1}
=-\max\{
\epsilon_{\scriptscriptstyle X},
\eta_{\scriptscriptstyle X}\theta_t
\}O_t^{\scriptscriptstyle X}$
and
$Y_t-Y_{t-1}
=\max\{
\epsilon_{\scriptscriptstyle Y},
\eta_{\scriptscriptstyle Y}\sigma_t
\}O_t^{\scriptscriptstyle Y}$,
together with
$\|O_t^{\scriptscriptstyle X}\|_F^2\le r$
and
$\|O_t^{\scriptscriptstyle Y}\|_F^2\le r$
under the global rank bound $r$.

Define
$D_t:=
\sum_{\tau=0}^{t-1}\beta^\tau
\big(
\nabla_Y f(X_{t-\tau},Y_{t-\tau})
-\nabla_Y f(X_{t-\tau-1},Y_{t-\tau-1})
\big)$
and
$U_t:=
\sum_{\tau=0}^{t-1}
\beta^\tau\tilde\Delta_{t-\tau}^{\scriptscriptstyle Y}$.
From \cref{eq:Fbar-unroll} and
$\|A+B\|^2\le2(\|A\|^2+\|B\|^2)$, we have
\begin{equation}
\begin{aligned}
\|\bar F_{t+1}^{\scriptscriptstyle Y}\|_F^2
&\le
2\beta^{2t}\|\bar F_1^{\scriptscriptstyle Y}\|_F^2
+2\|U_t\|_F^2\le
3\beta^{2t}\|\bar F_1^{\scriptscriptstyle Y}\|_F^2
+3(1-\beta)^2\|D_t\|_F^2
+3\|U_t\|_F^2.
\end{aligned}
\label{eq:Fbar-sq-split}
\end{equation}

Summing the first term on the right-hand side of
\cref{eq:Fbar-sq-split} over $t=0,\ldots,T-1$
and taking expectations gives
\begin{equation}
\begin{aligned}
\sum_{t=0}^{T-1}
3\beta^{2t}\,
\mathbb E[\|\bar F_1^{\scriptscriptstyle Y}\|_F^2]
&\le
\frac{3}{1-\beta^2}
\mathbb E[\|\bar F_1^{\scriptscriptstyle Y}\|_F^2]\\
&\le
\frac{3}{1-\beta}
\mathbb E[\|\bar F_1^{\scriptscriptstyle Y}\|_F^2]\\
&\le
\frac{3\beta^2G^2}{1-\beta}
+\frac{3(1-\beta)\kappa^2}{b_0}.
\end{aligned}
\label{F1upperbound}
\end{equation}
Here the initialization
$M_0^{\scriptscriptstyle Y}=0$,
$X_{-1}=X_0$, and $Y_{-1}=Y_0$ gives
$M_1^{\scriptscriptstyle Y}
=(1-\beta)\nabla_YF(X_0,Y_0;\xi_0^{\scriptscriptstyle Y})$.
Consequently, conditional unbiasedness and
\cref{stochasticgradient} imply
\[
\begin{aligned}
\mathbb E[\|\bar F_1^{\scriptscriptstyle Y}\|_F^2]
&=
\beta^2\|\nabla_Y f(X_0,Y_0)\|_F^2+
(1-\beta)^2
\mathbb E\left[
\left\|
\nabla_YF(X_0,Y_0;\xi_0^{\scriptscriptstyle Y})
-\nabla_Yf(X_0,Y_0)
\right\|_F^2
\right]\\
&\le
\beta^2G^2+\frac{(1-\beta)^2\kappa^2}{b_0}.
\end{aligned}
\]

Since we have
\(
D_t:=
\sum_{\tau=0}^{t-1}\beta^\tau
\big(
\nabla_Y f(X_{t-\tau},Y_{t-\tau})
-\nabla_Y f(X_{t-\tau-1},Y_{t-\tau-1})
\big),
\)
by the weighted Cauchy--Schwarz inequality,
\(
\left\Vert
\sum_{\tau=0}^{t-1}\beta^\tau V_\tau
\right\Vert_F^2
\le
\Big(\sum_{\tau=0}^{t-1}\beta^\tau\Big)
\Big(\sum_{\tau=0}^{t-1}
\beta^\tau\|V_\tau\|_F^2\Big)
\le
\frac{1}{1-\beta}
\sum_{\tau=0}^{t-1}
\beta^\tau\|V_\tau\|_F^2,
\)
we have
\begin{align}
\mathbb E[\|D_t\|_F^2]
&\le
\frac{1}{1-\beta}
\sum_{\tau=0}^{t-1}\beta^\tau\,
\mathbb E\left[
\left\|
\nabla_Y f(X_{t-\tau},Y_{t-\tau})
-\nabla_Y f(X_{t-\tau-1},Y_{t-\tau-1})
\right\|_F^2
\right].
\label{eq:Dt_weighted_single}
\end{align}
Summing \eqref{eq:Dt_weighted_single} over
$t=0,\dots,T-1$ and swapping the order of summation yields
\begin{align}
\sum_{t=0}^{T-1}\mathbb E[\|D_t\|_F^2]
&\le
\frac{1}{1-\beta}
\sum_{t=0}^{T-1}\sum_{\tau=0}^{t-1}\beta^\tau\,
\mathbb E\left[
\left\|
\nabla_Y f(X_{t-\tau},Y_{t-\tau})
-\nabla_Y f(X_{t-\tau-1},Y_{t-\tau-1})
\right\|_F^2
\right]
\notag\\
&=
\frac{1}{1-\beta}
\sum_{t=1}^{T-1}
\left(\sum_{j=0}^{T-1-t}\beta^j\right)
\mathbb E\left[
\left\|
\nabla_Y f(X_t,Y_t)
-\nabla_Y f(X_{t-1},Y_{t-1})
\right\|_F^2
\right]
\notag\\
&\le
\frac{1}{(1-\beta)^2}
\sum_{t=1}^{T-1}
\mathbb E\left[
\left\|
\nabla_Y f(X_t,Y_t)
-\nabla_Y f(X_{t-1},Y_{t-1})
\right\|_F^2
\right]
\notag\\
&\overset{(a)}{\le}
\frac{L^2}{(1-\beta)^2}
\sum_{t=1}^{T-1}
\mathbb E\left[
\|X_t-X_{t-1}\|_F^2
+\|Y_t-Y_{t-1}\|_F^2
\right]
\notag\\
&\overset{(b)}{\le}
\frac{L^2r}{(1-\beta)^2}
\sum_{t=1}^{T-1}
\left(
\mathbb E\left[
\big(\max\{
\epsilon_{\scriptscriptstyle X},
\eta_{\scriptscriptstyle X}\theta_t
\}\big)^2
\right]
+
\mathbb E\left[
\big(\max\{
\epsilon_{\scriptscriptstyle Y},
\eta_{\scriptscriptstyle Y}\sigma_t
\}\big)^2
\right]
\right)
\notag\\
&\overset{(c)}{\le}
\frac{T\epsilon_{\scriptscriptstyle X}^2L^2r}
{(1-\beta)^2}
+
\frac{\eta_{\scriptscriptstyle X}^2L^2r}
{(1-\beta)^2}
\ln\left(
1+\frac{\gamma^2T}
{[\max\{m_0^{\scriptscriptstyle X},
         m_0^{\scriptscriptstyle Y}\}]^2}
\right)
\notag\\
&\quad+
\frac{T\epsilon_{\scriptscriptstyle Y}^2L^2r}
{(1-\beta)^2}
+
\frac{\eta_{\scriptscriptstyle Y}^2L^2r}
{(1-\beta)^2}
\ln\left(
1+\frac{\gamma^2T}
{[m_0^{\scriptscriptstyle Y}]^2}
\right).
\label{eq:Dt-sum}
\end{align}
Here (a) follows from \cref{smoothness},
and (b) uses
$\|X_t-X_{t-1}\|_F^2
\le r(\max\{
\epsilon_{\scriptscriptstyle X},
\eta_{\scriptscriptstyle X}\theta_t
\})^2$
and
$\|Y_t-Y_{t-1}\|_F^2
\le r(\max\{
\epsilon_{\scriptscriptstyle Y},
\eta_{\scriptscriptstyle Y}\sigma_t
\})^2$
under the global rank bound $r$.
For the initialization
$m_0^{\scriptscriptstyle X}
=m_0^{\scriptscriptstyle Y}=c>0$,
(c) uses \cref{lem:log_cum_ratio}, applied to the sequence
beginning with the positive initialization $c^2$,
followed by the clipped squared gradient norms.

By conditional unbiasedness,
$\tilde\Delta_t^{\scriptscriptstyle Y}$
has zero conditional mean given the history before sampling
the current mini-batch.
By martingale orthogonality, we have
$\mathbb E[\|U_t\|_F^2]
=\sum_{\tau=0}^{t-1}
\beta^{2\tau}
\mathbb E[
\|\tilde\Delta_{t-\tau}^{\scriptscriptstyle Y}\|_F^2
]$.
Moreover, in what follows we invoke the global rank bound $r$
to upper bound
$\|X_t-X_{t-1}\|_F^2$
and
$\|Y_t-Y_{t-1}\|_F^2$ below.
Combining this identity with \cref{lem:log_cum_ratio} yields
\begin{align}
&\sum_{t=0}^{T-1}\mathbb E[\|U_t\|_F^2]
\notag\\
&=
\sum_{t=0}^{T-1}\sum_{\tau=0}^{t-1}
\beta^{2\tau}
\mathbb E\left[
\|\tilde\Delta_{t-\tau}^{\scriptscriptstyle Y}\|_F^2
\right]
\notag\\
&=
\sum_{k=1}^{T-1}
\left(\sum_{j=0}^{T-1-k}\beta^{2j}\right)
\mathbb E\left[
\|\tilde\Delta_k^{\scriptscriptstyle Y}\|_F^2
\right]
\notag\\
&\le
\frac{1}{1-\beta^2}
\sum_{t=1}^{T-1}
\mathbb E\left[
\|\tilde\Delta_t^{\scriptscriptstyle Y}\|_F^2
\right]
\notag\\
&\le
\frac{2\beta^2L^2}{1-\beta^2}
\sum_{t=1}^{T-1}
\frac{\mathbb E[\|X_t-X_{t-1}\|_F^2]}{b_t}
+
\frac{2\beta^2L^2}{1-\beta^2}
\sum_{t=1}^{T-1}
\frac{\mathbb E[\|Y_t-Y_{t-1}\|_F^2]}{b_t}
\notag\\
&\quad+
\frac{2(1-\beta)^2}{1-\beta^2}
\sum_{t=1}^{T-1}\frac{\kappa^2}{b_t}
\notag\\
&\le
\frac{2\beta^2L^2r}
{(1-\beta^2)\min_{0\le s<T}b_s}
\sum_{t=1}^{T-1}
\left(
\mathbb E\left[
\big(\max\{
\epsilon_{\scriptscriptstyle X},
\eta_{\scriptscriptstyle X}\theta_t
\}\big)^2
\right]
+
\mathbb E\left[
\big(\max\{
\epsilon_{\scriptscriptstyle Y},
\eta_{\scriptscriptstyle Y}\sigma_t
\}\big)^2
\right]
\right)
\notag\\
&\quad+
\frac{2(1-\beta)^2}{1-\beta^2}
\sum_{t=1}^{T-1}\frac{\kappa^2}{b_t}
\notag\\
&\le
\frac{2\beta^2T
\epsilon_{\scriptscriptstyle X}^2L^2r}
{(1-\beta^2)\min_{0\le s<T}b_s}
+
\frac{2\beta^2
\eta_{\scriptscriptstyle X}^2L^2r}
{(1-\beta^2)\min_{0\le s<T}b_s}
\ln\left(
1+\frac{\gamma^2T}
{[\max\{m_0^{\scriptscriptstyle X},
         m_0^{\scriptscriptstyle Y}\}]^2}
\right)
\notag\\
&\quad+
\frac{2\beta^2T
\epsilon_{\scriptscriptstyle Y}^2L^2r}
{(1-\beta^2)\min_{0\le s<T}b_s}
+
\frac{2\beta^2
\eta_{\scriptscriptstyle Y}^2L^2r}
{(1-\beta^2)\min_{0\le s<T}b_s}
\ln\left(
1+\frac{\gamma^2T}
{[m_0^{\scriptscriptstyle Y}]^2}
\right)+
\frac{2(1-\beta)^2}{1-\beta^2}
\sum_{t=1}^{T-1}\frac{\kappa^2}{b_t}.
\label{eq:St-sum}
\end{align}

Summing \cref{eq:Fbar-sq-split} over $t=0,\ldots,T-1$
and dividing by $T$, and subsequently substituting
\cref{F1upperbound}, \cref{eq:Dt-sum}, and
\cref{eq:St-sum} into the resulting bound, we obtain
\begin{align}
&\frac{1}{T}\sum_{t=0}^{T-1}
\mathbb E[
\|\bar F_{t+1}^{\scriptscriptstyle Y}\|_F^2
]
\notag\\
&\le
\frac{3\mathbb E[
\|\bar F_1^{\scriptscriptstyle Y}\|_F^2
]}{T(1-\beta)}
+
3\epsilon_{\scriptscriptstyle X}^2L^2r
+
\frac{3\eta_{\scriptscriptstyle X}^2L^2r}{T}
\ln\left(
1+\frac{\gamma^2T}
{[\max\{m_0^{\scriptscriptstyle X},
         m_0^{\scriptscriptstyle Y}\}]^2}
\right)
+
3\epsilon_{\scriptscriptstyle Y}^2L^2r
\notag\\
&\quad+
\frac{3\eta_{\scriptscriptstyle Y}^2L^2r}{T}
\ln\left(
1+\frac{\gamma^2T}
{[m_0^{\scriptscriptstyle Y}]^2}
\right)+
\frac{6\beta^2
\epsilon_{\scriptscriptstyle X}^2L^2r}
{(1-\beta^2)\min_{0\le s<T}b_s}
\notag\\
&\quad+
\frac{6\beta^2
\eta_{\scriptscriptstyle X}^2L^2r}
{T(1-\beta^2)\min_{0\le s<T}b_s}
\ln\left(
1+\frac{\gamma^2T}
{[\max\{m_0^{\scriptscriptstyle X},
         m_0^{\scriptscriptstyle Y}\}]^2}
\right)
\notag\\
&\quad+
\frac{6\beta^2
\epsilon_{\scriptscriptstyle Y}^2L^2r}
{(1-\beta^2)\min_{0\le s<T}b_s}+
\frac{6\beta^2
\eta_{\scriptscriptstyle Y}^2L^2r}
{T(1-\beta^2)\min_{0\le s<T}b_s}
\ln\left(
1+\frac{\gamma^2T}
{[m_0^{\scriptscriptstyle Y}]^2}
\right)
+
\frac{6(1-\beta)}{T(1+\beta)}
\sum_{t=1}^{T-1}\frac{\kappa^2}{b_t}
\notag\\
&\le
\frac{3\beta^2G^2}{T(1-\beta)}
+\frac{3(1-\beta)\kappa^2}{Tb_0}
+
3\epsilon_{\scriptscriptstyle X}^2L^2r
+
\frac{3\eta_{\scriptscriptstyle X}^2L^2r}{T}
\ln\left(
1+\frac{\gamma^2T}
{[\max\{m_0^{\scriptscriptstyle X},
         m_0^{\scriptscriptstyle Y}\}]^2}
\right)
+
3\epsilon_{\scriptscriptstyle Y}^2L^2r
\notag\\
&\quad+
\frac{3\eta_{\scriptscriptstyle Y}^2L^2r}{T}
\ln\left(
1+\frac{\gamma^2T}
{[m_0^{\scriptscriptstyle Y}]^2}
\right)+
\frac{6\beta^2
\epsilon_{\scriptscriptstyle X}^2L^2r}
{(1-\beta^2)\min_{0\le s<T}b_s}
\notag\\
&\quad+
\frac{6\beta^2
\eta_{\scriptscriptstyle X}^2L^2r}
{T(1-\beta^2)\min_{0\le s<T}b_s}
\ln\left(
1+\frac{\gamma^2T}
{[\max\{m_0^{\scriptscriptstyle X},
         m_0^{\scriptscriptstyle Y}\}]^2}
\right)+
\frac{6\beta^2
\epsilon_{\scriptscriptstyle Y}^2L^2r}
{(1-\beta^2)\min_{0\le s<T}b_s}
\notag\\
&\quad+
\frac{6\beta^2
\eta_{\scriptscriptstyle Y}^2L^2r}
{T(1-\beta^2)\min_{0\le s<T}b_s}
\ln\left(
1+\frac{\gamma^2T}
{[m_0^{\scriptscriptstyle Y}]^2}
\right)
+
\frac{6(1-\beta)}{T(1+\beta)}
\sum_{t=1}^{T-1}\frac{\kappa^2}{b_t}.
\label{eq:Fbar-second-avg-explicit-F1}
\end{align}
The last inequality follows from \cref{stochasticgradient}.

\paragraph{Upper Bound for $\frac{1}{T}\sum_{t=0}^{T-1}\mathbb E[\|Y_t-Y^\star(X_t)\|_F]$.} 
Substituting \cref{eq:Fbar-second-avg-explicit-F1} into \cref{firstordery} and applying the Cauchy--Schwarz inequality, the inequality \(\|A\|_*\le \sqrt r\,\|A\|_F\), and \cref{additionallemma}, we obtain
\begin{align}
&\frac{1}{T}\sum_{t=0}^{T-1}
\mathbb E[\|Y_t-Y^\star(X_t)\|_F]
\notag\\
&\le
\frac{\Delta_0^{\scriptscriptstyle Y}}
{T\mu\epsilon_{\scriptscriptstyle Y}}
+
\frac{G\sqrt r(2+\lambda)}
{T\mu\epsilon_{\scriptscriptstyle Y}}
\sum_{t=0}^{T-1}
\mathbb E[\|X_{t+1}-X_t\|_{\mathrm{op}}]
+
\frac{L}{T\mu\epsilon_{\scriptscriptstyle Y}}
\sum_{t=0}^{T-1}
\mathbb E\left[
\|X_{t+1}-X_t\|_{\mathrm{op}}
\|Y_{t+1}-Y_t\|_{\mathrm{op}}
\right]
\notag\\
&\quad+
\frac{L(2+\lambda^2)}
{2T\mu\epsilon_{\scriptscriptstyle Y}}
\sum_{t=0}^{T-1}
\mathbb E[\|X_{t+1}-X_t\|_{\mathrm{op}}^2]
+
\frac{Lr}{2T\mu\epsilon_{\scriptscriptstyle Y}}
\sum_{t=0}^{T-1}
\mathbb E\left[
\left(
\max\{
\epsilon_{\scriptscriptstyle Y},
\eta_{\scriptscriptstyle Y}\sigma_{t+1}
\}
\right)^2
\right]
\notag\\
&\quad+
\frac{2}{T\mu\epsilon_{\scriptscriptstyle Y}}
\sum_{t=0}^{T-1}
\mathbb E\left[
\max\{
\epsilon_{\scriptscriptstyle Y},
\eta_{\scriptscriptstyle Y}\sigma_{t+1}
\}
\|\bar F_{t+1}^{\scriptscriptstyle Y}\|_*
\right]
\notag\\
&\le
\frac{\Delta_0^{\scriptscriptstyle Y}}
{T\mu\epsilon_{\scriptscriptstyle Y}}
+
\frac{G\sqrt r(2+\lambda)}
{T\mu\epsilon_{\scriptscriptstyle Y}}
\sum_{t=0}^{T-1}
\mathbb E\left[
\max\{
\epsilon_{\scriptscriptstyle X},
\eta_{\scriptscriptstyle X}\theta_{t+1}
\}
\right]
\notag\\
&\quad+
\frac{L}{T\mu\epsilon_{\scriptscriptstyle Y}}
\sum_{t=0}^{T-1}
\mathbb E\left[
\max\{
\epsilon_{\scriptscriptstyle X},
\eta_{\scriptscriptstyle X}\theta_{t+1}
\}
\max\{
\epsilon_{\scriptscriptstyle Y},
\eta_{\scriptscriptstyle Y}\sigma_{t+1}
\}
\right]
+
\frac{L(2+\lambda^2)}
{2T\mu\epsilon_{\scriptscriptstyle Y}}
\sum_{t=0}^{T-1}
\mathbb E\left[
\left(
\max\{
\epsilon_{\scriptscriptstyle X},
\eta_{\scriptscriptstyle X}\theta_{t+1}
\}
\right)^2
\right]
\notag\\
&\quad+
\frac{Lr}{2T\mu\epsilon_{\scriptscriptstyle Y}}
\sum_{t=0}^{T-1}
\mathbb E\left[
\left(
\max\{
\epsilon_{\scriptscriptstyle Y},
\eta_{\scriptscriptstyle Y}\sigma_{t+1}
\}
\right)^2
\right]
+
\frac{2\sqrt r}{T\mu\epsilon_{\scriptscriptstyle Y}}
\sum_{t=0}^{T-1}
\mathbb E\left[
\max\{
\epsilon_{\scriptscriptstyle Y},
\eta_{\scriptscriptstyle Y}\sigma_{t+1}
\}
\|\bar F_{t+1}^{\scriptscriptstyle Y}\|_F
\right]
\notag\\
&\le
\frac{\Delta_0^{\scriptscriptstyle Y}}
{T\mu\epsilon_{\scriptscriptstyle Y}}
+
\frac{G\sqrt r(2+\lambda)}
{T\mu\epsilon_{\scriptscriptstyle Y}}
\sum_{t=0}^{T-1}
\mathbb E\left[
\max\{
\epsilon_{\scriptscriptstyle X},
\eta_{\scriptscriptstyle X}\theta_{t+1}
\}
\right]
\notag\\
&\quad+
\frac{L}{\mu\epsilon_{\scriptscriptstyle Y}}
\sqrt{
\frac1T\sum_{t=0}^{T-1}
\mathbb E\left[
\left(
\max\{
\epsilon_{\scriptscriptstyle X},
\eta_{\scriptscriptstyle X}\theta_{t+1}
\}
\right)^2
\right]
}
\sqrt{
\frac1T\sum_{t=0}^{T-1}
\mathbb E\left[
\left(
\max\{
\epsilon_{\scriptscriptstyle Y},
\eta_{\scriptscriptstyle Y}\sigma_{t+1}
\}
\right)^2
\right]
}
\notag\\
&\quad+
\frac{L(2+\lambda^2)}
{2T\mu\epsilon_{\scriptscriptstyle Y}}
\sum_{t=0}^{T-1}
\mathbb E\left[
\left(
\max\{
\epsilon_{\scriptscriptstyle X},
\eta_{\scriptscriptstyle X}\theta_{t+1}
\}
\right)^2
\right]
+
\frac{Lr}{2T\mu\epsilon_{\scriptscriptstyle Y}}
\sum_{t=0}^{T-1}
\mathbb E\left[
\left(
\max\{
\epsilon_{\scriptscriptstyle Y},
\eta_{\scriptscriptstyle Y}\sigma_{t+1}
\}
\right)^2
\right]
\notag\\
&\quad+
\frac{2\sqrt r}{\mu\epsilon_{\scriptscriptstyle Y}}
\sqrt{
\frac1T\sum_{t=0}^{T-1}
\mathbb E\left[
\left(
\max\{
\epsilon_{\scriptscriptstyle Y},
\eta_{\scriptscriptstyle Y}\sigma_{t+1}
\}
\right)^2
\right]
}
\sqrt{
\frac1T\sum_{t=0}^{T-1}
\mathbb E[
\|\bar F_{t+1}^{\scriptscriptstyle Y}\|_F^2
]
}
\notag\\
&\le
\frac{\Delta_0^{\scriptscriptstyle Y}}
{T\mu\epsilon_{\scriptscriptstyle Y}}
+
\frac{G\epsilon_{\scriptscriptstyle X}
\sqrt r(2+\lambda)}
{\mu\epsilon_{\scriptscriptstyle Y}}
+
\frac{G\eta_{\scriptscriptstyle X}
\sqrt r(2+\lambda)}
{\mu\epsilon_{\scriptscriptstyle Y}\sqrt T}
\sqrt{
\ln\left(
1+\frac{\gamma^2T}
{[\max\{m_0^{\scriptscriptstyle X},
         m_0^{\scriptscriptstyle Y}\}]^2}
\right)
}
\notag\\
&\quad+
\frac{L\epsilon_{\scriptscriptstyle X}^2
(2+\lambda^2)}
{2\mu\epsilon_{\scriptscriptstyle Y}}
+
\frac{L\eta_{\scriptscriptstyle X}^2
(2+\lambda^2)}
{2T\mu\epsilon_{\scriptscriptstyle Y}}
\ln\left(
1+\frac{\gamma^2T}
{[\max\{m_0^{\scriptscriptstyle X},
         m_0^{\scriptscriptstyle Y}\}]^2}
\right)
\notag\\
&\quad+
\frac{L}{\mu\epsilon_{\scriptscriptstyle Y}}
\sqrt{
\left(
\epsilon_{\scriptscriptstyle X}^2
+\frac{\eta_{\scriptscriptstyle X}^2}{T}
\ln\left(
1+\frac{\gamma^2T}
{[\max\{m_0^{\scriptscriptstyle X},
         m_0^{\scriptscriptstyle Y}\}]^2}
\right)
\right)
\left(
\epsilon_{\scriptscriptstyle Y}^2
+\frac{\eta_{\scriptscriptstyle Y}^2}{T}
\ln\left(
1+\frac{\gamma^2T}
{[m_0^{\scriptscriptstyle Y}]^2}
\right)
\right)
}
\notag\\
&\quad+
\frac{Lr\epsilon_{\scriptscriptstyle Y}}{2\mu}
+
\frac{Lr\eta_{\scriptscriptstyle Y}^2}
{2T\mu\epsilon_{\scriptscriptstyle Y}}
\ln\left(
1+\frac{\gamma^2T}
{[m_0^{\scriptscriptstyle Y}]^2}
\right)
+
\frac{2\sqrt r}{\mu\epsilon_{\scriptscriptstyle Y}}
\sqrt{
\epsilon_{\scriptscriptstyle Y}^2
+\frac{\eta_{\scriptscriptstyle Y}^2}{T}
\ln\left(
1+\frac{\gamma^2T}
{[m_0^{\scriptscriptstyle Y}]^2}
\right)
}
\sqrt{
\frac1T\sum_{t=0}^{T-1}
\mathbb E[
\|\bar F_{t+1}^{\scriptscriptstyle Y}\|_F^2
]
}
\notag\\
&\le
\frac{\Delta_0^{\scriptscriptstyle Y}}
{T\mu\epsilon_{\scriptscriptstyle Y}}
+
\frac{G\epsilon_{\scriptscriptstyle X}
\sqrt r(2+\lambda)}
{\mu\epsilon_{\scriptscriptstyle Y}}
+
\frac{G\eta_{\scriptscriptstyle X}
\sqrt r(2+\lambda)}
{\mu\epsilon_{\scriptscriptstyle Y}\sqrt T}
\sqrt{
\ln\left(
1+\frac{\gamma^2T}
{[\max\{m_0^{\scriptscriptstyle X},
         m_0^{\scriptscriptstyle Y}\}]^2}
\right)
}
\notag\\
&\quad+
\frac{L\epsilon_{\scriptscriptstyle X}^2
(2+\lambda^2)}
{2\mu\epsilon_{\scriptscriptstyle Y}}
+
\frac{L\eta_{\scriptscriptstyle X}^2
(2+\lambda^2)}
{2T\mu\epsilon_{\scriptscriptstyle Y}}
\ln\left(
1+\frac{\gamma^2T}
{[\max\{m_0^{\scriptscriptstyle X},
         m_0^{\scriptscriptstyle Y}\}]^2}
\right)
\notag\\
&\quad+
\frac{L}{\mu\epsilon_{\scriptscriptstyle Y}}
\sqrt{
\left(
\epsilon_{\scriptscriptstyle X}^2
+\frac{\eta_{\scriptscriptstyle X}^2}{T}
\ln\left(
1+\frac{\gamma^2T}
{[\max\{m_0^{\scriptscriptstyle X},
         m_0^{\scriptscriptstyle Y}\}]^2}
\right)
\right)
\left(
\epsilon_{\scriptscriptstyle Y}^2
+\frac{\eta_{\scriptscriptstyle Y}^2}{T}
\ln\left(
1+\frac{\gamma^2T}
{[m_0^{\scriptscriptstyle Y}]^2}
\right)
\right)
}
\notag\\
&\quad+
\frac{Lr\epsilon_{\scriptscriptstyle Y}}{2\mu}
+
\frac{Lr\eta_{\scriptscriptstyle Y}^2}
{2T\mu\epsilon_{\scriptscriptstyle Y}}
\ln\left(
1+\frac{\gamma^2T}
{[m_0^{\scriptscriptstyle Y}]^2}
\right)+
\frac{2\sqrt r}{\mu\epsilon_{\scriptscriptstyle Y}}
\sqrt{
\epsilon_{\scriptscriptstyle Y}^2
+\frac{\eta_{\scriptscriptstyle Y}^2}{T}
\ln\left(
1+\frac{\gamma^2T}
{[m_0^{\scriptscriptstyle Y}]^2}
\right)
}
\notag\\
&\quad\cdot
\Bigg(
\frac{3\beta^2G^2}{T(1-\beta)}
+
\frac{3(1-\beta)\kappa^2}{Tb_0}
+
3\epsilon_{\scriptscriptstyle X}^2L^2r
+
\frac{3\eta_{\scriptscriptstyle X}^2L^2r}{T}
\ln\left(
1+\frac{\gamma^2T}
{[\max\{m_0^{\scriptscriptstyle X},
         m_0^{\scriptscriptstyle Y}\}]^2}
\right)
+
3\epsilon_{\scriptscriptstyle Y}^2L^2r
\notag\\
&\qquad+
\frac{3\eta_{\scriptscriptstyle Y}^2L^2r}{T}
\ln\left(
1+\frac{\gamma^2T}
{[m_0^{\scriptscriptstyle Y}]^2}
\right)+
\frac{6\beta^2
\epsilon_{\scriptscriptstyle X}^2L^2r}
{(1-\beta^2)\min_{0\le s<T}b_s}
\notag\\
&\qquad+
\frac{6\beta^2
\eta_{\scriptscriptstyle X}^2L^2r}
{T(1-\beta^2)\min_{0\le s<T}b_s}
\ln\left(
1+\frac{\gamma^2T}
{[\max\{m_0^{\scriptscriptstyle X},
         m_0^{\scriptscriptstyle Y}\}]^2}
\right)
+
\frac{6\beta^2
\epsilon_{\scriptscriptstyle Y}^2L^2r}
{(1-\beta^2)\min_{0\le s<T}b_s}
\notag\\
&\qquad+
\frac{6\beta^2
\eta_{\scriptscriptstyle Y}^2L^2r}
{T(1-\beta^2)\min_{0\le s<T}b_s}
\ln\left(
1+\frac{\gamma^2T}
{[m_0^{\scriptscriptstyle Y}]^2}
\right)
+
\frac{6(1-\beta)}{T(1+\beta)}
\sum_{t=1}^{T-1}\frac{\kappa^2}{b_t}
\Bigg)^{\frac12}.
\label{firstordery1}
\end{align}

\paragraph{Upper Bound for $\frac{1}{T}\sum_{t=0}^{T-1}\mathbb E[\|Y_t-Y^\star(X_t)\|_F^2]$.} Similarly, substituting \cref{eq:Fbar-second-avg-explicit-F1} into \cref{eq:dual-avg-dt2-final-noVR}, we have
\begin{align}
&\frac1T\sum_{t=0}^{T-1}
\mathbb E[\|Y_t-Y^\star(X_t)\|_F^2]
\notag\\
&\le
\frac{8G\Delta_0^{\scriptscriptstyle Y}}
{7\mu^2\epsilon_{\scriptscriptstyle Y}T}
+
\left(
\frac{G}{7\mu\epsilon_{\scriptscriptstyle Y}}
+\frac{4GLr}{7\mu^2\epsilon_{\scriptscriptstyle Y}}
\right)
\frac1T\sum_{t=0}^{T-1}
\mathbb E\left[
\left(
\max\{
\epsilon_{\scriptscriptstyle Y},
\eta_{\scriptscriptstyle Y}\sigma_{t+1}
\}
\right)^2
\right]
\notag\\
&\quad+
\left(
\frac{4Gr(L_\Phi+L)}
{7\mu^2\epsilon_{\scriptscriptstyle Y}}
+\frac{16r^2L^2G}
{7\mu^3\epsilon_{\scriptscriptstyle Y}}
+\frac{16rG^2L^2}
{7\mu^4\epsilon_{\scriptscriptstyle Y}^2}
\right)
\frac1T\sum_{t=0}^{T-1}
\mathbb E\left[
\left(
\max\{
\epsilon_{\scriptscriptstyle X},
\eta_{\scriptscriptstyle X}\theta_{t+1}
\}
\right)^2
\right]
\notag\\
&\quad+
\frac{16G\sqrt r}
{7\mu^2\epsilon_{\scriptscriptstyle Y}}
\sqrt{
\frac1T\sum_{t=0}^{T-1}
\mathbb E\left[
\left(
\max\{
\epsilon_{\scriptscriptstyle Y},
\eta_{\scriptscriptstyle Y}\sigma_{t+1}
\}
\right)^2
\right]
}
\sqrt{
\frac1T\sum_{t=0}^{T-1}
\mathbb E[
\|\bar F_{t+1}^{\scriptscriptstyle Y}\|_F^2
]
}
\notag\\
&\overset{(a)}{\le}
\frac{8G\Delta_0^{\scriptscriptstyle Y}}
{7\mu^2\epsilon_{\scriptscriptstyle Y}T}
+
\left(
\frac{G}{7\mu\epsilon_{\scriptscriptstyle Y}}
+\frac{4GLr}{7\mu^2\epsilon_{\scriptscriptstyle Y}}
\right)
\frac1T\sum_{t=0}^{T-1}
\mathbb E\left[
\left(
\max\{
\epsilon_{\scriptscriptstyle Y},
\eta_{\scriptscriptstyle Y}\sigma_{t+1}
\}
\right)^2
\right]
\notag\\
&\quad+
\left(
\frac{4Gr(L_\Phi+L)}
{7\mu^2\epsilon_{\scriptscriptstyle Y}}
+\frac{16r^2L^2G}
{7\mu^3\epsilon_{\scriptscriptstyle Y}}
+\frac{16rG^2L^2}
{7\mu^4\epsilon_{\scriptscriptstyle Y}^2}
\right)
\frac1T\sum_{t=0}^{T-1}
\mathbb E\left[
\left(
\max\{
\epsilon_{\scriptscriptstyle X},
\eta_{\scriptscriptstyle X}\theta_{t+1}
\}
\right)^2
\right]
\notag\\
&\quad+
\frac{16G\sqrt r}
{7\mu^2\epsilon_{\scriptscriptstyle Y}}
\sqrt{
\frac1T\sum_{t=0}^{T-1}
\mathbb E\left[
\left(
\max\{
\epsilon_{\scriptscriptstyle Y},
\eta_{\scriptscriptstyle Y}\sigma_{t+1}
\}
\right)^2
\right]
}
\notag\\
&\quad\cdot
\Bigg(
\frac{3\beta^2G^2}{T(1-\beta)}
+\frac{3(1-\beta)\kappa^2}{Tb_0}
+
3\epsilon_{\scriptscriptstyle X}^2L^2r
+
\frac{3\eta_{\scriptscriptstyle X}^2L^2r}{T}
\ln\left(
1+\frac{\gamma^2T}
{[\max\{m_0^{\scriptscriptstyle X},
         m_0^{\scriptscriptstyle Y}\}]^2}
\right)
+
3\epsilon_{\scriptscriptstyle Y}^2L^2r
\notag\\
&\qquad+
\frac{3\eta_{\scriptscriptstyle Y}^2L^2r}{T}
\ln\left(
1+\frac{\gamma^2T}
{[m_0^{\scriptscriptstyle Y}]^2}
\right)
+
\frac{6\beta^2
\epsilon_{\scriptscriptstyle X}^2L^2r}
{(1-\beta^2)\min_{0\le s<T}b_s}
\notag\\
&\qquad+
\frac{6\beta^2
\eta_{\scriptscriptstyle X}^2L^2r}
{T(1-\beta^2)\min_{0\le s<T}b_s}
\ln\left(
1+\frac{\gamma^2T}
{[\max\{m_0^{\scriptscriptstyle X},
         m_0^{\scriptscriptstyle Y}\}]^2}
\right)+
\frac{6\beta^2
\epsilon_{\scriptscriptstyle Y}^2L^2r}
{(1-\beta^2)\min_{0\le s<T}b_s}
\notag\\
&\qquad+
\frac{6\beta^2
\eta_{\scriptscriptstyle Y}^2L^2r}
{T(1-\beta^2)\min_{0\le s<T}b_s}
\ln\left(
1+\frac{\gamma^2T}
{[m_0^{\scriptscriptstyle Y}]^2}
\right)
+
\frac{6(1-\beta)}{T(1+\beta)}
\sum_{t=1}^{T-1}\frac{\kappa^2}{b_t}
\Bigg)^{\frac12}
\notag\\
&\overset{(b)}{\le}
\frac{8G\Delta_0^{\scriptscriptstyle Y}}
{7\mu^2\epsilon_{\scriptscriptstyle Y}T}
+
\frac{4Gr\epsilon_{\scriptscriptstyle X}^2
(L_\Phi+L)}
{7\mu^2\epsilon_{\scriptscriptstyle Y}}
+
\frac{16r^2L^2G
\epsilon_{\scriptscriptstyle X}^2}
{7\mu^3\epsilon_{\scriptscriptstyle Y}}
+
\frac{16rG^2L^2
\epsilon_{\scriptscriptstyle X}^2}
{7\mu^4\epsilon_{\scriptscriptstyle Y}^2}
\notag\\
&\quad+
\left(
\frac{4Gr\eta_{\scriptscriptstyle X}^2
(L_\Phi+L)}
{7T\mu^2\epsilon_{\scriptscriptstyle Y}}
+
\frac{16r^2L^2G
\eta_{\scriptscriptstyle X}^2}
{7T\mu^3\epsilon_{\scriptscriptstyle Y}}
+
\frac{16rG^2L^2
\eta_{\scriptscriptstyle X}^2}
{7T\mu^4\epsilon_{\scriptscriptstyle Y}^2}
\right)
\ln\left(
1+\frac{\gamma^2T}
{[\max\{m_0^{\scriptscriptstyle X},
         m_0^{\scriptscriptstyle Y}\}]^2}
\right)
\notag\\
&\quad+
\frac{G\epsilon_{\scriptscriptstyle Y}}{7\mu}
+
\frac{4GLr\epsilon_{\scriptscriptstyle Y}}{7\mu^2}+
\left(
\frac{G\eta_{\scriptscriptstyle Y}^2}
{7T\mu\epsilon_{\scriptscriptstyle Y}}
+
\frac{4GLr\eta_{\scriptscriptstyle Y}^2}
{7T\mu^2\epsilon_{\scriptscriptstyle Y}}
\right)
\ln\left(
1+\frac{\gamma^2T}
{[m_0^{\scriptscriptstyle Y}]^2}
\right)
\notag\\
&\quad+
\frac{16G\sqrt r}
{7\mu^2\epsilon_{\scriptscriptstyle Y}}
\sqrt{
\epsilon_{\scriptscriptstyle Y}^2
+\frac{\eta_{\scriptscriptstyle Y}^2}{T}
\ln\left(
1+\frac{\gamma^2T}
{[m_0^{\scriptscriptstyle Y}]^2}
\right)
}
\notag\\
&\quad\cdot
\Bigg(
\frac{3\beta^2G^2}{T(1-\beta)}
+\frac{3(1-\beta)\kappa^2}{Tb_0}
+
3\epsilon_{\scriptscriptstyle X}^2L^2r
+
\frac{3\eta_{\scriptscriptstyle X}^2L^2r}{T}
\ln\left(
1+\frac{\gamma^2T}
{[\max\{m_0^{\scriptscriptstyle X},
         m_0^{\scriptscriptstyle Y}\}]^2}
\right)
+
3\epsilon_{\scriptscriptstyle Y}^2L^2r
\notag\\
&\qquad+
\frac{3\eta_{\scriptscriptstyle Y}^2L^2r}{T}
\ln\left(
1+\frac{\gamma^2T}
{[m_0^{\scriptscriptstyle Y}]^2}
\right)
+
\frac{6\beta^2
\epsilon_{\scriptscriptstyle X}^2L^2r}
{(1-\beta^2)\min_{0\le s<T}b_s}
\notag\\
&\qquad+
\frac{6\beta^2
\eta_{\scriptscriptstyle X}^2L^2r}
{T(1-\beta^2)\min_{0\le s<T}b_s}
\ln\left(
1+\frac{\gamma^2T}
{[\max\{m_0^{\scriptscriptstyle X},
         m_0^{\scriptscriptstyle Y}\}]^2}
\right)
+
\frac{6\beta^2
\epsilon_{\scriptscriptstyle Y}^2L^2r}
{(1-\beta^2)\min_{0\le s<T}b_s}
\notag\\
&\qquad+
\frac{6\beta^2
\eta_{\scriptscriptstyle Y}^2L^2r}
{T(1-\beta^2)\min_{0\le s<T}b_s}
\ln\left(
1+\frac{\gamma^2T}
{[m_0^{\scriptscriptstyle Y}]^2}
\right)
+
\frac{6(1-\beta)}{T(1+\beta)}
\sum_{t=1}^{T-1}\frac{\kappa^2}{b_t}
\Bigg)^{\frac12}.
\label{eq:dual-avg-dt2-final-noVR222}
\end{align}
Here (a) uses \cref{eq:Fbar-second-avg-explicit-F1}.
Inequality (b) uses the squared-step bounds above, which follow
from \cref{lem:log_cum_ratio} with the positive initialization.
The proof is thus completed.
\end{proof}

\subsection{Proof of \cref{e3lemma}}\label{sece3}
From the update rule, we have
\begin{align}
&M_{t+1}^{\scriptscriptstyle X}
\notag\\
&=
\beta M_t^{\scriptscriptstyle X}
+(1-\beta)
\nabla_X F(X_t,Y_t;\xi_t^{\scriptscriptstyle X})+
\beta\big(
\nabla_X F(X_t,Y_t;\xi_t^{\scriptscriptstyle X})
-\nabla_X F(X_{t-1},Y_{t-1};
             \xi_t^{\scriptscriptstyle X})
\big)
\notag\\
&=
\beta M_t^{\scriptscriptstyle X}
+\nabla_X f(X_t,Y_t)
-\beta\nabla_X f(X_{t-1},Y_{t-1})
\notag\\
&\quad+
\left(
\big(
\nabla_X F(X_t,Y_t;\xi_t^{\scriptscriptstyle X})
-\nabla_X f(X_t,Y_t)
\big)
-\beta\big(
\nabla_X F(X_{t-1},Y_{t-1};
             \xi_t^{\scriptscriptstyle X})
-\nabla_X f(X_{t-1},Y_{t-1})
\big)
\right)
\notag\\
&=
\beta M_t^{\scriptscriptstyle X}
+\nabla_X f(X_t,Y_t)
-\beta\nabla_X f(X_{t-1},Y_{t-1})
+\tilde\Delta_t,
\label{eq:storm-tracker-x}
\end{align}
where
\[
\tilde\Delta_t
:=
\big(
\nabla_X F(X_t,Y_t;\xi_t^{\scriptscriptstyle X})
-\nabla_X f(X_t,Y_t)
\big)
-\beta\big(
\nabla_X F(X_{t-1},Y_{t-1};
             \xi_t^{\scriptscriptstyle X})
-\nabla_X f(X_{t-1},Y_{t-1})
\big).
\]
Let us define
$\tilde E_{t+1}
:=M_{t+1}^{\scriptscriptstyle X}
-\nabla_X f(X_t,Y_t)$.
It then follows that, for $t\ge1$,
\begin{align}
\tilde E_{t+1}
&=
\beta\big(
M_t^{\scriptscriptstyle X}
-\nabla_X f(X_{t-1},Y_{t-1})
\big)
+\tilde\Delta_t
=
\beta\tilde E_t+\tilde\Delta_t.
\label{eq:Ebar-rec}
\end{align}
Applying \cref{eq:Ebar-rec} recursively yields,
for any $t\ge0$,
\(
\tilde E_{t+1}
=
\beta^t\tilde E_1
+\sum_{\tau=0}^{t-1}
\beta^\tau\tilde\Delta_{t-\tau}.
\)
Using
$\|A+B\|_F^2\le2\|A\|_F^2+2\|B\|_F^2$,
we have
\begin{align}
&\mathbb E[\|\tilde\Delta_t\|_F^2]
\notag\\
&=
\mathbb E\Big[
\Big\|
\big(
\nabla_X F(X_t,Y_t;\xi_t^{\scriptscriptstyle X})
-\nabla_X f(X_t,Y_t)
\big)
-\beta\big(
\nabla_X F(X_{t-1},Y_{t-1};
             \xi_t^{\scriptscriptstyle X})
-\nabla_X f(X_{t-1},Y_{t-1})
\big)
\Big\|_F^2
\Big]
\notag\\
&=
\mathbb E\Big[
\Big\|
(1-\beta)\big(
\nabla_X F(X_t,Y_t;\xi_t^{\scriptscriptstyle X})
-\nabla_X f(X_t,Y_t)
\big)
\notag\\
&\qquad+
\beta\Big(
\nabla_X F(X_t,Y_t;\xi_t^{\scriptscriptstyle X})
-\nabla_X F(X_{t-1},Y_{t-1};
             \xi_t^{\scriptscriptstyle X})
-\nabla_X f(X_t,Y_t)
+\nabla_X f(X_{t-1},Y_{t-1})
\Big)
\Big\|_F^2
\Big]
\notag\\
&\le
2(1-\beta)^2
\mathbb E\Big[
\big\|
\nabla_X F(X_t,Y_t;\xi_t^{\scriptscriptstyle X})
-\nabla_X f(X_t,Y_t)
\big\|_F^2
\Big]
\notag\\
&\quad+
2\beta^2
\mathbb E\Big[
\Big\|
\nabla_X F(X_t,Y_t;\xi_t^{\scriptscriptstyle X})
-\nabla_X F(X_{t-1},Y_{t-1};
             \xi_t^{\scriptscriptstyle X})
-\nabla_X f(X_t,Y_t)
+\nabla_X f(X_{t-1},Y_{t-1})
\Big\|_F^2
\Big]
\notag\\
&=
2(1-\beta)^2
\mathbb E\Big[
\big\|
\nabla_X F(X_t,Y_t;\xi_t^{\scriptscriptstyle X})
-\nabla_X f(X_t,Y_t)
\big\|_F^2
\Big]
\notag\\
&\quad+
\frac{2\beta^2}{b_t}
\mathbb E\Big[
\Big\|
\nabla_X F(X_t,Y_t;
             \xi_{t,1}^{\scriptscriptstyle X})
-\nabla_X F(X_{t-1},Y_{t-1};
             \xi_{t,1}^{\scriptscriptstyle X})
-\nabla_X f(X_t,Y_t)
+\nabla_X f(X_{t-1},Y_{t-1})
\Big\|_F^2
\Big]
\notag\\
&\overset{(a)}{\le}
2(1-\beta)^2
\mathbb E\Big[
\big\|
\nabla_X F(X_t,Y_t;\xi_t^{\scriptscriptstyle X})
-\nabla_X f(X_t,Y_t)
\big\|_F^2
\Big]
\notag\\
&\quad+
\frac{2\beta^2}{b_t}
\mathbb E\Big[
\big\|
\nabla_X F(X_t,Y_t;
             \xi_{t,1}^{\scriptscriptstyle X})
-\nabla_X F(X_{t-1},Y_{t-1};
             \xi_{t,1}^{\scriptscriptstyle X})
\big\|_F^2
\Big]
\notag\\
&\overset{(b)}{\le}
\frac{2(1-\beta)^2\kappa^2}{b_t}
+
\frac{2\beta^2L^2}{b_t}
\mathbb E\Big[
\|X_t-X_{t-1}\|_F^2
+\|Y_t-Y_{t-1}\|_F^2
\Big]
\notag\\
&=
\frac{2\beta^2L^2}{b_t}
\mathbb E[\|X_t-X_{t-1}\|_F^2]
+
\frac{2\beta^2L^2}{b_t}
\mathbb E[\|Y_t-Y_{t-1}\|_F^2]
+
\frac{2(1-\beta)^2\kappa^2}{b_t}
\notag\\
&\overset{(c)}{\le}
\frac{2\beta^2L^2r}{b_t}
\mathbb E\left[
\left(
\max\{
\epsilon_{\scriptscriptstyle X},
\eta_{\scriptscriptstyle X}\theta_t
\}
\right)^2
\right]+
\frac{2\beta^2L^2r}{b_t}
\mathbb E\left[
\left(
\max\{
\epsilon_{\scriptscriptstyle Y},
\eta_{\scriptscriptstyle Y}\sigma_t
\}
\right)^2
\right]
+
\frac{2(1-\beta)^2\kappa^2}{b_t},
\label{eq:Delta-sq-split}
\end{align}
where (a) follows from
$\mathbb E[\|Z\|^2]-\|\mathbb E[Z]\|^2
\le\mathbb E[\|Z\|^2]$,
applied conditionally on the history before sampling
the current mini-batch;
(b) follows from the mean-square smoothness in
\cref{smoothness} and the variance bound in
\cref{stochasticgradient};
and (c) uses the update expressions together with the bounds
$\|O_t^{\scriptscriptstyle X}\|_F^2\le r$
and
$\|O_t^{\scriptscriptstyle Y}\|_F^2\le r$,
which hold under the global rank bound $r$.

Moreover, $\{\tilde\Delta_{t+1}\}_{t\ge0}$
forms a martingale difference sequence with respect to
the corresponding pre-sampling histories, so that
$\mathbb E\langle
\tilde\Delta_{i+1},\tilde\Delta_{j+1}
\rangle=0$
for $i\ne j$.
Consequently, we obtain
\begin{align}
&\mathbb E\left[
\left\|
\sum_{\tau=0}^{t-1}
\beta^\tau\tilde\Delta_{t-\tau}
\right\|_*
\right]
\notag\\
&\le
\sqrt r\,
\sqrt{
\mathbb E\left[
\left\|
\sum_{\tau=0}^{t-1}
\beta^\tau\tilde\Delta_{t-\tau}
\right\|_F^2
\right]
}
\notag\\
&=
\sqrt r\,
\sqrt{
\sum_{\tau=0}^{t-1}
\beta^{2\tau}
\mathbb E[
\|\tilde\Delta_{t-\tau}\|_F^2
]
}
\notag\\
&\le
\sqrt r\,
\Bigg(
2\beta^2L^2r
\sum_{\tau=0}^{t-1}
\frac{\beta^{2\tau}}{b_{t-\tau}}
\mathbb E\left[
\left(
\max\{
\epsilon_{\scriptscriptstyle X},
\eta_{\scriptscriptstyle X}\theta_{t-\tau}
\}
\right)^2
\right]+
2\beta^2L^2r
\sum_{\tau=0}^{t-1}
\frac{\beta^{2\tau}}{b_{t-\tau}}
\mathbb E\left[
\left(
\max\{
\epsilon_{\scriptscriptstyle Y},
\eta_{\scriptscriptstyle Y}\sigma_{t-\tau}
\}
\right)^2
\right]
\notag\\
&\qquad+
2(1-\beta)^2\kappa^2
\sum_{\tau=0}^{t-1}
\frac{\beta^{2\tau}}{b_{t-\tau}}
\Bigg)^{\frac12}.
\label{eq:St-star-pointwise}
\end{align}

Then, by Jensen's inequality, we have
\begin{align}
&\frac1T\sum_{t=0}^{T-1}
\mathbb E\left[
\left\|
\sum_{\tau=0}^{t-1}
\beta^\tau\tilde\Delta_{t-\tau}
\right\|_*
\right]
\notag\\
&\le
\sqrt r\,
\sqrt{
\frac1T\sum_{t=0}^{T-1}
\mathbb E\left[
\left\|
\sum_{\tau=0}^{t-1}
\beta^\tau\tilde\Delta_{t-\tau}
\right\|_F^2
\right]
}
\notag\\
&=
\sqrt r\,
\sqrt{
\frac1T\sum_{t=0}^{T-1}
\sum_{\tau=0}^{t-1}
\beta^{2\tau}
\mathbb E[
\|\tilde\Delta_{t-\tau}\|_F^2
]
}
\notag\\
&=
\sqrt r\,
\sqrt{
\frac1T\sum_{k=1}^{T-1}
\left(
\sum_{j=0}^{T-1-k}\beta^{2j}
\right)
\mathbb E[\|\tilde\Delta_k\|_F^2]
}
\notag\\
&\le
\sqrt r\,
\sqrt{
\frac{1}{T(1-\beta^2)}
\sum_{t=1}^{T-1}
\mathbb E[\|\tilde\Delta_t\|_F^2]
}
\notag\\
&=
\sqrt{\frac{r}{1-\beta^2}}\,
\sqrt{
\frac1T\sum_{t=1}^{T-1}
\mathbb E[\|\tilde\Delta_t\|_F^2]
}
\notag\\
&\le
\sqrt{\frac{r}{1-\beta^2}}\,
\Bigg(
\frac{2\beta^2L^2r}{T}
\sum_{t=1}^{T-1}
\frac{1}{b_t}
\mathbb E\left[
\left(
\max\{
\epsilon_{\scriptscriptstyle X},
\eta_{\scriptscriptstyle X}\theta_t
\}
\right)^2
\right]
\notag\\
&\qquad+
\frac{2\beta^2L^2r}{T}
\sum_{t=1}^{T-1}
\frac{1}{b_t}
\mathbb E\left[
\left(
\max\{
\epsilon_{\scriptscriptstyle Y},
\eta_{\scriptscriptstyle Y}\sigma_t
\}
\right)^2
\right]
+
\frac{2(1-\beta)^2}{T}
\sum_{t=1}^{T-1}\frac{\kappa^2}{b_t}
\Bigg)^{\frac12}
\notag\\
&\le
\sqrt{\frac{r}{1-\beta^2}}\,
\Bigg(
\frac{2\beta^2L^2r}
{T\min_{0\le s<T}b_s}
\sum_{t=1}^{T-1}
\mathbb E\left[
\left(
\max\{
\epsilon_{\scriptscriptstyle X},
\eta_{\scriptscriptstyle X}\theta_t
\}
\right)^2
\right]
\notag\\
&\qquad+
\frac{2\beta^2L^2r}
{T\min_{0\le s<T}b_s}
\sum_{t=1}^{T-1}
\mathbb E\left[
\left(
\max\{
\epsilon_{\scriptscriptstyle Y},
\eta_{\scriptscriptstyle Y}\sigma_t
\}
\right)^2
\right]+
\frac{2(1-\beta)^2}{T}
\sum_{t=1}^{T-1}\frac{\kappa^2}{b_t}
\Bigg)^{\frac12}
\notag\\
&\le
\sqrt{\frac{r}{1-\beta^2}}\,
\Bigg(
\frac{2\beta^2L^2r
\epsilon_{\scriptscriptstyle X}^2}
{\min_{0\le s<T}b_s}+
\frac{2\beta^2L^2r
\eta_{\scriptscriptstyle X}^2}
{T\min_{0\le s<T}b_s}
\ln\left(
1+\frac{\gamma^2T}
{[\max\{m_0^{\scriptscriptstyle X},
         m_0^{\scriptscriptstyle Y}\}]^2}
\right)
\notag\\
&\qquad+
\frac{2\beta^2L^2r
\epsilon_{\scriptscriptstyle Y}^2}
{\min_{0\le s<T}b_s}
+
\frac{2\beta^2L^2r
\eta_{\scriptscriptstyle Y}^2}
{T\min_{0\le s<T}b_s}
\ln\left(
1+\frac{\gamma^2T}
{[m_0^{\scriptscriptstyle Y}]^2}
\right)+
\frac{2(1-\beta)^2}{T}
\sum_{t=1}^{T-1}\frac{\kappa^2}{b_t}
\Bigg)^{\frac12}.
\label{eq:St-star-average}
\end{align}
The last inequality follows from \cref{lem:log_cum_ratio},
applied to the sequence beginning with the positive
initialization $c^2$ and followed by the clipped squared
gradient norms, where
$m_0^{\scriptscriptstyle X}
=m_0^{\scriptscriptstyle Y}=c>0$.

Then we have
\begin{align}
&\mathbb E[\|\tilde E_{t+1}\|_*]
\notag\\
&=
\mathbb E\left[
\left\|
\beta^t\tilde E_1
+\sum_{\tau=0}^{t-1}
\beta^\tau\tilde\Delta_{t-\tau}
\right\|_*
\right]
\notag\\
&\le
\beta^t\mathbb E[\|\tilde E_1\|_*]
+
\mathbb E\left[
\left\|
\sum_{\tau=0}^{t-1}
\beta^\tau\tilde\Delta_{t-\tau}
\right\|_*
\right]
\notag\\
&\le
\beta^t\mathbb E[\|\tilde E_1\|_*]
+
\sqrt r\,
\sqrt{
\mathbb E\left[
\left\|
\sum_{\tau=0}^{t-1}
\beta^\tau\tilde\Delta_{t-\tau}
\right\|_F^2
\right]
}
\notag\\
&=
\beta^t\mathbb E[\|\tilde E_1\|_*]
+
\sqrt r\,
\sqrt{
\sum_{\tau=0}^{t-1}
\beta^{2\tau}
\mathbb E[
\|\tilde\Delta_{t-\tau}\|_F^2
]
}
\notag\\
&\le
\beta^t\mathbb E[\|\tilde E_1\|_*]
+
\sqrt r\,
\Bigg(
2\beta^2L^2
\sum_{\tau=0}^{t-1}
\frac{\beta^{2\tau}}{b_{t-\tau}}
\mathbb E[
\|X_{t-\tau}-X_{t-\tau-1}\|_F^2
]
\notag\\
&\qquad+
2\beta^2L^2
\sum_{\tau=0}^{t-1}
\frac{\beta^{2\tau}}{b_{t-\tau}}
\mathbb E[
\|Y_{t-\tau}-Y_{t-\tau-1}\|_F^2
]+
2(1-\beta)^2\kappa^2
\sum_{\tau=0}^{t-1}
\frac{\beta^{2\tau}}{b_{t-\tau}}
\Bigg)^{\frac12}
\notag\\
&\le
\beta^t\mathbb E[\|\tilde E_1\|_*]
+
\sqrt r\,
\Bigg(
2\beta^2L^2r
\sum_{\tau=0}^{t-1}
\frac{\beta^{2\tau}}{b_{t-\tau}}
\mathbb E\left[
\left(
\max\{
\epsilon_{\scriptscriptstyle X},
\eta_{\scriptscriptstyle X}\theta_{t-\tau}
\}
\right)^2
\right]
\notag\\
&\qquad+
2\beta^2L^2r
\sum_{\tau=0}^{t-1}
\frac{\beta^{2\tau}}{b_{t-\tau}}
\mathbb E\left[
\left(
\max\{
\epsilon_{\scriptscriptstyle Y},
\eta_{\scriptscriptstyle Y}\sigma_{t-\tau}
\}
\right)^2
\right]+
2(1-\beta)^2\kappa^2
\sum_{\tau=0}^{t-1}
\frac{\beta^{2\tau}}{b_{t-\tau}}
\Bigg)^{\frac12}.
\label{eq:Ebar-pointwise}
\end{align}

We now consider
$E_{t+1}
=M_{t+1}^{\scriptscriptstyle X}-\nabla\Phi(X_t)$.
By \cref{smoothness,additionallemma}, we obtain
\begin{align}
\|E_{t+1}\|_*
&=
\|M_{t+1}^{\scriptscriptstyle X}
-\nabla\Phi(X_t)\|_*
\notag\\
&\le
\|M_{t+1}^{\scriptscriptstyle X}
-\nabla_X f(X_t,Y_t)\|_*
+
\|\nabla_X f(X_t,Y_t)-\nabla\Phi(X_t)\|_*
\notag\\
&=
\underbrace{
\|M_{t+1}^{\scriptscriptstyle X}
-\nabla_X f(X_t,Y_t)\|_*
}_{=\|\tilde E_{t+1}\|_*}
+
\|\nabla_X f(X_t,Y_t)
-\nabla_X f(X_t,Y^\star(X_t))\|_*
\notag\\
&\le
\|\tilde E_{t+1}\|_*
+L\|Y_t-Y^\star(X_t)\|_{\mathrm{op}}.
\label{eq:E-via-Ebar}
\end{align}
Using
$\|A\|_{\mathrm{op}}\le\|A\|_F$,
taking expectations, and applying
\cref{eq:Ebar-pointwise} yield the following bound:
\begin{align}
&\mathbb E[\|E_{t+1}\|_*]
\notag\\
&\le
\mathbb E[\|\tilde E_{t+1}\|_*]
+
L\mathbb E[
\|Y_t-Y^\star(X_t)\|_{\mathrm{op}}
]
\notag\\
&\le
\mathbb E[\|\tilde E_{t+1}\|_*]
+
L\mathbb E[
\|Y_t-Y^\star(X_t)\|_F
]
\notag\\
&\le
\beta^t\mathbb E[\|\tilde E_1\|_*]
+
L\mathbb E[
\|Y_t-Y^\star(X_t)\|_F
]
+
\sqrt r\,
\sqrt{
\sum_{\tau=0}^{t-1}
\beta^{2\tau}
\mathbb E[
\|\tilde\Delta_{t-\tau}\|_F^2
]
}
\notag\\
&\le
\beta^t\mathbb E[\|\tilde E_1\|_*]
+
L\mathbb E[
\|Y_t-Y^\star(X_t)\|_F
]+
\sqrt r\,
\Bigg(
2\beta^2L^2r
\sum_{\tau=0}^{t-1}
\frac{\beta^{2\tau}}{b_{t-\tau}}
\mathbb E\left[
\left(
\max\{
\epsilon_{\scriptscriptstyle X},
\eta_{\scriptscriptstyle X}\theta_{t-\tau}
\}
\right)^2
\right]
\notag\\
&\qquad+
2\beta^2L^2r
\sum_{\tau=0}^{t-1}
\frac{\beta^{2\tau}}{b_{t-\tau}}
\mathbb E\left[
\left(
\max\{
\epsilon_{\scriptscriptstyle Y},
\eta_{\scriptscriptstyle Y}\sigma_{t-\tau}
\}
\right)^2
\right]+
2(1-\beta)^2\kappa^2
\sum_{\tau=0}^{t-1}
\frac{\beta^{2\tau}}{b_{t-\tau}}
\Bigg)^{\frac12}.
\label{eq:E-pointwise-final}
\end{align}

We next derive an upper bound on
$\sum_{t=0}^{T-1}\mathbb E[\|E_{t+1}\|_*]$.
Summing \cref{eq:E-via-Ebar} over $t=0,\dots,T-1$,
using the recursive expansion of $\tilde E_{t+1}$,
and applying \cref{eq:St-star-average,firstordery1},
we obtain
\begin{align}
&\sum_{t=0}^{T-1}
\mathbb E[\|E_{t+1}\|_*]
\notag\\
&\le
\sum_{t=0}^{T-1}
\mathbb E[\|\tilde E_{t+1}\|_*]
+
L\sum_{t=0}^{T-1}
\mathbb E[\|Y_t-Y^\star(X_t)\|_F]
\notag\\
&\le
\sum_{t=0}^{T-1}
\beta^t\mathbb E[\|\tilde E_1\|_*]+
\sum_{t=0}^{T-1}
\mathbb E\left[
\left\|
\sum_{\tau=0}^{t-1}
\beta^\tau\tilde\Delta_{t-\tau}
\right\|_*
\right]
+
L\sum_{t=0}^{T-1}
\mathbb E[\|Y_t-Y^\star(X_t)\|_F]
\notag\\
&=
\frac{1-\beta^T}{1-\beta}
\mathbb E[\|\tilde E_1\|_*]+
\sum_{t=0}^{T-1}
\mathbb E\left[
\left\|
\sum_{\tau=0}^{t-1}
\beta^\tau\tilde\Delta_{t-\tau}
\right\|_*
\right]
+
L\sum_{t=0}^{T-1}
\mathbb E[\|Y_t-Y^\star(X_t)\|_F]
\notag\\
&\le
\frac{\mathbb E[\|\tilde E_1\|_*]}{1-\beta}
+
T\sqrt r\,
\sqrt{
\frac{1}{T(1-\beta^2)}
\sum_{t=1}^{T-1}
\mathbb E[\|\tilde\Delta_t\|_F^2]
}+
L\sum_{t=0}^{T-1}
\mathbb E[\|Y_t-Y^\star(X_t)\|_F]
\notag\\
&\le
\frac{\mathbb E[\|\tilde E_1\|_*]}{1-\beta}
+
T\sqrt{\frac{r}{1-\beta^2}}\,
\Bigg(
\frac{2\beta^2L^2r
\epsilon_{\scriptscriptstyle X}^2}
{\min_{0\le s<T}b_s}+
\frac{2\beta^2L^2r
\eta_{\scriptscriptstyle X}^2}
{T\min_{0\le s<T}b_s}
\ln\left(
1+\frac{\gamma^2T}
{[\max\{m_0^{\scriptscriptstyle X},
         m_0^{\scriptscriptstyle Y}\}]^2}
\right)
\notag\\
&\quad+
\frac{2\beta^2L^2r
\epsilon_{\scriptscriptstyle Y}^2}
{\min_{0\le s<T}b_s}+
\frac{2\beta^2L^2r
\eta_{\scriptscriptstyle Y}^2}
{T\min_{0\le s<T}b_s}
\ln\left(
1+\frac{\gamma^2T}
{[m_0^{\scriptscriptstyle Y}]^2}
\right)+
\frac{2(1-\beta)^2}{T}
\sum_{t=1}^{T-1}\frac{\kappa^2}{b_t}
\Bigg)^{\frac12}
\notag\\
&\quad+
LT\Bigg[
\frac{\Delta_0^{\scriptscriptstyle Y}}
{T\mu\epsilon_{\scriptscriptstyle Y}}
+
\frac{G\epsilon_{\scriptscriptstyle X}
\sqrt r(2+\lambda)}
{\mu\epsilon_{\scriptscriptstyle Y}}+
\frac{G\eta_{\scriptscriptstyle X}
\sqrt r(2+\lambda)}
{\mu\epsilon_{\scriptscriptstyle Y}\sqrt T}
\sqrt{
\ln\left(
1+\frac{\gamma^2T}
{[\max\{m_0^{\scriptscriptstyle X},
         m_0^{\scriptscriptstyle Y}\}]^2}
\right)
}
\notag\\
&\quad+
\frac{L\epsilon_{\scriptscriptstyle X}^2
(2+\lambda^2)}
{2\mu\epsilon_{\scriptscriptstyle Y}}
+
\frac{L\eta_{\scriptscriptstyle X}^2
(2+\lambda^2)}
{2T\mu\epsilon_{\scriptscriptstyle Y}}
\ln\left(
1+\frac{\gamma^2T}
{[\max\{m_0^{\scriptscriptstyle X},
         m_0^{\scriptscriptstyle Y}\}]^2}
\right)
\notag\\
&\quad+
\frac{L}{\mu\epsilon_{\scriptscriptstyle Y}}
\sqrt{
\left(
\epsilon_{\scriptscriptstyle X}^2
+\frac{\eta_{\scriptscriptstyle X}^2}{T}
\ln\left(
1+\frac{\gamma^2T}
{[\max\{m_0^{\scriptscriptstyle X},
         m_0^{\scriptscriptstyle Y}\}]^2}
\right)
\right)
\left(
\epsilon_{\scriptscriptstyle Y}^2
+\frac{\eta_{\scriptscriptstyle Y}^2}{T}
\ln\left(
1+\frac{\gamma^2T}
{[m_0^{\scriptscriptstyle Y}]^2}
\right)
\right)
}
\notag\\
&\quad+
\frac{Lr\epsilon_{\scriptscriptstyle Y}}{2\mu}
+
\frac{Lr\eta_{\scriptscriptstyle Y}^2}
{2T\mu\epsilon_{\scriptscriptstyle Y}}
\ln\left(
1+\frac{\gamma^2T}
{[m_0^{\scriptscriptstyle Y}]^2}
\right)+
\frac{2\sqrt r}{\mu\epsilon_{\scriptscriptstyle Y}}
\sqrt{
\epsilon_{\scriptscriptstyle Y}^2
+\frac{\eta_{\scriptscriptstyle Y}^2}{T}
\ln\left(
1+\frac{\gamma^2T}
{[m_0^{\scriptscriptstyle Y}]^2}
\right)
}
\notag\\
&\quad\cdot
\Bigg(
\frac{3\beta^2G^2}{T(1-\beta)}
+
\frac{3(1-\beta)\kappa^2}{Tb_0}
+
3\epsilon_{\scriptscriptstyle X}^2L^2r+
\frac{3\eta_{\scriptscriptstyle X}^2L^2r}{T}
\ln\left(
1+\frac{\gamma^2T}
{[\max\{m_0^{\scriptscriptstyle X},
         m_0^{\scriptscriptstyle Y}\}]^2}
\right)
+
3\epsilon_{\scriptscriptstyle Y}^2L^2r
\notag\\
&\qquad+
\frac{3\eta_{\scriptscriptstyle Y}^2L^2r}{T}
\ln\left(
1+\frac{\gamma^2T}
{[m_0^{\scriptscriptstyle Y}]^2}
\right)+
\frac{6\beta^2
\epsilon_{\scriptscriptstyle X}^2L^2r}
{(1-\beta^2)\min_{0\le s<T}b_s}
\notag\\
&\qquad+
\frac{6\beta^2
\eta_{\scriptscriptstyle X}^2L^2r}
{T(1-\beta^2)\min_{0\le s<T}b_s}
\ln\left(
1+\frac{\gamma^2T}
{[\max\{m_0^{\scriptscriptstyle X},
         m_0^{\scriptscriptstyle Y}\}]^2}
\right)+
\frac{6\beta^2
\epsilon_{\scriptscriptstyle Y}^2L^2r}
{(1-\beta^2)\min_{0\le s<T}b_s}
\notag\\
&\qquad+
\frac{6\beta^2
\eta_{\scriptscriptstyle Y}^2L^2r}
{T(1-\beta^2)\min_{0\le s<T}b_s}
\ln\left(
1+\frac{\gamma^2T}
{[m_0^{\scriptscriptstyle Y}]^2}
\right)+
\frac{6(1-\beta)}{T(1+\beta)}
\sum_{t=1}^{T-1}\frac{\kappa^2}{b_t}
\Bigg)^{\frac12}
\Bigg].
\label{eq:E0-sum-split-again}
\end{align}

With the initialization
$M_0^{\scriptscriptstyle X}=0$,
$X_{-1}=X_0$, and $Y_{-1}=Y_0$, we have
\(
M_1^{\scriptscriptstyle X}
=
(1-\beta)
\nabla_XF(X_0,Y_0;\xi_0^{\scriptscriptstyle X})\) and \(\tilde E_1
=
M_1^{\scriptscriptstyle X}
-\nabla_Xf(X_0,Y_0)
=
-\beta\nabla_Xf(X_0,Y_0)
+(1-\beta)\big(
\nabla_XF(X_0,Y_0;\xi_0^{\scriptscriptstyle X})
-\nabla_Xf(X_0,Y_0)
\big).\)
Consequently, conditional unbiasedness,
$\|A\|_*\le\sqrt r\,\|A\|_F$,
Jensen's inequality, and \cref{stochasticgradient} imply
\begin{align*}
\mathbb E[\|\tilde E_1\|_*]
&\le
\beta\mathbb E[
\|\nabla_Xf(X_0,Y_0)\|_*
]+
(1-\beta)
\mathbb E\left[
\left\|
\nabla_XF(X_0,Y_0;\xi_0^{\scriptscriptstyle X})
-\nabla_Xf(X_0,Y_0)
\right\|_*
\right]
\\
&\le
\beta\sqrt r\,
\mathbb E[
\|\nabla_Xf(X_0,Y_0)\|_F
]+
(1-\beta)\sqrt r\,
\sqrt{
\mathbb E\left[
\left\|
\nabla_XF(X_0,Y_0;\xi_0^{\scriptscriptstyle X})
-\nabla_Xf(X_0,Y_0)
\right\|_F^2
\right]
}
\\
&\le
\beta\sqrt r\,G
+\frac{(1-\beta)\kappa\sqrt r}{\sqrt{b_0}}.
\end{align*}
Therefore,
\(
\frac{\mathbb E[\|\tilde E_1\|_*]}
{T(1-\beta)}
\le
\frac{\beta\sqrt r\,G}{T(1-\beta)}
+\frac{\kappa\sqrt r}{T\sqrt{b_0}}.
\)
Dividing both sides of \cref{eq:E0-sum-split-again}
by $T$ and using the initialization bound above,
we obtain
\begin{align}
&\frac1T\sum_{t=0}^{T-1}
\mathbb E[\|E_{t+1}\|_*]
\notag\\
&\le
\frac{\beta\sqrt r\,G}{T(1-\beta)}
+
\frac{\kappa\sqrt r}{T\sqrt{b_0}}+
\sqrt{\frac{r}{1-\beta^2}}\,
\Bigg(
\frac{2\beta^2L^2r
\epsilon_{\scriptscriptstyle X}^2}
{\min_{0\le s<T}b_s}+
\frac{2\beta^2L^2r
\eta_{\scriptscriptstyle X}^2}
{T\min_{0\le s<T}b_s}
\ln\left(
1+\frac{\gamma^2T}
{[\max\{m_0^{\scriptscriptstyle X},
         m_0^{\scriptscriptstyle Y}\}]^2}
\right)
\notag\\
&\quad+
\frac{2\beta^2L^2r
\epsilon_{\scriptscriptstyle Y}^2}
{\min_{0\le s<T}b_s}+
\frac{2\beta^2L^2r
\eta_{\scriptscriptstyle Y}^2}
{T\min_{0\le s<T}b_s}
\ln\left(
1+\frac{\gamma^2T}
{[m_0^{\scriptscriptstyle Y}]^2}
\right)+
\frac{2(1-\beta)^2}{T}
\sum_{t=1}^{T-1}\frac{\kappa^2}{b_t}
\Bigg)^{\frac12}
\notag\\
&\quad+
L\Bigg[
\frac{\Delta_0^{\scriptscriptstyle Y}}
{T\mu\epsilon_{\scriptscriptstyle Y}}
+
\frac{G\epsilon_{\scriptscriptstyle X}
\sqrt r(2+\lambda)}
{\mu\epsilon_{\scriptscriptstyle Y}}+
\frac{G\eta_{\scriptscriptstyle X}
\sqrt r(2+\lambda)}
{\mu\epsilon_{\scriptscriptstyle Y}\sqrt T}
\sqrt{
\ln\left(
1+\frac{\gamma^2T}
{[\max\{m_0^{\scriptscriptstyle X},
         m_0^{\scriptscriptstyle Y}\}]^2}
\right)
}
\notag\\
&\quad+
\frac{L\epsilon_{\scriptscriptstyle X}^2
(2+\lambda^2)}
{2\mu\epsilon_{\scriptscriptstyle Y}}
+
\frac{L\eta_{\scriptscriptstyle X}^2
(2+\lambda^2)}
{2T\mu\epsilon_{\scriptscriptstyle Y}}
\ln\left(
1+\frac{\gamma^2T}
{[\max\{m_0^{\scriptscriptstyle X},
         m_0^{\scriptscriptstyle Y}\}]^2}
\right)
\notag\\
&\quad+
\frac{L}{\mu\epsilon_{\scriptscriptstyle Y}}
\sqrt{
\left(
\epsilon_{\scriptscriptstyle X}^2
+\frac{\eta_{\scriptscriptstyle X}^2}{T}
\ln\left(
1+\frac{\gamma^2T}
{[\max\{m_0^{\scriptscriptstyle X},
         m_0^{\scriptscriptstyle Y}\}]^2}
\right)
\right)
\left(
\epsilon_{\scriptscriptstyle Y}^2
+\frac{\eta_{\scriptscriptstyle Y}^2}{T}
\ln\left(
1+\frac{\gamma^2T}
{[m_0^{\scriptscriptstyle Y}]^2}
\right)
\right)
}
\notag\\
&\quad+
\frac{Lr\epsilon_{\scriptscriptstyle Y}}{2\mu}
+
\frac{Lr\eta_{\scriptscriptstyle Y}^2}
{2T\mu\epsilon_{\scriptscriptstyle Y}}
\ln\left(
1+\frac{\gamma^2T}
{[m_0^{\scriptscriptstyle Y}]^2}
\right)+
\frac{2\sqrt r}{\mu\epsilon_{\scriptscriptstyle Y}}
\sqrt{
\epsilon_{\scriptscriptstyle Y}^2
+\frac{\eta_{\scriptscriptstyle Y}^2}{T}
\ln\left(
1+\frac{\gamma^2T}
{[m_0^{\scriptscriptstyle Y}]^2}
\right)
}
\notag\\
&\quad\cdot
\Bigg(
\frac{3\beta^2G^2}{T(1-\beta)}
+
\frac{3(1-\beta)\kappa^2}{Tb_0}
+
3\epsilon_{\scriptscriptstyle X}^2L^2r+
\frac{3\eta_{\scriptscriptstyle X}^2L^2r}{T}
\ln\left(
1+\frac{\gamma^2T}
{[\max\{m_0^{\scriptscriptstyle X},
         m_0^{\scriptscriptstyle Y}\}]^2}
\right)
+
3\epsilon_{\scriptscriptstyle Y}^2L^2r
\notag\\
&\qquad+
\frac{3\eta_{\scriptscriptstyle Y}^2L^2r}{T}
\ln\left(
1+\frac{\gamma^2T}
{[m_0^{\scriptscriptstyle Y}]^2}
\right)+
\frac{6\beta^2
\epsilon_{\scriptscriptstyle X}^2L^2r}
{(1-\beta^2)\min_{0\le s<T}b_s}
\notag\\
&\qquad+
\frac{6\beta^2
\eta_{\scriptscriptstyle X}^2L^2r}
{T(1-\beta^2)\min_{0\le s<T}b_s}
\ln\left(
1+\frac{\gamma^2T}
{[\max\{m_0^{\scriptscriptstyle X},
         m_0^{\scriptscriptstyle Y}\}]^2}
\right)+
\frac{6\beta^2
\epsilon_{\scriptscriptstyle Y}^2L^2r}
{(1-\beta^2)\min_{0\le s<T}b_s}
\notag\\
&\qquad+
\frac{6\beta^2
\eta_{\scriptscriptstyle Y}^2L^2r}
{T(1-\beta^2)\min_{0\le s<T}b_s}
\ln\left(
1+\frac{\gamma^2T}
{[m_0^{\scriptscriptstyle Y}]^2}
\right)+
\frac{6(1-\beta)}{T(1+\beta)}
\sum_{t=1}^{T-1}\frac{\kappa^2}{b_t}
\Bigg)^{\frac12}
\Bigg].
\label{eq:E-sum-split-again}
\end{align}

The proof for \cref{e3lemma} is thus completed.

\subsection{Proof of \cref{e4lemma}}\label{sece4}
Starting from the decomposition
\begin{align}
&M_{t+1}^{\scriptscriptstyle X}
\notag\\
&=
\beta M_t^{\scriptscriptstyle X}
+\nabla_X f(X_t,Y_t)
-\beta\nabla_X f(X_{t-1},Y_{t-1})
\notag\\
&\quad+
\left[
\big(
\nabla_X F(X_t,Y_t;\xi_t^{\scriptscriptstyle X})
-\nabla_X f(X_t,Y_t)
\big)
-\beta\big(
\nabla_X F(X_{t-1},Y_{t-1};
             \xi_t^{\scriptscriptstyle X})
-\nabla_X f(X_{t-1},Y_{t-1})
\big)
\right],
\label{eq:Mx-storm-decomp-again}
\end{align}
it follows that, for all $t\ge1$,
\begin{align}
&E_{t+1}
\notag\\
&=
\beta\big(
M_t^{\scriptscriptstyle X}
-\nabla\Phi(X_{t-1})
\big)
\notag\\
&\quad+
\beta\big(
\nabla\Phi(X_{t-1})-\nabla\Phi(X_t)
\big)
+\nabla_X f(X_t,Y_t)
-\beta\nabla_X f(X_{t-1},Y_{t-1})
-(1-\beta)\nabla\Phi(X_t)
\notag\\
&\quad+
\left[
\big(
\nabla_X F(X_t,Y_t;\xi_t^{\scriptscriptstyle X})
-\nabla_X f(X_t,Y_t)
\big)
-\beta\big(
\nabla_X F(X_{t-1},Y_{t-1};
             \xi_t^{\scriptscriptstyle X})
-\nabla_X f(X_{t-1},Y_{t-1})
\big)
\right]
\notag\\
&=
\beta E_t
+\beta\big(
\nabla\Phi(X_{t-1})-\nabla\Phi(X_t)
\big)
\notag\\
&\quad+
\nabla_X f(X_t,Y_t)
-\beta\nabla_X f(X_{t-1},Y_{t-1})
-(1-\beta)\nabla\Phi(X_t)
\notag\\
&\quad+
\left[
\big(
\nabla_X F(X_t,Y_t;\xi_t^{\scriptscriptstyle X})
-\nabla_X f(X_t,Y_t)
\big)
-\beta\big(
\nabla_X F(X_{t-1},Y_{t-1};
             \xi_t^{\scriptscriptstyle X})
-\nabla_X f(X_{t-1},Y_{t-1})
\big)
\right].
\label{eq:E-rec-storm-correct}
\end{align}

Taking squared Frobenius norms on both sides of
\cref{eq:E-rec-storm-correct} and then taking expectations,
we obtain
\begin{align}
&\mathbb E[\|E_{t+1}\|_F^2]
\notag\\
&=
\mathbb E\Bigg[
\Bigg\|
\beta E_t
+\beta\big(
\nabla\Phi(X_{t-1})-\nabla\Phi(X_t)
\big)+
\nabla_X f(X_t,Y_t)
-\beta\nabla_X f(X_{t-1},Y_{t-1})
-(1-\beta)\nabla\Phi(X_t)
\notag\\
&\qquad+
\big(
\nabla_X F(X_t,Y_t;\xi_t^{\scriptscriptstyle X})
-\nabla_X f(X_t,Y_t)
\big)-
\beta\big(
\nabla_X F(X_{t-1},Y_{t-1};
             \xi_t^{\scriptscriptstyle X})
-\nabla_X f(X_{t-1},Y_{t-1})
\big)
\Bigg\|_F^2
\Bigg]
\notag\\
&=
\mathbb E\Bigg[
\Bigg\|
\beta E_t
+\beta\big(
\nabla\Phi(X_{t-1})-\nabla\Phi(X_t)
\big)
\notag\\
&\qquad+
\nabla_X f(X_t,Y_t)
-\beta\nabla_X f(X_{t-1},Y_{t-1})
-(1-\beta)\nabla\Phi(X_t)
\Bigg\|_F^2
\Bigg]
\notag\\
&\quad+
2\mathbb E\Bigg\langle
\beta E_t
+\beta\big(
\nabla\Phi(X_{t-1})-\nabla\Phi(X_t)
\big)
\notag\\
&\qquad+
\nabla_X f(X_t,Y_t)
-\beta\nabla_X f(X_{t-1},Y_{t-1})
-(1-\beta)\nabla\Phi(X_t),
\notag\\
&\qquad
\big(
\nabla_X F(X_t,Y_t;\xi_t^{\scriptscriptstyle X})
-\nabla_X f(X_t,Y_t)
\big)-
\beta\big(
\nabla_X F(X_{t-1},Y_{t-1};
             \xi_t^{\scriptscriptstyle X})
-\nabla_X f(X_{t-1},Y_{t-1})
\big)
\Bigg\rangle
\notag\\
&\quad+
\mathbb E\Bigg[
\Bigg\|
\big(
\nabla_X F(X_t,Y_t;\xi_t^{\scriptscriptstyle X})
-\nabla_X f(X_t,Y_t)
\big)
\notag\\
&\qquad-
\beta\big(
\nabla_X F(X_{t-1},Y_{t-1};
             \xi_t^{\scriptscriptstyle X})
-\nabla_X f(X_{t-1},Y_{t-1})
\big)
\Bigg\|_F^2
\Bigg]
\notag\\
&=
\mathbb E\Bigg[
\Bigg\|
\beta E_t
+\beta\big(
\nabla\Phi(X_{t-1})-\nabla\Phi(X_t)
\big)
\notag\\
&\qquad+
\nabla_X f(X_t,Y_t)
-\beta\nabla_X f(X_{t-1},Y_{t-1})
-(1-\beta)\nabla\Phi(X_t)
\Bigg\|_F^2
\Bigg]
\notag\\
&\quad+
\mathbb E\Bigg[
\Bigg\|
\big(
\nabla_X F(X_t,Y_t;\xi_t^{\scriptscriptstyle X})
-\nabla_X f(X_t,Y_t)
\big)
\notag\\
&\qquad-
\beta\big(
\nabla_X F(X_{t-1},Y_{t-1};
             \xi_t^{\scriptscriptstyle X})
-\nabla_X f(X_{t-1},Y_{t-1})
\big)
\Bigg\|_F^2
\Bigg]
\notag\\
&\le
\beta\mathbb E[\|E_t\|_F^2]+
\frac{1}{1-\beta}
\mathbb E\Bigg[
\Bigg\|
\beta\big(
\nabla\Phi(X_{t-1})-\nabla\Phi(X_t)
\big)
\notag\\
&\qquad+
\nabla_X f(X_t,Y_t)
-\beta\nabla_X f(X_{t-1},Y_{t-1})
-(1-\beta)\nabla\Phi(X_t)
\Bigg\|_F^2
\Bigg]
\notag\\
&\quad+
\mathbb E\Bigg[
\Bigg\|
\big(
\nabla_X F(X_t,Y_t;\xi_t^{\scriptscriptstyle X})
-\nabla_X f(X_t,Y_t)
\big)
\notag\\
&\qquad-
\beta\big(
\nabla_X F(X_{t-1},Y_{t-1};
             \xi_t^{\scriptscriptstyle X})
-\nabla_X f(X_{t-1},Y_{t-1})
\big)
\Bigg\|_F^2
\Bigg].
\label{eq:Et1-second-cond}
\end{align}
The last inequality follows from Young's inequality:
$\|\beta A+B\|_F^2
\le\beta\|A\|_F^2
+\frac{1}{1-\beta}\|B\|_F^2$,
where $\beta\in[0,1)$.

To bound the second term in \cref{eq:Et1-second-cond},
note that
\[
\begin{aligned}
&\nabla_X f(X_t,Y_t)
-\beta\nabla_X f(X_{t-1},Y_{t-1})
-(1-\beta)\nabla\Phi(X_t)
\\
&=
(1-\beta)\big(
\nabla_X f(X_t,Y_t)-\nabla\Phi(X_t)
\big)+
\beta\big(
\nabla_X f(X_t,Y_t)
-\nabla_X f(X_{t-1},Y_{t-1})
\big).
\end{aligned}
\]
Therefore, by the inequality
$\|A+B+C\|_F^2
\le3(\|A\|_F^2+\|B\|_F^2+\|C\|_F^2)$,
we obtain
\begin{align}
&\mathbb E\Bigg[
\Bigg\|
\beta\big(
\nabla\Phi(X_{t-1})-\nabla\Phi(X_t)
\big)+
\nabla_X f(X_t,Y_t)
-\beta\nabla_X f(X_{t-1},Y_{t-1})
-(1-\beta)\nabla\Phi(X_t)
\Bigg\|_F^2
\Bigg]
\notag\\
&=
\mathbb E\Bigg[
\Bigg\|
\beta\big(
\nabla\Phi(X_{t-1})-\nabla\Phi(X_t)
\big)+
(1-\beta)\big(
\nabla_X f(X_t,Y_t)-\nabla\Phi(X_t)
\big)
\notag\\
&\qquad+
\beta\big(
\nabla_X f(X_t,Y_t)
-\nabla_X f(X_{t-1},Y_{t-1})
\big)
\Bigg\|_F^2
\Bigg]
\notag\\
&\le
3\beta^2
\mathbb E\left[
\|\nabla\Phi(X_t)-\nabla\Phi(X_{t-1})\|_F^2
\right]+
3(1-\beta)^2
\mathbb E\left[
\|\nabla_X f(X_t,Y_t)-\nabla\Phi(X_t)\|_F^2
\right]
\notag\\
&\quad+
3\beta^2
\mathbb E\left[
\|\nabla_X f(X_t,Y_t)
-\nabla_X f(X_{t-1},Y_{t-1})\|_F^2
\right]
\notag\\
&\le
3\beta^2L_\Phi^2r
\mathbb E\left[
\left(
\max\{
\epsilon_{\scriptscriptstyle X},
\eta_{\scriptscriptstyle X}\theta_t
\}
\right)^2
\right]+
3(1-\beta)^2L^2
\mathbb E[
\|Y_t-Y^\star(X_t)\|_F^2
]
\notag\\
&\quad+
3\beta^2L^2r
\left(
\mathbb E\left[
\left(
\max\{
\epsilon_{\scriptscriptstyle X},
\eta_{\scriptscriptstyle X}\theta_t
\}
\right)^2
\right]
+
\mathbb E\left[
\left(
\max\{
\epsilon_{\scriptscriptstyle Y},
\eta_{\scriptscriptstyle Y}\sigma_t
\}
\right)^2
\right]
\right).
\label{eq:bt-second-bound}
\end{align}
The last inequality follows from
\cref{smoothness}, \cref{additionallemma},
and the global rank bound $r$, which yields
$\|X_t-X_{t-1}\|_F^2
\le r(\max\{
\epsilon_{\scriptscriptstyle X},
\eta_{\scriptscriptstyle X}\theta_t
\})^2$
and
$\|Y_t-Y_{t-1}\|_F^2
\le r(\max\{
\epsilon_{\scriptscriptstyle Y},
\eta_{\scriptscriptstyle Y}\sigma_t
\})^2$.

Next, according to \cref{eq:Delta-sq-split},
the third term in \cref{eq:Et1-second-cond}
can be upper bounded by
\begin{align}
&\mathbb E\Bigg[
\Bigg\|
\big(
\nabla_X F(X_t,Y_t;\xi_t^{\scriptscriptstyle X})
-\nabla_X f(X_t,Y_t)
\big)-
\beta\big(
\nabla_X F(X_{t-1},Y_{t-1};
             \xi_t^{\scriptscriptstyle X})
-\nabla_X f(X_{t-1},Y_{t-1})
\big)
\Bigg\|_F^2
\Bigg]
\notag\\
&\le
\frac{2\beta^2L^2}{b_t}
\mathbb E[\|X_t-X_{t-1}\|_F^2]
+
\frac{2\beta^2L^2}{b_t}
\mathbb E[\|Y_t-Y_{t-1}\|_F^2]
+
\frac{2(1-\beta)^2\kappa^2}{b_t}
\notag\\
&\le
\frac{2\beta^2L^2r}{b_t}
\mathbb E\left[
\left(
\max\{
\epsilon_{\scriptscriptstyle X},
\eta_{\scriptscriptstyle X}\theta_t
\}
\right)^2
\right]+
\frac{2\beta^2L^2r}{b_t}
\mathbb E\left[
\left(
\max\{
\epsilon_{\scriptscriptstyle Y},
\eta_{\scriptscriptstyle Y}\sigma_t
\}
\right)^2
\right]
+
\frac{2(1-\beta)^2\kappa^2}{b_t}.
\label{eq:Delta-second-refined-x}
\end{align}

Summing \cref{eq:Et1-second-cond} over
$t=1,\ldots,T-1$, keeping the initial term separately,
and substituting \cref{eq:bt-second-bound} and
\cref{eq:Delta-second-refined-x}, we obtain
\begin{align*}
&(1-\beta)
\sum_{t=0}^{T-1}\mathbb E[\|E_{t+1}\|_F^2]
\\
&\le
\mathbb E[\|E_1\|_F^2]
-\beta\mathbb E[\|E_T\|_F^2]+
\left(
\frac{3\beta^2L_\Phi^2r}{1-\beta}
+\frac{3\beta^2L^2r}{1-\beta}
\right)
\sum_{t=1}^{T-1}
\mathbb E\left[
\left(
\max\{
\epsilon_{\scriptscriptstyle X},
\eta_{\scriptscriptstyle X}\theta_t
\}
\right)^2
\right]
\\
&\quad+
\frac{3\beta^2L^2r}{1-\beta}
\sum_{t=1}^{T-1}
\mathbb E\left[
\left(
\max\{
\epsilon_{\scriptscriptstyle Y},
\eta_{\scriptscriptstyle Y}\sigma_t
\}
\right)^2
\right]+
2\beta^2L^2r
\sum_{t=1}^{T-1}\frac1{b_t}
\mathbb E\left[
\left(
\max\{
\epsilon_{\scriptscriptstyle X},
\eta_{\scriptscriptstyle X}\theta_t
\}
\right)^2
\right]
\\
&\quad+
2\beta^2L^2r
\sum_{t=1}^{T-1}\frac1{b_t}
\mathbb E\left[
\left(
\max\{
\epsilon_{\scriptscriptstyle Y},
\eta_{\scriptscriptstyle Y}\sigma_t
\}
\right)^2
\right]
\\
&\quad+
3(1-\beta)L^2
\sum_{t=1}^{T-1}
\mathbb E[\|Y_t-Y^\star(X_t)\|_F^2]
+
2(1-\beta)^2
\sum_{t=1}^{T-1}\frac{\kappa^2}{b_t}.
\end{align*}
Dividing by $1-\beta$, using
$b_t\ge\min_{0\le s<T}b_s$, and applying
\cref{lem:log_cum_ratio} with the positive initialization
yield
\begin{align}
&\sum_{t=0}^{T-1}\mathbb E[\|E_{t+1}\|_F^2]
\notag\\
&\le
\frac{\mathbb E[\|E_1\|_F^2]}{1-\beta}
-\frac{\beta}{1-\beta}
\mathbb E[\|E_T\|_F^2]+
\sum_{t=1}^{T-1}
\left(
\frac{3\beta^2L_\Phi^2r}{(1-\beta)^2}
+\frac{3\beta^2L^2r}{(1-\beta)^2}
+\frac{2\beta^2L^2r}{(1-\beta)b_t}
\right)
\mathbb E\left[
\left(
\max\{
\epsilon_{\scriptscriptstyle X},
\eta_{\scriptscriptstyle X}\theta_t
\}
\right)^2
\right]
\notag\\
&\quad+
\sum_{t=1}^{T-1}
\left(
\frac{3\beta^2L^2r}{(1-\beta)^2}
+\frac{2\beta^2L^2r}{(1-\beta)b_t}
\right)
\mathbb E\left[
\left(
\max\{
\epsilon_{\scriptscriptstyle Y},
\eta_{\scriptscriptstyle Y}\sigma_t
\}
\right)^2
\right]
\notag\\
&\quad+
3L^2\sum_{t=1}^{T-1}
\mathbb E[\|Y_t-Y^\star(X_t)\|_F^2]
+
2(1-\beta)\sum_{t=1}^{T-1}\frac{\kappa^2}{b_t}
\notag\\
&\le
\frac{\mathbb E[\|E_1\|_F^2]}{1-\beta}+
\left(
\frac{3\beta^2L_\Phi^2r}{(1-\beta)^2}
+\frac{3\beta^2L^2r}{(1-\beta)^2}
+\frac{2\beta^2L^2r}
{(1-\beta)\min_{0\le s<T}b_s}
\right)
\sum_{t=1}^{T-1}
\mathbb E\left[
\left(
\max\{
\epsilon_{\scriptscriptstyle X},
\eta_{\scriptscriptstyle X}\theta_t
\}
\right)^2
\right]
\notag\\
&\quad+
\left(
\frac{3\beta^2L^2r}{(1-\beta)^2}
+\frac{2\beta^2L^2r}
{(1-\beta)\min_{0\le s<T}b_s}
\right)
\sum_{t=1}^{T-1}
\mathbb E\left[
\left(
\max\{
\epsilon_{\scriptscriptstyle Y},
\eta_{\scriptscriptstyle Y}\sigma_t
\}
\right)^2
\right]
\notag\\
&\quad+
3L^2\sum_{t=0}^{T-1}
\mathbb E[\|Y_t-Y^\star(X_t)\|_F^2]
+
2(1-\beta)\sum_{t=1}^{T-1}\frac{\kappa^2}{b_t}
\notag\\
&\le
\frac{\mathbb E[\|E_1\|_F^2]}{1-\beta}+
3L^2\sum_{t=0}^{T-1}
\mathbb E[\|Y_t-Y^\star(X_t)\|_F^2]
+
2(1-\beta)\sum_{t=1}^{T-1}\frac{\kappa^2}{b_t}
\notag\\
&\quad+
\left(
\frac{3\beta^2L_\Phi^2r}{(1-\beta)^2}
+\frac{3\beta^2L^2r}{(1-\beta)^2}
+\frac{2\beta^2L^2r}
{(1-\beta)\min_{0\le s<T}b_s}
\right)
\left(
T\epsilon_{\scriptscriptstyle X}^2
+
\eta_{\scriptscriptstyle X}^2
\ln\left(
1+\frac{\gamma^2T}
{[\max\{m_0^{\scriptscriptstyle X},
         m_0^{\scriptscriptstyle Y}\}]^2}
\right)
\right)
\notag\\
&\quad+
\left(
\frac{3\beta^2L^2r}{(1-\beta)^2}
+\frac{2\beta^2L^2r}
{(1-\beta)\min_{0\le s<T}b_s}
\right)
\left(
T\epsilon_{\scriptscriptstyle Y}^2
+
\eta_{\scriptscriptstyle Y}^2
\ln\left(
1+\frac{\gamma^2T}
{[m_0^{\scriptscriptstyle Y}]^2}
\right)
\right).
\label{eq:sumEt1-second-final}
\end{align}

We next bound the initialization error.
For fixed initial iterates, the initialization
$M_0^{\scriptscriptstyle X}=0$,
$X_{-1}=X_0$, and $Y_{-1}=Y_0$ gives
\(
E_1
=
(1-\beta)
\nabla_XF(X_0,Y_0;\xi_0^{\scriptscriptstyle X})
-\nabla\Phi(X_0).
\)
By strong concavity and inner optimality,
\(
\frac{\mu}{2}
\|Y_0-Y^\star(X_0)\|_F^2
\le
\Phi(X_0)-f(X_0,Y_0)
=
\Delta_0^{\scriptscriptstyle Y}.
\)
Moreover, \cref{smoothness,additionallemma} imply
\[
\begin{aligned}
&\|\nabla_Xf(X_0,Y_0)-\nabla\Phi(X_0)\|_F^2=
\|\nabla_Xf(X_0,Y_0)
-\nabla_Xf(X_0,Y^\star(X_0))\|_F^2\le
L^2\|Y_0-Y^\star(X_0)\|_F^2.
\end{aligned}
\]
Using conditional unbiasedness of the initial batch and
\cref{stochasticgradient}, we obtain
\begin{align*}
\mathbb E[\|E_1\|_F^2]
&=
\left\|
(1-\beta)\nabla_Xf(X_0,Y_0)
-\nabla\Phi(X_0)
\right\|_F^2\\
&\quad+
(1-\beta)^2
\mathbb E\left[
\left\|
\nabla_XF(X_0,Y_0;\xi_0^{\scriptscriptstyle X})
-\nabla_Xf(X_0,Y_0)
\right\|_F^2
\right]\\
&=
\left\|
-\beta\nabla_Xf(X_0,Y_0)
+\big(
\nabla_Xf(X_0,Y_0)-\nabla\Phi(X_0)
\big)
\right\|_F^2\\
&\quad+
(1-\beta)^2
\mathbb E\left[
\left\|
\nabla_XF(X_0,Y_0;\xi_0^{\scriptscriptstyle X})
-\nabla_Xf(X_0,Y_0)
\right\|_F^2
\right]\\
&\le
2\beta^2\|\nabla_Xf(X_0,Y_0)\|_F^2
+
2\|\nabla_Xf(X_0,Y_0)-\nabla\Phi(X_0)\|_F^2\\
&\quad+
(1-\beta)^2
\mathbb E\left[
\left\|
\nabla_XF(X_0,Y_0;\xi_0^{\scriptscriptstyle X})
-\nabla_Xf(X_0,Y_0)
\right\|_F^2
\right]\\
&\le
2\beta^2G^2
+
2L^2\|Y_0-Y^\star(X_0)\|_F^2
+
\frac{(1-\beta)^2\kappa^2}{b_0}\\
&\le
2\beta^2G^2
+
\frac{4L^2\Delta_0^{\scriptscriptstyle Y}}{\mu}
+
\frac{(1-\beta)^2\kappa^2}{b_0}.
\end{align*}

Then, applying \cref{stochasticgradient} and invoking
\cref{eq:dual-avg-dt2-final-noVR222}, we obtain
\begin{align}
&\sum_{t=0}^{T-1}
\mathbb E[\|E_{t+1}\|_F^2]
\notag\\
&\le
\frac{2\beta^2G^2}{1-\beta}
+
\frac{4L^2\Delta_0^{\scriptscriptstyle Y}}
{\mu(1-\beta)}
+
\frac{(1-\beta)\kappa^2}{b_0}
\notag\\
&\quad+
\left(
\frac{3\beta^2L_\Phi^2r}{(1-\beta)^2}
+
\frac{3\beta^2L^2r}{(1-\beta)^2}
+
\frac{2\beta^2L^2r}
{(1-\beta)\min_{0\le s<T}b_s}
\right)
\left(
T\epsilon_{\scriptscriptstyle X}^2
+
\eta_{\scriptscriptstyle X}^2
\ln\left(
1+\frac{\gamma^2T}
{[\max\{m_0^{\scriptscriptstyle X},
         m_0^{\scriptscriptstyle Y}\}]^2}
\right)
\right)
\notag\\
&\quad+
\left(
\frac{3\beta^2L^2r}{(1-\beta)^2}
+
\frac{2\beta^2L^2r}
{(1-\beta)\min_{0\le s<T}b_s}
\right)
\left(
T\epsilon_{\scriptscriptstyle Y}^2
+
\eta_{\scriptscriptstyle Y}^2
\ln\left(
1+\frac{\gamma^2T}
{[m_0^{\scriptscriptstyle Y}]^2}
\right)
\right)
+
2(1-\beta)
\sum_{t=1}^{T-1}\frac{\kappa^2}{b_t}
\notag\\
&\quad+
\frac{24GL^2\Delta_0^{\scriptscriptstyle Y}}
{7\mu^2\epsilon_{\scriptscriptstyle Y}}+
\frac{12GrL^2(L_\Phi+L)
\epsilon_{\scriptscriptstyle X}^2T}
{7\mu^2\epsilon_{\scriptscriptstyle Y}}
+
\frac{48r^2L^4G
\epsilon_{\scriptscriptstyle X}^2T}
{7\mu^3\epsilon_{\scriptscriptstyle Y}}
+
\frac{48rG^2L^4
\epsilon_{\scriptscriptstyle X}^2T}
{7\mu^4\epsilon_{\scriptscriptstyle Y}^2}
\notag\\
&\quad+
\left(
\frac{12GrL^2(L_\Phi+L)
\eta_{\scriptscriptstyle X}^2}
{7\mu^2\epsilon_{\scriptscriptstyle Y}}
+
\frac{48r^2L^4G
\eta_{\scriptscriptstyle X}^2}
{7\mu^3\epsilon_{\scriptscriptstyle Y}}
+
\frac{48rG^2L^4
\eta_{\scriptscriptstyle X}^2}
{7\mu^4\epsilon_{\scriptscriptstyle Y}^2}
\right)
\ln\left(
1+\frac{\gamma^2T}
{[\max\{m_0^{\scriptscriptstyle X},
         m_0^{\scriptscriptstyle Y}\}]^2}
\right)
\notag\\
&\quad+
\frac{3GL^2\epsilon_{\scriptscriptstyle Y}T}
{7\mu}
+
\frac{12GL^3r\epsilon_{\scriptscriptstyle Y}T}
{7\mu^2}+
\left(
\frac{3GL^2\eta_{\scriptscriptstyle Y}^2}
{7\mu\epsilon_{\scriptscriptstyle Y}}
+
\frac{12GL^3r\eta_{\scriptscriptstyle Y}^2}
{7\mu^2\epsilon_{\scriptscriptstyle Y}}
\right)
\ln\left(
1+\frac{\gamma^2T}
{[m_0^{\scriptscriptstyle Y}]^2}
\right)
\notag\\
&\quad+
\frac{48G\sqrt r\,L^2T}
{7\mu^2\epsilon_{\scriptscriptstyle Y}}
\sqrt{
\epsilon_{\scriptscriptstyle Y}^2
+\frac{\eta_{\scriptscriptstyle Y}^2}{T}
\ln\left(
1+\frac{\gamma^2T}
{[m_0^{\scriptscriptstyle Y}]^2}
\right)
}\cdot
\Bigg(
\frac{3\beta^2G^2}{T(1-\beta)}
+
\frac{3(1-\beta)\kappa^2}{Tb_0}
+
3\epsilon_{\scriptscriptstyle X}^2L^2r
\notag\\
&\quad+
\frac{3\eta_{\scriptscriptstyle X}^2L^2r}{T}
\ln\left(
1+\frac{\gamma^2T}
{[\max\{m_0^{\scriptscriptstyle X},
         m_0^{\scriptscriptstyle Y}\}]^2}
\right)
+
3\epsilon_{\scriptscriptstyle Y}^2L^2r+
\frac{3\eta_{\scriptscriptstyle Y}^2L^2r}{T}
\ln\left(
1+\frac{\gamma^2T}
{[m_0^{\scriptscriptstyle Y}]^2}
\right)
\notag\\
&\quad+
\frac{6\beta^2
\epsilon_{\scriptscriptstyle X}^2L^2r}
{(1-\beta^2)\min_{0\le s<T}b_s}+
\frac{6\beta^2
\eta_{\scriptscriptstyle X}^2L^2r}
{T(1-\beta^2)\min_{0\le s<T}b_s}
\ln\left(
1+\frac{\gamma^2T}
{[\max\{m_0^{\scriptscriptstyle X},
         m_0^{\scriptscriptstyle Y}\}]^2}
\right)
\notag\\
&\quad+
\frac{6\beta^2
\epsilon_{\scriptscriptstyle Y}^2L^2r}
{(1-\beta^2)\min_{0\le s<T}b_s}+
\frac{6\beta^2
\eta_{\scriptscriptstyle Y}^2L^2r}
{T(1-\beta^2)\min_{0\le s<T}b_s}
\ln\left(
1+\frac{\gamma^2T}
{[m_0^{\scriptscriptstyle Y}]^2}
\right)+
\frac{6(1-\beta)}{T(1+\beta)}
\sum_{t=1}^{T-1}\frac{\kappa^2}{b_t}
\Bigg)^{\frac12}.
\label{eq:sumEt10-second-final}
\end{align}
The proof for \cref{e4lemma} is thus completed.

If we set
$1-\beta=T^{-\frac{1}{3}}$,
$\epsilon_{\scriptscriptstyle X}=T^{-\frac{2}{3}}$,
$\epsilon_{\scriptscriptstyle Y}=T^{-\frac{1}{3}}$,
$\eta_{\scriptscriptstyle X}=T^{-(\frac{2}{3}+q)}$,
$\eta_{\scriptscriptstyle Y}=T^{-(\frac{1}{2}+q)}$,
where $q\ge0$ is fixed, together with
$T\ge2$, $\gamma>0$,
$m_0^{\scriptscriptstyle X}
=m_0^{\scriptscriptstyle Y}>0$,
and
$b_t\geq T^{\frac{1}{3}}$
for every $t\in\{0,\ldots,T-1\}$,
then
\(
1-\beta^2
=
T^{-\frac{1}{3}}(1+\beta)\), \(\frac{\beta^2}{1+\beta}\le\frac12\), and \(\frac1{1+\beta}\le1.\)
Moreover,
\(
\frac1{b_0}\le T^{-\frac{1}{3}}\), \(\frac1{\min_{0\le s<T}b_s}\le T^{-\frac{1}{3}}\), and \(\sum_{t=1}^{T-1}\frac{\kappa^2}{b_t}
\le
(T-1)\kappa^2T^{-\frac{1}{3}}
\le
\kappa^2T^{\frac{2}{3}}\).
Therefore, substituting the schedules into
\cref{eq:E-sum-split-again}, we obtain
\begin{align}
&\frac1T\sum_{t=0}^{T-1}
\mathbb E[\|E_{t+1}\|_*]
\notag\\
&\le
\frac{\sqrt r\,G}{T^{\frac{2}{3}}}
+
\frac{\kappa\sqrt r}{T^{\frac{7}{6}}}+
\Bigg(
\frac{r^2L^2+2r\kappa^2}{T^{\frac{2}{3}}}
+
\frac{r^2L^2}{T^{\frac{4}{3}}}+
\frac{r^2L^2}{T^{\frac{7}{3}+2q}}
\ln\left(
1+\frac{\gamma^2T}
{[\max\{m_0^{\scriptscriptstyle X},
         m_0^{\scriptscriptstyle Y}\}]^2}
\right)
\notag\\
&\quad+
\frac{r^2L^2}{T^{2+2q}}
\ln\left(
1+\frac{\gamma^2T}
{[m_0^{\scriptscriptstyle Y}]^2}
\right)
\Bigg)^{\frac12}+
\frac{L\Delta_0^{\scriptscriptstyle Y}}
{\mu T^{\frac{2}{3}}}
+
\frac{GL\sqrt r(2+\lambda)}
{\mu T^{\frac{1}{3}}}
\notag\\
&\quad+
\frac{GL\sqrt r(2+\lambda)}
{\mu T^{\frac{5}{6}+q}}
\sqrt{
\ln\left(
1+\frac{\gamma^2T}
{[\max\{m_0^{\scriptscriptstyle X},
         m_0^{\scriptscriptstyle Y}\}]^2}
\right)
}
\notag\\
&\quad+
\frac{L^2(2+\lambda^2)}{2\mu T}
+
\frac{L^2(2+\lambda^2)}
{2\mu T^{2+2q}}
\ln\left(
1+\frac{\gamma^2T}
{[\max\{m_0^{\scriptscriptstyle X},
         m_0^{\scriptscriptstyle Y}\}]^2}
\right)
\notag\\
&\quad+
\frac{L^2T^{\frac{1}{3}}}{\mu}
\sqrt{
\left(
\frac1{T^{\frac{4}{3}}}
+
\frac1{T^{\frac{7}{3}+2q}}
\ln\left(
1+\frac{\gamma^2T}
{[\max\{m_0^{\scriptscriptstyle X},
         m_0^{\scriptscriptstyle Y}\}]^2}
\right)
\right)
\left(
\frac1{T^{\frac{2}{3}}}
+
\frac1{T^{2+2q}}
\ln\left(
1+\frac{\gamma^2T}
{[m_0^{\scriptscriptstyle Y}]^2}
\right)
\right)
}
\notag\\
&\quad+
\frac{L^2r}{2\mu T^{\frac{1}{3}}}
+
\frac{L^2r}{2\mu T^{\frac{5}{3}+2q}}
\ln\left(
1+\frac{\gamma^2T}
{[m_0^{\scriptscriptstyle Y}]^2}
\right)+
\frac{2L\sqrt r\,T^{\frac{1}{3}}}{\mu}
\sqrt{
\frac1{T^{\frac{2}{3}}}
+
\frac1{T^{2+2q}}
\ln\left(
1+\frac{\gamma^2T}
{[m_0^{\scriptscriptstyle Y}]^2}
\right)
}
\notag\\
&\quad\cdot
\Bigg(
\frac{3G^2+6L^2r+6\kappa^2}{T^{\frac{2}{3}}}
+
\frac{6L^2r}{T^{\frac{4}{3}}}
+
\frac{3\kappa^2}{T^{\frac{5}{3}}}+
\frac{6L^2r}{T^{\frac{7}{3}+2q}}
\ln\left(
1+\frac{\gamma^2T}
{[\max\{m_0^{\scriptscriptstyle X},
         m_0^{\scriptscriptstyle Y}\}]^2}
\right)
\notag\\
&\qquad+
\frac{6L^2r}{T^{2+2q}}
\ln\left(
1+\frac{\gamma^2T}
{[m_0^{\scriptscriptstyle Y}]^2}
\right)
\Bigg)^{\frac12}
\notag\\
&\lesssim \frac{
rL+\kappa\sqrt r
}{T^{\frac{1}{3}}}+
\frac{
{L\sqrt r}
G(1+\lambda)+L^2 r+\kappa L\sqrt r
}{\mu T^{\frac{1}{3}}}
+
\frac{
\sqrt r\,G
}{T^{\frac{2}{3}}}+
\frac{
{L}
\Delta_0^{\scriptscriptstyle Y}+L^2
}{\mu T^{\frac{2}{3}}}.
\label{final111}
\end{align}

Similarly, for \cref{e4lemma}, we have
\begin{align}
&\frac{2\sqrt r\,\eta_{\scriptscriptstyle X}}
{T\epsilon_{\scriptscriptstyle X}}
\sqrt{
\ln\left(
1+\frac{\gamma^2T}
{[\max\{m_0^{\scriptscriptstyle X},
         m_0^{\scriptscriptstyle Y}\}]^2}
\right)
}
\sqrt{
\sum_{t=0}^{T-1}
\mathbb E[\|E_{t+1}\|_F^2]
}
\notag\\
&=
\frac{2\sqrt r}{T^{1+q}}
\sqrt{
\ln\left(
1+\frac{\gamma^2T}
{[\max\{m_0^{\scriptscriptstyle X},
         m_0^{\scriptscriptstyle Y}\}]^2}
\right)
}
\sqrt{
\sum_{t=0}^{T-1}
\mathbb E[\|E_{t+1}\|_F^2]
}
\notag\\
&\le
\frac{2\sqrt r}{T^{1+q}}
\sqrt{
\ln\left(
1+\frac{\gamma^2T}
{[\max\{m_0^{\scriptscriptstyle X},
         m_0^{\scriptscriptstyle Y}\}]^2}
\right)
}\cdot
\Bigg[
3rL^2T
+
\left(
\frac{3GL^2}{7\mu}
+\frac{12GL^3r}{7\mu^2}
\right)T^{\frac{2}{3}}
\notag\\
&\quad+
\left(
2G^2+2\kappa^2+3rL_\Phi^2+5rL^2
+\frac{4L^2\Delta_0^{\scriptscriptstyle Y}}{\mu}
+\frac{24GL^2\Delta_0^{\scriptscriptstyle Y}}
{7\mu^2}
+\frac{48rG^2L^4}{7\mu^4}
\right)T^{\frac{1}{3}}
\notag\\
&\quad+
\frac{2rL^2}{T^{\frac{1}{3}}}
+\frac{\kappa^2}{T^{\frac{2}{3}}}
+\frac{12GrL^2(L_\Phi+L)}{7\mu^2}
+\frac{48r^2GL^4}{7\mu^3}
+
\Bigg(
\frac{
3r(L_\Phi^2+L^2)
+\dfrac{48rG^2L^4}{7\mu^4}
}
{T^{\frac{2}{3}+2q}}
+
\frac{2rL^2}{T^{\frac{4}{3}+2q}}
\notag\\
&\quad+
\frac{
\dfrac{12GrL^2(L_\Phi+L)}{7\mu^2}
+\dfrac{48r^2GL^4}{7\mu^3}
}
{T^{1+2q}}
\Bigg)
\ln\left(
1+\frac{\gamma^2T}
{[\max\{m_0^{\scriptscriptstyle X},
         m_0^{\scriptscriptstyle Y}\}]^2}
\right)
\notag\\
&\quad+
\Bigg(
\frac{3rL^2}{T^{\frac{1}{3}+2q}}
+
\frac{2rL^2}{T^{1+2q}}
+
\frac{
\dfrac{3GL^2}{7\mu}
+\dfrac{12GL^3r}{7\mu^2}
}
{T^{\frac{2}{3}+2q}}
\Bigg)
\ln\left(
1+\frac{\gamma^2T}
{[m_0^{\scriptscriptstyle Y}]^2}
\right)
\notag\\
&\quad+
\frac{48G\sqrt r\,L^2T^{\frac{4}{3}}}{7\mu^2}
\sqrt{
\frac1{T^{\frac{2}{3}}}
+
\frac1{T^{2+2q}}
\ln\left(
1+\frac{\gamma^2T}
{[m_0^{\scriptscriptstyle Y}]^2}
\right)
}
\notag\\
&\quad\cdot
\Bigg(
\frac{3G^2+6L^2r+6\kappa^2}{T^{\frac{2}{3}}}
+
\frac{6L^2r}{T^{\frac{4}{3}}}
+
\frac{3\kappa^2}{T^{\frac{5}{3}}}+
\frac{6L^2r}{T^{\frac{7}{3}+2q}}
\ln\left(
1+\frac{\gamma^2T}
{[\max\{m_0^{\scriptscriptstyle X},
         m_0^{\scriptscriptstyle Y}\}]^2}
\right)
\notag\\
&\quad+
\frac{6L^2r}{T^{2+2q}}
\ln\left(
1+\frac{\gamma^2T}
{[m_0^{\scriptscriptstyle Y}]^2}
\right)
\Bigg)^{\frac12}
\Bigg]^{\frac12}
\notag\\
&\lesssim
\Bigg[
\frac{rL}{T^{\frac{1}{2}+q}}
+
\frac{
GL\left(
\mu+rL
+\sqrt r\sqrt{G^2+L^2r+\kappa^2}
\right)
}
{\mu^2T^{\frac{5}{6}+q}}
\Bigg]\sqrt{
\ln\left(
1+\frac{\gamma^2T}
{[\max\{m_0^{\scriptscriptstyle X},
         m_0^{\scriptscriptstyle Y}\}]^2}
\right)
}.
\label{final222}
\end{align}

Substituting \cref{final111} and \cref{final222} into \cref{descentlemma}, we obtain
\begin{align}
&\frac{1}{T} \sum_{t=1}^T \mathbb{E}[\|\nabla\Phi(X_{t})\|_*]\notag\\
&\leq \frac{\Delta}{T^\frac{1}{3}} \!+\! 2\Bigg(\frac{\sqrt r\,G}{T^{\frac{2}{3}}}
+
\frac{\kappa\sqrt r}{T^{\frac{7}{6}}}+
\Bigg(
\frac{r^2L^2+2r\kappa^2}{T^{\frac{2}{3}}}
+
\frac{r^2L^2}{T^{\frac{4}{3}}}+
\frac{r^2L^2}{T^{\frac{7}{3}+2q}}
\ln\left(
1+\frac{\gamma^2T}
{[\max\{m_0^{\scriptscriptstyle X},
         m_0^{\scriptscriptstyle Y}\}]^2}
\right)
\notag\\
&\quad+
\frac{r^2L^2}{T^{2+2q}}
\ln\left(
1+\frac{\gamma^2T}
{[m_0^{\scriptscriptstyle Y}]^2}
\right)
\Bigg)^{\frac12}+
\frac{L\Delta_0^{\scriptscriptstyle Y}}
{\mu T^{\frac{2}{3}}}
+
\frac{GL\sqrt r(2+\lambda)}
{\mu T^{\frac{1}{3}}}
\notag\\
&\quad+
\frac{GL\sqrt r(2+\lambda)}
{\mu T^{\frac{5}{6}+q}}
\sqrt{
\ln\left(
1+\frac{\gamma^2T}
{[\max\{m_0^{\scriptscriptstyle X},
         m_0^{\scriptscriptstyle Y}\}]^2}
\right)
}
\notag\\
&\quad+
\frac{L^2(2+\lambda^2)}{2\mu T}
+
\frac{L^2(2+\lambda^2)}
{2\mu T^{2+2q}}
\ln\left(
1+\frac{\gamma^2T}
{[\max\{m_0^{\scriptscriptstyle X},
         m_0^{\scriptscriptstyle Y}\}]^2}
\right)
\notag\\
&\quad+
\frac{L^2T^{\frac{1}{3}}}{\mu}
\sqrt{
\left(
\frac1{T^{\frac{4}{3}}}
+
\frac1{T^{\frac{7}{3}+2q}}
\ln\left(
1+\frac{\gamma^2T}
{[\max\{m_0^{\scriptscriptstyle X},
         m_0^{\scriptscriptstyle Y}\}]^2}
\right)
\right)
\left(
\frac1{T^{\frac{2}{3}}}
+
\frac1{T^{2+2q}}
\ln\left(
1+\frac{\gamma^2T}
{[m_0^{\scriptscriptstyle Y}]^2}
\right)
\right)
}
\notag\\
&\quad+
\frac{L^2r}{2\mu T^{\frac{1}{3}}}
+
\frac{L^2r}{2\mu T^{\frac{5}{3}+2q}}
\ln\left(
1+\frac{\gamma^2T}
{[m_0^{\scriptscriptstyle Y}]^2}
\right)+
\frac{2L\sqrt r\,T^{\frac{1}{3}}}{\mu}
\sqrt{
\frac1{T^{\frac{2}{3}}}
+
\frac1{T^{2+2q}}
\ln\left(
1+\frac{\gamma^2T}
{[m_0^{\scriptscriptstyle Y}]^2}
\right)
}
\notag\\
&\quad\cdot
\Bigg(
\frac{3G^2+6L^2r+6\kappa^2}{T^{\frac{2}{3}}}
+
\frac{6L^2r}{T^{\frac{4}{3}}}
+
\frac{3\kappa^2}{T^{\frac{5}{3}}}+
\frac{6L^2r}{T^{\frac{7}{3}+2q}}
\ln\left(
1+\frac{\gamma^2T}
{[\max\{m_0^{\scriptscriptstyle X},
         m_0^{\scriptscriptstyle Y}\}]^2}
\right)
\notag\\
&\qquad+
\frac{6L^2r}{T^{2+2q}}
\ln\left(
1+\frac{\gamma^2T}
{[m_0^{\scriptscriptstyle Y}]^2}
\right)
\Bigg)^{\frac12}\Bigg)+\frac{2\sqrt{r}}{T^{1+q}}\left(\sqrt{\ln\!\left(\frac{\gamma^2T}{[\max\{m_0^{\scriptscriptstyle X},m_0^{\scriptscriptstyle Y}\}]^2}\right)+1} \right)\notag\\
&\quad\cdot\Bigg[
3rL^2T
+
\left(
\frac{3GL^2}{7\mu}
+\frac{12GL^3r}{7\mu^2}
\right)T^{\frac{2}{3}}
\notag\\
&\quad+
\left(
2G^2+2\kappa^2+3rL_\Phi^2+5rL^2
+\frac{4L^2\Delta_0^{\scriptscriptstyle Y}}{\mu}
+\frac{24GL^2\Delta_0^{\scriptscriptstyle Y}}
{7\mu^2}
+\frac{48rG^2L^4}{7\mu^4}
\right)T^{\frac{1}{3}}
\notag\\
&\quad+
\frac{2rL^2}{T^{\frac{1}{3}}}
+\frac{\kappa^2}{T^{\frac{2}{3}}}
+\frac{12GrL^2(L_\Phi+L)}{7\mu^2}
+\frac{48r^2GL^4}{7\mu^3}
+
\Bigg(
\frac{
3r(L_\Phi^2+L^2)
+\dfrac{48rG^2L^4}{7\mu^4}
}
{T^{\frac{2}{3}+2q}}
+
\frac{2rL^2}{T^{\frac{4}{3}+2q}}
\notag\\
&\quad+
\frac{
\dfrac{12GrL^2(L_\Phi+L)}{7\mu^2}
+\dfrac{48r^2GL^4}{7\mu^3}
}
{T^{1+2q}}
\Bigg)
\ln\left(
1+\frac{\gamma^2T}
{[\max\{m_0^{\scriptscriptstyle X},
         m_0^{\scriptscriptstyle Y}\}]^2}
\right)
\notag\\
&\quad+
\Bigg(
\frac{3rL^2}{T^{\frac{1}{3}+2q}}
+
\frac{2rL^2}{T^{1+2q}}
+
\frac{
\dfrac{3GL^2}{7\mu}
+\dfrac{12GL^3r}{7\mu^2}
}
{T^{\frac{2}{3}+2q}}
\Bigg)
\ln\left(
1+\frac{\gamma^2T}
{[m_0^{\scriptscriptstyle Y}]^2}
\right)
\notag\\
&\quad+
\frac{48G\sqrt r\,L^2T^{\frac{4}{3}}}{7\mu^2}
\sqrt{
\frac1{T^{\frac{2}{3}}}
+
\frac1{T^{2+2q}}
\ln\left(
1+\frac{\gamma^2T}
{[m_0^{\scriptscriptstyle Y}]^2}
\right)
}
\notag\\
&\quad\cdot
\Bigg(
\frac{3G^2+6L^2r+6\kappa^2}{T^{\frac{2}{3}}}
+
\frac{6L^2r}{T^{\frac{4}{3}}}
+
\frac{3\kappa^2}{T^{\frac{5}{3}}}+
\frac{6L^2r}{T^{\frac{7}{3}+2q}}
\ln\left(
1+\frac{\gamma^2T}
{[\max\{m_0^{\scriptscriptstyle X},
         m_0^{\scriptscriptstyle Y}\}]^2}
\right)
\notag\\
&\quad+
\frac{6L^2r}{T^{2+2q}}
\ln\left(
1+\frac{\gamma^2T}
{[m_0^{\scriptscriptstyle Y}]^2}
\right)
\Bigg)^{\frac12}\Bigg]^\frac{1}{2} +\frac{L_\Phi}{T^\frac{2}{3}} + \frac{L_\Phi}{T^{\frac{5}{3}+2q}}\left(\ln\left(\frac{\gamma^2 T}{[\max\{m_0^{\scriptscriptstyle X},m_0^{\scriptscriptstyle Y}\}]^2}\right) + 1\right) \notag\\
&\lesssim\frac{
\Delta+rL+\kappa\sqrt r
}{T^{\frac{1}{3}}}+
\frac{
{L\sqrt r}
G(1+\lambda)+L^2 r+\kappa L\sqrt r
}{\mu T^{\frac{1}{3}}}
+
\frac{
L_\Phi+\sqrt r\,G
}{T^{\frac{2}{3}}}+
\frac{
{L}
\Delta_0^{\scriptscriptstyle Y}+L^2
}{\mu T^{\frac{2}{3}}}
\notag\\
&\quad+\Bigg[
\frac{rL}{T^{\frac{1}{2}+q}}
+
\frac{
GL\left(
\mu+rL
+\sqrt r\sqrt{G^2+L^2r+\kappa^2}
\right)
}
{\mu^2T^{\frac{5}{6}+q}}
\Bigg]\sqrt{
\ln\left(
1+\frac{\gamma^2T}
{[\max\{m_0^{\scriptscriptstyle X},
         m_0^{\scriptscriptstyle Y}\}]^2}
\right)
}\notag\\
&\leq  \mathcal{O}\left(\frac{\Delta+rL+\kappa\sqrt r}{T^\frac{1}{3}} +\frac{GL\sqrt r(1+\lambda)+rL^2+\kappa L\sqrt{r}}{{\mu}T^\frac{1}{3}}+\frac{{r}L}{T^{\frac{1}{2}+q}}\sqrt{\ln\!\left(\frac{\gamma^2T}{z_0^2}\right)+1}\right),
\end{align}
where $\Delta:=\Phi(X_0) - \Phi_*$ and $z_0:=\max\{m_0^{\scriptscriptstyle X},m_0^{\scriptscriptstyle Y}\}$.

The proof for \cref{theoremVR} is thus completed.

\newpage
\section{Additional Experiments}\label{experimentsec}
\begin{figure*}[ht]
    \centering
    \begin{subfigure}[t]{0.43\textwidth}
        \centering
        \includegraphics[width=\linewidth]{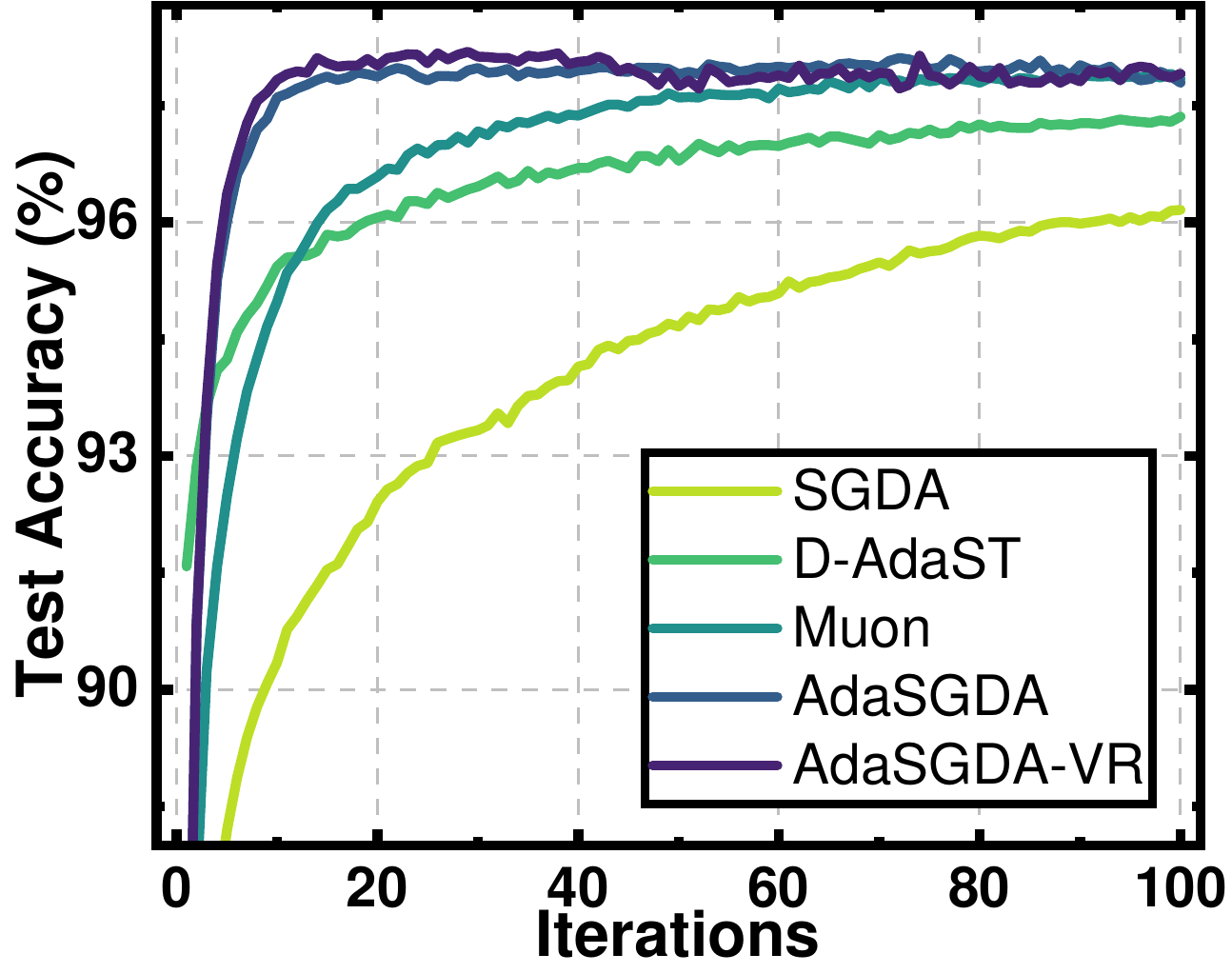}
        \caption{Test accuracy}
        \label{mnisttest}
    \end{subfigure}
    \begin{subfigure}[t]{0.43\textwidth}
        \centering
        \includegraphics[width=\linewidth]{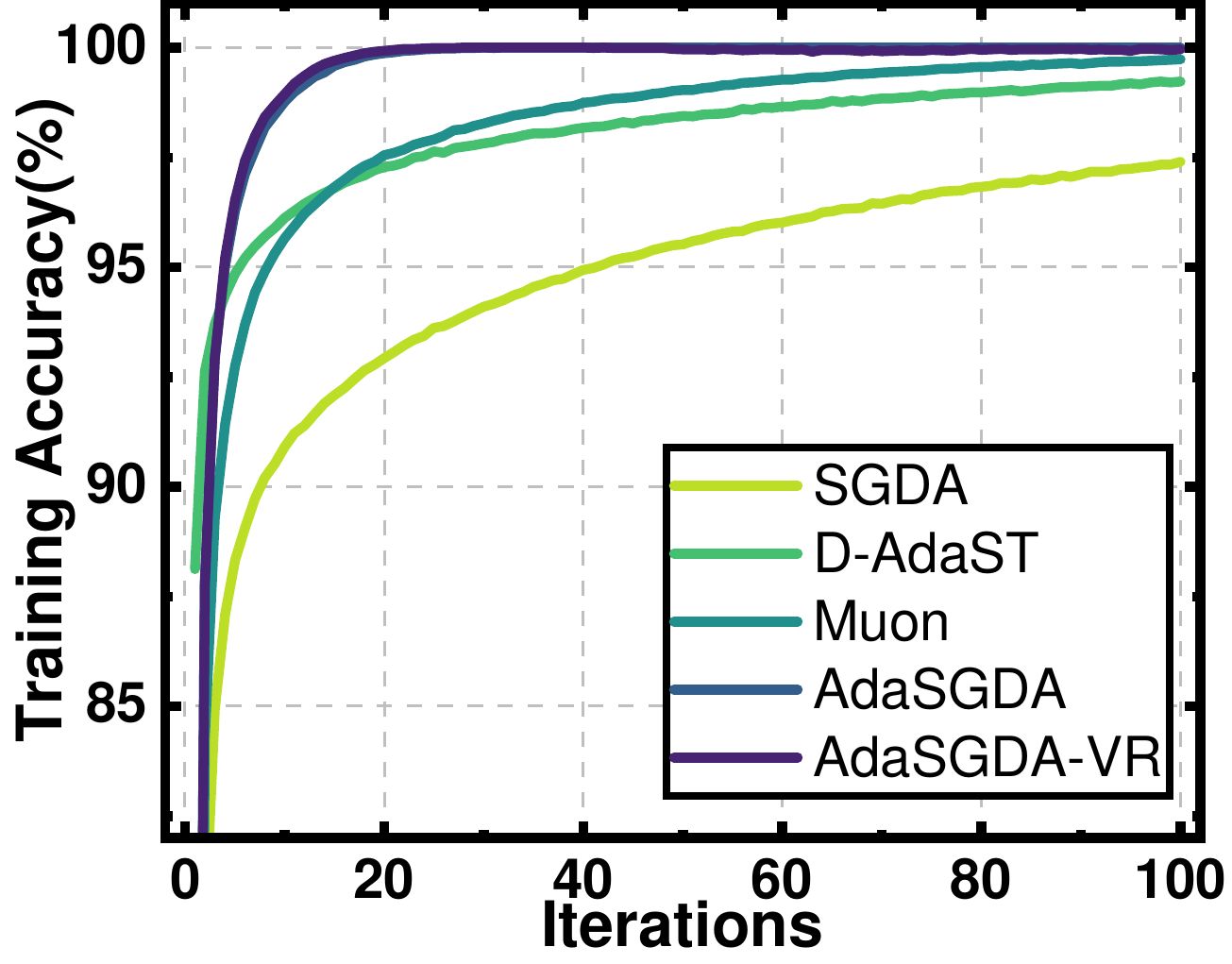}
        \caption{Training accuracy}
        \label{mnisttrain}
    \end{subfigure}
    \caption{Test and training accuracy on the MNIST dataset.}
    \label{mnisttt}
\end{figure*}

\begin{figure*}[ht]
    \centering
    \begin{subfigure}[t]{0.43\textwidth}
        \centering
        \includegraphics[width=\linewidth]{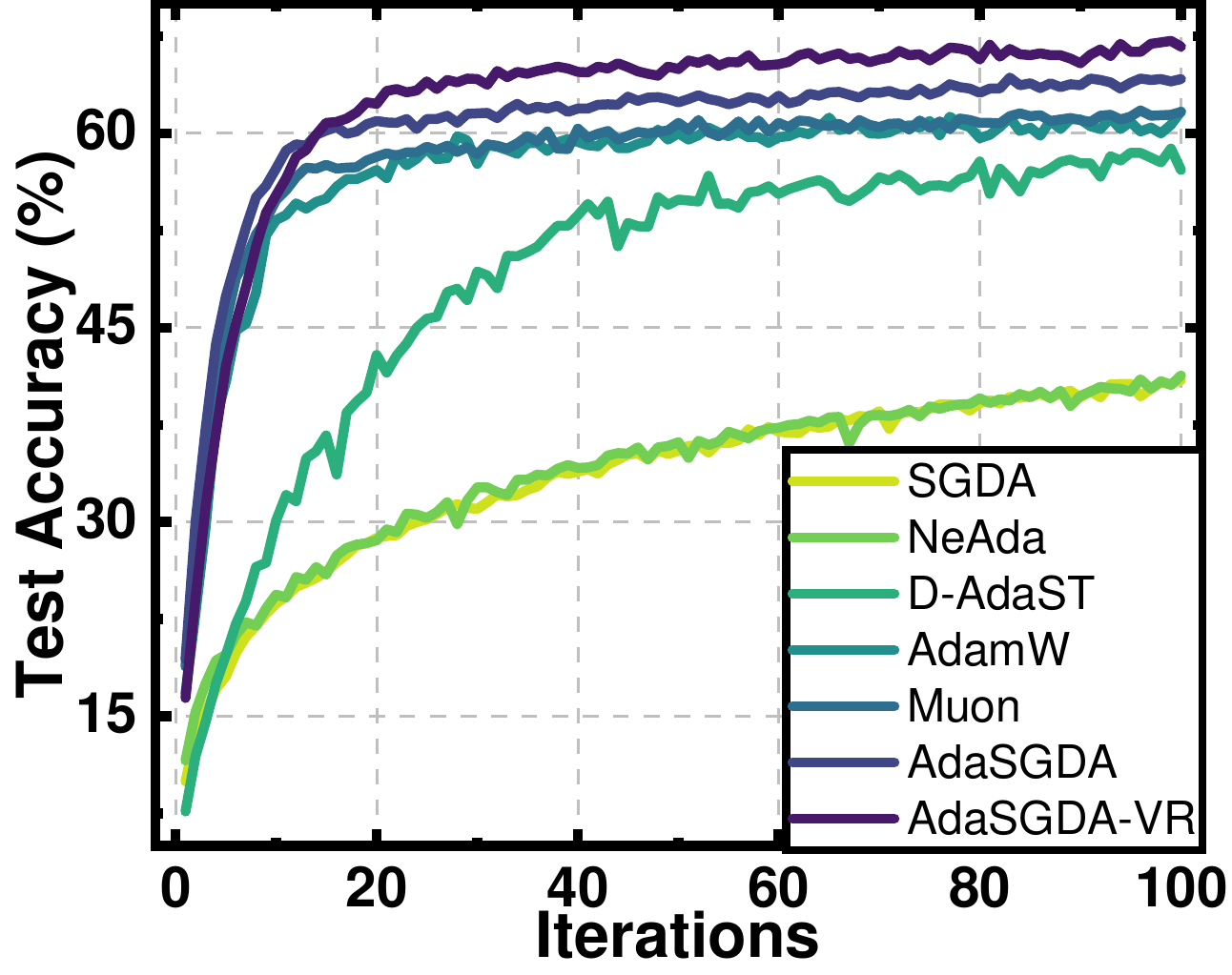}
        \caption{Test accuracy}
        \label{cifar100test}
    \end{subfigure}
    \begin{subfigure}[t]{0.43\textwidth}
        \centering
        \includegraphics[width=\linewidth]{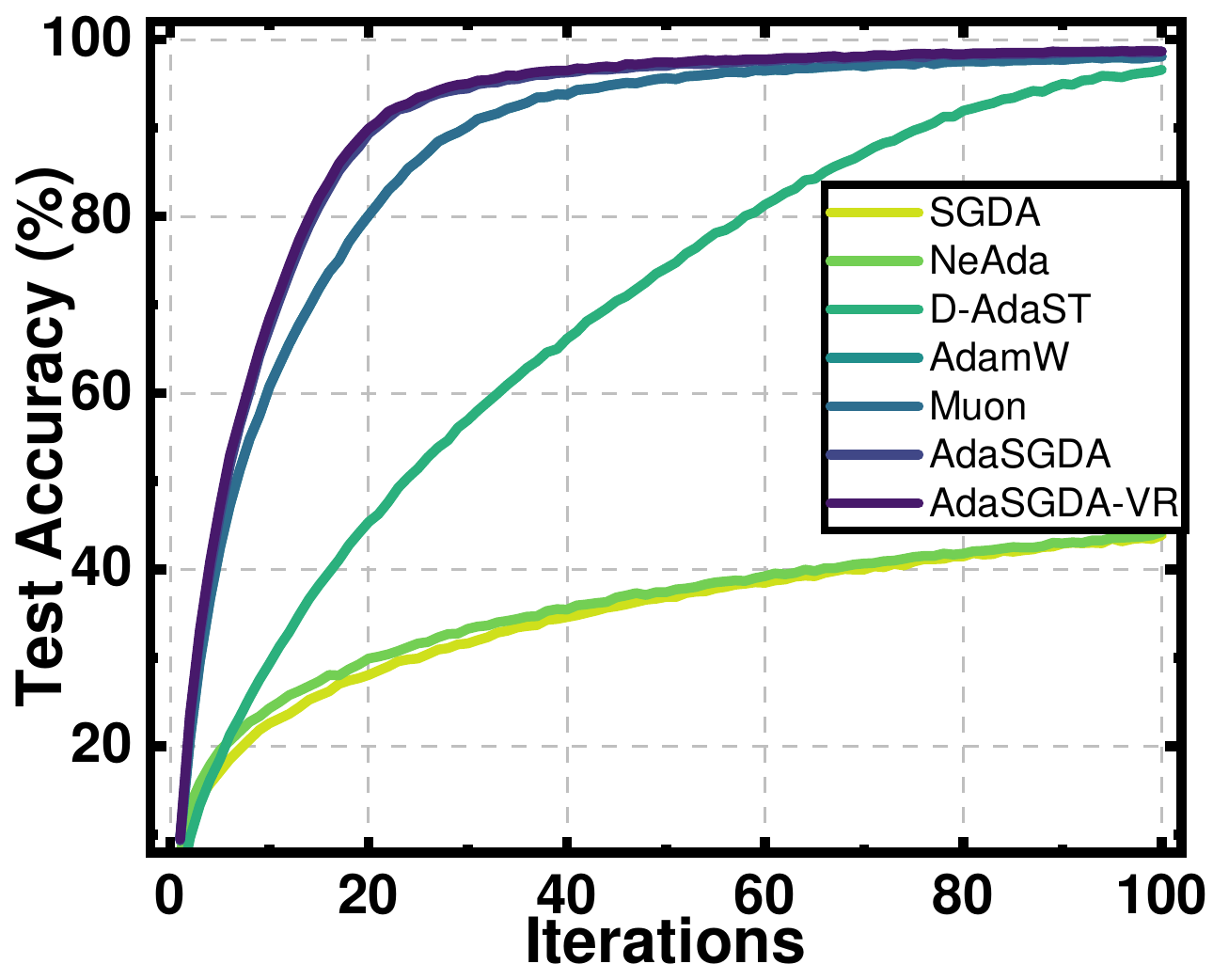}
        \caption{Training accuracy}
        \label{cifar100train}
    \end{subfigure}
    \caption{Test and training accuracy on the CIFAR-100 dataset.}
    \label{cifar100tt}
\end{figure*}
\subsection{Experimental Details}
This section summarizes the experimental setups and hyperparameter choices used throughout our experiments. For our methods, hyperparameter selection is guided by the theoretical analysis in \cref{theorem1,theoremVR}. Throughout all experiments, the additional exponent offset is fixed to $q=0$. We fix $\gamma$ and the initial accumulators $m_0^{\scriptscriptstyle X},m_0^{\scriptscriptstyle Y}$ to $0.9$ in all experiments and do not tune them. For both Muon and our methods, the number of Newton--Schulz (NS) iterations is set to $3$ by default.
For all other baselines, we tune hyperparameters to report their best performance. In particular, we first locate a reasonable stepsize scale by searching over $\{10^{-1},10^{-2},10^{-3},10^{-4}\}$, and then perform a finer grid search around the best scale. For Muon, we additionally tune the momentum coefficient over $\{0.1,0.3,0.5,0.7,0.9,0.95\}$ and report the best result.

\paragraph{Distributional Robustness Optimization.}
For the DRO experiments, we follow the same setup as \citet{sebbouh2022randomized} and \citet{sinha2017certifiable}, and fix the robustness parameter to $\rho=1.3$. We use different model architectures across datasets: an MLP (0.5M parameters) for MNIST \citep{lecun2002gradient} and FMNIST \citep{xiao2017fashion}, CifarNet (2M) for CIFAR-10 \citep{krizhevsky2009learning}, ResNet-34 (21.8M) for CIFAR-100 \citep{krizhevsky2009learning}, and three Transformer language models for FineWeb \citep{penedo2024fineweb}, namely OPT-125M, OPT-350M, and OPT-1.3B.

\paragraph{Generative Adversarial Network.}
For the GAN experiments, we use an implementation adapted from \citet{green9}. The discriminator is a four-layer convolutional neural network (CNN) and the generator is a four-layer transposed-convolution CNN. Following the experimental protocol in \citet{daskalakis2017training}, we set the latent dimension to $50$ and use a gradient-penalty weight of $10^{-4}$. Inception Score is computed by feeding $8{,}000$ generated samples into a pretrained Inception network.

\begin{figure*}[t]
    \centering
    \begin{subfigure}[t]{0.32\textwidth}
        \centering
        \includegraphics[width=\linewidth]{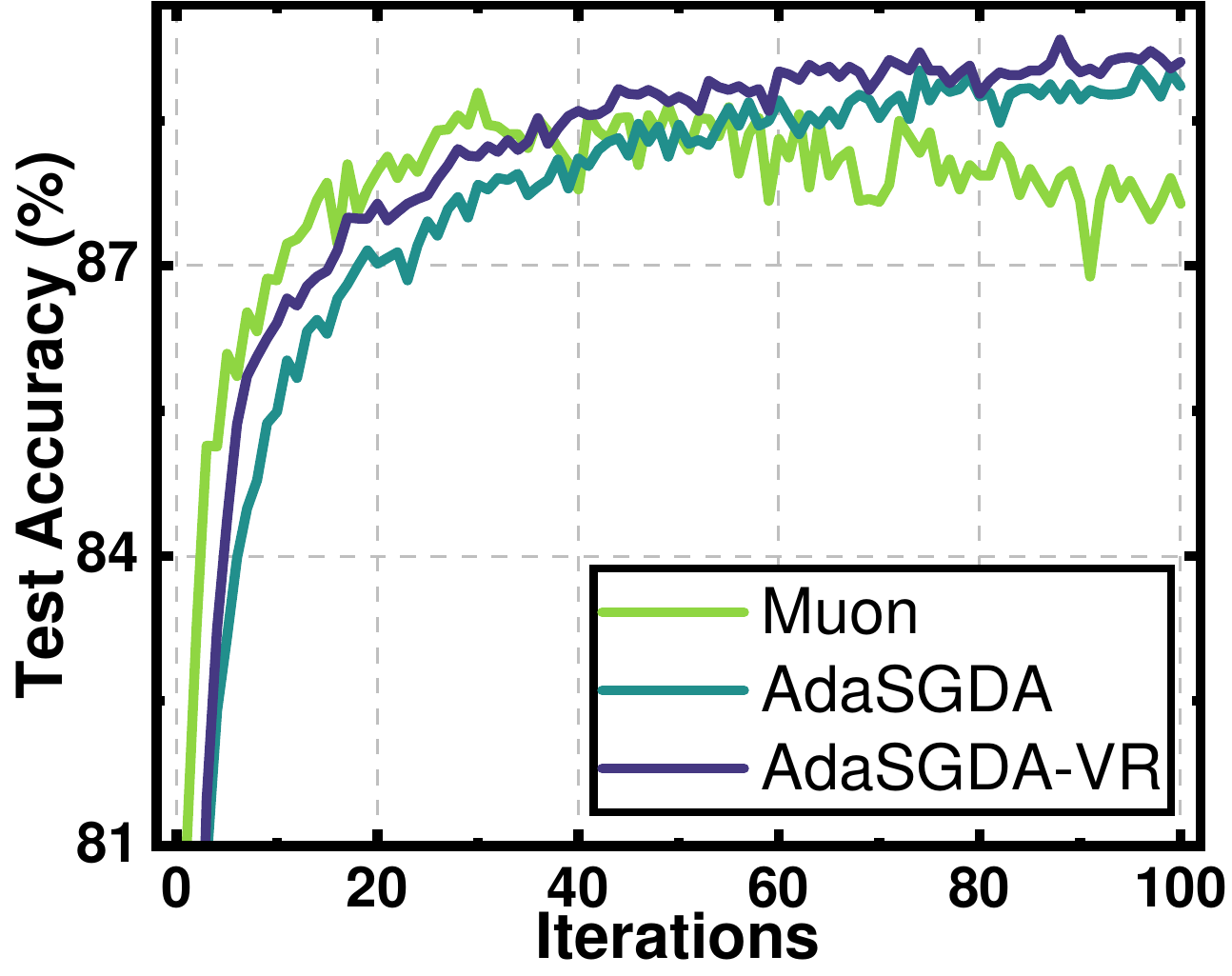}
        \caption{\texttt{Newton–Schulz} steps: 1}
        \label{fmnistmuonstep1}
    \end{subfigure}\hfill
    \begin{subfigure}[t]{0.32\textwidth}
        \centering
        \includegraphics[width=\linewidth]{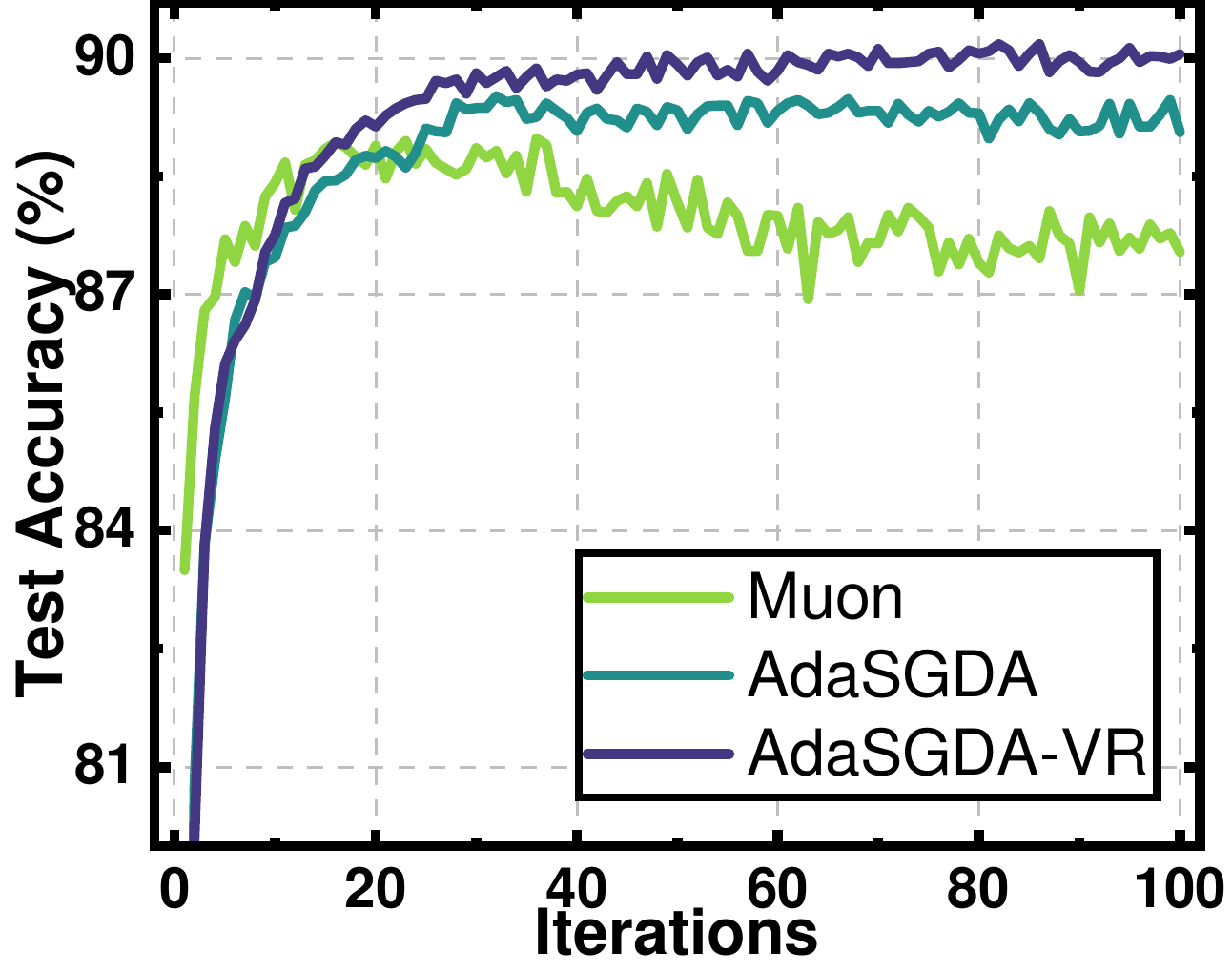}
        \caption{\texttt{Newton–Schulz} steps: 2}
        \label{fmnistmuonstep2}
    \end{subfigure}\hfill
    \begin{subfigure}[t]{0.32\textwidth}
        \centering
        \includegraphics[width=\linewidth]{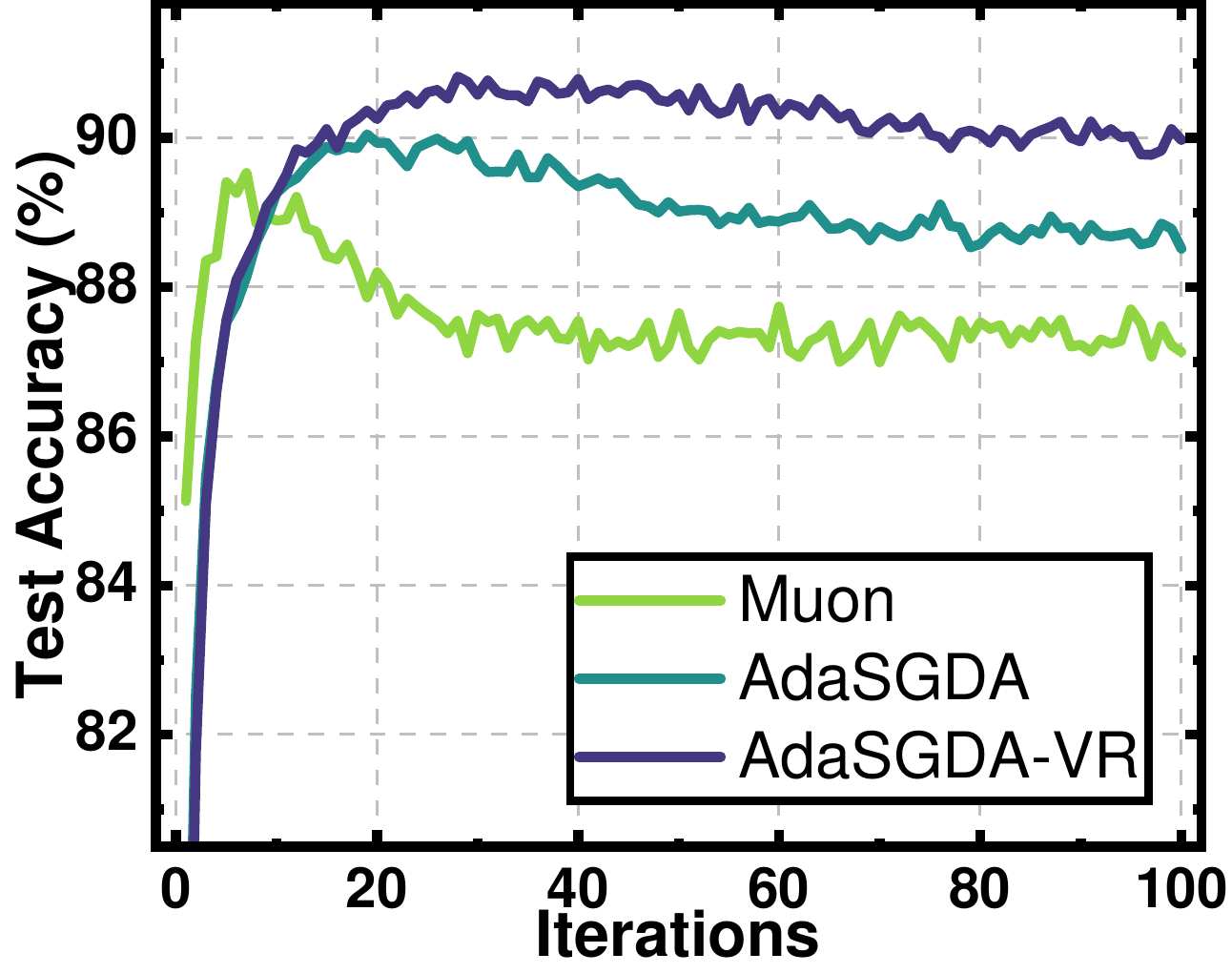}
        \caption{\texttt{Newton–Schulz} steps: 5}
        \label{fmnistmuonstep5}
    \end{subfigure}
    \caption{Test accuracy on the FMNIST dataset with different \texttt{Newton–Schulz} steps.}
    \label{fmnistmuonstep}
\end{figure*}

\begin{figure*}[t]
    \centering
    \begin{subfigure}[t]{0.32\textwidth}
        \centering
        \includegraphics[width=\linewidth]{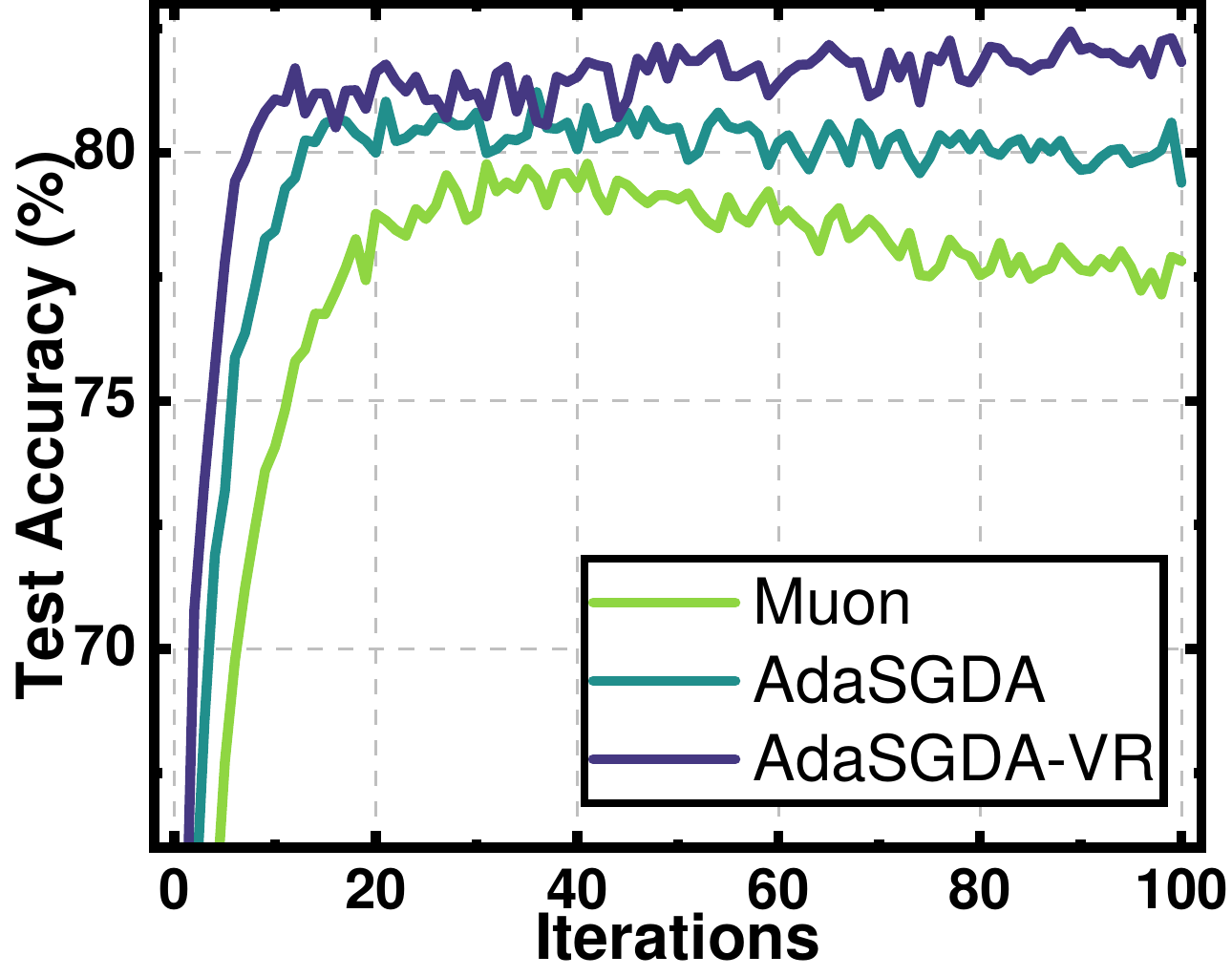}
        \caption{\texttt{Newton–Schulz} steps: 1}
        \label{cifar10muonstep1}
    \end{subfigure}\hfill
    \begin{subfigure}[t]{0.32\textwidth}
        \centering
        \includegraphics[width=\linewidth]{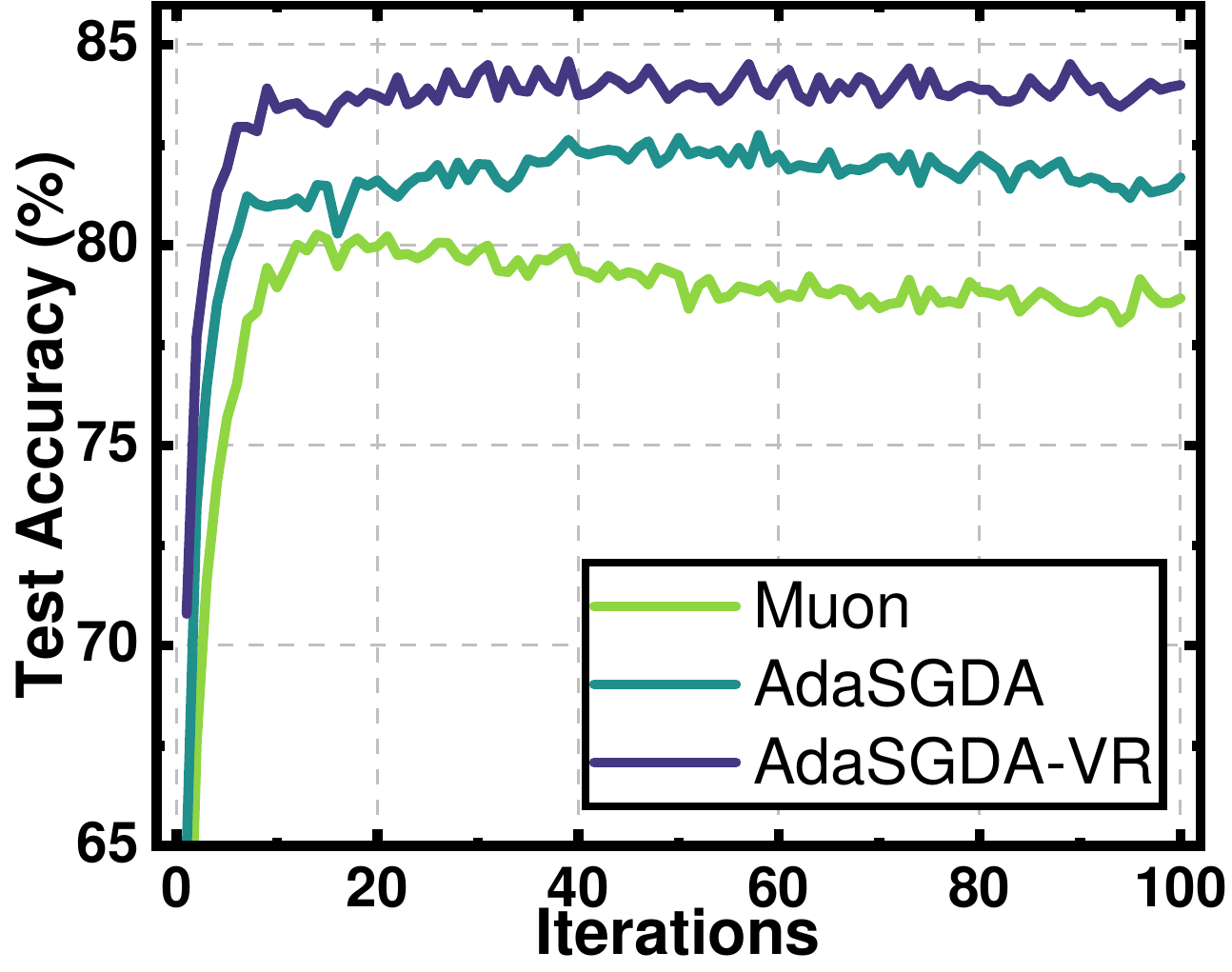}
        \caption{\texttt{Newton–Schulz} steps: 2}
        \label{cifar10muonstep2}
    \end{subfigure}\hfill
    \begin{subfigure}[t]{0.32\textwidth}
        \centering
        \includegraphics[width=\linewidth]{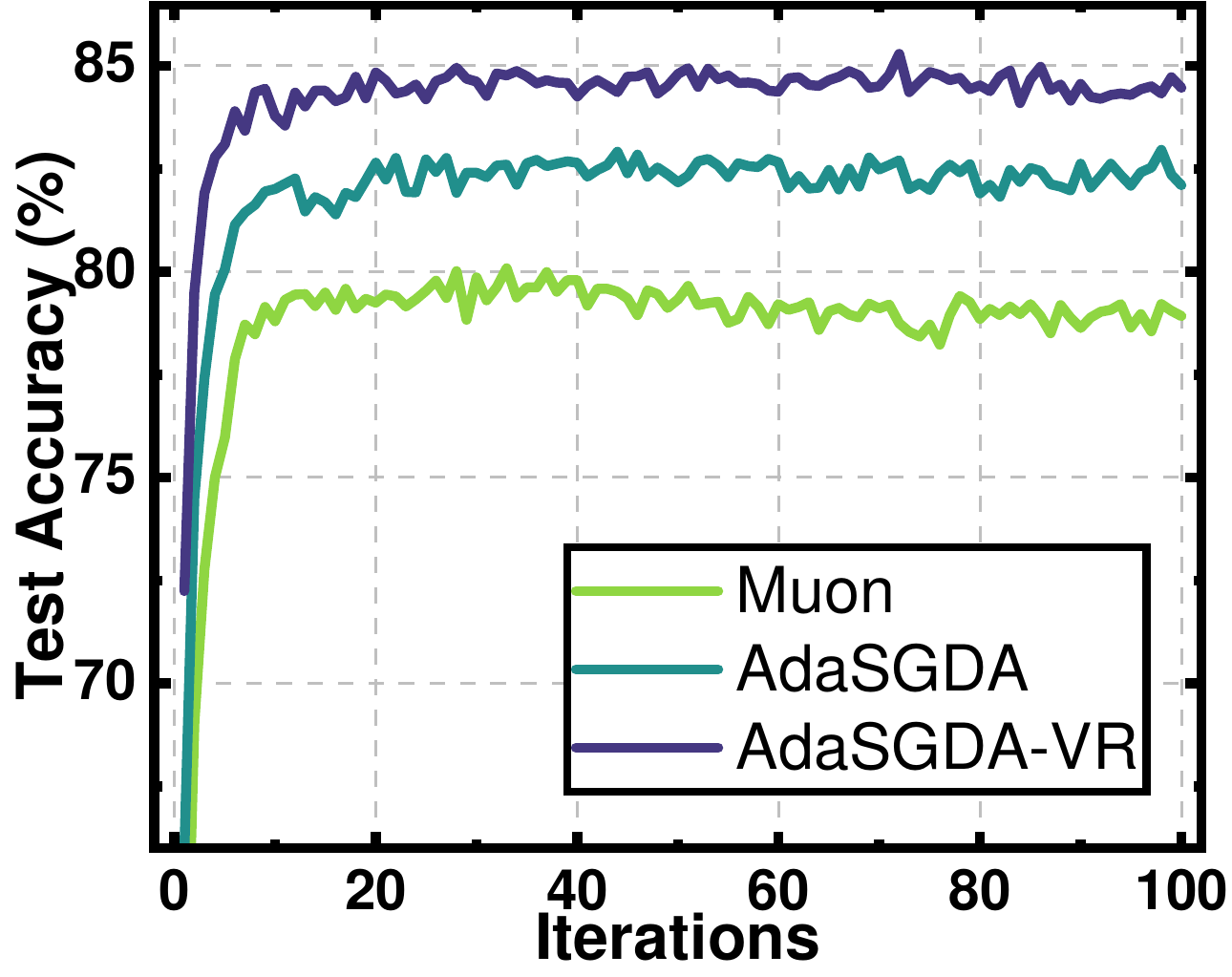}
        \caption{\texttt{Newton–Schulz} steps: 5}
        \label{cifar10muonstep5}
    \end{subfigure}
    \caption{Test accuracy on the CIFAR-10 dataset with different \texttt{Newton–Schulz} steps.}
    \label{cifar10muonstep}
\end{figure*}

\subsection{Additional Results on MNIST and CIFAR-100 Datasets}
As shown in \cref{mnisttt,cifar100tt}, which report both training and test accuracies on the MNIST and CIFAR-100 datasets, respectively, \textbf{AdaSGDA} and \textbf{AdaSGDA-VR} significantly outperform SGDA and D-AdaST, which rely on conventional gradient directions. Compared with the orthogonal-update baseline Muon, our proposed methods also achieve stronger performance. This improvement stems from our parameter-free design: stepsizes are automatically adjusted based on accumulated gradient norms, calibrating the update magnitude throughout training and avoiding brittle manually tuned schedules.
Notably, the advantage of our methods becomes more pronounced as the datasets and models grow more challenging (e.g., CIFAR-100 with a deeper network). In these settings, orthogonal updates can provide stability but also make the effective update scale more sensitive to the evolving geometry; our adaptive stepsize rule compensates for this effect and maintains well-balanced progress throughout training. Finally, we emphasize that all baselines require substantial manual tuning to reach their best performance, whereas our methods can be run with theory-guided hyperparameters, greatly simplifying deployment in practice.

\begin{figure*}[t]
    \centering
    \begin{subfigure}[t]{0.32\textwidth}
        \centering
        \includegraphics[width=\linewidth]{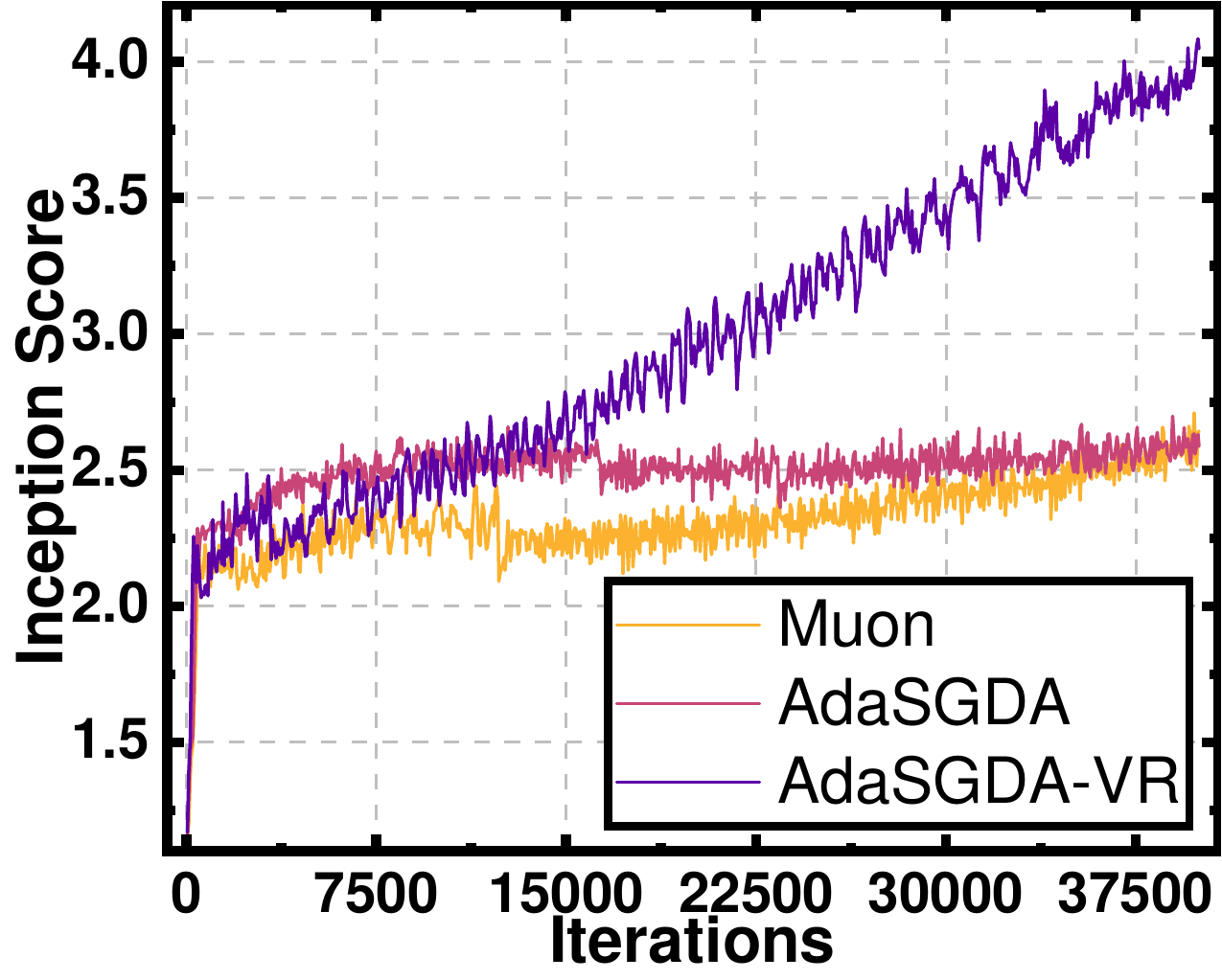}
        \caption{\texttt{Newton–Schulz} steps: 1}
        \label{ganmuonstep1}
    \end{subfigure}\hfill
    \begin{subfigure}[t]{0.32\textwidth}
        \centering
        \includegraphics[width=\linewidth]{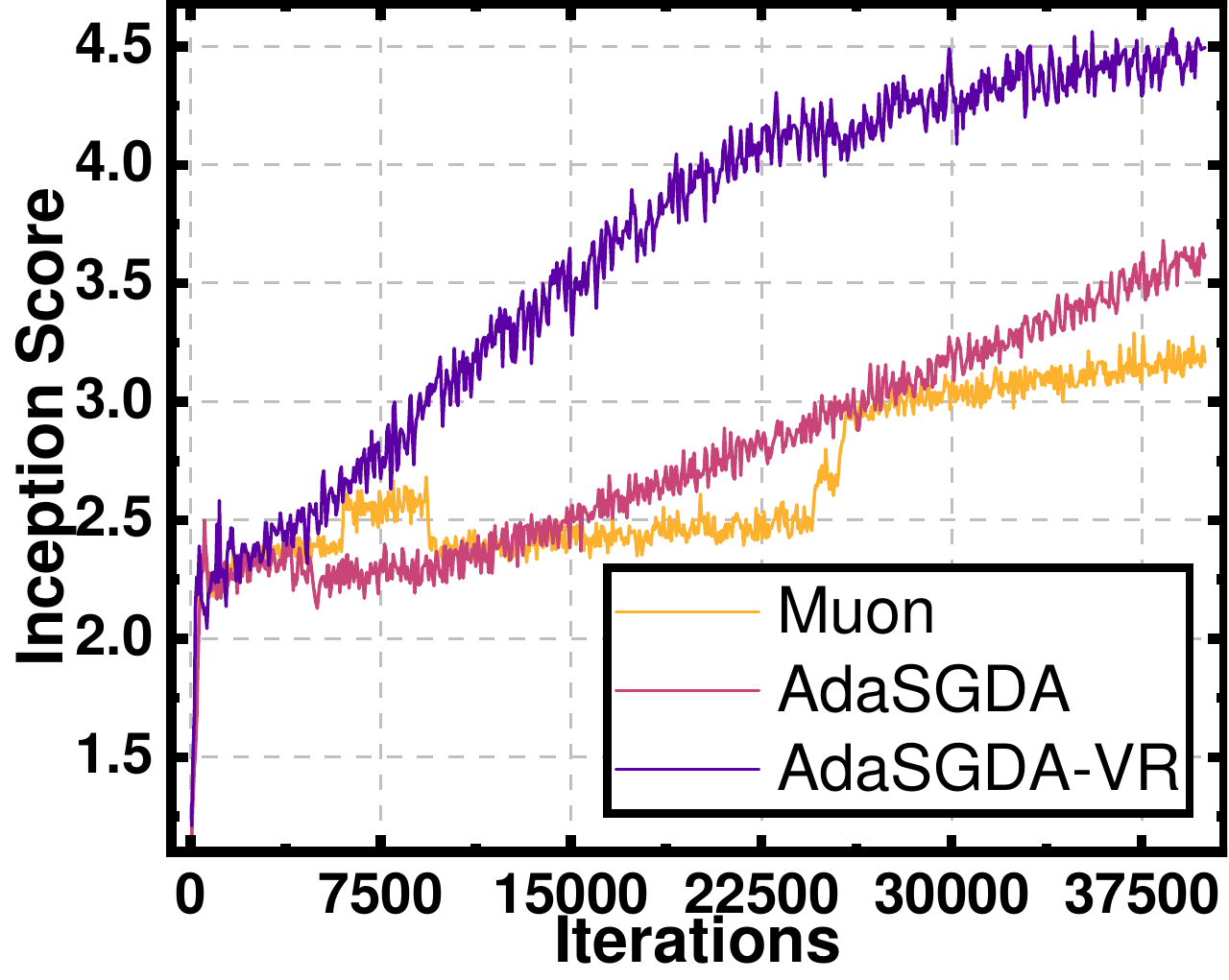}
        \caption{\texttt{Newton–Schulz} steps: 2}
        \label{ganmuonstep2}
    \end{subfigure}\hfill
    \begin{subfigure}[t]{0.32\textwidth}
        \centering
        \includegraphics[width=\linewidth]{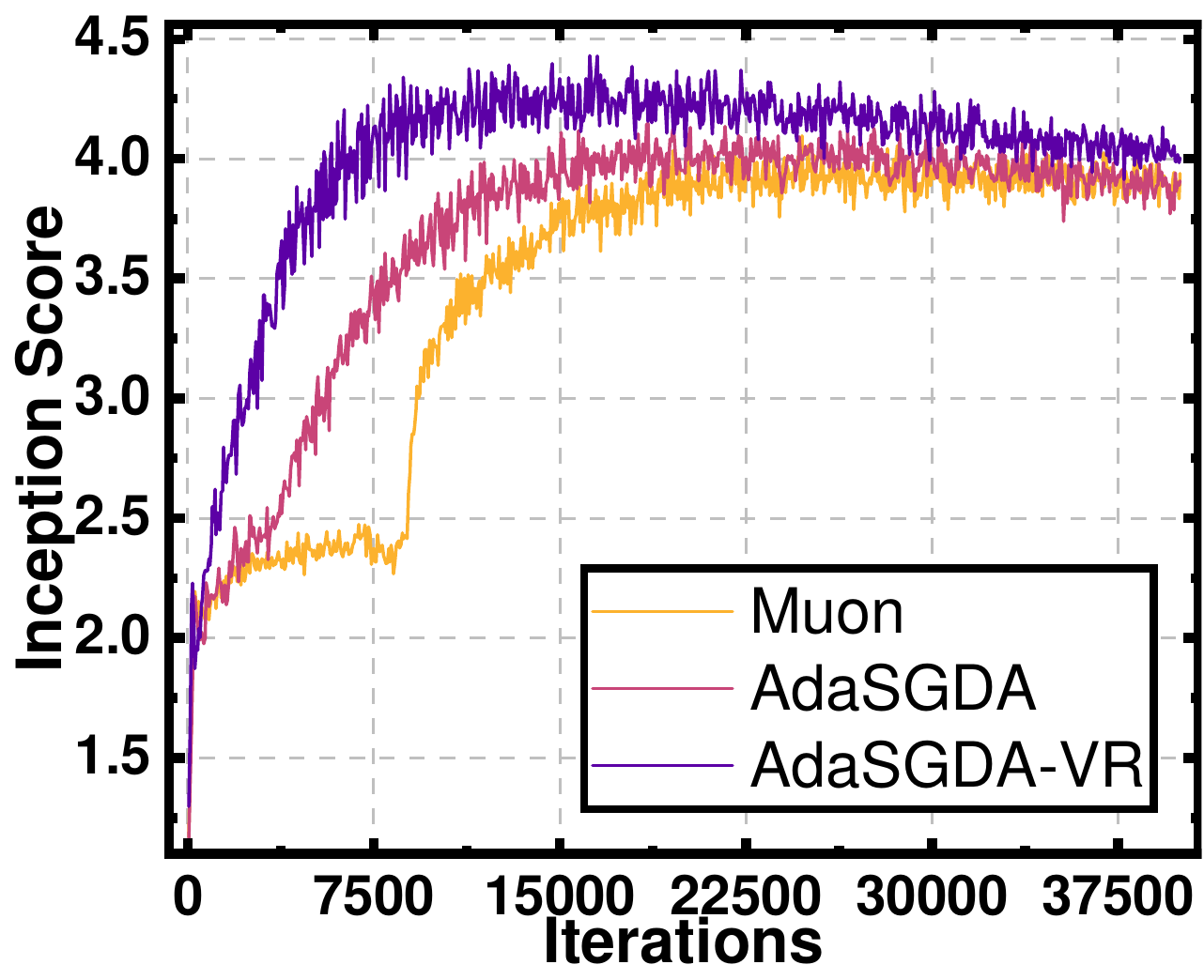}
        \caption{\texttt{Newton–Schulz} steps: 5}
        \label{ganmuonstep5}
    \end{subfigure}
    \caption{Inception Score on WGAN-GP with different \texttt{Newton–Schulz} steps.}
    \label{ganmuonstep}
\end{figure*}

\begin{figure}[t]
    \centering
    \begin{subfigure}[b]{0.43\textwidth}
    \includegraphics[width=\linewidth]{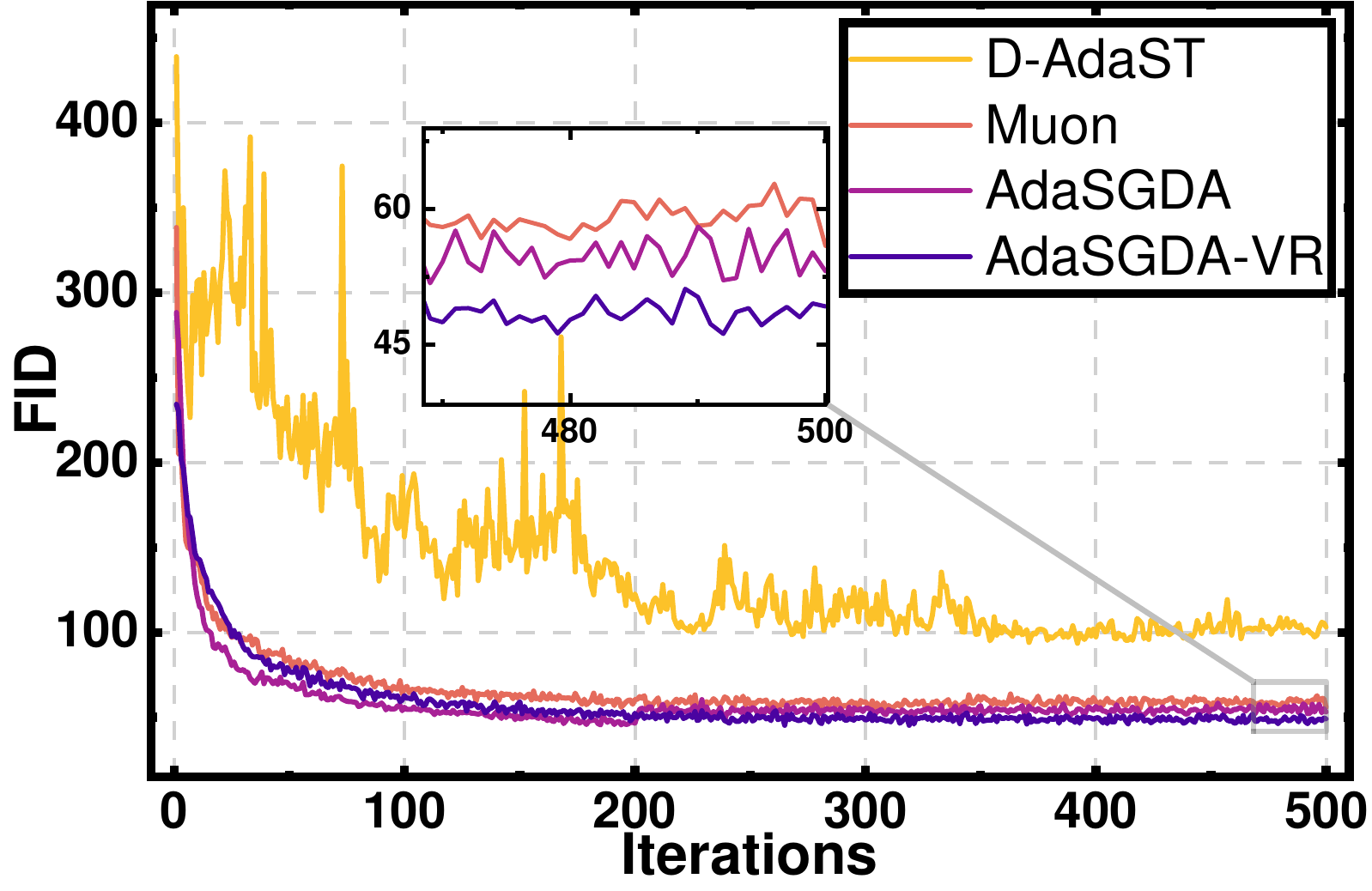}
    \caption{CIFAR-10 dataset}
  \end{subfigure}
  \hspace{0.04\textwidth}
  \begin{subfigure}[b]{0.43\textwidth}
    \includegraphics[width=\linewidth]{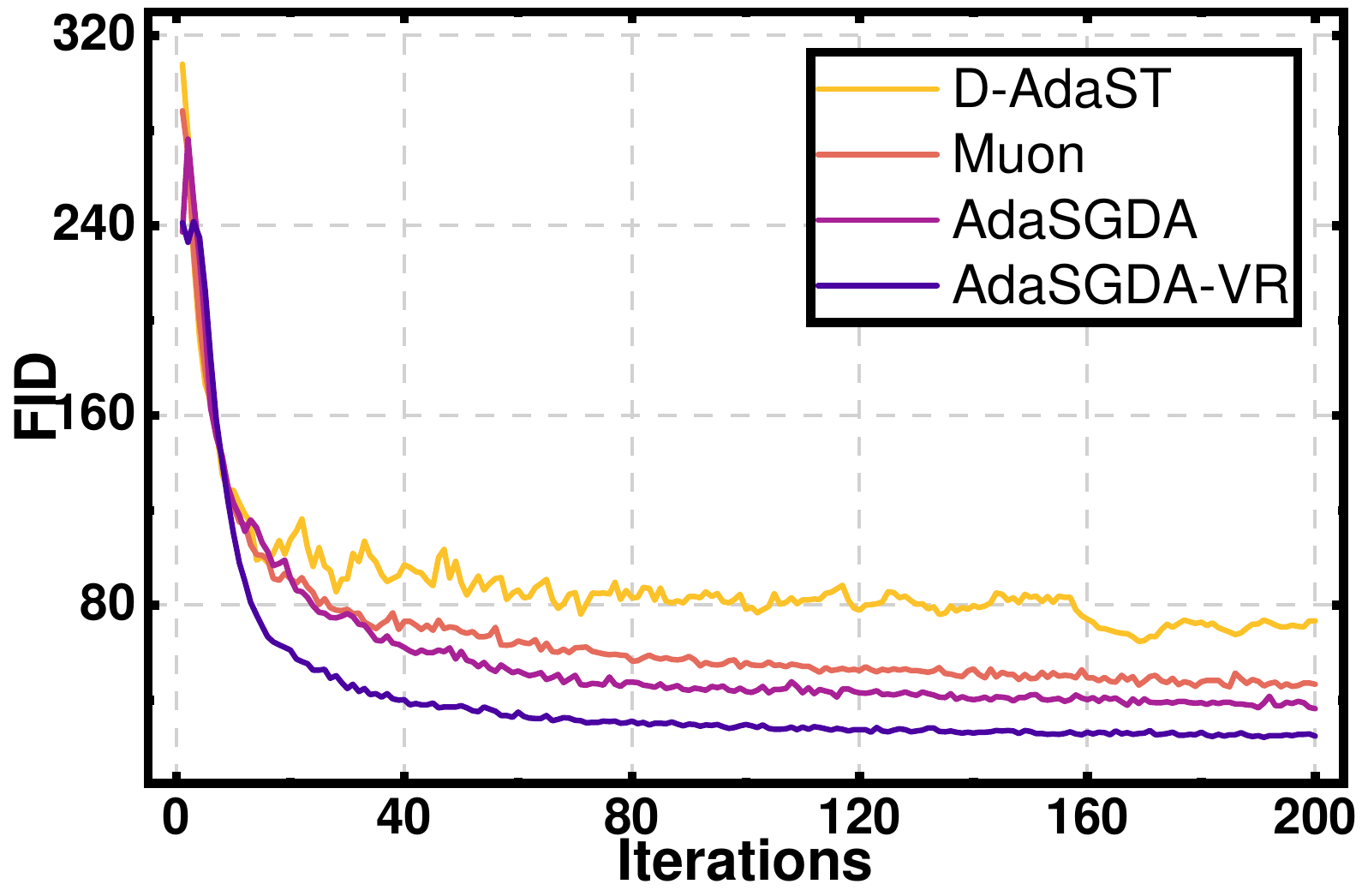}
    \caption{CIFAR-100 dataset}
  \end{subfigure}
\caption{FID on WGAN-GP with CIFAR-10 and CIFAR-100 datasets.}
\label{fidcifar}
\end{figure}

\subsection{Sensitivity Analysis to Newton--Schulz Steps}
Beyond the default setting of $3$ NS steps, we vary the number of NS steps in $\{1,2,5\}$ and evaluate on the FMNIST and CIFAR-10 datasets, as shown in \cref{fmnistmuonstep,cifar10muonstep}, respectively. The NS iterations determine the orthogonalization accuracy, i.e., how well the update direction approximates the orthogonal (polar) factor; increasing NS steps typically yields a more accurate orthogonal direction, which can improve stability and training effectiveness.
As shown in the results, our methods consistently outperform Muon under all NS choices. Moreover, increasing the number of NS steps generally improves performance for all methods, as orthogonalization becomes more accurate. Notably, our methods exhibit robust behavior across different NS steps and achieve consistently strong results, whereas Muon shows performance fluctuations when the NS steps are not well chosen. This further demonstrates the effectiveness of our adaptive design under orthogonal updates.

\begin{table}[t]
\centering
\setlength{\tabcolsep}{6pt}
\renewcommand{\arraystretch}{1.12}
\caption{Wall-clock time inception score of WGAN-GP for different algorithms.}
\label{tab:wallclockgan}
\begin{tabular}{lccccccc}
\specialrule{1.9pt}{0pt}{0pt}
 & \multicolumn{6}{c}{\textbf{Wall-clock time inception score}} \\
\cmidrule(lr){2-8}
\textbf{Algorithm} & \textbf{600 sec} & \textbf{1200 sec} & \textbf{1800 sec} & \textbf{2400 sec} & \textbf{3000 sec} & \textbf{3600 sec}& \textbf{4200 sec} \\
\midrule
SGDA           & 2.2465 & 2.3326 & 2.3479 & 2.3512 & 2.3546 & 2.3787 & 2.4095 \\
Muon   & 2.0232 & 2.2709 & 2.3299 & 2.3385 & 2.3552 & 2.3711 & 2.3752 \\
AdaSGDA  & 2.2542  & 2.3227 & 2.3751 & 2.4293 & 2.4438 & 2.4487 & 2.4758 \\
AdaSGDA-VR & 2.3062  & 2.3452 & 2.4475 & 2.4748 & 2.4832 & 2.5507& 2.6897
 \\
\specialrule{1.9pt}{0pt}{0pt}
\end{tabular}
\end{table}

\begin{table}[!t]
\centering
\setlength{\tabcolsep}{6pt}
\renewcommand{\arraystretch}{1.12}
\caption{Wall-clock time FID of WGAN-GP on CIFAR-10 dataset for different algorithms.}
\label{wallclockfidcifar10}
\begin{tabular}{lccccccc}
\specialrule{1.9pt}{0pt}{0pt}
 & \multicolumn{6}{c}{\textbf{Wall-clock time FID}} \\
\cmidrule(lr){2-8}
\textbf{Algorithm} & \textbf{600 sec} & \textbf{1200 sec} & \textbf{1800 sec} & \textbf{2400 sec} & \textbf{3000 sec} & \textbf{3600 sec}& \textbf{4200 sec} \\
\midrule
D-AdaST   & 271.75 & 242.48 & 226.89 & 199.81 & 171.55 & 160.87 & 134.67 \\
Muon   & 167.84 & 128.28 & 107.90 & 97.04 & 92.60 & 84.79 & 79.24 \\
AdaSGDA  & 164.24  & 127.06 & 105.06 & 96.09 & 89.58 & 83.41 & 75.94 \\
AdaSGDA-VR & 131.39  & 106.77 & 92.88 & 86.52 & 79.56 & 74.00& 66.86
 \\
\specialrule{1.9pt}{0pt}{0pt}
\end{tabular}
\end{table}

\begin{table}[!t]
\centering
\setlength{\tabcolsep}{6pt}
\renewcommand{\arraystretch}{1.12}
\caption{Wall-clock time FID of WGAN-GP on CIFAR-100 dataset for different algorithms.}
\label{wallclockfidcifar100}
\begin{tabular}{lccccccc}
\specialrule{1.9pt}{0pt}{0pt}
 & \multicolumn{6}{c}{\textbf{Wall-clock time FID}} \\
\cmidrule(lr){2-8}
\textbf{Algorithm} & \textbf{600 sec} & \textbf{1200 sec} & \textbf{1800 sec} & \textbf{2400 sec} & \textbf{3000 sec} & \textbf{3600 sec}& \textbf{4200 sec} \\
\midrule
D-AdaST   & 237.42 & 167.50 & 147.39 & 124.18 & 111.55 & 98.31 & 89.68 \\
Muon   & 198.96  & 129.08 & 114.99 & 100.96 & 91.21 & 82.71 & 76.52 \\
AdaSGDA  & 163.15  & 122.94 & 106.38 & 96.62 & 86.25 & 76.97 & 71.60 \\
AdaSGDA-VR & 158.19  & 115.35 & 97.81 & 89.03 & 81.25 & 71.14& 66.70
 \\
\specialrule{1.9pt}{0pt}{0pt}
\end{tabular}
\end{table}

\begin{figure*}[t]
    \centering
    \begin{subfigure}[t]{0.43\textwidth}
        \centering
        \includegraphics[width=\linewidth]{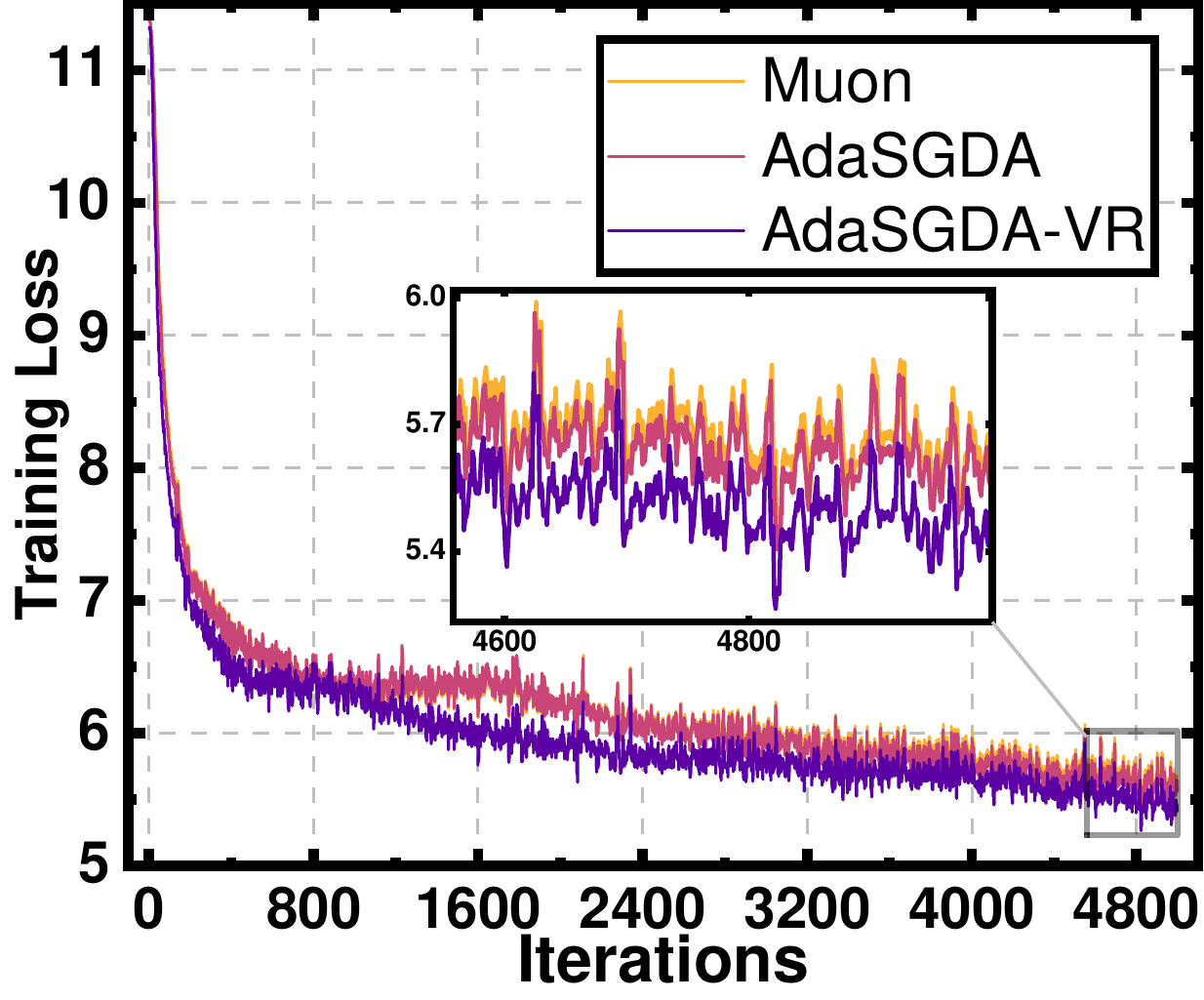}
        \caption{Training loss}
        \label{opt350train}
    \end{subfigure}
    \begin{subfigure}[t]{0.43\textwidth}
        \centering
        \includegraphics[width=\linewidth]{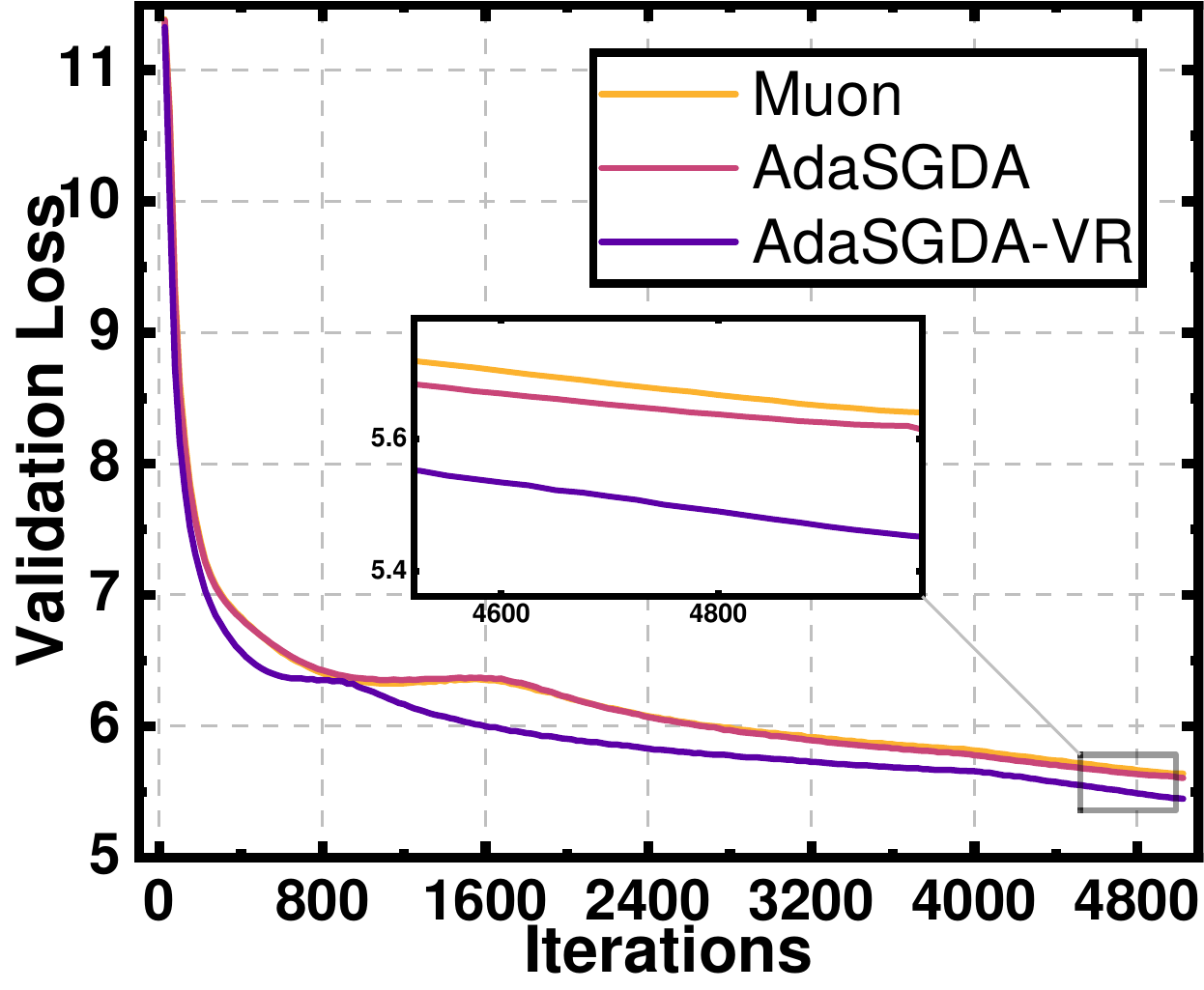}
        \caption{Validation loss}
        \label{opt350val}
    \end{subfigure}
    \caption{Training and validation loss curves for OPT-350M on the FineWeb dataset.}
    \label{opt350tt}
\end{figure*}

\begin{figure*}[ht]
    \centering
    \begin{subfigure}[t]{0.43\textwidth}
        \centering
        \includegraphics[width=\linewidth]{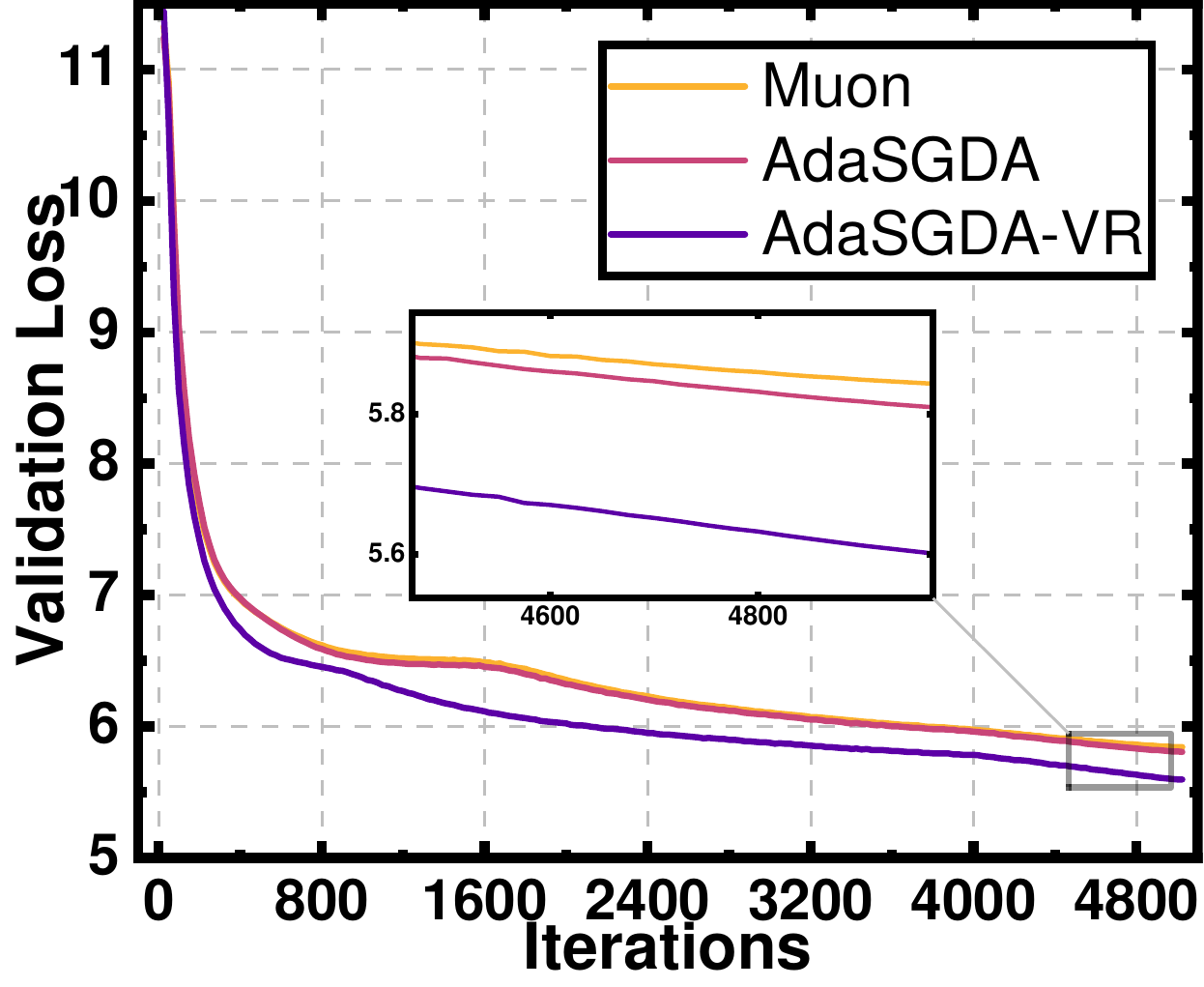}
        \caption{Validation loss (OPT-125M)}
        \label{opt125val}
    \end{subfigure}
    \begin{subfigure}[t]{0.43\textwidth}
        \centering
        \includegraphics[width=\linewidth]{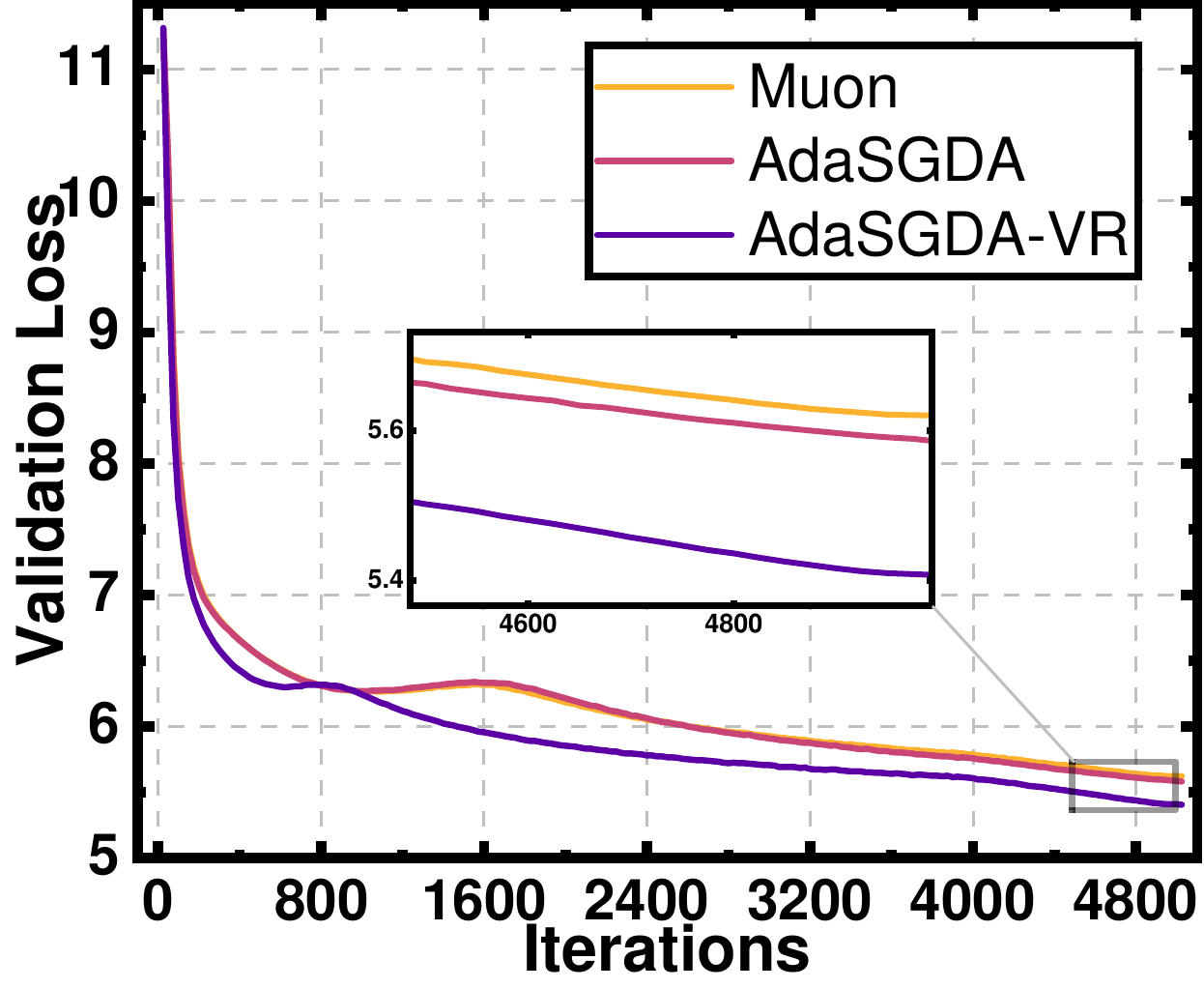}
        \caption{Validation loss (OPT-1.3B)}
        \label{opt12bval}
    \end{subfigure}
    \caption{Validation loss curves for OPT-125M and OPT-1.3B on the FineWeb dataset.}
    \label{opt12512bOPtt}
\end{figure*}

\subsection{Additional Results on WGAN-GP}
In \cref{ganmuonstep}, we report the Inception Score under different numbers of NS steps. When NS steps are small (e.g., $1$ or $2$), the orthogonalization is less accurate, which can distort the update direction and amplify training instability. In this regime, AdaSGDA-VR shows a clear advantage over Muon, achieving higher inception scores with faster improvements over training. More broadly, AdaSGDA and AdaSGDA-VR consistently outperform Muon across all NS choices, while Muon exhibits visible performance fluctuations when NS steps are not well chosen. These results highlight the robustness of our adaptive orthogonal-update design, which remains stable even when the orthogonalization accuracy varies.
We further report wall-clock time performance in \cref{tab:wallclockgan}. Our methods achieve better time-to-quality than all baselines: AdaSGDA-VR attains the highest inception score at every time budget, and AdaSGDA also consistently improves over SGDA and Muon. This demonstrates that our gains are not only in iteration complexity but also translate into practical efficiency in terms of wall-clock time.

Following \citep{shmelkov2018good}, we additionally evaluate WGAN-GP using the Fr\'echet Inception Distance (FID), which compares the feature statistics of generated and real images and is a standard metric for GAN evaluation. We report the iteration-wise FID curves on CIFAR-10 and CIFAR-100 in \cref{fidcifar}. On CIFAR-10, both AdaSGDA and AdaSGDA-VR consistently achieve lower FID than Muon and D-AdaST throughout training, while also exhibiting noticeably more stable trajectories; in particular, D-AdaST shows substantial fluctuations, whereas our methods remain much more stable over the entire course of training. On the more challenging CIFAR-100 task, the same trend becomes even more pronounced: AdaSGDA and especially AdaSGDA-VR maintain a clearer advantage over the baselines, indicating that our adaptive orthogonal methods produce samples that are closer to the real data distribution and that the benefit of adaptivity becomes stronger in harder generative settings.
We further report wall-clock FID in \cref{wallclockfidcifar10,wallclockfidcifar100}. The same ranking is preserved under equal time budgets on both datasets: AdaSGDA-VR consistently achieves the best time-to-quality tradeoff, and AdaSGDA also remains uniformly better than Muon and D-AdaST. Therefore, the improvement of our methods is not only reflected by iteration-based comparisons, but also by stronger practical efficiency and better generative quality in real running time.

\begin{table}[p]
\centering
\begin{minipage}{\linewidth}
\centering
\setlength{\tabcolsep}{6pt}
\renewcommand{\arraystretch}{1.12}
\vspace{-0.7cm}
\captionof{table}{Wall-clock training loss of OPT-125M.}
\label{tab:wallclock125}
\begin{tabular}{lccccccc}
\specialrule{1.9pt}{0pt}{0pt}
 & \multicolumn{7}{c}{\textbf{Wall-clock training loss}} \\
\cmidrule(lr){2-8}
\textbf{Algorithm} & \textbf{600s} & \textbf{1200s} & \textbf{1800s} & \textbf{2400s} & \textbf{3000s} & \textbf{3600s} & \textbf{4200s} \\
\midrule
Muon        & 6.6598 & 6.4266 & 6.3381 & 6.1984 & 6.0494 & 5.9709 & 5.8419 \\
AdaSGDA     & 6.5628 & 6.3523 & 6.2203 & 6.0404 & 5.9412 & 5.8038 & 5.6744 \\
AdaSGDA-VR  & 6.4094 & 6.2918 & 6.0617 & 5.9695 & 5.8521 & 5.7178 & 5.4577 \\
\specialrule{1.9pt}{0pt}{0pt}
\end{tabular}
\end{minipage}

\vspace{0.41cm}

\begin{minipage}{\linewidth}
\centering
\setlength{\tabcolsep}{6pt}
\renewcommand{\arraystretch}{1.12}
\captionof{table}{Wall-clock training loss of OPT-350M.}
\label{tab:wallclock350}
\begin{tabular}{lccccccc}
\specialrule{1.9pt}{0pt}{0pt}
 & \multicolumn{7}{c}{\textbf{Wall-clock training loss}} \\
\cmidrule(lr){2-8}
\textbf{Algorithm} & \textbf{600s} & \textbf{1200s} & \textbf{1800s} & \textbf{2400s} & \textbf{3000s} & \textbf{3600s} & \textbf{4200s} \\
\midrule
Muon        & 7.0506 & 6.7895 & 6.4186 & 6.3847 & 6.2401 & 6.1844 & 6.1175 \\
AdaSGDA     & 6.9492 & 6.5542 & 6.3820 & 6.2238 & 6.1351 & 6.0878 & 5.9382 \\
AdaSGDA-VR  & 6.7569 & 6.5519 & 6.3191 & 6.2176 & 6.0907 & 5.9691 & 5.8582 \\
\specialrule{1.9pt}{0pt}{0pt}
\end{tabular}
\end{minipage}

\vspace{0.41cm}

\begin{minipage}{\linewidth}
\centering
\setlength{\tabcolsep}{6pt}
\renewcommand{\arraystretch}{1.12}
\captionof{table}{Wall-clock training loss of OPT-1.3B.}
\label{tab:wallclock13}
\begin{tabular}{lccccccc}
\specialrule{1.9pt}{0pt}{0pt}
 & \multicolumn{7}{c}{\textbf{Wall-clock training loss}} \\
\cmidrule(lr){2-8}
\textbf{Algorithm} & \textbf{600s} & \textbf{1200s} & \textbf{1800s} & \textbf{2400s} & \textbf{3000s} & \textbf{3600s} & \textbf{4200s} \\
\midrule
Muon        & 7.8490 & 7.1454 & 6.7588 & 6.6044 & 6.3795 & 6.2671 & 6.1316 \\
AdaSGDA     & 7.5296 & 6.9236 & 6.4627 & 6.3201 & 6.2450 & 6.1573 & 6.0156 \\
AdaSGDA-VR  & 7.3432 & 6.6373 & 6.4266 & 6.2602 & 6.1029 & 6.0209 & 5.8820 \\
\specialrule{1.9pt}{0pt}{0pt}
\end{tabular}
\end{minipage}

\vspace{0.41cm}

\begin{minipage}{\linewidth}
\centering
\setlength{\tabcolsep}{6pt}
\renewcommand{\arraystretch}{1.12}
\captionof{table}{Test accuracy with stepsizes smaller than $\eta^*$ on FMNIST.}
\label{tab:stepsize}
\begin{tabular}{lcccc}
\specialrule{1.9pt}{0pt}{0pt}
\textbf{Algorithm} & \textbf{$\eta^*\times 0.05$} & \textbf{$\eta^*\times 0.1$} & \textbf{$\eta^*\times 0.5$} & \textbf{$\eta^*$} \\
\midrule
AdaSGDA     & 88.85\textsubscript{$\pm$0.07} &  88.83\textsubscript{$\pm$0.07} &  88.16\textsubscript{$\pm$0.08} &  89.53\textsubscript{$\pm$0.06}\\
AdaSGDA-VR  & 90.15\textsubscript{$\pm$0.05} &  90.12\textsubscript{$\pm$0.07}  &  90.22\textsubscript{$\pm$0.06} &  90.26\textsubscript{$\pm$0.04} \\
Muon        &  87.90\textsubscript{$\pm$0.08} &  88.01\textsubscript{$\pm$0.06} &  88.11\textsubscript{$\pm$0.08} &  88.19\textsubscript{$\pm$0.09}  \\
D-AdaST     &  81.82\textsubscript{$\pm$0.03} &  83.65\textsubscript{$\pm$0.04} &  86.76\textsubscript{$\pm$0.08} &  87.79\textsubscript{$\pm$0.08} \\
SGDA        &  80.33\textsubscript{$\pm$0.07} &  82.03\textsubscript{$\pm$0.08} &  86.55\textsubscript{$\pm$0.06} &  87.32\textsubscript{$\pm$0.06}  \\
\specialrule{1.9pt}{0pt}{0pt}
\end{tabular}
\end{minipage}

\vspace{0.41cm}

\begin{minipage}{\linewidth}
\centering
\setlength{\tabcolsep}{6pt}
\renewcommand{\arraystretch}{1.12}
\captionof{table}{Test accuracy with stepsizes larger than $\eta^*$ on FMNIST.}
\label{tab:stepsize2}
\begin{tabular}{lcccc}
\specialrule{1.9pt}{0pt}{0pt}
\textbf{Algorithm}  & \textbf{$\eta^*\times 5$} & \textbf{$\eta^*\times 10$} & \textbf{$\eta^*\times 50$}&\textbf{$\eta^*\times 100$} \\
\midrule
AdaSGDA   & 88.88\textsubscript{$\pm$0.10} & 88.94\textsubscript{$\pm$0.08} & 88.93\textsubscript{$\pm$0.09}& 88.96\textsubscript{$\pm$0.11} \\
AdaSGDA-VR   & 90.10\textsubscript{$\pm$0.09} & 90.12\textsubscript{$\pm$0.09} & 90.05\textsubscript{$\pm$0.12}& 89.99\textsubscript{$\pm$0.07} \\
Muon     & 84.21\textsubscript{$\pm$0.11} & 80.86\textsubscript{$\pm$0.12} & 10.09\textsubscript{$\pm$0.03}& 10.09\textsubscript{$\pm$0.03} \\
D-AdaST    & 87.52\textsubscript{$\pm$0.11} & 87.30\textsubscript{$\pm$0.11} & 86.51\textsubscript{$\pm$0.16} & 9.59\textsubscript{$\pm$0.05} \\
SGDA      & 85.03\textsubscript{$\pm$0.13} & 82.26\textsubscript{$\pm$0.15} & 81.98\textsubscript{$\pm$0.12}& 19.63\textsubscript{$\pm$0.06} \\
\specialrule{1.9pt}{0pt}{0pt}
\end{tabular}
\end{minipage}

\vspace{0.41cm}

\begin{minipage}{\linewidth}
\centering
\setlength{\tabcolsep}{6pt}
\renewcommand{\arraystretch}{1.12}
\captionof{table}{Test accuracy versus momentum coefficient $\beta$ on the FMNIST dataset.}
\label{tab:main-resultsbeta}
\begin{tabular}{lccccc}
\specialrule{1.9pt}{0pt}{0pt}
\textbf{Algorithm} & \textbf{$\beta=0.1$} & \textbf{$\beta=0.3$} & \textbf{$\beta=0.5$} & \textbf{$\beta=0.7$} & \textbf{$\beta=0.9$} \\
\midrule
AdaSGDA     &  89.20\textsubscript{$\pm$0.07} & 89.32\textsubscript{$\pm$0.06} & 89.48\textsubscript{$\pm$0.09} & 89.46\textsubscript{$\pm$0.08} &  89.49\textsubscript{$\pm$0.06}  \\
AdaSGDA-VR  &  89.48\textsubscript{$\pm$0.05} & 90.10\textsubscript{$\pm$0.07} & 90.06\textsubscript{$\pm$0.10} & 90.24\textsubscript{$\pm$0.09} &  90.25\textsubscript{$\pm$0.06}  \\
Muon        &  87.46\textsubscript{$\pm$0.09} & 87.34\textsubscript{$\pm$0.10} & 86.68\textsubscript{$\pm$0.12} & 86.82\textsubscript{$\pm$0.10} & 87.96\textsubscript{$\pm$0.10}  \\
\specialrule{1.9pt}{0pt}{0pt}
\end{tabular}
\end{minipage}
\end{table}

\subsection{Additional Results on the FineWeb Dataset}
We further evaluate our methods on the FineWeb dataset under an embedding-level adversarial DRO formulation. 
For each training sample (token sequence) $S_i$, let $V_i = E_X(S_i)$ denote its input embedding sequence. 
We introduce an adversarial variable $Y_i$ of the same shape, representing a perturbed embedding, and consider the following minimax objective with a quadratic penalty:
\(
\min_{X}\ \max_{Y_1,\ldots,Y_n}\ \frac{1}{n}\sum_{i=1}^n \ell(X; Y_i, S_i)\;-\;\rho\|Y_i-V_i\|_F^2,
\)
where $\ell(X; Y_i, S_i)$ is the next-token prediction loss when using $Y_i$ as the input embedding, and $\rho>0$ controls the perturbation strength. 
This formulation trains the model against adversarial embedding perturbations in a local neighborhood around $V_i$, while discouraging overly large deviations through the quadratic regularization.

In \cref{opt350tt}, we report the training loss and validation loss for OPT-350M. 
Both {AdaSGDA} and {AdaSGDA-VR} achieve lower losses than {Muon} under the same DRO training setup, demonstrating consistently better empirical performance. 
In \cref{opt12512bOPtt}, we further compare validation loss on FineWeb for OPT-125M and OPT-1.3B. 
Across model scales, all three methods exhibit stable training trends, while the benefits of orthogonal updates become more pronounced as model size increases. 
This behavior can be attributed to the fact that, as model size increases, gradients and momentum directions tend to become more ill-conditioned and highly correlated across layers and parameters, so successive updates can interfere with each other and amplify unstable components in minimax training. 
Orthogonalized updates help alleviate this issue by reshaping the update directions to reduce such interference, thereby enabling more stable and reliable optimization at scale.
Notably, the performance gap between our methods and {Muon} widens at OPT-1.3B compared with OPT-125M and OPT-350M. 
Meanwhile, unlike {Muon}, our proposed methods remain parameter-free and thus eliminate extensive manual tuning, further supporting the effectiveness of our adaptive design for large-scale training.
Finally, {AdaSGDA-VR} achieves the best performance across all settings, consistent with our theory that variance reduction provides tighter stochastic error control and faster convergence.

We further report wall-clock training loss in \cref{tab:wallclock125,tab:wallclock350,tab:wallclock13} for OPT-125M, OPT-350M, and OPT-1.3B. Across all three model scales, the same ranking is observed at every reported time budget: AdaSGDA-VR achieves the lowest training loss at every reported wall-clock checkpoint across all three model scales, while AdaSGDA is uniformly better than Muon. This demonstrates that the empirical advantage of our methods is not limited to loss-versus-iteration comparisons, but also yields consistently better time-to-quality tradeoffs in real training time. This consistent trend across model scales suggests that our adaptive orthogonal design continues to remain effective as the model size increases. Overall, these wall-clock results provide additional evidence that our proposed methods are practically more efficient for large-scale minimax-based language pre-training.

\subsection{Sensitivity Analysis on Initial Stepsize and Momentum Coefficient}
We further conduct sensitivity analyses on the FMNIST dataset for the initial stepsize and the momentum coefficient $\beta$, with results summarized in Tables~\ref{tab:stepsize} and~\ref{tab:main-resultsbeta}.

\begin{itemize}[leftmargin=*,topsep=2pt,itemsep=2pt]
   \item \textbf{Initial stepsize.} As shown in \cref{tab:stepsize,tab:stepsize2}, $\eta^\star$ denotes the best-tuned stepsize for each baseline, while for our methods $\eta^\star$ is selected following the theoretical guidelines in \cref{theorem1,theoremVR}. We then scale $\eta^\star$ by $\{0.05,\,0.1,\,0.5,\,1,\,5,\,10,\,50,\,100\}$ to test robustness. All algorithms are evaluated for $100$ rounds for a fair comparison. The results show that our methods achieve the best accuracy across {all} stepsize scalings and exhibit remarkable resilience to stepsize selection, maintaining stable performance over a much broader range than the baselines.
   In sharp contrast, Muon fails to converge at the $50\times$ scale and all baselines break down at $100\times$, whereas our methods remain accurate and stable even with overly aggressive stepsizes, continuing to make steady progress without divergence or severe degradation. This highlights not only the strong stepsize robustness of our approach, but also its practical value: the method can be deployed with minimal tuning effort while remaining reliable across a wide range of hyperparameter choices.

    \item \textbf{Momentum coefficient.} Table~\ref{tab:main-resultsbeta} varies the momentum coefficient over $\beta\in\{0.1,0.3,0.5,0.7,0.9\}$. Across all choices, our methods consistently outperform Muon and show much smaller performance variation, indicating strong robustness to momentum tuning. This is expected: our adaptive stepsizes continuously calibrate the effective update magnitude using accumulated gradient information, which reduces the impact of momentum-induced overshooting and oscillations.
\end{itemize}

\subsection{Ablation Study}
\label{app:component-ablation}

We conduct ablation studies on FMNIST, CIFAR-10, and
CIFAR-100 to examine the contributions of the individual components of
AdaSGDA. Each ablated variant removes one component while keeping the
remaining settings unchanged:
\begin{itemize}[leftmargin=*,topsep=2pt,itemsep=2pt]
    \item \textbf{w/o orthogonalization:} replace the polar direction with a
    Frobenius-normalized direction while preserving the update magnitude;
    \item \textbf{w/o adaptivity:} use fixed stepsizes selected through the
    same validation protocol;
    \item \textbf{w/o hierarchical scaling:} scale the primal update using
    only its own accumulator;
    \item \textbf{w/o clamping:} accumulate the original, unclipped gradient
    norms;
    \item \textbf{w/o momentum:} set the momentum coefficient to zero.
\end{itemize}

\begin{table}[t]
    \begin{minipage}{\linewidth}
    \centering
    \setlength{\tabcolsep}{6pt}
    \renewcommand{\arraystretch}{1.12}
    \captionof{table}{Test accuracy for the ablation study on FMNIST, CIFAR-10, and CIFAR-100.}
    \label{tab:component-ablation}
    \begin{tabular}{lccc}
        \specialrule{1.9pt}{0pt}{0pt}
        \textbf{Algorithm} & \textbf{Fashion-MNIST} & \textbf{CIFAR-10} & \textbf{CIFAR-100} \\
        \midrule
        w/o Orthogonalization & 88.09\textsubscript{$\pm$0.10} & 75.56\textsubscript{$\pm$0.11} & 32.76\textsubscript{$\pm$0.11} \\
        w/o Adaptivity        & 89.08\textsubscript{$\pm$0.08} & 80.86\textsubscript{$\pm$0.08} & 38.73\textsubscript{$\pm$0.09} \\
        w/o Hierarchical Scaling & 89.05\textsubscript{$\pm$0.08} & 80.52\textsubscript{$\pm$0.11} & 38.82\textsubscript{$\pm$0.10} \\
        w/o Clamping          & 89.02\textsubscript{$\pm$0.06} & 80.56\textsubscript{$\pm$0.06} & 39.01\textsubscript{$\pm$0.08} \\
        w/o Momentum          & 86.42\textsubscript{$\pm$0.06} & 75.97\textsubscript{$\pm$0.09} & 33.52\textsubscript{$\pm$0.11} \\
        AdaSGDA               & 89.53\textsubscript{$\pm$0.06} & 80.91\textsubscript{$\pm$0.08} & 40.33\textsubscript{$\pm$0.08} \\
        AdaSGDA-VR            & {90.26}\textsubscript{$\pm$0.04} & {81.56}\textsubscript{$\pm$0.07} & {44.85}\textsubscript{$\pm$0.08} \\
        \specialrule{1.9pt}{0pt}{0pt}
    \end{tabular}
    \end{minipage}
\end{table}

Table~\ref{tab:component-ablation} shows that removing any individual component consistently leads to a degradation in performance, indicating that each design choice contributes meaningfully to the overall performance of the algorithm. In particular, orthogonalization and momentum play significant roles, as their removal results in the largest accuracy drops across all datasets. The adaptive stepsize scheme, hierarchical scaling, and gradient clamping also provide consistent and non-negligible improvements, further stabilizing and enhancing optimization across tasks. Moreover, variance reduction yields additional gains, and AdaSGDA-VR achieves the best overall performance, which is consistent with its improved theoretical convergence rate. Overall, these results demonstrate that all components are indispensable and jointly contribute to the strong empirical performance of the proposed method.

\end{document}